\documentclass{article}
\usepackage[utf8]{inputenc}
\usepackage{color}
\usepackage[colorlinks=true,urlcolor=black,citecolor=black,linkcolor=black]{hyperref}
\usepackage{amsmath}
\usepackage{amssymb}
\usepackage{amsfonts}
\usepackage{tikz}
\usepackage{svg}
\usetikzlibrary{shapes.geometric, arrows.meta, positioning, fit, calc}
\usepackage{algorithm}
\usepackage{algpseudocode}
\usepackage[a4paper, margin={2cm,2cm}]{geometry}
\usepackage{hyperref}

\usepackage{float}
\usepackage{subcaption}

\usepackage[
backend=biber,
style=authoryear,
sorting=nty,
]{biblatex}
\title{Probabilistic Inversion with Flow Matching}

\author{Baldur Paulwitz\footnote{\href{mailto:baldur.paulwitz@geophysik.tu-freiberg.de}{baldur.paulwitz@geophysik.tu-freiberg.de}}, \hspace{0.2cm}Stefan Buske \footnote{\href{mailto:stefan.buske@geophysik.tu-freiberg.de}{stefan.buske@geophysik.tu-freiberg.de}}\\
	Institute of Geophysics and Geoinformatics, TU Bergakademie Freiberg
}
\date{}

\begin{document}

\newcommand{\todo}[1]{\textcolor{red}{TODO: #1}}
\newcommand{\tocite}[1]{\textcolor{red}{TO CITE: #1}}

\label{firstpage}
\maketitle

\begin{abstract}
	We demonstrate the application of Flow Matching, a technique originating from generative Artificial Intelligence, to probabilistic inversion in geophysical settings, such as seismic Full-Waveform inversion. We adapt the well-established mathematical theory of Flow Matching from generative Artificial Intelligence to the context of probabilistic inversion. We evaluate the approach with two case studies: a simple 2D velocity model to illustrate the general features of the method, and the OpenFWI dataset to show its capabilities for probabilistic inversion of more complex seismic velocity models.
\end{abstract}

\section{Introduction }
Let $\mathbf{m} \in \mathcal{M}$ define a geophysical model (e.g. a seismic velocity model in the subsurface together with a certain acquisition geometry). Let $\mathbf{d_\text{obs}}$ be the observed data associated with $\mathbf{m}$. A forward operator $f: \mathcal{M} \rightarrow \mathcal{D}$ maps the model space $\mathcal{M}$ to the data space $\mathcal{D}$ and describes the underlying physical system, such that $f(\mathbf{m}) = \mathbf{d_\text{syn}}$ where $\mathbf{d_\text{syn}} \in \mathcal{D}$ and $\mathbf{d_\text{obs}} = \mathbf{d_\text{syn}} + \varepsilon$. Therefore, the observed data $\mathbf{d_\text{obs}}$ corresponds to the synthetic data $\mathbf{d_\text{syn}}$ plus some small uncorrelated noise $\varepsilon > 0$ from the measurement. The \textit{forward problem} refers to the task of determining the forward operator $f$. The \textit{inverse problem} addresses finding the inverse operator $f^{-1}: \mathcal{D} \rightarrow \mathcal{M}$, which maps the data space to the model space, thereby recovering the model parameters, i.e. the seismic velocity distribution, from the observed data.

In deterministic physics, the forward operator has a unique solution \parencite{tarantola_inverse_2005}. However, this is generally not the case for the inverse operator. Nevertheless, most modern inversion frameworks use \textit{deterministic inversion}, which pragmatically assumes $f^{-1}$ to be a unique mapping, because it is computationally easier to solve. As the inverse problem is ill-posed (in contrast to a well-posed problem, where a unique solution exists and the behavior changes continuously with respect to input changes), deterministic inversion requires a set of simplifications to obtain a reliable method. At first, an initial guess of the solution needs to be given as a starting point for the iterative solver. The solver needs to use a Taylor series approximation of the non-linear forward operator in order to compute a model update. Furthermore, significant regularization is required, usually by enforcing smoothness constraints. Lastly, the observed data may need to be pre-processed, e.g. in Full-Waveform Inversion of seismic data, where often multiple inversions are done on increasing frequency content to avoid cycle skipping \parencite{bunks_multiscale_1995,virieux_overview_2009}.

\textit{Probabilistic inversion} tries to express the non-uniqueness of the inverse operator by finding a probability distribution $p(\mathbf{m} | \mathbf{d_\text{syn}})$, from which samples $\mathbf{z} \sim p(\mathbf{m} | \mathbf{d_\text{syn}})$ are possible solutions with $f(\mathbf{z}) = \mathbf{d_\text{syn}}$. In probabilistic inversion, no initial guess for the solution or regularization is required and the forward operator does not need to be approximated with a Taylor series. Although this approach is arguably more elegant and allows the assessments of uncertainty, it is not often used today. The reason for this is, that finding the distribution $p(\mathbf{m} | \mathbf{d_\text{syn}})$ with traditional sampling-based methods requires evaluating the forward operator in each step, which scales roughly exponentially with the amount of model parameters \parencite{fichtner_full_2011}. This is computationally too demanding for real-world applications with up to millions of model parameters.

Recently, there have been significant improvements in probabilistic inversion, where the problem is reframed as an optimization problem. This was mostly driven by research in Deep Learning methods. Variational Inference with Normalizing Flows \parencite{zhao_bayesian_2021} constructs a (discrete) series of Normalizing Flows that convert a known initial distribution into the target distribution. The Flows are trained by maximizing the Evidence Lower Bound and hence minimizing the Kullback-Leibler divergence between the prior and the target distribution. In a different area of research, Diffusion Models are used to obtain the target distribution \parencite{zhang_diffusionvel_2024}. In Diffusion Models, a stochastic process converts a known initial distribution into the target distribution.

Flow Matching \parencite{lipman_flow_2023} can be seen as a bridge between Variational Inference with Normalizing Flows and Diffusion Models. It is also a Deep Learning method that iteratively transforms a known initial distribution, such as the d-dimensional Normal distribution, into either an unconditioned target distribution, like $p(\mathbf{m})$, or a conditioned target distribution, like $p(\mathbf{m} | \mathbf{d_\text{syn}})$. In the literature, the first case is referred to as \textit{unguided Flow Matching} and the second case as \textit{guided Flow Matching}.

This transformation of probability distribution is achieved by learning Continuous Normalizing Flows (CNF) during the training process. At inference, a simple Ordinary Differential Equation (ODE) is solved.\\
In the following, the Flow Matching theory will be described in more detail with the help of a simple case study of a 2D velocity model. The mathematical framework to be used for that is largely based on \parencite{lipman_flow_2023,holderrieth_introduction_2026,lipman_flow_2024} and adapted for the use case of probabilistic inversion of geophysical data, because  Flow Matching is generally used for generative AI \parencite{davtyan_efficient_2023,song_equivariant_2023,wu_fast_2022,chen_flow_2024,le_voicebox_2023} and is not commonly used for probabilistic inversion within the geophysics community yet.

\section{Case Study: Simple 2D Velocity Model}

\begin{figure}
    \centering
    \includegraphics[width=\linewidth]{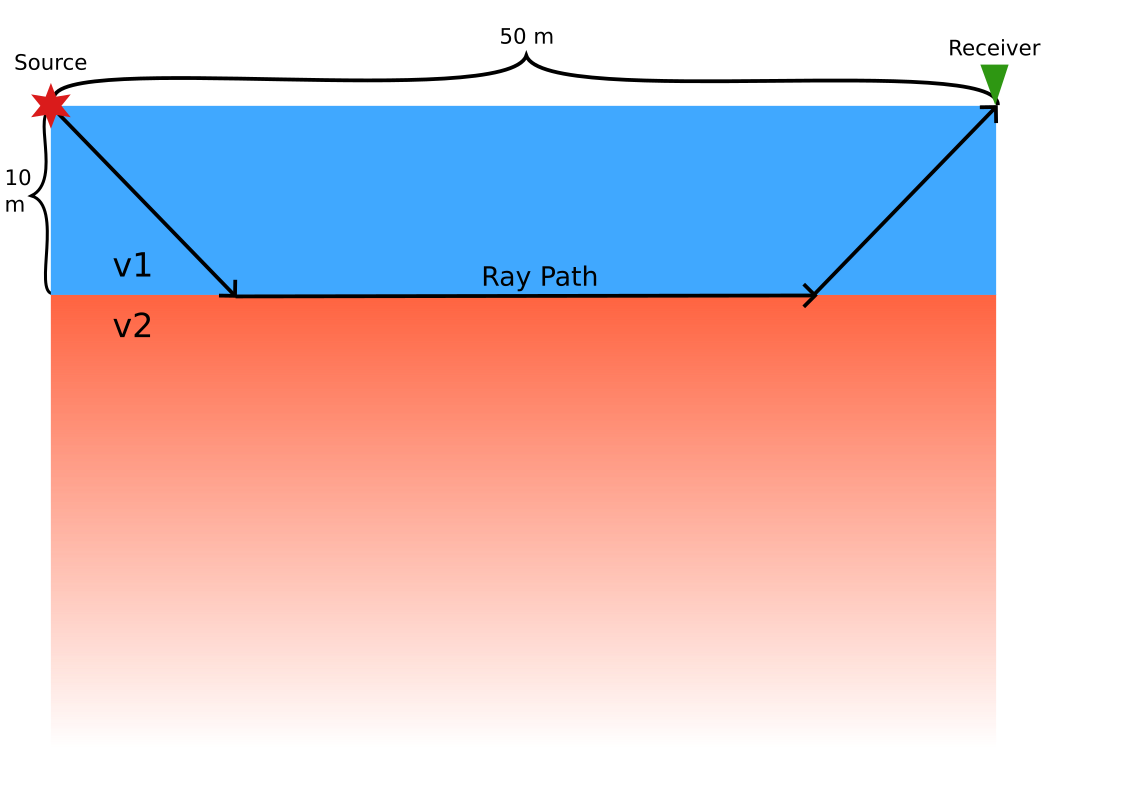}
    \caption{Simple 2D velocity model.}
    \label{fig-setup_toy_example}
\end{figure}

As a first step, we illustrate the Flow Matching method on a minimalistic probabilistic inversion scenario. We consider a simple 2D velocity model with two layers, as shown in \autoref{fig-setup_toy_example}. The first layer has a known thickness of 10 metres and the second layer is a homogeneous half space. The velocity model can therefore be described by the two layer velocities $v_1 \in [300, 700] \frac{\text{m}}{\text{s}}$ and $v_2\in [s \cdot v_1, 5000] \frac{\text{m}}{\text{s}}$ where $s$ is the minimum factor for which the ray of the refracted wave travels at least 1 m with the speed of $v_2$.\
To condition (guide) the probabilistic inversion by some observed data, the traveltime of the refracted wave is computed for a source and receiver pair in a fixed distance of 50 metres. The forward problem (measuring the traveltime of the refracted wave for a given velocity model) is unique, but the inverse problem (obtaining a possible velocity field given some observed traveltime of the refracted wave) is not unique, because different sets of $v_1, v_2$ may result in the same traveltime of the refracted wave.

\subsection{Initial and Target distribution}
While any initial distribution may be used where a Continuous Normalizing Flow to the target distribution can be constructed, the d-dimensional Normal distribution is most popular in practice, as it is easy to sample from it and to construct the Flows. Therefore, we will focus on the initial distribution $p_\text{init} = \mathcal{N}(0, \mathbf{I}_d)$. In the case of the current scenario, where the seismic velocity model has two parameters, $d = 2$.

The target distribution is unknown, but we assume that we have a finite set of samples from it, which will later serve as the training dataset. In our example case, the target distribution would be a distribution of all possible seismic velocity models (for unguided Flow Matching) or the joint distribution of all possible seismic velocity fields and all possible observed data points (for guided Flow Matching). We denote the target distribution with $p_\text{target}$ and its samples as $\mathbf{m}\sim p_\text{target}$ with $\mathbf{m}\in \mathbb{R}^d$.

\subsection{Continuous Normalizing Flows}
In Flow Matching, samples of the initial distribution are transformed into samples of the target distribution along some time-dependent trajectory $X_t: [0, 1] \rightarrow \mathbb{R}^d$. For any given time $t\in[0,1]$, $X_t$ returns a location in $\mathbb{R}^d$. Time $t$ represents a time step on the trajectory in this case, where at $t=0$ the sample on the trajectory lies in the initial distribution and at $t=1$, it lies in the target distribution. This trajectory is the solution to the ODE

\begin{equation}
\label{eq-ode-trajectory}
\frac{\text{d}}{\text{d}t}X_t = u_t(X_t)
\end{equation}

where $u_t(X_t)$ is a time-dependent vector field $u_t: [0, 1]\times \mathbb{R}^d \rightarrow \mathbb{R}^d$ that specifies the velocity in space, given some time $t$ and a location $\mathbf{x}\in \mathbb{R}^d$. The initial condition of the ODE in \autoref{eq-ode-trajectory} is $X_0 = \mathbf{\epsilon}$ where $\mathbf{\epsilon} \sim p_\text{init}$ is the starting point of the trajectory.

In general, the trajectory needs to depend on the specific sample from the initial distribution, yielding different trajectories for different samples. For this, we use a Flow $\psi_t: [0,1]\times \mathbb{R}^d\rightarrow \mathbb{R}^d$, which is a function that requires a time $t$ and some location $\mathbf{x} \in \mathbb{R}^d$ and returns, like the previously introduced trajectory, a location in $\mathbb{R}^d$. As $t\in [0,1]$ is continuous, we obtain a Continuous Normalizing Flow (CNF).

With this, we need to solve the ODE

\begin{equation}
\label{eq-ode-flow}
\frac{\text{d}}{\text{d}t}\psi_t(\mathbf{x}) = u_t(\psi_t(\mathbf{x}))
\end{equation}

instead, where the initial condition is $\psi_0(\mathbf{x}) = \mathbf{x}$. The time-dependent vector field $u_t$ will be approximated with a Neural Network (NN) $u^\theta_t$ with the parameters $\theta$.

\subsection{Gaussian Conditional Probability Path}
In Flow Matching, the CNFs follow \textit{Probability Paths}, as they transport samples from one probability distribution to samples from another probability distribution. First, let's consider the (time-dependent) Gaussian Conditional Probability Path $p_t(\cdot| \mathbf{m})$ with $\mathbf{m} \sim p_\text{target}$. It is conditioned on a sample of the target distribution and corresponds to the Flow from the Normal distribution $p_\text{init} = \mathcal{N}(0, \mathbf{I}_d)$ to the Dirac distribution $\delta_\mathbf{m}$ (sampling from $\delta_\mathbf{m}$ always returns $\mathbf{m}$). Therefore $p_0(\cdot | \mathbf{m}) = p_\text{init}$ and $p_1(\cdot | \mathbf{m}) = \delta_\mathbf{m}$.

Such a Gaussian Conditional Probability Path can be constructed with

\begin{equation}
\label{eq-gcpp}
p_t(\cdot | \mathbf{m}) = \mathcal{N}(\alpha_t \mathbf{m}, \beta_t^2 \mathbf{I}_d)
\end{equation}

where $\alpha_t$ and $\beta_t$ are time-dependent differentiable monotonic functions (noise schedulers) with $\alpha_0 = \beta_1 = 0$ and $\alpha_1 = \beta_0 = 1$. In the most basic form, they can be chosen as $\alpha_t = t$ and $\beta_t = 1 - t$, which we do in this work. To sample from the distribution in \autoref{eq-gcpp}, we can use the reparametrization trick:

\begin{equation}
\label{eq-reparametrization}
\mathbf{m}\sim p_\text{target}, \mathbf{\epsilon}\sim p_\text{init} = \mathcal{N}(0, \mathbf{I}_d) \Rightarrow \alpha_t \mathbf{m} + \beta_t \mathbf{\epsilon} = \mathbf{x} \sim p_t
\end{equation}

In order to train a NN to learn the vector field for the Gaussian Conditional Probability Path, the target vector field $u_t^\text{target}$ needs to be computed. For this, the conditional Flow is constructed as $\psi_t^\text{target}(\mathbf{x}|\mathbf{m}) = \alpha_t \mathbf{m} + \beta_t \mathbf{\epsilon}$ and then the ODE is solved for the conditional vector field:

\begin{equation}
\label{eq-cond-train}
\begin{split}
&\frac{\text{d}}{\text{d}t}\psi_t^\text{target}(\mathbf{\epsilon} | \mathbf{m}) = u_t^\text{target}(\psi_t^\text{target}(\mathbf{\epsilon} | \mathbf{m}) | \mathbf{m})\\
&\underbrace{\frac{\text{d}}{\text{d}t}\alpha_t}_{=1}\mathbf{\mathbf{m}} + \underbrace{\frac{\text{d}}{\text{d}t}\beta_t}_{= -1}\mathbf{\mathbf{\epsilon}} = u_t^\text{target}(\underbrace{\alpha_t}_{=t} \mathbf{m} + \underbrace{\beta_t}_{=1-t} \mathbf{\epsilon} | \mathbf{m})\\
&\text{reparametrizing } \mathbf{x} = \alpha_t \mathbf{m} + \beta_t \mathbf{\epsilon} = t \mathbf{m} + (1 - t) \mathbf{\epsilon}\\
&\mathbf{m} - \mathbf{\epsilon} = u_t^\text{target}(\mathbf{x} | \mathbf{m})\\
\end{split}
\end{equation}

This leads to the Conditional Flow Matching loss:

\begin{equation}
\label{eq-cond-fm-loss}
\mathcal{L}_\text{CFM}(\theta) = \mathbb{E}_{t\sim \text{Unif}_{[0, 1]}, \mathbf{m}\sim p_\text{target},\mathbf{\epsilon} \sim \mathcal{N}(0, \mathbf{I}_d)}\left[||u_t^\theta(\mathbf{x}) - (\mathbf{m} - \mathbf{\epsilon})||^2\right]
\end{equation}

We can now implement a minimalistic function to train a NN (in this case, a simple Multi-Layer Perceptron was used) on the objective function given in \autoref{eq-cond-fm-loss} for the case when the condition is sampled from the Dirac distribution $\mathbf{m}\sim \delta_\mathbf{m}$:

\begin{algorithm}
\caption{Pseudocode for Conditional Flow Matching Loss conditioned on a single model from the target distribution.}\label{alg-cond}
\begin{algorithmic}
\Require $\mathbf{m}\sim \delta_\mathbf{m}$, Neural Network $u^\theta(t, \mathbf{x})$, learning rate $\eta$, amount of steps $n$, index $i$.
\State $i\gets 0$
\While{$i < n$}
    \State $\mathbf{\epsilon} \sim \mathcal{N}(0, \mathbf{I}_2)$ \Comment{($\mathbf{m} \in \mathbb{R}^2 \Rightarrow \mathbf{I}_2$)}
    \State $t \sim \text{Unif}_{[0, 1]}$ \Comment{Sample time $t$}
    \State $\mathbf{x} \gets t\cdot \mathbf{m} + (1 - t)\cdot \mathbf{\epsilon}$ \Comment{$\mathbf{x} \sim p_t$}
    \State $u_t^\text{target} \gets \mathbf{m} - \mathbf{\epsilon}$
    \State $u_t^\text{prediction} \gets u^\theta(t, \mathbf{x})$
    \State $\mathcal{L}_\text{CFM} \gets \mathbb{E}\left[||u_t^\text{prediction} - u_t^\text{target}||^2\right]$ \Comment{Conditional Flow Matching objective}
    \State $\theta \gets \theta - \eta \nabla_\theta \mathcal{L}_\text{CFM}$ \Comment{Update parameters via gradient descent}
    \State $i \gets i + 1$
\EndWhile
\end{algorithmic}
\end{algorithm}

\begin{figure}
    \centering
    %\includesvg[width=\linewidth]{img/toy_example.svg}
    \includegraphics[width=0.5\linewidth]{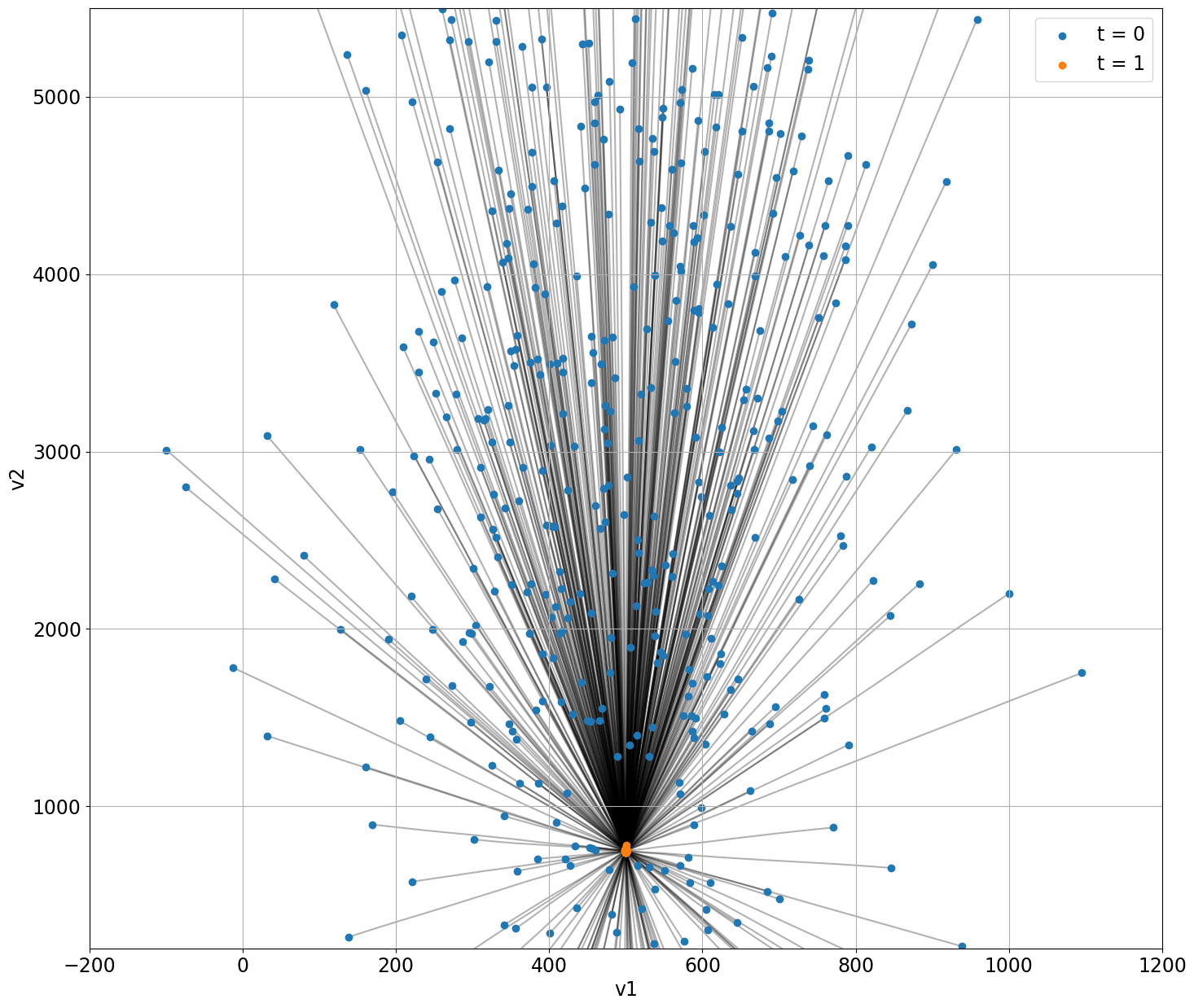}
    \caption{Conditional Flow Matching for a Dirac distribution $\delta_\mathbf{m}$ with $\mathbf{m}_{v1} = 500$ and $\mathbf{m}_{v2} = 750$ for $\mathbf{m}\sim \delta_\mathbf{m}$. Blue dots are samples from the Normal distribution (after projecting them into the parameter range for the velocity model), the black lines show the Flow over time $t$ and the orange dots are samples from the target distribution $\delta_\mathbf{m}$.}
    \label{fig-conditional-fm}
\end{figure}

Sampling from the target distribution $\delta_\mathbf{m}$ is then only a matter of solving the ODE in \autoref{eq-ode-flow} with the initial condition of $\psi_0(\mathbf{x}) = \mathbf{\epsilon}\sim p_\text{init}$, which is shown in \autoref{fig-conditional-fm}.

\subsection{Marginal Probability Path}
While the Dirac distribution is of little use in practical applications, the Conditional Flow Matching serves as a building block for Marginal Flow Matching, where the marginal distribution is learned. A Gaussian Conditional Probability Path induces a Marginal Probability Path. Sampling from the Marginal Probability Path can be done by first sampling from the target distribution, then using it as a condition for the Gaussian Conditional Probability Path, and finally sampling from the Gaussian Conditional Probability Path: $\mathbf{m}\sim p_\text{target}, \mathbf{x}\sim p_t(\cdot | \mathbf{m}) \Rightarrow \mathbf{x} \sim p_t(\mathbf{m})$.

The density can theoretically be computed by marginalization:

\begin{equation}
\label{eq-marginalization}
p_t(\mathbf{x}) = \int p_t(\mathbf{x} | \mathbf{m}) p_\text{target}(\mathbf{m}) \text{d}\mathbf{m}.
\end{equation}

However, \autoref{eq-marginalization} is generally intractable.

Just as for the Gaussian Conditional Probability Path, the target vector field $u_t^\text{target}$ needs to be computed to train the NN. As shown in \parencite{holderrieth_introduction_2026}, $u_t^\text{target}$ is only valid if and only if it satisfies the continuity equation:

\begin{equation}
\label{eq-continuity}
\frac{\text{d}}{\text{d}t}p_t(\mathbf{x}) = -\nabla\cdot\left(p_t u_t^\text{target}\right)(\mathbf{x}) \quad\forall \mathbf{x}\in \mathbb{R}^d, 0\le t\le 1
\end{equation}

where $\nabla \cdot$ is the divergence operator. With \autoref{eq-continuity} and \autoref{eq-marginalization}, $u_t^\text{target}$ can be derived as:

\begin{equation*}
\begin{split}
\frac{\text{d}}{\text{d}t}p_t(\mathbf{x}) &= \frac{\text{d}}{\text{d}t}\int p_t(\mathbf{x} | \mathbf{m}) p_\text{target}(\mathbf{m}) \text{d}\mathbf{m} = \int\frac{\text{d}}{\text{d}t} p_t(\mathbf{x} | \mathbf{m}) p_\text{target}(\mathbf{m}) \text{d}\mathbf{m} =\\
&= \int -\nabla\cdot\left(p_t(\cdot | \mathbf{m}) u_t^\text{target}(\cdot | \mathbf{m})\right)(\mathbf{x}) p_\text{target}\text{d}\mathbf{m} =\\
&= -\nabla\cdot\left(\int p_t(\mathbf{x} | \mathbf{m}) u_t^\text{target}(\mathbf{x} | \mathbf{m}) p_\text{target}\text{d}\mathbf{m}\right) =\\
&= -\nabla\cdot p_t(\mathbf{x}) \left(\underbrace{\int u_t^\text{target}(\mathbf{x} | \mathbf{m}) \frac{p_t(\mathbf{x} | \mathbf{m}) p_\text{target}(\mathbf{m})}{p_t(\mathbf{x})}\text{d}\mathbf{m}}_{\overset{!}{=} u_t^\text{target}}\right)(\mathbf{x}) =\\
&= -\nabla\cdot\left(p_tu_t^\text{target}\right)(\mathbf{x})\\
\end{split}
\end{equation*}

Therefore, the target marginal vector field is defined as:

\begin{equation}
\label{eq-marginal-vecfield}
u_t^\text{target}(\mathbf{x}) = \int u_t^\text{target}(\mathbf{x} | \mathbf{m})\frac{p_t(\mathbf{x} | \mathbf{m}) p_\text{target}(\mathbf{m})}{p_t(\mathbf{x})}\text{d} \mathbf{m}.
\end{equation}

With \autoref{eq-marginal-vecfield}, the Flow Matching loss for the marginal distribution can be derived:

\begin{equation}
\label{eq-marginal-fm-loss}
\mathcal{L}_\text{FM}(\theta) = \mathbb{E}_{t\sim \text{Unif}_{[0, 1]}, \mathbf{x}\sim p_t}\left[||u_t^\theta(\mathbf{\mathbf{x}}) - u_t^\text{target}(\mathbf{x}))||^2\right].
\end{equation}

Unfortunately, $u_t^\text{target}$ is still intractable, but as shown by \parencite{lipman_flow_2023}, the (marginal) Flow Matching loss is equivalent to the Conditional Flow Matching Loss up to a constant, when the condition is not sampled from the Dirac distribution $\delta_\mathbf{m}$, but from $p_\text{target}$:

\begin{equation*}
\begin{split}
\mathcal{L}_\text{FM}(\theta) &= \mathbb{E}_{t\sim \text{Unif}_{[0, 1]}, \mathbf{x}\sim p_t}\left[||u_t^\theta(\mathbf{x}) - u_t^\text{target}(\mathbf{x}))||^2\right] \\
&= \mathbb{E}_{t\sim \text{Unif}_{[0, 1]}, \mathbf{x}\sim p_t}\left[||u_t^\theta(\mathbf{x}) ||^2 - 2u_t^\theta(\mathbf{x})^Tu_t^\text{target}(\mathbf{x}) + ||u_t^\text{target}(\mathbf{x}))||^2\right] \\
&= \mathbb{E}_{t\sim \text{Unif}_{[0, 1]}, \mathbf{x}\sim p_t}\left[||u_t^\theta(\mathbf{x}) ||^2\right] - 2\mathbb{E}_{t\sim \text{Unif}_{[0, 1]}, \mathbf{x}\sim p_t}\left[u_t^\theta(\mathbf{x})^Tu_t^\text{target}(\mathbf{x})\right] +\\
&\quad +\underbrace{\mathbb{E}_{t\sim \text{Unif}_{[0, 1]}, \mathbf{x}\sim p_t}\left[||u_t^\text{target}(\mathbf{x}))||^2\right]}_{=: C_1} \\
&= \mathbb{E}_{t\sim \text{Unif}_{[0, 1]}, \mathbf{x}\sim p_t}\left[||u_t^\theta(\mathbf{x}) ||^2\right] - 2\mathbb{E}_{t\sim \text{Unif}_{[0, 1]}, \mathbf{x}\sim p_t}\left[u_t^\theta(\mathbf{x})^Tu_t^\text{target}(\mathbf{x})\right] + C_1\\
\end{split}
\end{equation*}

where the second summand is:

\begin{equation*}
\begin{split}
\mathbb{E}_{t\sim \text{Unif}_{[0, 1]}, \mathbf{x}\sim p_t}\left[u_t^\theta(\mathbf{x})^Tu_t^\text{target}(\mathbf{x})\right] &= \int_0^1 \int p_t(\mathbf{x}) u_t^\theta(\mathbf{x})^T u_t^\text{target}(\mathbf{x})\text{ d} \mathbf{x} \text{ d} t\\
&= \int_0^1 \int p_t(\mathbf{x}) u_t^\theta(\mathbf{x})^T \left[\int u_t^\text{target}(\mathbf{x} | \mathbf{m})\frac{p_t(\mathbf{x} | \mathbf{m}) p_\text{target}(\mathbf{m})}{p_t(\mathbf{x})}\text{ d}\mathbf{m}\right] \text{ d} \mathbf{x} \text{ d} t\\
&= \int_0^1 \int \int u_t^\theta(\mathbf{x})^T  u_t^\text{target}(\mathbf{x} | \mathbf{m}) p_t(\mathbf{x} | \mathbf{m}) p_\text{target}(\mathbf{m})\text{ d}\mathbf{m} \text{ d} \mathbf{x} \text{ d} t\\
&= \mathbb{E}_{t\sim\text{Unif}_{[0,1]},\mathbf{m}\sim p_\text{target}, \mathbf{x}\sim p_t(\cdot|\mathbf{m})}\left[u_t^\theta(\mathbf{x})^T u_t^\text{target}(\mathbf{x}|\mathbf{m})\right].
\end{split}
\end{equation*}

This can be used again in the previous equation:

\begin{equation*}
\begin{split}
\mathcal{L}_\text{FM}(\theta) &= \mathbb{E}_{t\sim \text{Unif}_{[0, 1]}, \mathbf{x}\sim p_t}\left[||u_t^\theta(\mathbf{x}) ||^2\right] - 2\mathbb{E}_{t\sim \text{Unif}_{[0, 1]}, \mathbf{x}\sim p_t}\left[u_t^\theta(\mathbf{x})^Tu_t^\text{target}(\mathbf{x})\right] + C_1\\
&= \mathbb{E}_{t\sim \text{Unif}_{[0, 1]}, \mathbf{x}\sim p_t}\left[||u_t^\theta(\mathbf{x}) ||^2\right] - 2\mathbb{E}_{t\sim\text{Unif}_{[0,1]},\mathbf{m}\sim p_\text{target}, \mathbf{x}\sim p_t(\cdot|\mathbf{m})}\left[u_t^\theta(\mathbf{x})^T u_t^\text{target}(\mathbf{x}|\mathbf{m})\right] + C_1\\
&= \mathbb{E}_{t\sim \text{Unif}_{[03, 1]}, \mathbf{m}\sim p_\text{target}, \mathbf{x}\sim p_t(\cdot | \mathbf{m})}\Bigl[||u_t^\theta(\mathbf{x}) ||^2 - 2u_t^\theta(\mathbf{x})^T u_t^\text{target}(\mathbf{x}|\mathbf{m}) +\\
&\quad +||u_t^\text{target}(\mathbf{x}|\mathbf{m})||^2 - ||u_t^\text{target}(\mathbf{x}|\mathbf{m})||^2 \Bigr] + C_1\\
&= \mathbb{E}_{t\sim \text{Unif}_{[0, 1]}, \mathbf{m}\sim p_\text{target}, \mathbf{x}\sim p_t(\cdot | \mathbf{m})}\left[||u_t^\theta(\mathbf{x}) - u_t^\text{target}(\mathbf{x}|\mathbf{m}) ||^2 \right] +\\
&\quad +\underbrace{\mathbb{E}_{t\sim \text{Unif}_{[0, 1]}, \mathbf{m}\sim p_\text{target}, \mathbf{x}\sim p_t(\cdot | \mathbf{m})}\left[- ||u_t^\text{target}(\mathbf{x}|\mathbf{m})||^2 \right]}_{=: C_2} + C_1\\
&= \mathcal{L}_\text{CFM}(\theta) + \underbrace{C_2 + C_1}_{=: C}.
\end{split}
\end{equation*}

To train the NN, the gradient of the loss function with respect to the parameters of the NN is computed:

\begin{equation}
\label{eq-grad-equality-cfm-fm}
\nabla \mathcal{L}_\text{FM}(\theta) = \nabla\left(\mathcal{L}_\text{CFM}(\theta) + C\right) = \nabla\mathcal{L}_\text{CFM}(\theta).
\end{equation}

Therefore, if the NN is trained with a Stochastic Gradient Descent algorithm, minimizing the Conditional Flow Matching loss is equivalent to minimizing the (marginal) Flow Matching loss.

To demonstrate this in our simple seismic velocity model example, we therefore only need to change the algorithm such that different $\mathbf{m} \sim p_\text{target}$ are sampled in each training iteration, as in algorithm \autoref{alg-marginal}.

\begin{algorithm}
\caption{Pseudocode for Marginal Flow Matching.}\label{alg-marginal}
\begin{algorithmic}
\Require Neural Network $u^\theta(t, \mathbf{x})$, learning rate $\eta$, amount of steps $n$, index $i$.
\State $i\gets 0$
\While{$i < n$}
    \State $\mathbf{m}\sim p_\text{target}$
    \State $\mathbf{\epsilon} \sim \mathcal{N}(0, \mathbf{I}_2)$ \Comment{($\mathbf{m} \in \mathbb{R}^2 \Rightarrow \mathbf{I}_2$)}
    \State $t \sim \text{Unif}_{[0, 1]}$ \Comment{Sample time $t$}
    \State $\mathbf{x} \gets t\cdot \mathbf{m} + (1 - t)\cdot \mathbf{\epsilon}$ \Comment{$\mathbf{x} \sim p_t$}
    \State $u_t^\text{target} \gets \mathbf{m} - \mathbf{\epsilon}$
    \State $u_t^\text{prediction} \gets u^\theta(t, \mathbf{x})$
    \State $\mathcal{L}_\text{CFM} \gets \mathbb{E}\left[||u_t^\text{prediction} - u_t^\text{target}||^2\right]$ \Comment{Conditional Flow Matching objective}
    \State $\theta \gets \theta - \eta \nabla_\theta \mathcal{L}_\text{CFM}$ \Comment{Update parameters via gradient descent}
    \State $i \gets i + 1$
\EndWhile
\end{algorithmic}
\end{algorithm}

After training the NN, samples of the marginal distribution can again be obtained by sampling from the initial (Normal) distribution and solving the ODE with the trained NN:

\begin{figure}
    \centering
    \includegraphics[width=0.5\linewidth]{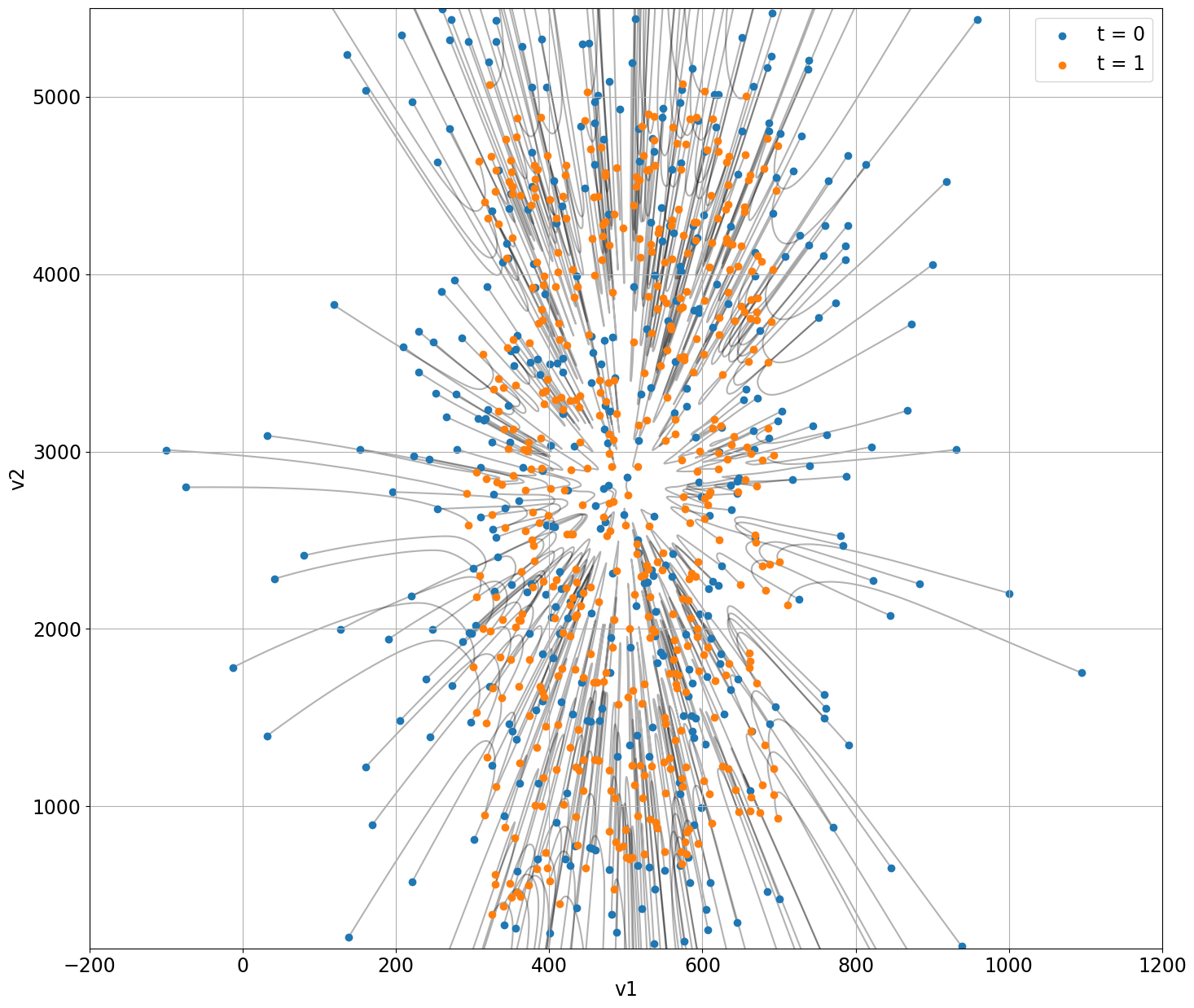}
    \caption{(Marginal) Flow Matching. Blue dots are samples from the Normal distribution (after projecting them into the parameter range for the velocity model), the black lines show the Flow over time and the orange dots are samples from the target distribution.}
    \label{fig-marginal-fm}
\end{figure}

\subsection{Guided Flow Matching and Probabilistic Inversion}
For probabilistic inversion, we are interested in sampling from the marginal distribution conditioned on some observed data. To avoid confusion with the 'Conditional Flow Matching', we refer to this scenario as 'Guided Flow Matching', which is common in the literature \parencite{lipman_flow_2024,holderrieth_introduction_2026}.

In addition to a 'model distribution' $p_\text{model}$ (e.g. a distribution of seismic velocity models), a 'data distribution' $p_\text{data}$ exists, which represents the distribution of all possible data outputs for all models. Until now, we only used the model distribution with $p_\text{target} = p_\text{model}$. In Guided Flow Matching, we need to use the joint distribution $p_\text{model,data}$ as training data.

Each sample $(\mathbf{m}, \mathbf{d}) \sim p_\text{model,data}$ contains some model $\mathbf{m}$ and the synthetic or observed data $\mathbf{d} = f(\mathbf{m})$ where $f$ is the forward operator. It is important to note that not even the forward operator $f$ needs to be deterministic (e.g. in text-guided Flow Matching for image generation, where the same photo may have different text captions).

As the probabilistic inversion result (the model distribution, given some data) depends on the data, the CNF, and therefore the NN, needs to depend on the data as well. This leads to the Guided Flow Matching Loss:

\begin{equation}
\label{eq-guided-fm-loss}
\mathcal{L}_\text{CFM}^\text{guided}(\theta) = \mathbb{E}_{(\mathbf{m},\mathbf{d})\sim p_\text{model,data},t\sim \text{Unif}_{[0,1]},\mathbf{x}\sim p_t(\cdot| \mathbf{m})}\left[||u_t^\theta(\mathbf{x}|\mathbf{d}) - u_t^\text{target}(\mathbf{x}|\mathbf{m})||^2\right]
\end{equation}

For our small seismic velocity model example, we have to update the sampling from the target distribution, which now needs to be a joint distribution and returns both the model and the corresponding data. Then, we also need to update the NN to also take the data as an input:

\begin{algorithm}
\caption{Pseudocode for Guided Flow Matching.}\label{alg-guided}
\begin{algorithmic}
\Require Neural Network $u^\theta(t, \mathbf{x}, \mathbf{d})$, learning rate $\eta$, amount of steps $n$, index $i$.
\State $i\gets 0$
\While{$i < n$}
    \State $(\mathbf{m}, \mathbf{d})\sim p_\text{model,data}$
    \State $\mathbf{\epsilon} \sim \mathcal{N}(0, \mathbf{I}_2)$ \Comment{($\mathbf{m} \in \mathbb{R}^2 \Rightarrow \mathbf{I}_2$)}
    \State $t \sim \text{Unif}_{[0, 1]}$ \Comment{Sample time $t$}
    \State $\mathbf{x} \gets t\cdot \mathbf{m} + (1 - t)\cdot \mathbf{\epsilon}$ \Comment{$\mathbf{x} \sim p_t$}
    \State $u_t^\text{target} \gets \mathbf{m} - \mathbf{\epsilon}$
    \State $u_t^\text{prediction} \gets u^\theta(t, \mathbf{x}, \mathbf{d})$
    \State $\mathcal{L}_\text{CFM} \gets \mathbb{E}\left[||u_t^\text{prediction} - u_t^\text{target}||^2\right]$ \Comment{Conditional Flow Matching objective}
    \State $\theta \gets \theta - \eta \nabla_\theta \mathcal{L}_\text{CFM}$ \Comment{Update parameters via gradient descent}
    \State $i \gets i + 1$
\EndWhile
\end{algorithmic}
\end{algorithm}

\begin{figure}
    \centering
    \includegraphics[width=0.5\linewidth]{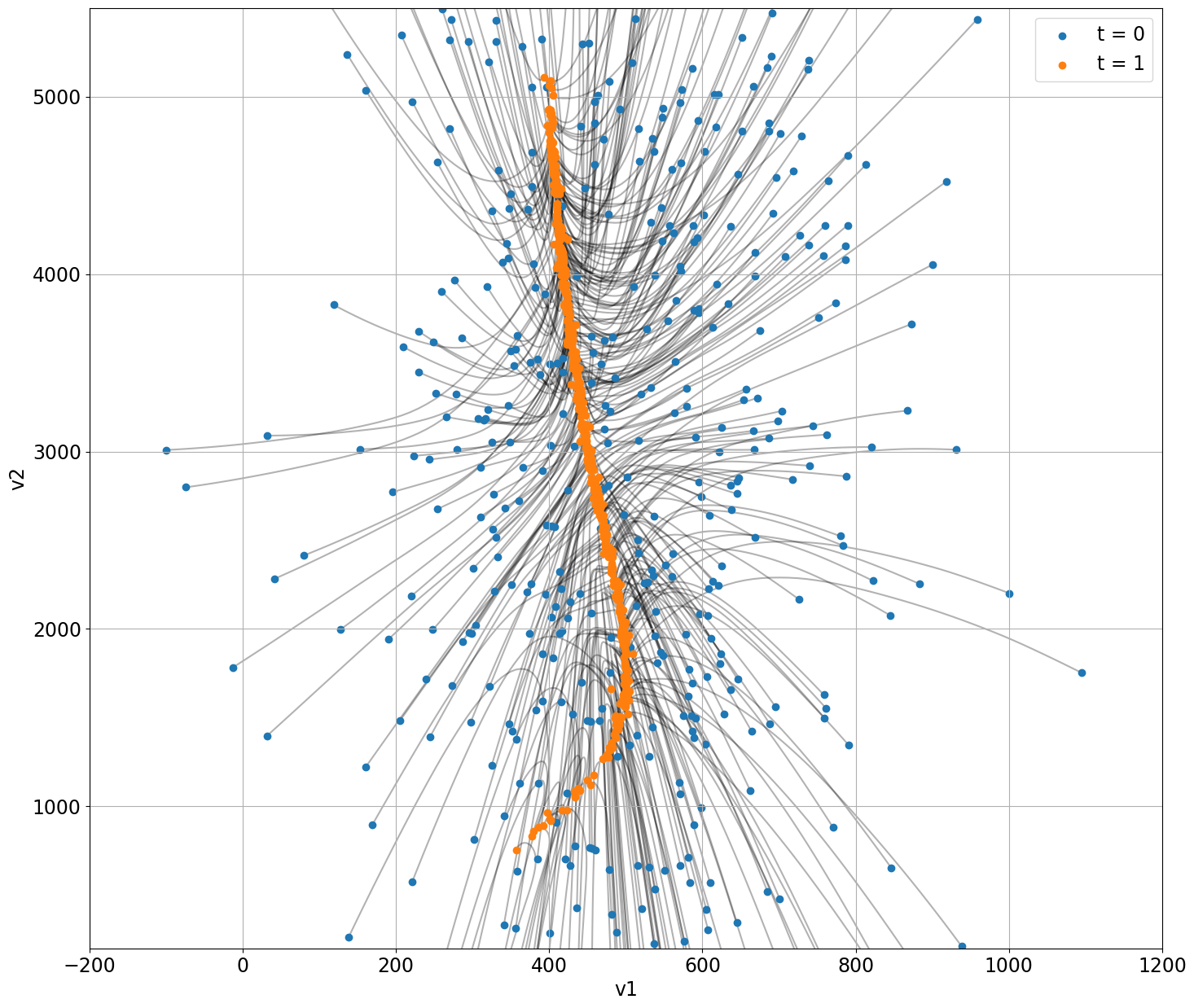}
    \caption{Probabilistic inversion result for given traveltime t = 0.06 s}
    \label{fig-guided-single}
\end{figure}

\begin{figure}
    \centering
    \includegraphics[width=\linewidth]{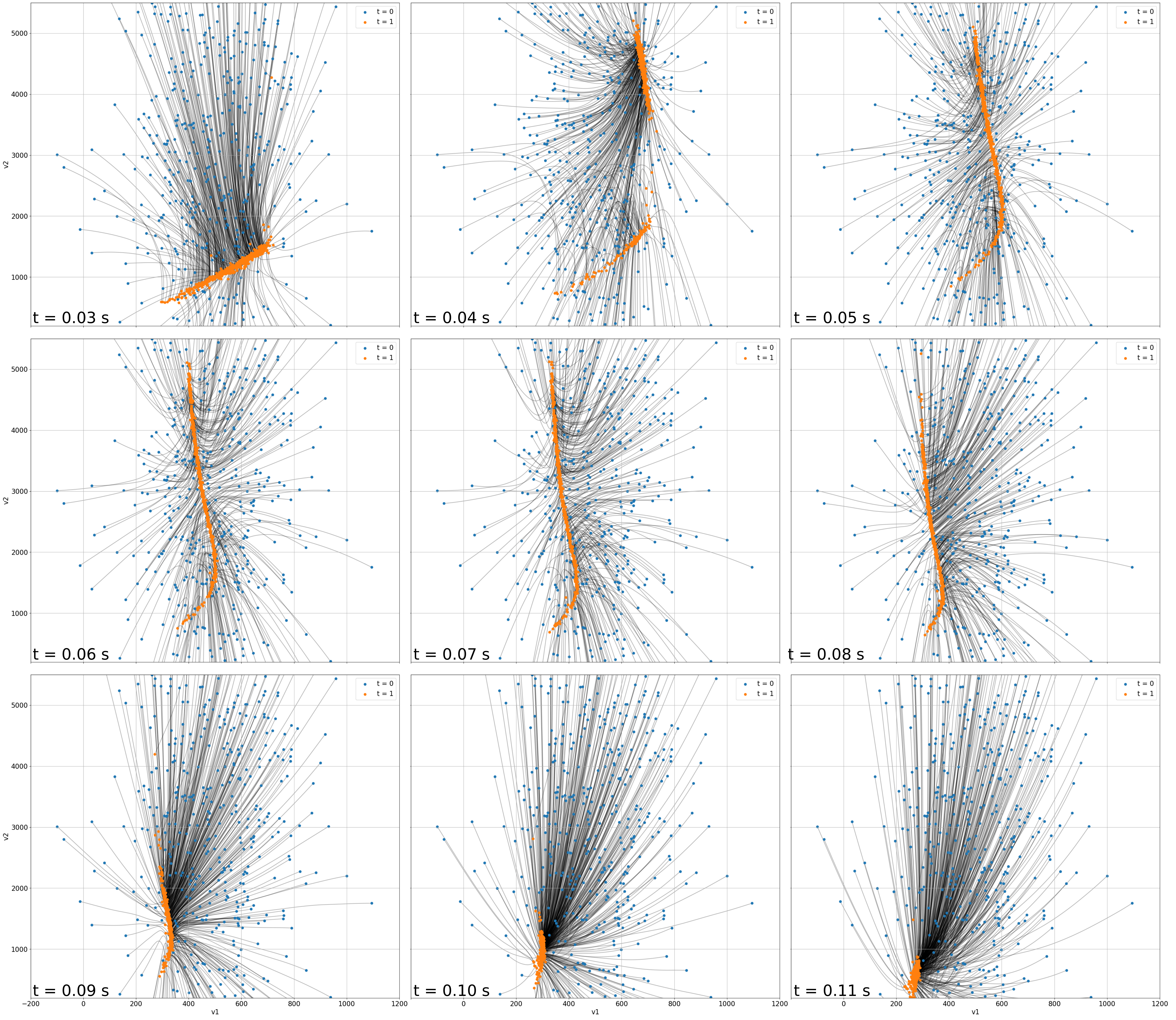}
    \caption{Guided Flow Matching. Blue dots are samples from the Normal distribution (after projecting them into the parameter range for the velocity model), the black lines show the Flow over time and the orange dots are probabilistic inversion results conditioned (guided) on the given traveltime of the refracted wave, which is shown on the bottom left corner of each image.}
    \label{fig-guided-all}
\end{figure}

Now we can perform probabilistic inversion by sampling from the initial (Normal) distribution and solving the ODE, guiding the NN with the data during inference. \autoref{fig-guided-single} shows a probabilistic inversion result with Guided Flow Matching for a traveltime of the refracted wave $t = 0.06$ s. \autoref{fig-guided-all} shows more results for different traveltimes. Note that the inversion results always lie in the model distribution given in the training dataset. For $t = 0.03$ s or $t = 0.04$ s, no valid models outside the training distribution were sampled. For practical applications, it is therefore crucial that the training dataset captures the distribution of all realistic models of interest.

\subsection{Classifier-Free Guidance}
\begin{figure}
    \centering
    \begin{subfigure}{0.48\textwidth}
        \includegraphics[width=\textwidth]{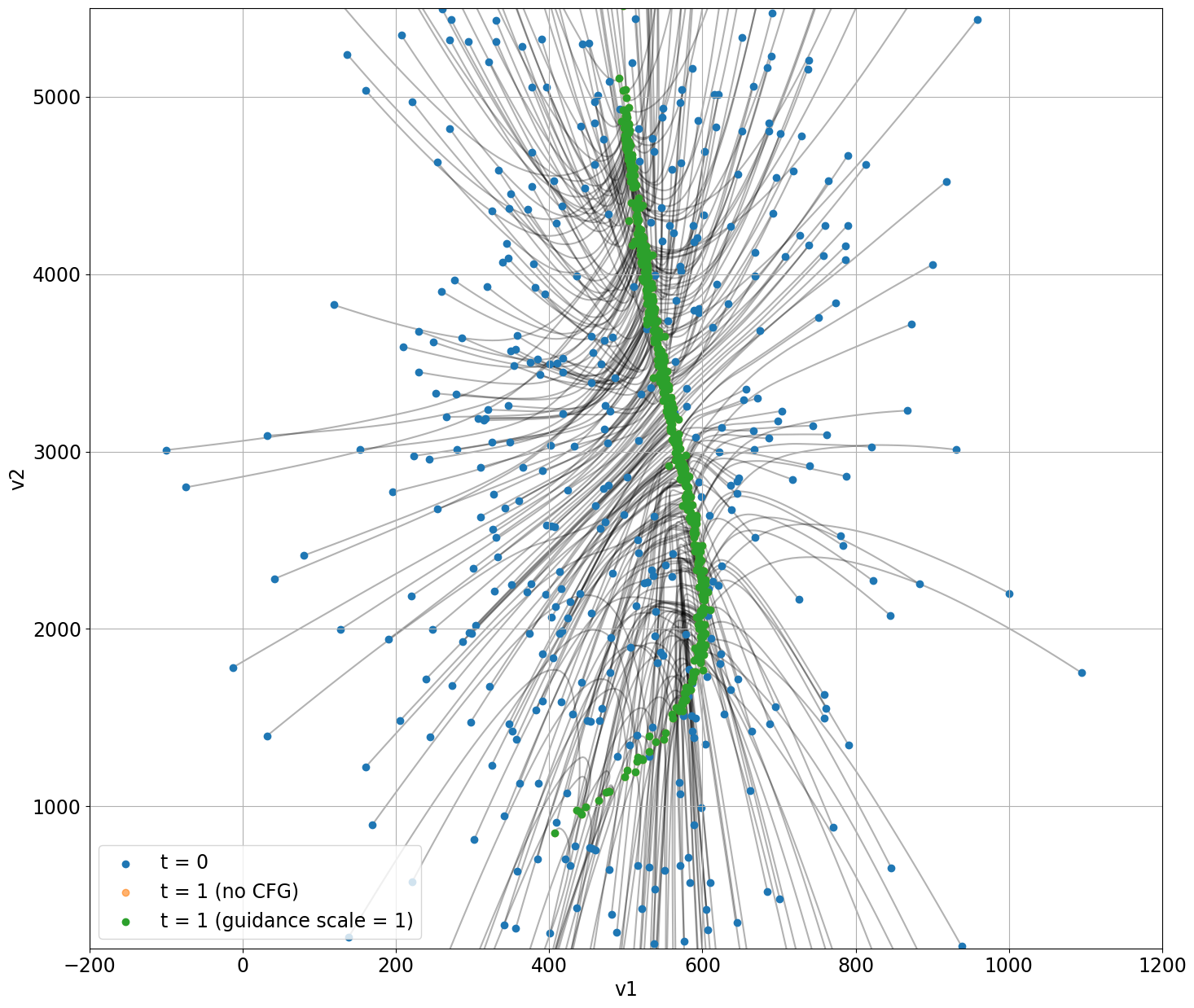}
        \caption{guidance scale = 1}
    \end{subfigure}
    \hfill
    \begin{subfigure}{0.48\textwidth}
        \includegraphics[width=\textwidth]{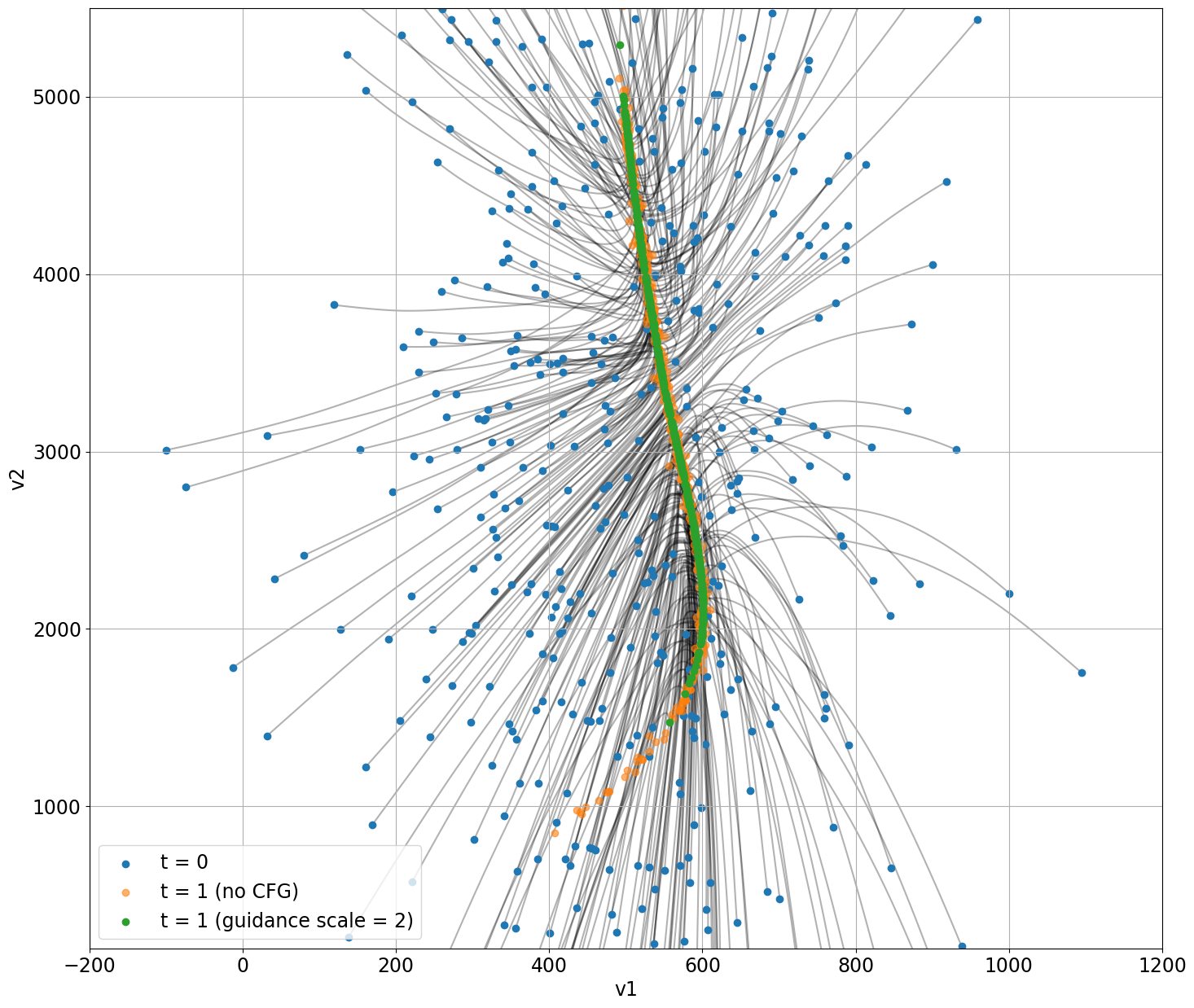}
        \caption{guidance scale = 2}
    \end{subfigure}
    \hfill
    \begin{subfigure}{0.48\textwidth}
        \includegraphics[width=\textwidth]{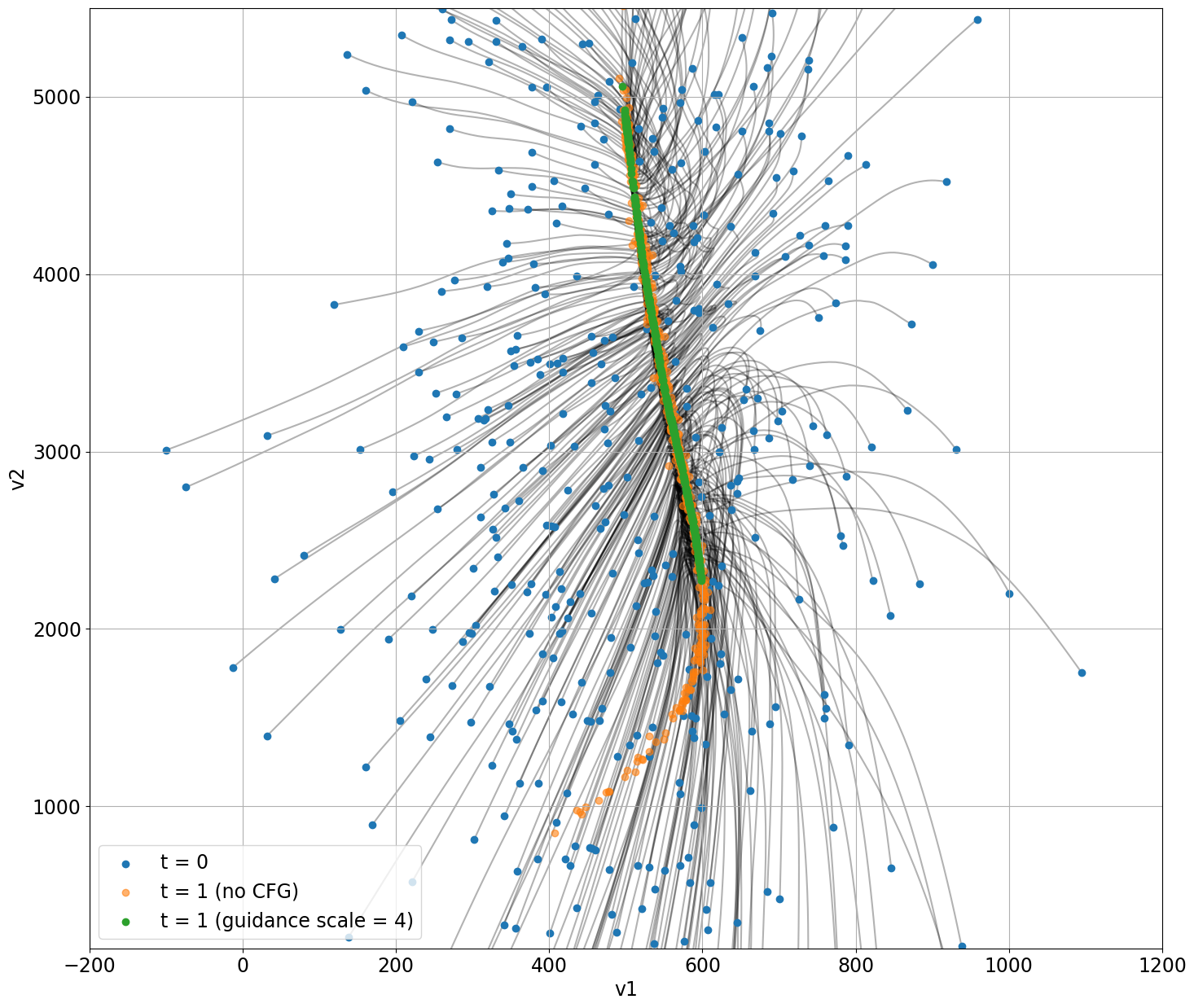}
        \caption{guidance scale = 4}
    \end{subfigure}
    \hfill
    \begin{subfigure}{0.48\textwidth}
        \includegraphics[width=\textwidth]{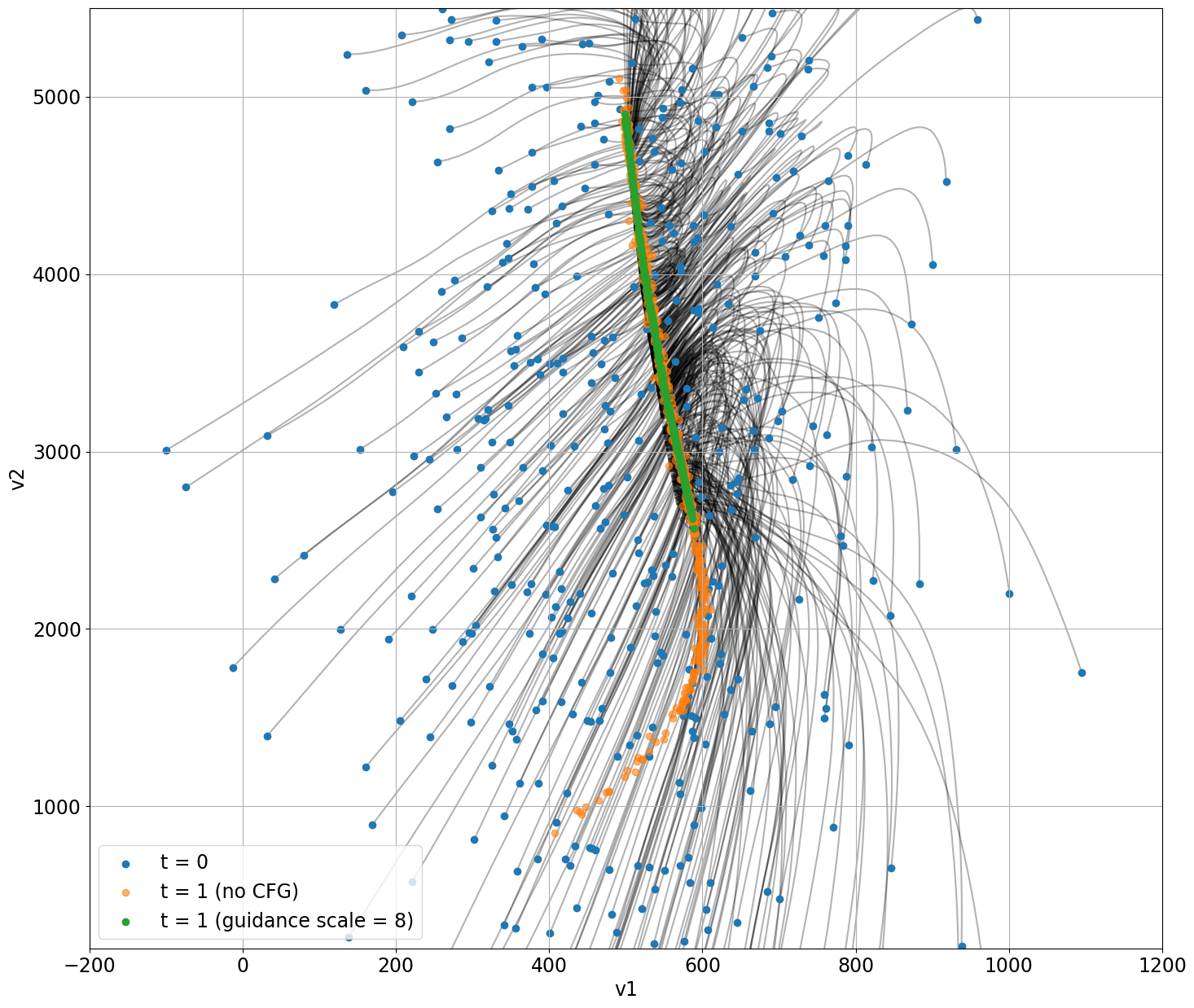}
        \caption{guidance scale = 8}
    \end{subfigure}
    \hfill
    \caption{Guided Flow Matching with Classifier-Free Guidance for $t = 0.05$ s. Blue dots are samples from the Normal distribution (after projecting them into the parameter range for the velocity model), the black lines show the Flow over time, the orange dots are probabilistic inversion results conditioned (guided) on the given traveltime of the refracted wave without CFG. The green dots are probabilistic inversion results shown for varying guidance scales.}
    \label{fig-cfg-005}
\end{figure}

When Flow Matching is used in generative AI (e.g. image or video generation), usually Classifier-Free Guidance (CFG) is employed \parencite{zheng_guided_2023}. In CFG, the Neural Network is trained to accept inputs both with and without guidance. Then, to compute the vector for the ODE during inference, the NN predicts both the vector for the guided and the unguided input. The vector for the ODE is then computed as:

\begin{equation}
\label{eq-cfg}
\tilde{u}_t(\mathbf{x}|\mathbf{d}) = (1 - w)u_t^\theta(\mathbf{x}) + wu_t^\theta(\mathbf{x}|\mathbf{d})
\end{equation}

where $w$ is the guidance scale. For $w=1$, CFG is equivalent to the previous Guided Flow Matching at inference. Nevertheless, usually $w>1$ is chosen for generative, as it leads to empirically better results. Intuitively, the output vector may then be interpreted as weighting the direction away from the marginal target distribution and the direction towards the target distribution conditioned/ guided on the observed data. As a result, the vector points more into the direction of the target distribution (guided on the observed data) than when using $w = 1$. We can see this effect also in our seismic velocity field example in \autoref{fig-cfg-005}.

\begin{figure}
    \centering
    \begin{subfigure}{0.48\textwidth}
        \includegraphics[width=\textwidth]{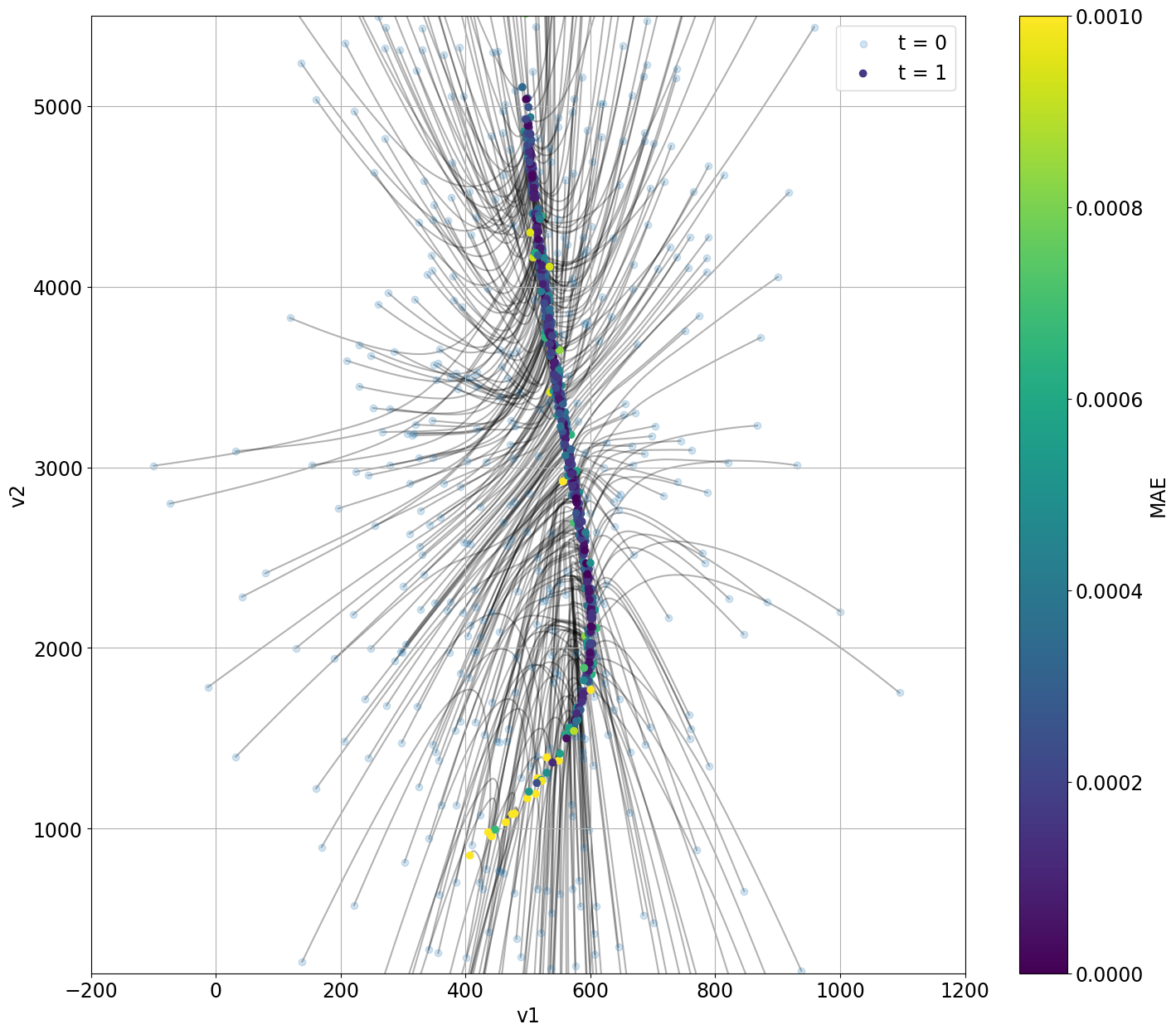}
        \caption{Errors for guidance scale $w=1$ (no CFG), mean error = $27.7\cdot 10^{-5}$ s}
    \end{subfigure}
    \hfill
    \begin{subfigure}{0.48\textwidth}
        \includegraphics[width=\textwidth]{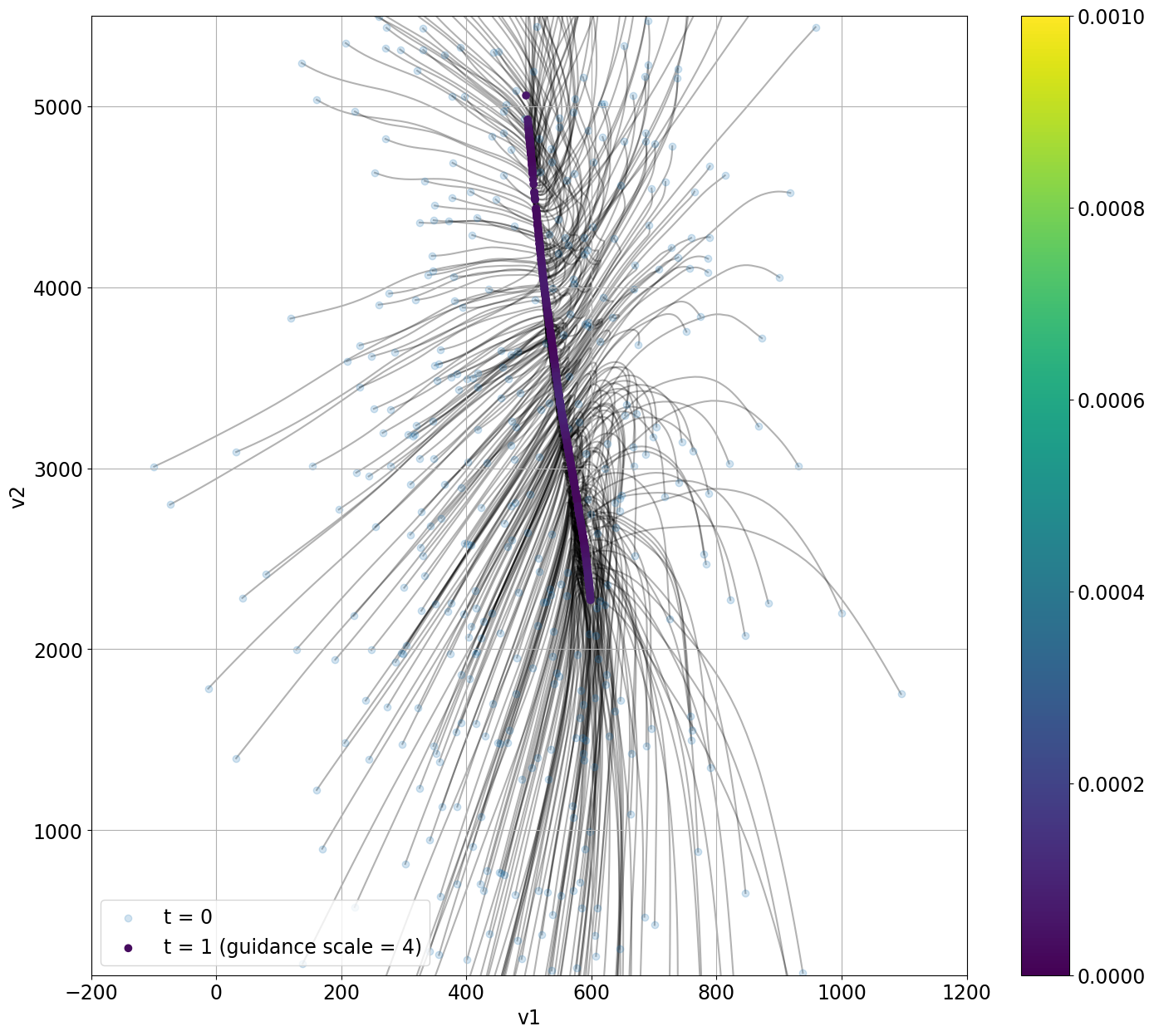}
        \caption{Guided Flow Matching with guidance scale $w=4$, mean error = $0.52\cdot 10^{-5}$ s}
    \end{subfigure}
    \hfill
    \caption{On the left: the absolute error of the predicted samples in the target distribution without CFG (guided on t = 0.05 s). On the right: with guidance scale $w = 4$.}
    \label{fig-cfg-error}
\end{figure}

The resulting distribution differs significantly from the one obtained without CFG. To understand why, we can examine the data misfit of the predicted samples in the target distribution, as shown in \autoref{fig-cfg-error}. It shows that without CFG, there is a number of samples in the range of $v_1\in [400,600]$ and $v_2\in[900, 2000]$ that show a high absolute error compared to other samples. In contrast, when using CFG, the vector field points more into the direction of the samples with the highest likelihood, resulting in a significantly lower absolute error of the predicted samples compared to not using CFG.

\section{Case Study: Seismic Full-Waveform Inversion}
After showing the theory and methodology of Flow Matching for the simple 2D velocity model with seismic refraction traveltimes and only two variables, we now show the general applicability of probabilistic inversion with Flow Matching for settings with significantly more parameters. For sampling-based methods like Markov-Chain Monte Carlo, this is generally infeasible due to the "Curse of Dimensionality".

In this case study, we explore seismic Full-Waveform inversion in 2D, similar to what \parencite{zhang_diffusionvel_2024} did for Diffusion models, but we only use the observed data given in the dataset.

\subsection{Dataset}
\begin{figure}
    \centering
    \includegraphics[width=\linewidth]{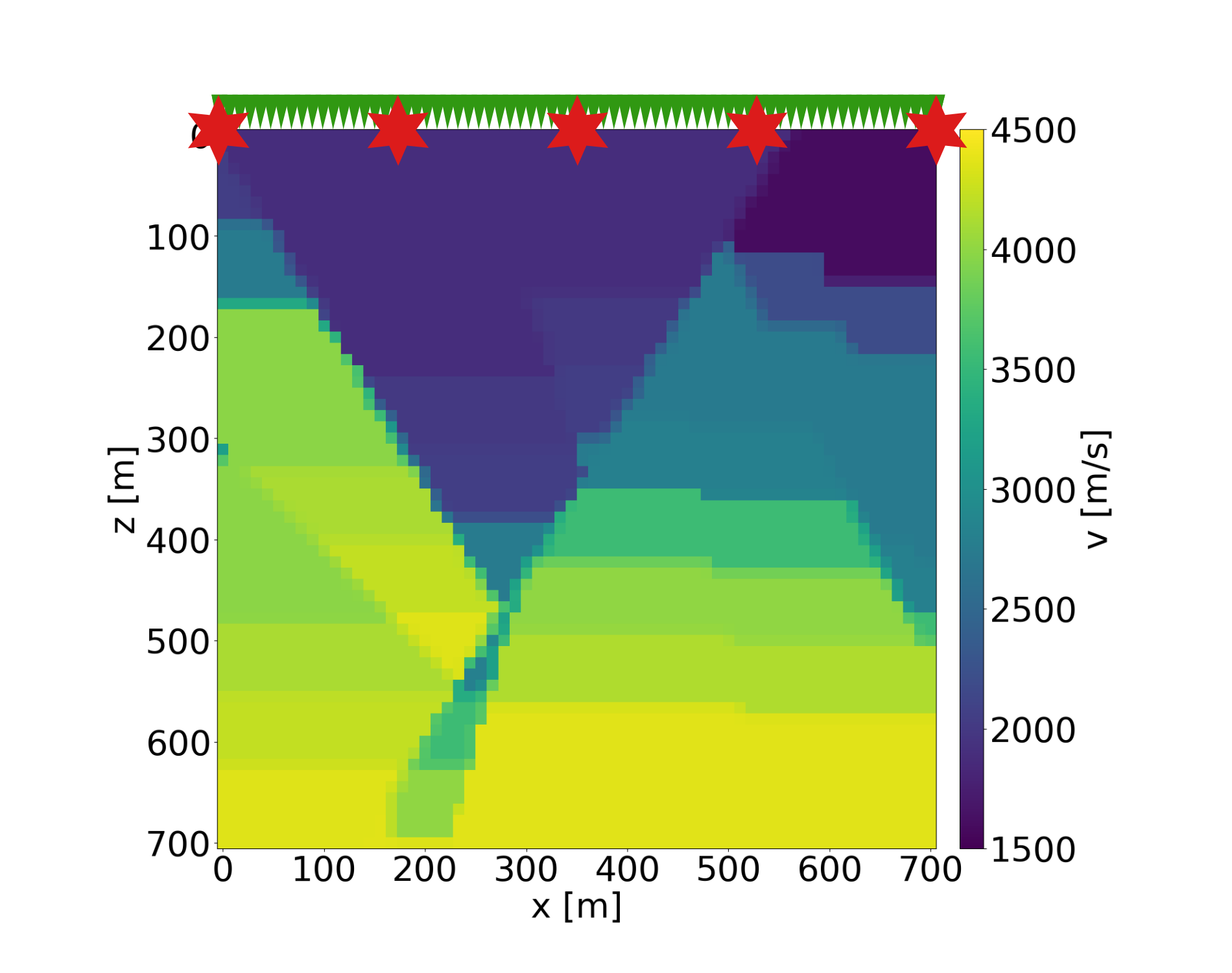}
    \caption{A representative velocity model from the FlatFaultB dataset. Source positions are marked as red stars and receivers as green triangles on the surface.}
    \label{fig-openfwi}
\end{figure}

For the training data, we use the OpenFWI FlatFaultB dataset \parencite{deng_openfwi_2023}, which is a collection of 54000 seismic velocity models with horizontal layers and several dipping faults. For each velocity model, five source positions are equidistantly placed along the 700 m long profile on the surface together with 70 receivers with a 10 m spacing. Therefore, five seismogram sections with 101 receivers each are given for each velocity model as observed data, which will serve as guidance during Flow Matching. We use 90\% of the dataset for training and 10\% for validation. A representative velocity model from the OpenFWI FlatFaultB dataset is shown in \autoref{fig-openfwi}. Due to the U-Net architecture of our NN, we interpolate the seismic velocity models (originally 70x70) to 64x64.

\subsection{Neural Network Architecture}
\begin{figure}
    \centering
    \includegraphics[width=\linewidth]{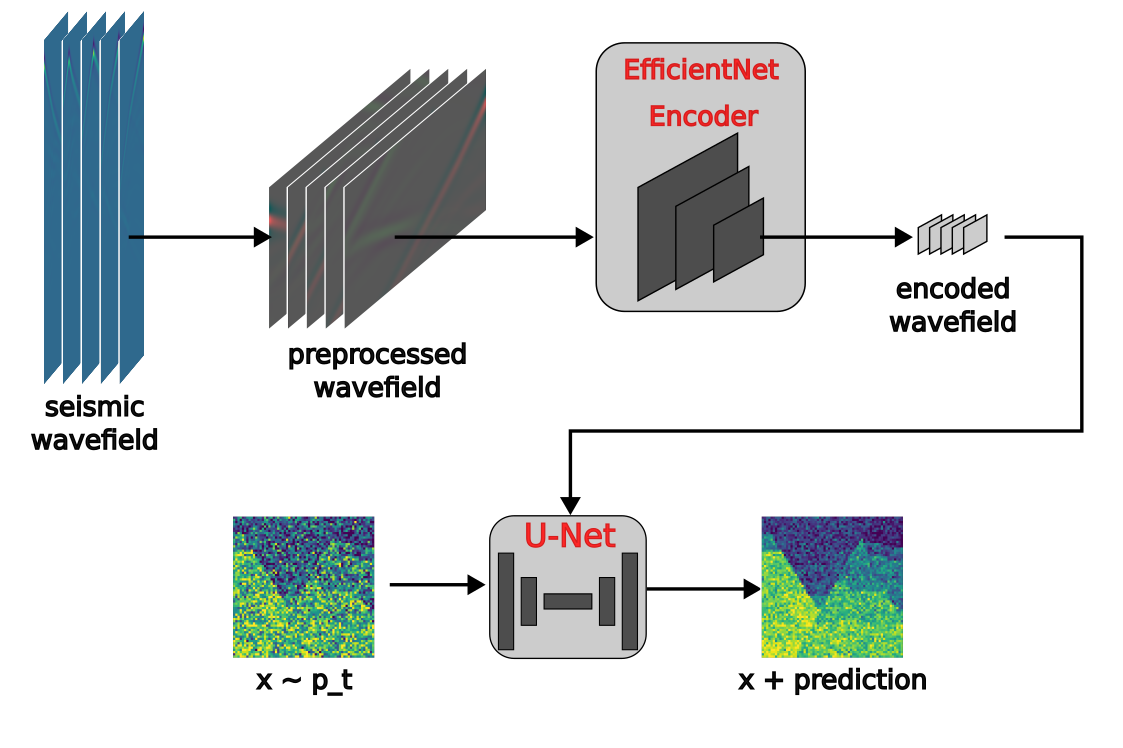}
    \caption{Our Flow Matching approach. First the observed data (seismic wavefield) is preprocessed to obtain five three-channeled images with the dimensions of 224x224, which is the required shape for the pretrained EfficientNet Encoder. Then, the EfficientNet encodes the observed data into five compressed tensors in latent space. We employ a U-Net that takes a sample from the Gaussian Conditional Probability path and incorporates the information from the compressed observed data when predicting the output vector.}
    \label{fig-architecture}
\end{figure}

For the architecture of the NN, we were inspired by Würstchen \parencite{pernias_wuerstchen_2023}, where an EfficientNet \parencite{tan_efficientnet_2020} was used as semantic compressor (in our case for the observed wavefield). With this, we could build upon a pretrained feature extractor for images, which only needed to be fine-tuned for our case. To approximate the vector field $u_t^\text{target}(\mathbf{m} | \mathbf{d})$, we use a U-Net, which is also a popular architecture for Diffusion models. We used attention blocks in the U-Net to incorporate the information of the observed data features from the EfficientNet encoder into the prediction of the output vector. During training, we dropped the observed data in 20\% of the iterations to also learn the unguided vector field, which is necessary to do CFG at inference.

Aside from interpolating the seismic velocity models to 64x64, we also needed to preprocess the given observed data (i.e. the seismogram sections), as the pretrained EfficientNet encoder expects three-channel images with the spatial dimensions of 224x224. The observed data for each source shot is given as a tensor of shape [1000, 70] (recorded on 70 receivers for a duration of one second with 1ms sampling interval). To obtain the required three-channel image of shape [3, 224, 224], we omit the last time sample and chunk the observed data along the time dimension into [3, 333, 70]. Then, we use bilinear interpolation to obtain the shape [3, 224, 224]. After extracting the features of the observed data with the EfficientNet encoder, we get a compressed tensor in latent space with the shape [16, 7, 7] for each of the five shots. This means, that the observed data is compressed to about 1.1\% of its original size.

While using an EfficientNet encoder allows us to drastically compress the data size, it always requires the data to be in the same geometry. That implies that our NN architecture will only work for seismic data where the acquisition geometry corresponds to the acquisition geometry in the training dataset. For the case study, that does not pose any issues, as both the training and validation dataset always have the same acquisition geometry. For real-world applications however, more general NN architectures are required, which are subject of ongoing research.

We trained the Neural Network using the framework of \parencite{lipman_flow_2024}. We chose the AdamW optimizer \parencite{loshchilov_decoupled_2019} and trained for 500 epochs on a single NVIDIA A100 GPU with a batch size of 180, which resulted in a total training time of 37 hours.

\subsection{Results}
\begin{figure}
    \centering
    \begin{subfigure}{0.48\textwidth}
        \includegraphics[width=\textwidth]{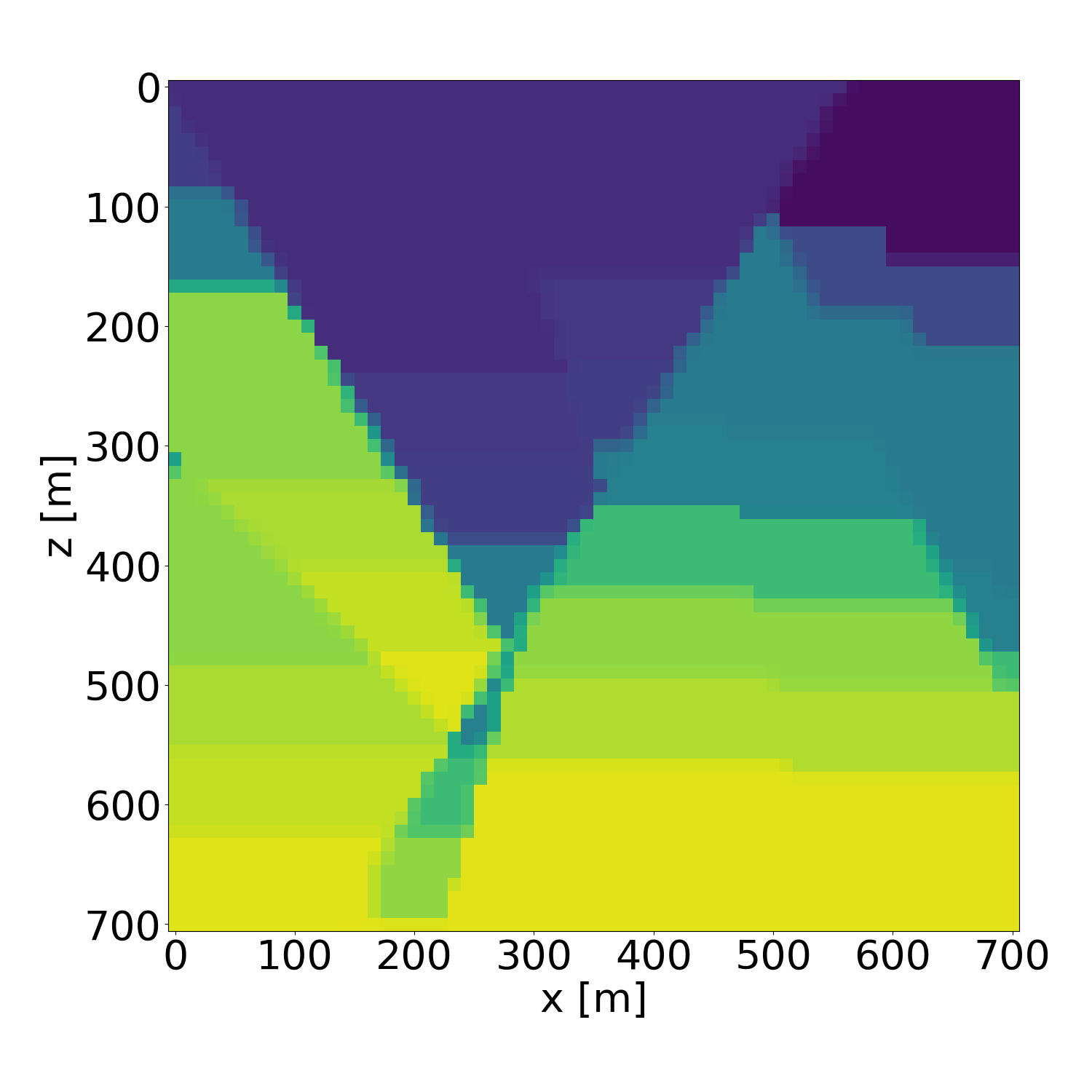}
        \caption{Ground truth}
    \end{subfigure}
    \hfill
    \begin{subfigure}{0.48\textwidth}
        \includegraphics[width=\textwidth]{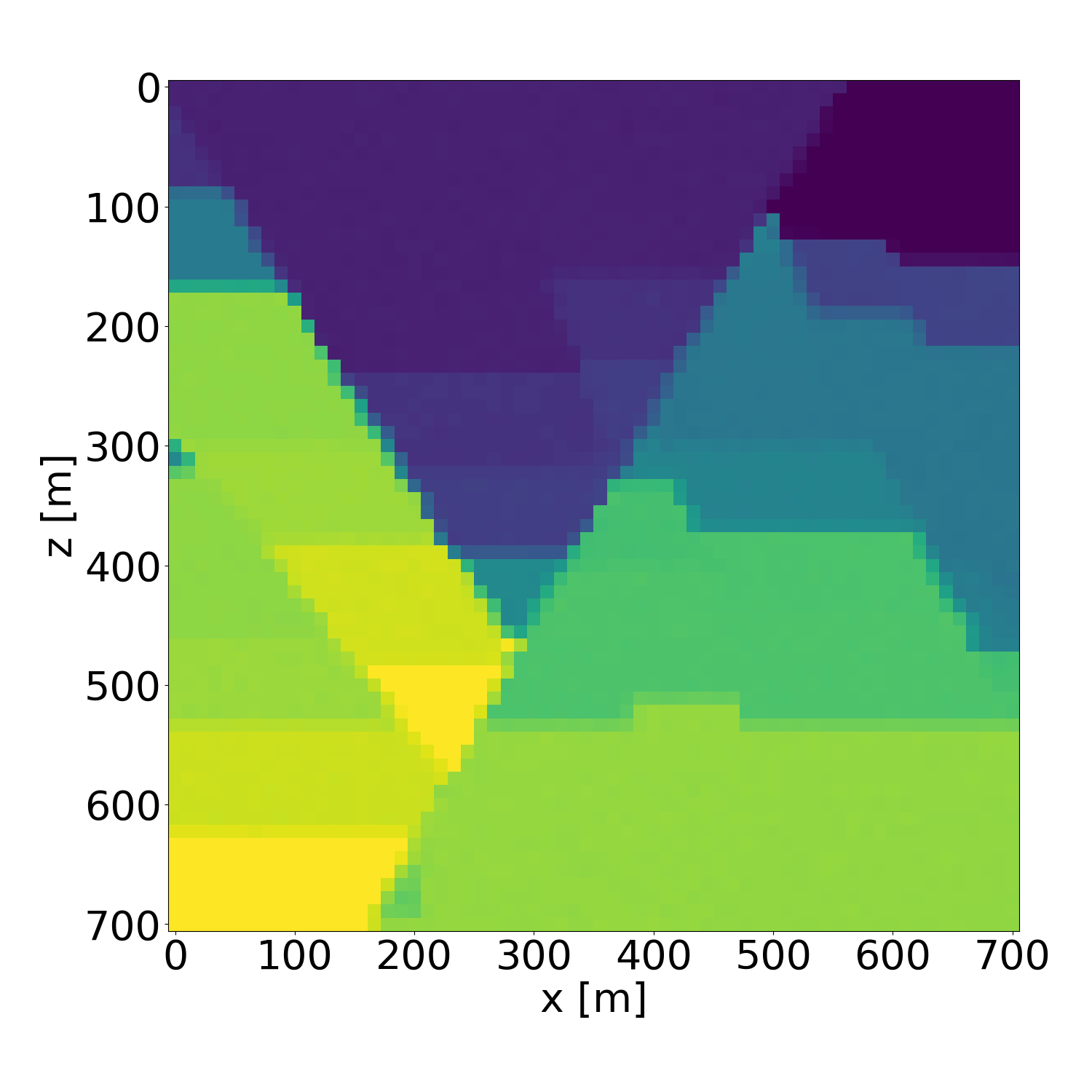}
        \caption{Probabilistic inversion result}
    \end{subfigure}
    \hfill
    \caption{Probabilistic inversion of a velocity model in the validation dataset. Left: target velocity model. Middle: a probabilistic inversion result using the trained NN with a guidance scale of $w=4$.}
    \label{fig-result1}
\end{figure}

A single probabilistic inversion result, obtained by sampling from the distribution obtained with guided Flow Matching with the trained NN, is shown in \autoref{fig-result1}. The visualization of a trajectory during probabilistic inversion is shown in \autoref{fig-result-gcpp}. More examples of inversions are given in the Appendix.

\begin{figure}
    \centering
    \begin{subfigure}{0.40\textwidth}
        \includegraphics[width=\textwidth]{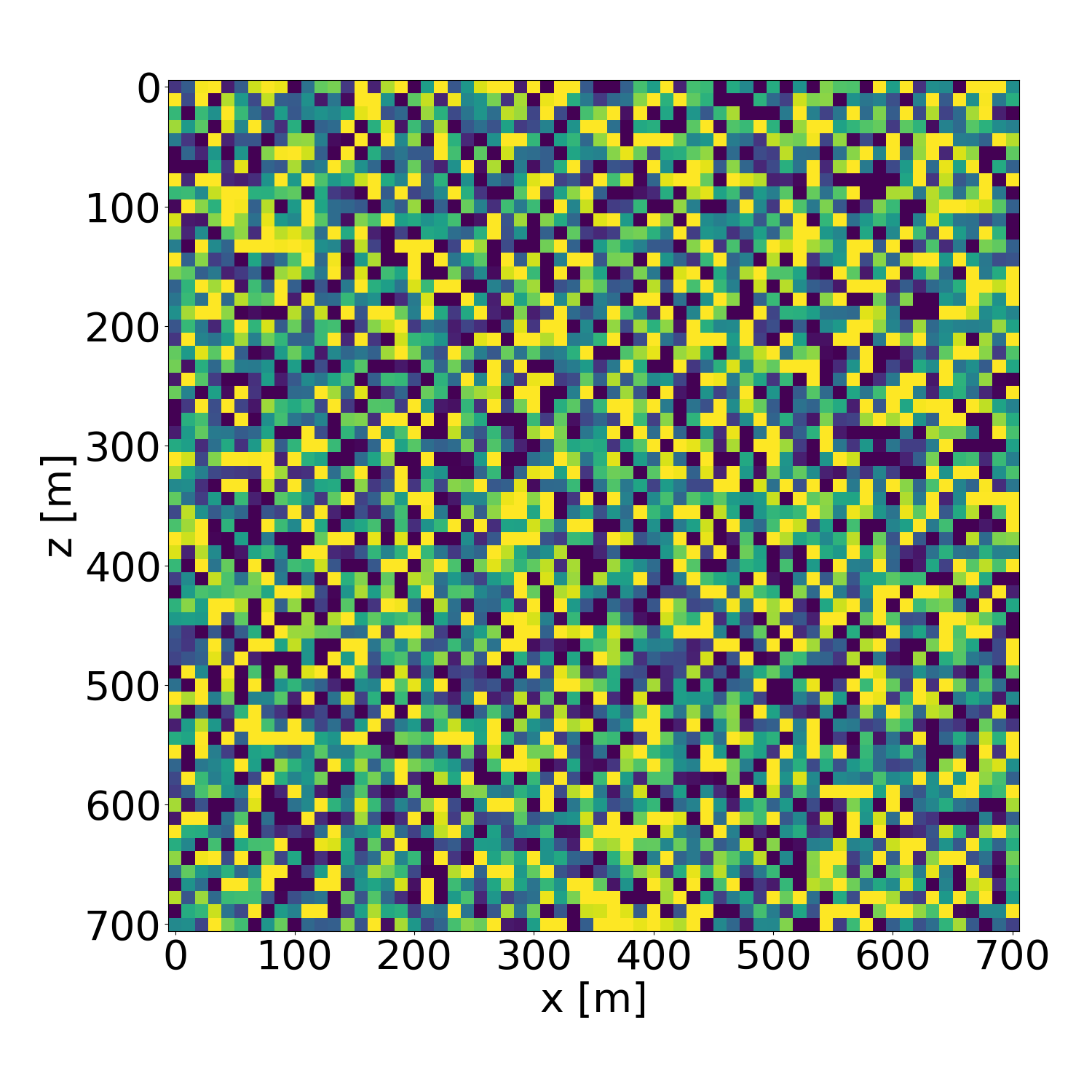}
        \caption{$t = 0$}
    \end{subfigure}
    \hfill
    \begin{subfigure}{0.40\textwidth}
        \includegraphics[width=\textwidth]{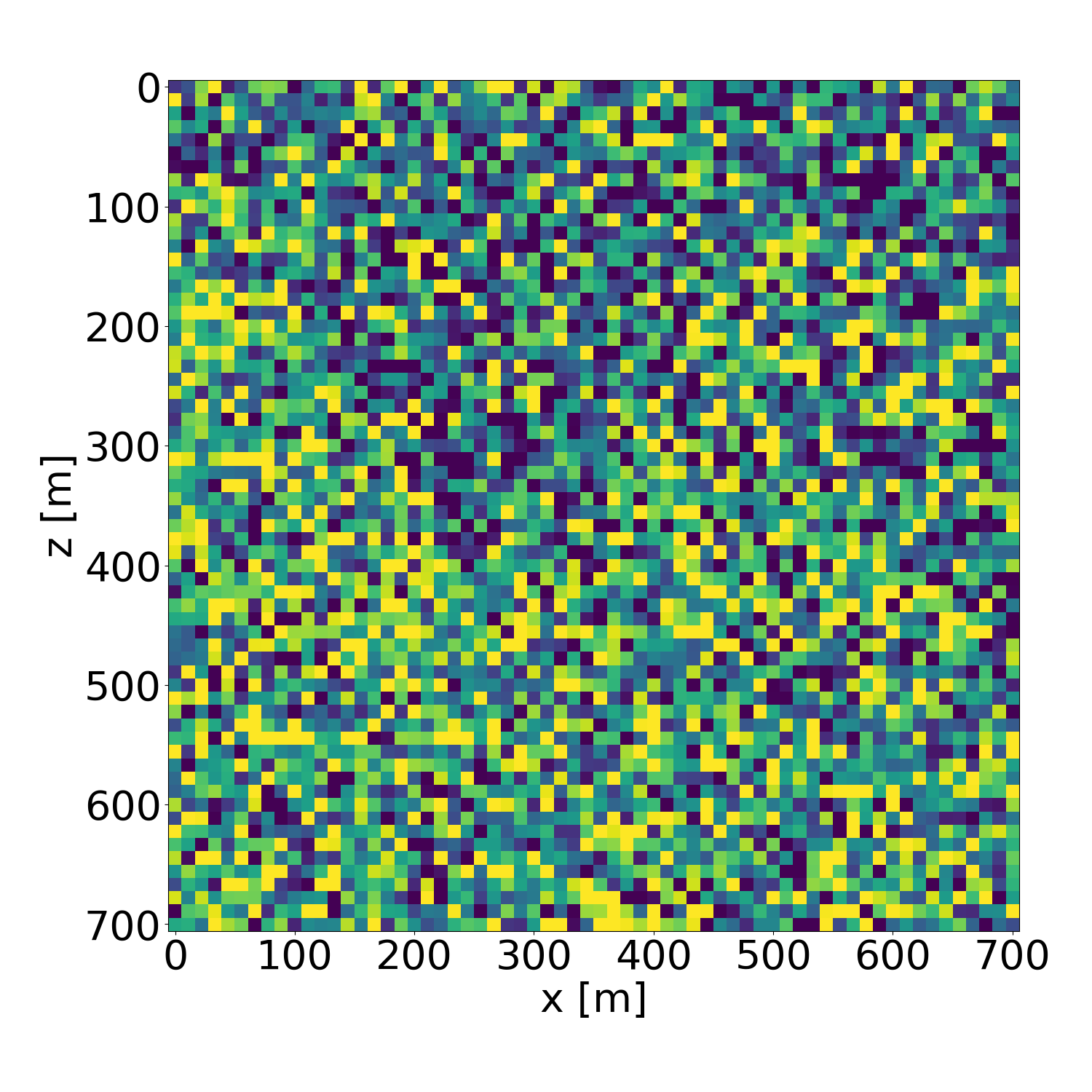}
        \caption{$t = 0.2$}
    \end{subfigure}
    \hfill
    \begin{subfigure}{0.40\textwidth}
        \includegraphics[width=\textwidth]{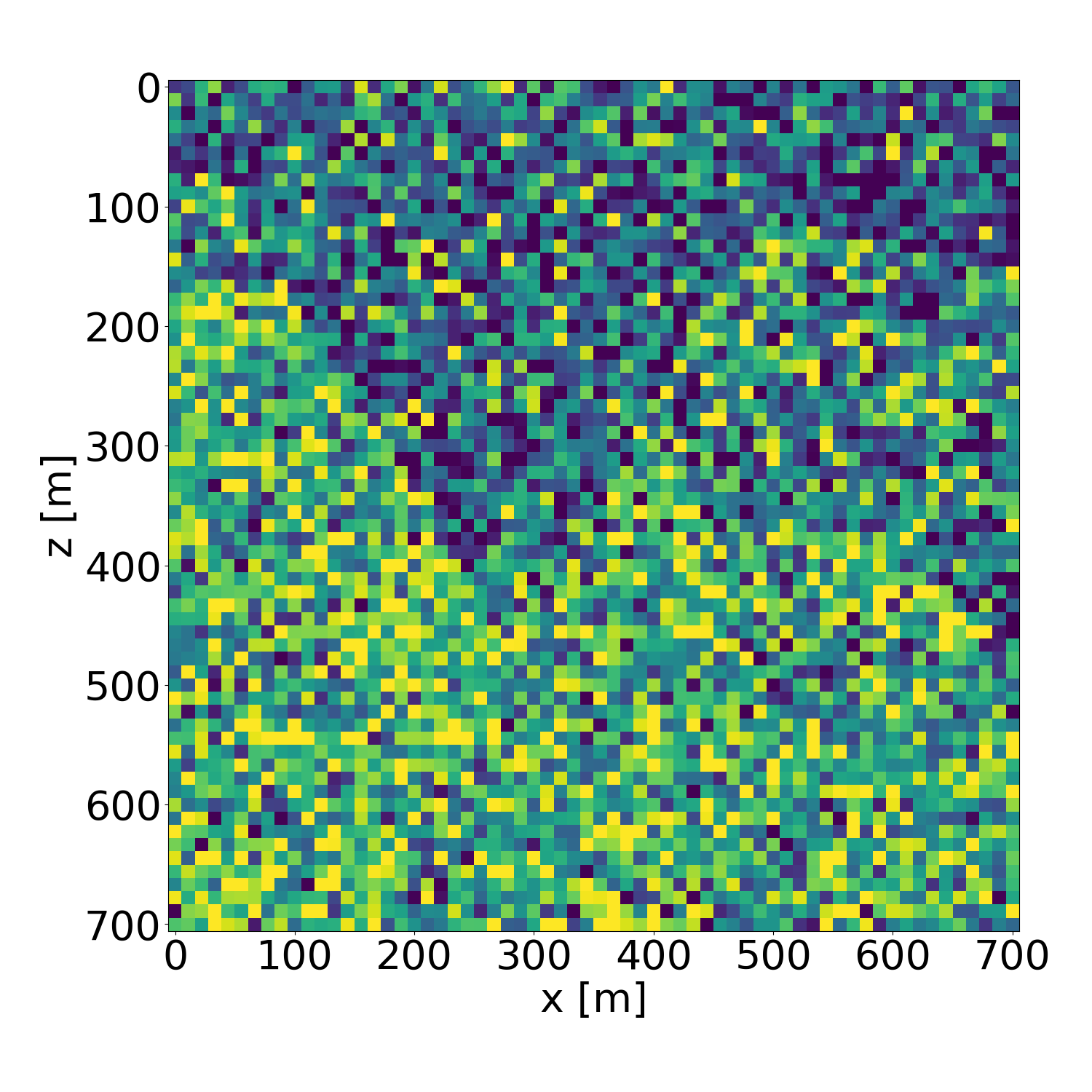}
        \caption{$t = 0.4$}
    \end{subfigure}
    \hfill
    \begin{subfigure}{0.40\textwidth}
        \includegraphics[width=\textwidth]{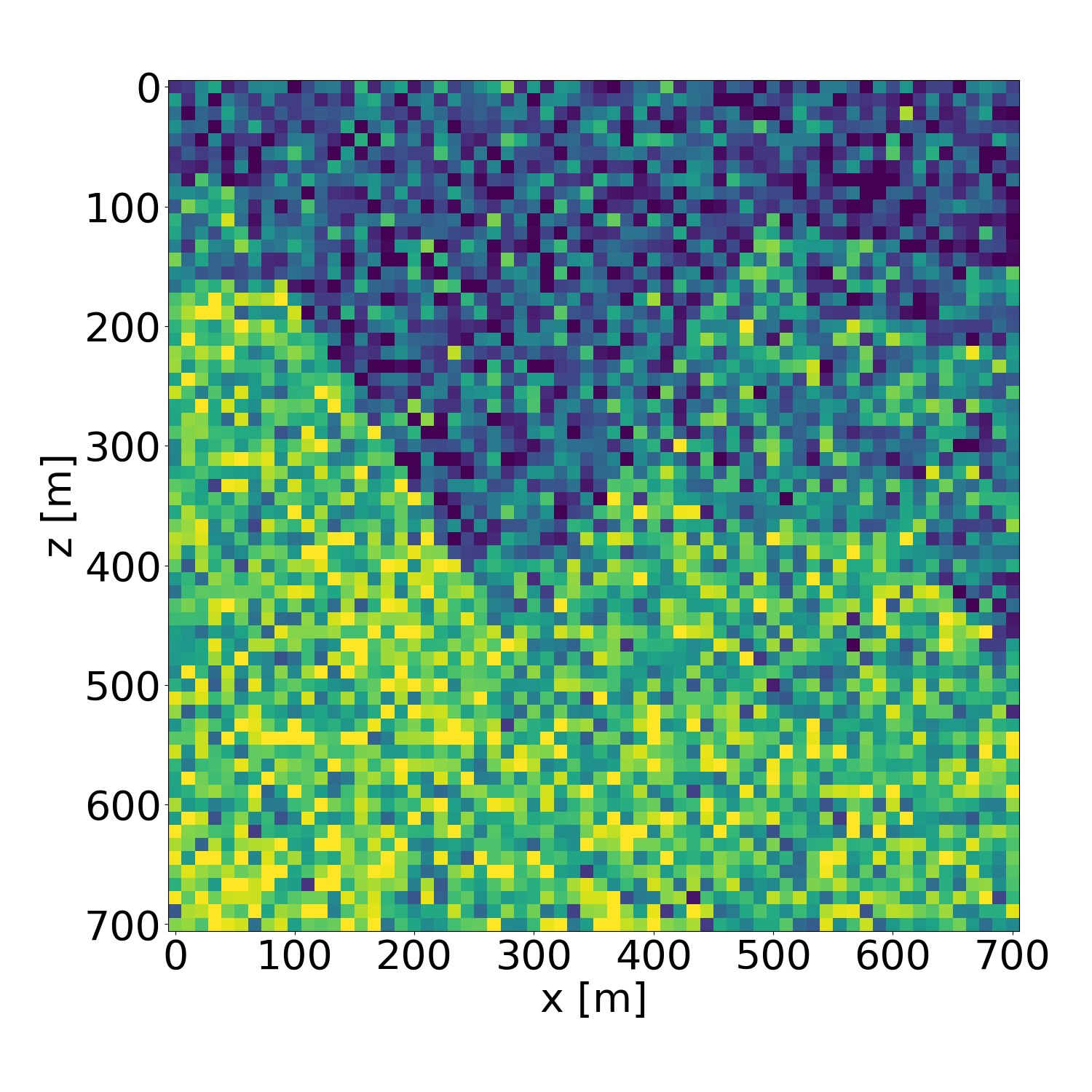}
        \caption{$t = 0.6$}
    \end{subfigure}
    \hfill
    \begin{subfigure}{0.40\textwidth}
        \includegraphics[width=\textwidth]{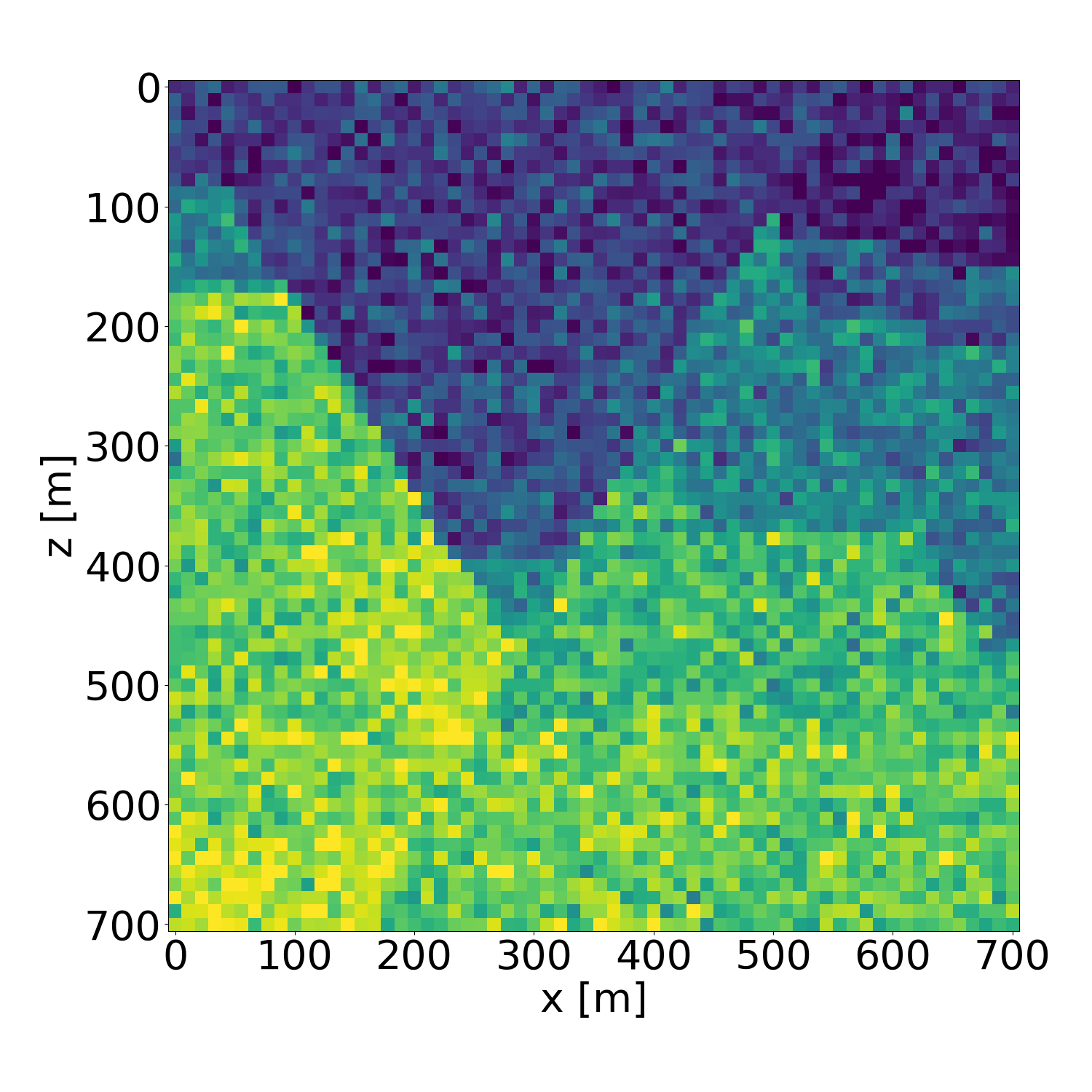}
        \caption{$t = 0.8$}
    \end{subfigure}
    \hfill
    \begin{subfigure}{0.40\textwidth}
        \includegraphics[width=\textwidth]{img/CodeOpenFWI/sample_000_step100_batch_1.png}
        \caption{$t = 1$}
    \end{subfigure}
    \hfill
    \caption{Probabilistic inversion results as a function of time along the trajectory.}
    \label{fig-result-gcpp}
\end{figure}

\begin{figure}
    \centering\
    \begin{subfigure}{0.48\textwidth}
        \includegraphics[width=\textwidth]{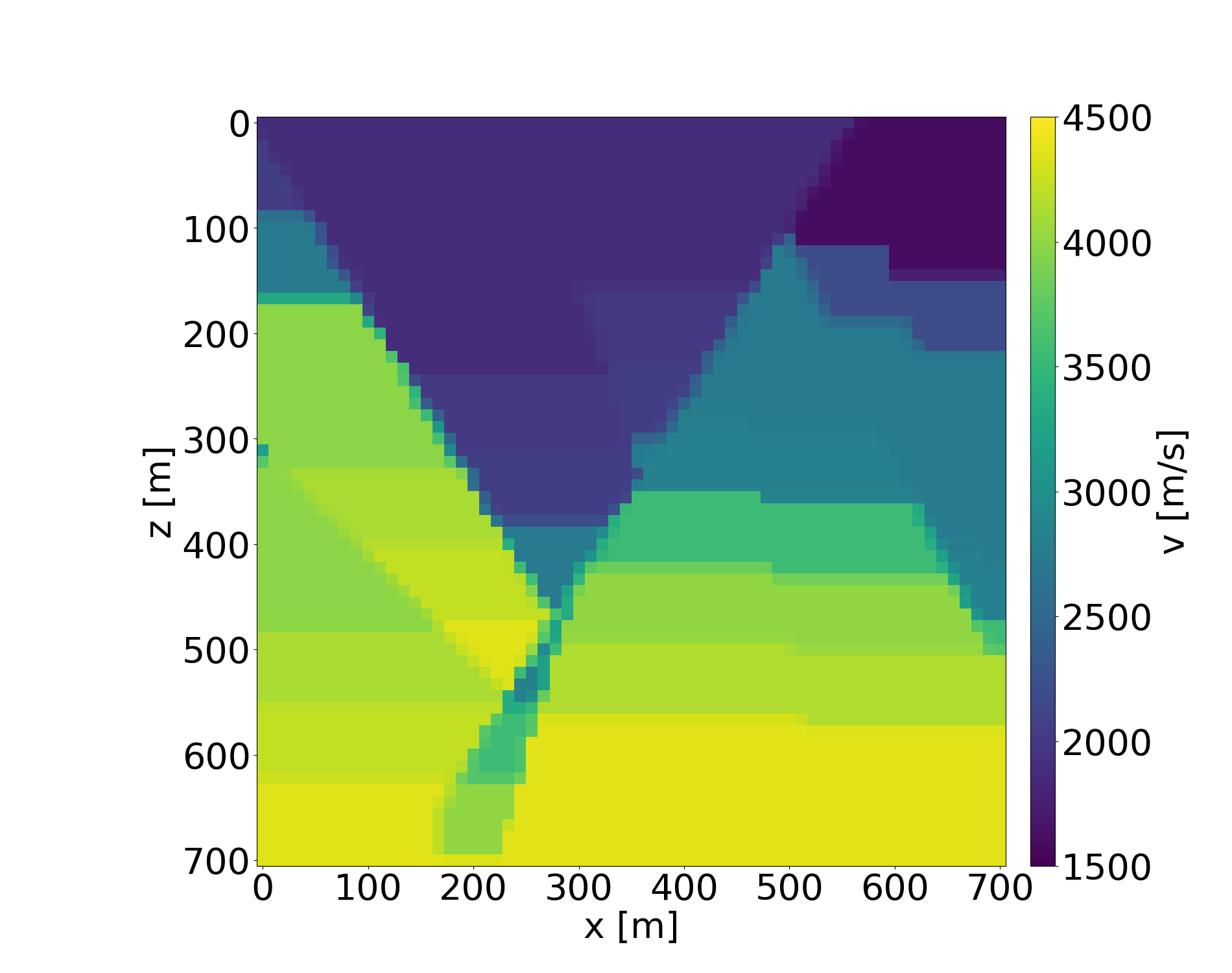}
        \caption{Ground truth}
    \end{subfigure}
    \hfill
    \begin{subfigure}{0.48\textwidth}
        \includegraphics[width=\textwidth]{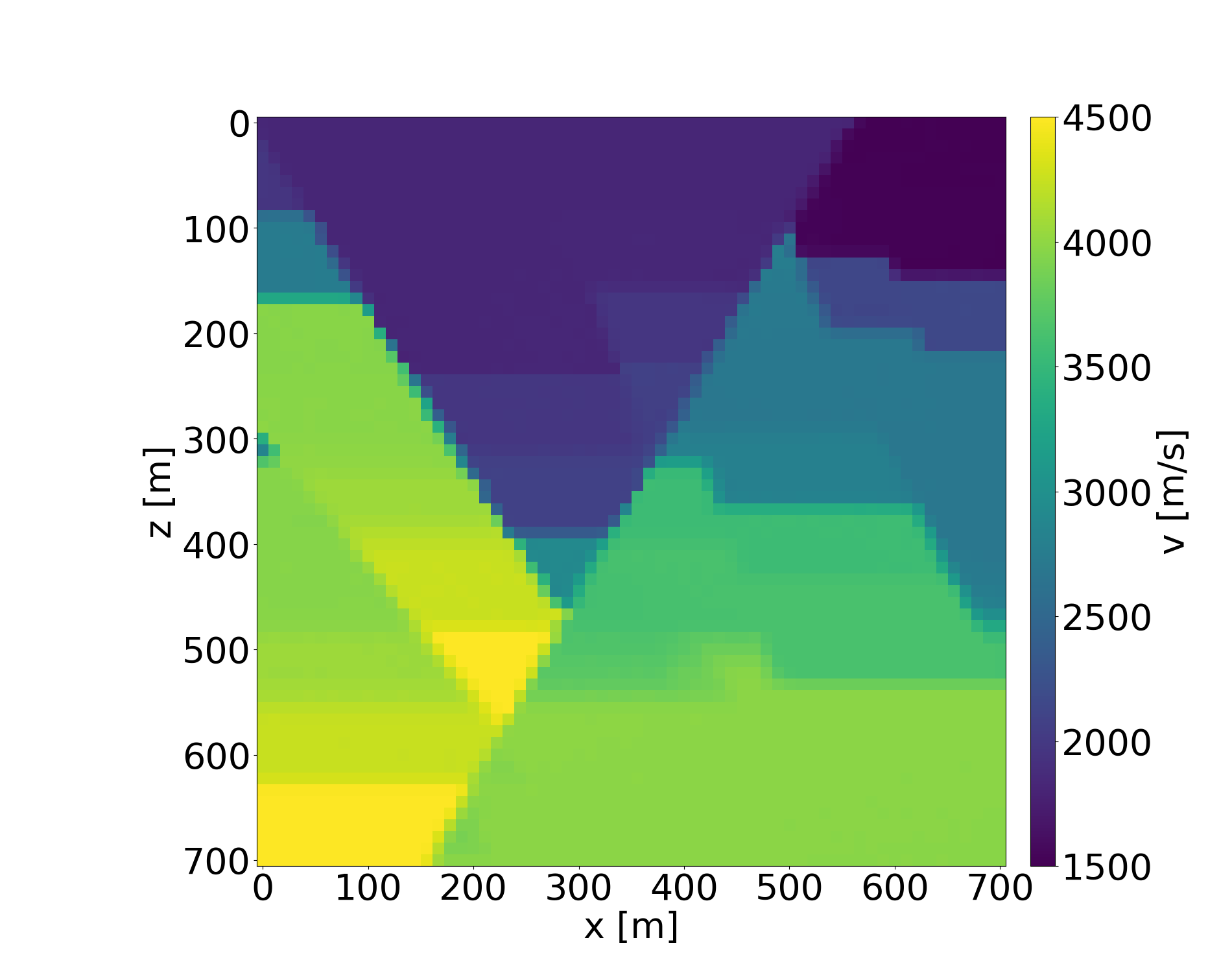}
        \caption{Average velocity model}
    \end{subfigure}
    \hfill
    \begin{subfigure}{0.48\textwidth}
        \includegraphics[width=\textwidth]{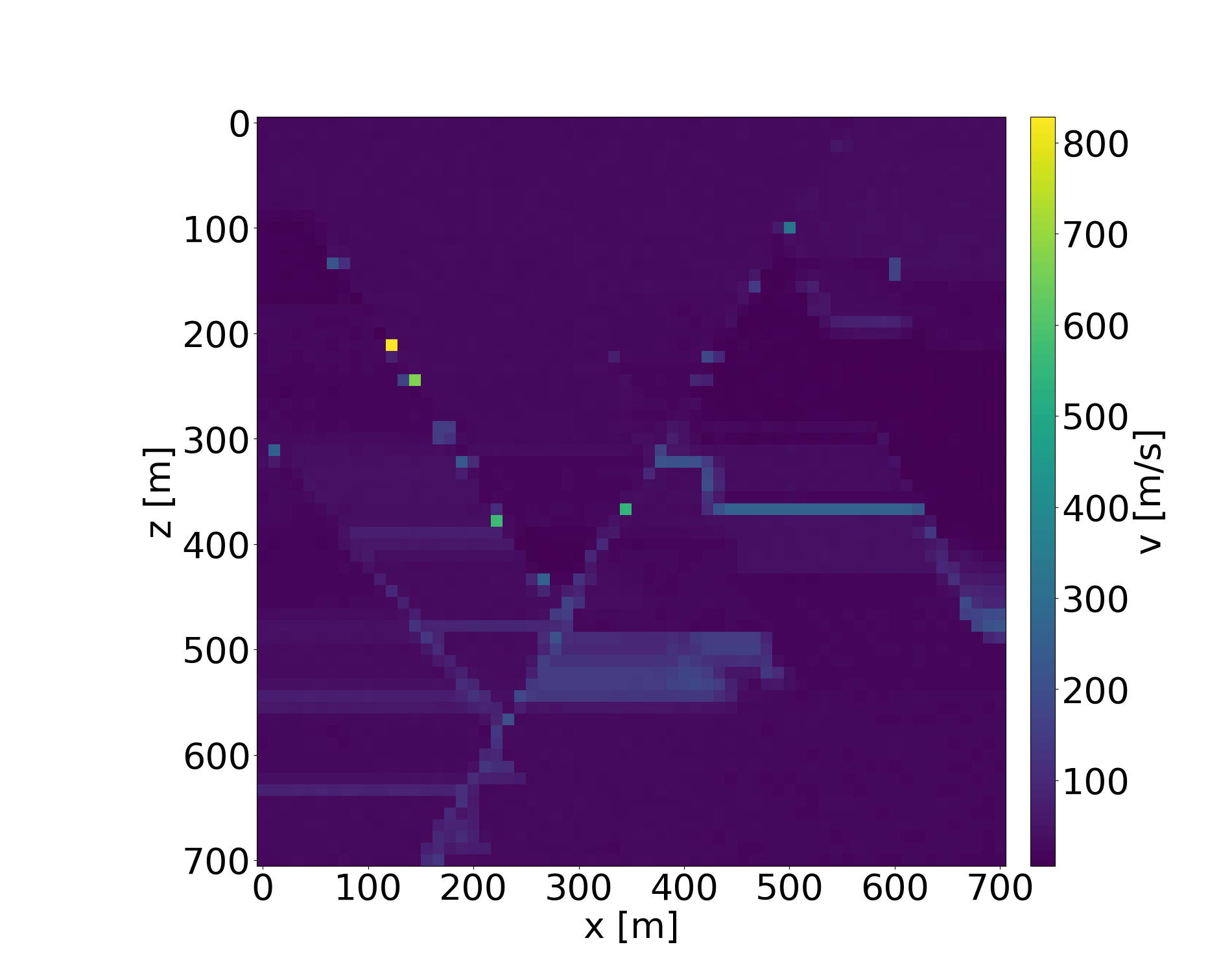}
        \caption{Standard deviation}
    \end{subfigure}
    \hfill
    \begin{subfigure}{0.48\textwidth}
        \includegraphics[width=\textwidth]{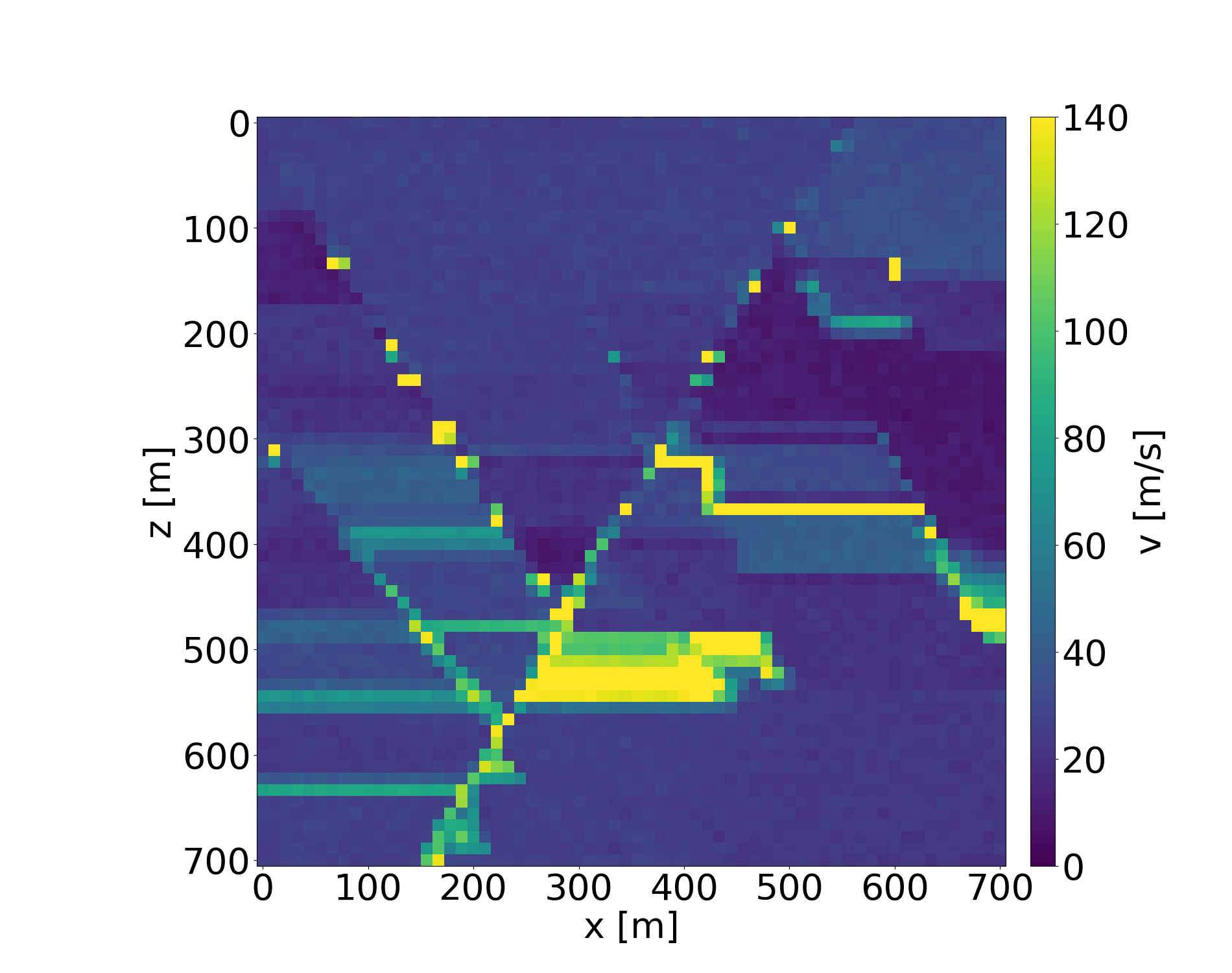}
        \caption{Standard deviation, clipped to [0, 150]}
    \end{subfigure}
    \hfill
    \caption{Assessing the uncertainty of a probabilistic inversion result. Upper row: target velocity model (left) and average velocity model of the inversion result (right). Bottom: standard deviation of the inversion result unscaled (left) and clipped to [0, 150] (right).}
    \label{fig-result-std}
\end{figure}

To better assess the quality as well as the uncertainty of the probabilistic inversion, we computed the standard deviation of 50 inversion results, as depicted in \autoref{fig-result-std}. It is obvious that largest uncertainty is at the faults and layer boundaries. This is to be expected, because when the velocity models in the training dataset were interpolated to 64x64, the sharp edges are affected by anti-aliasing. While the overall slope in the predicted velocity models is nearly the same, some pixels at the faults are sometimes considered to be in the left layer and some to be in the right layer. If there is a large velocity contrast in the two layers, it therefore results in a high standard deviation.

Overall, the probabilistic inversion results show a good agreement with the target velocity model and recover the general shape of the structures well.

\begin{figure}
    \centering
    \begin{subfigure}{0.3\textwidth}
        \includegraphics[width=\textwidth]{img/CodeOpenFWI/sample_000_target_cbar.png}
        \caption{Ground truth (target)}
    \end{subfigure}
    \hfill
    \begin{subfigure}{0.3\textwidth}
        \includegraphics[width=\textwidth]{img/CodeOpenFWI/sample_000_mean.png}
        \caption{Average inversion result}
    \end{subfigure}
    \hfill
    \begin{subfigure}{0.3\textwidth}
        \includegraphics[width=\textwidth]{img/CodeOpenFWI/sample000_std_scaled.png}
        \caption{Standard deviation}
    \end{subfigure}
    \hfill
    \begin{subfigure}{0.3\textwidth}
        \includegraphics[width=\textwidth]{img/CodeOpenFWI/sample_000_step100_batch_1.png}
        \caption{Probabilistic inversion result 1}
    \end{subfigure}
    \hfill
    \begin{subfigure}{0.3\textwidth}
        \includegraphics[width=\textwidth]{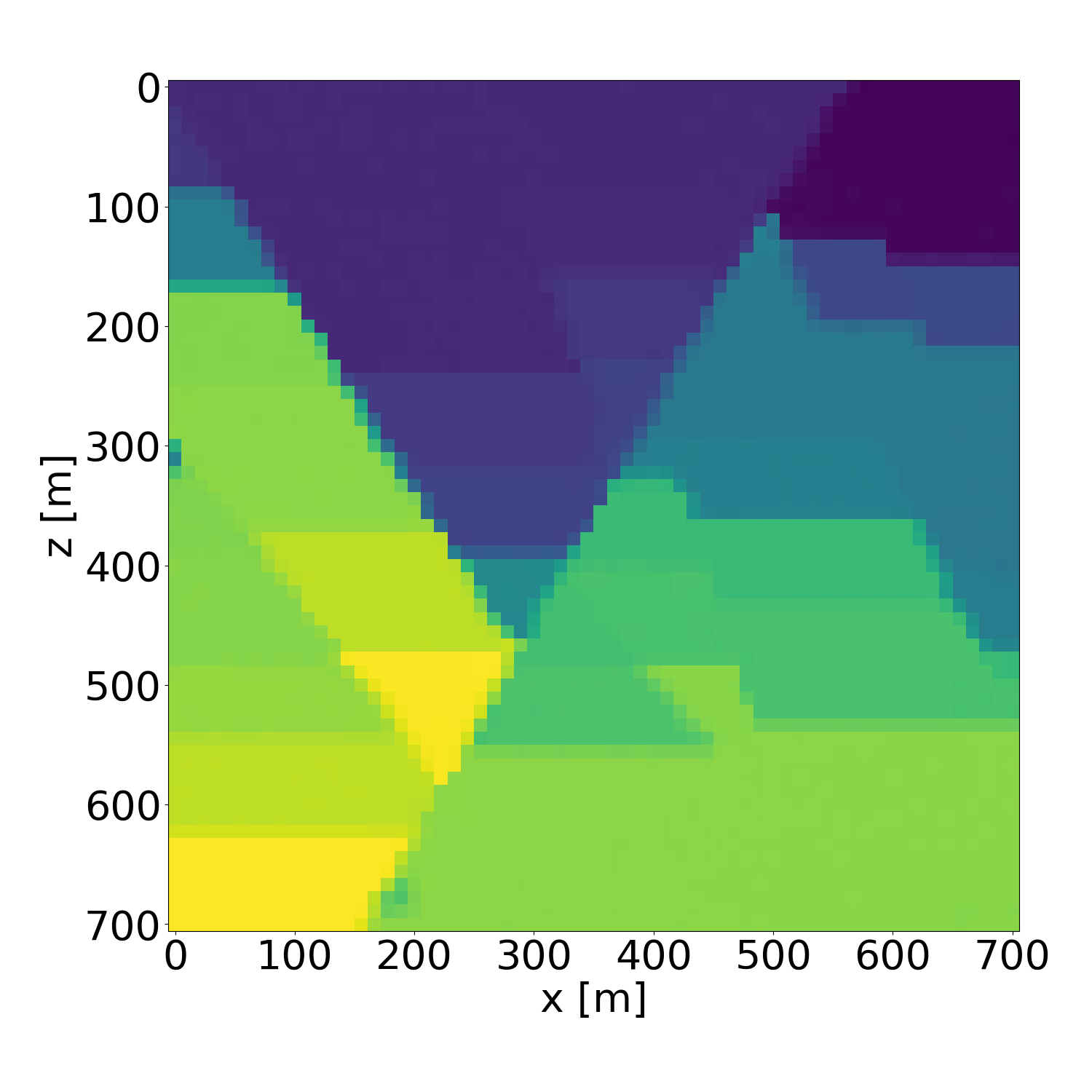}
        \caption{Probabilistic inversion result 2}
    \end{subfigure}
    \hfill
    \begin{subfigure}{0.3\textwidth}
        \includegraphics[width=\textwidth]{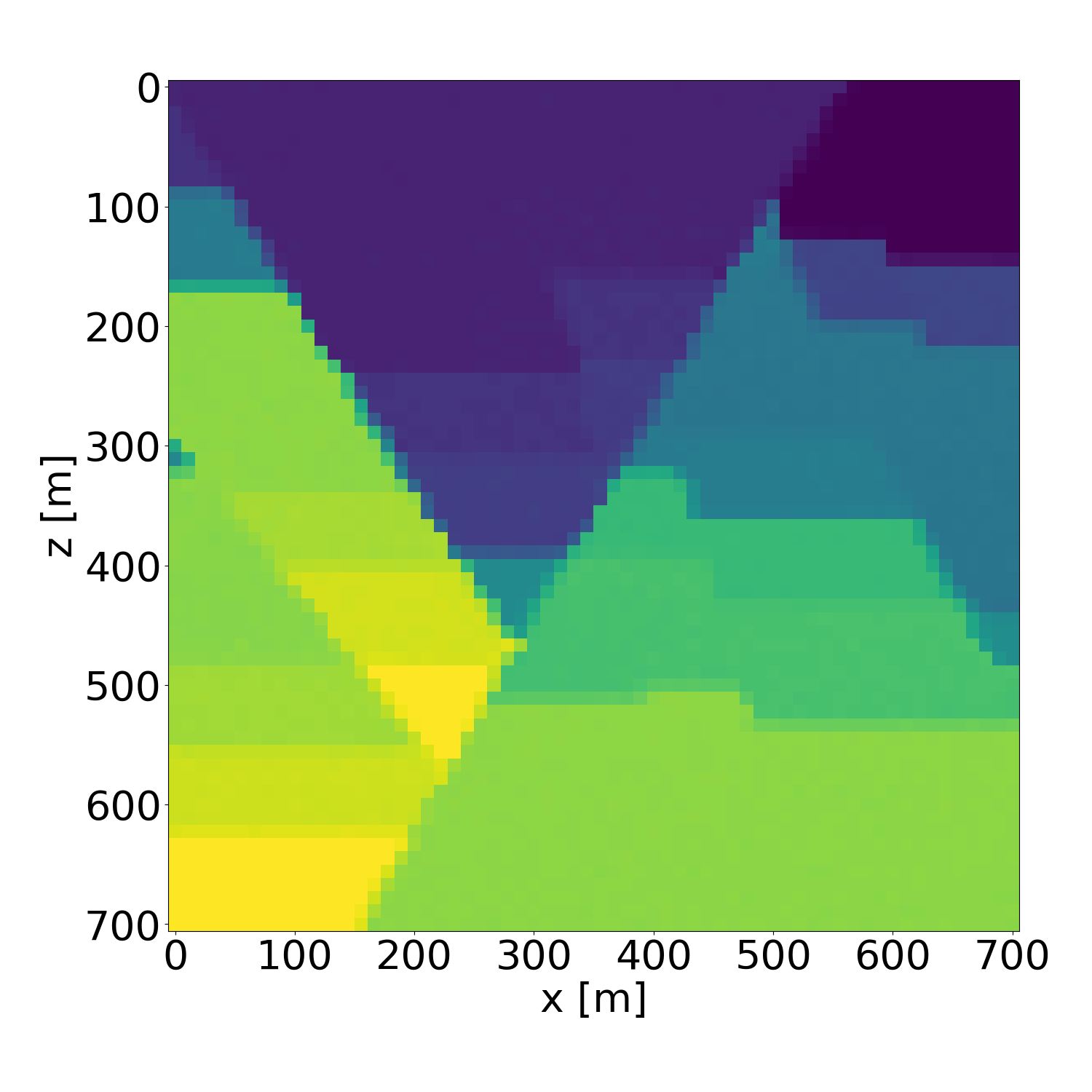}
        \caption{Probabilistic inversion result 3}
    \end{subfigure}
    \hfill
    \begin{subfigure}{0.3\textwidth}
        \includegraphics[width=\textwidth]{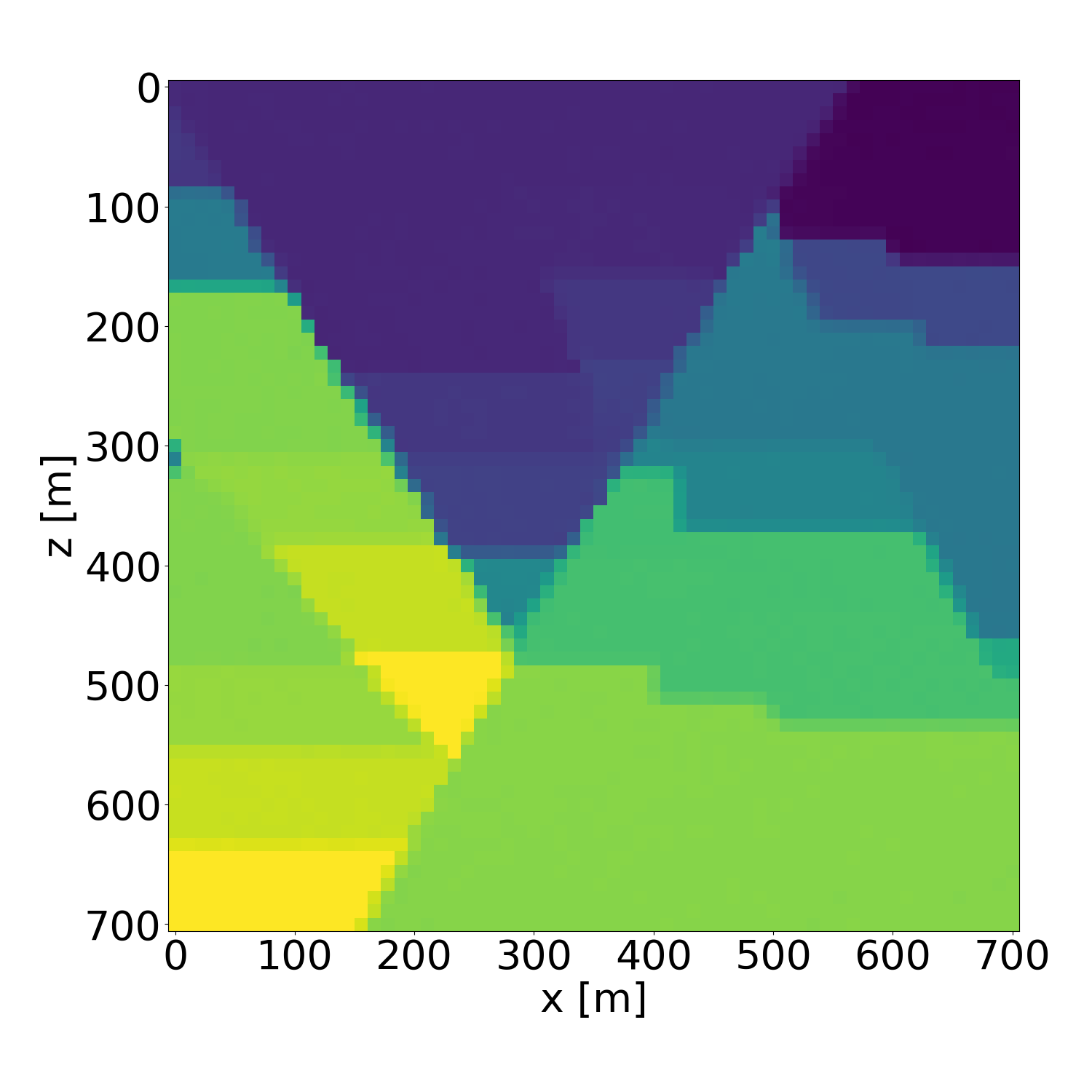}
        \caption{Probabilistic inversion result 4}
    \end{subfigure}
    \hfill
    \begin{subfigure}{0.3\textwidth}
        \includegraphics[width=\textwidth]{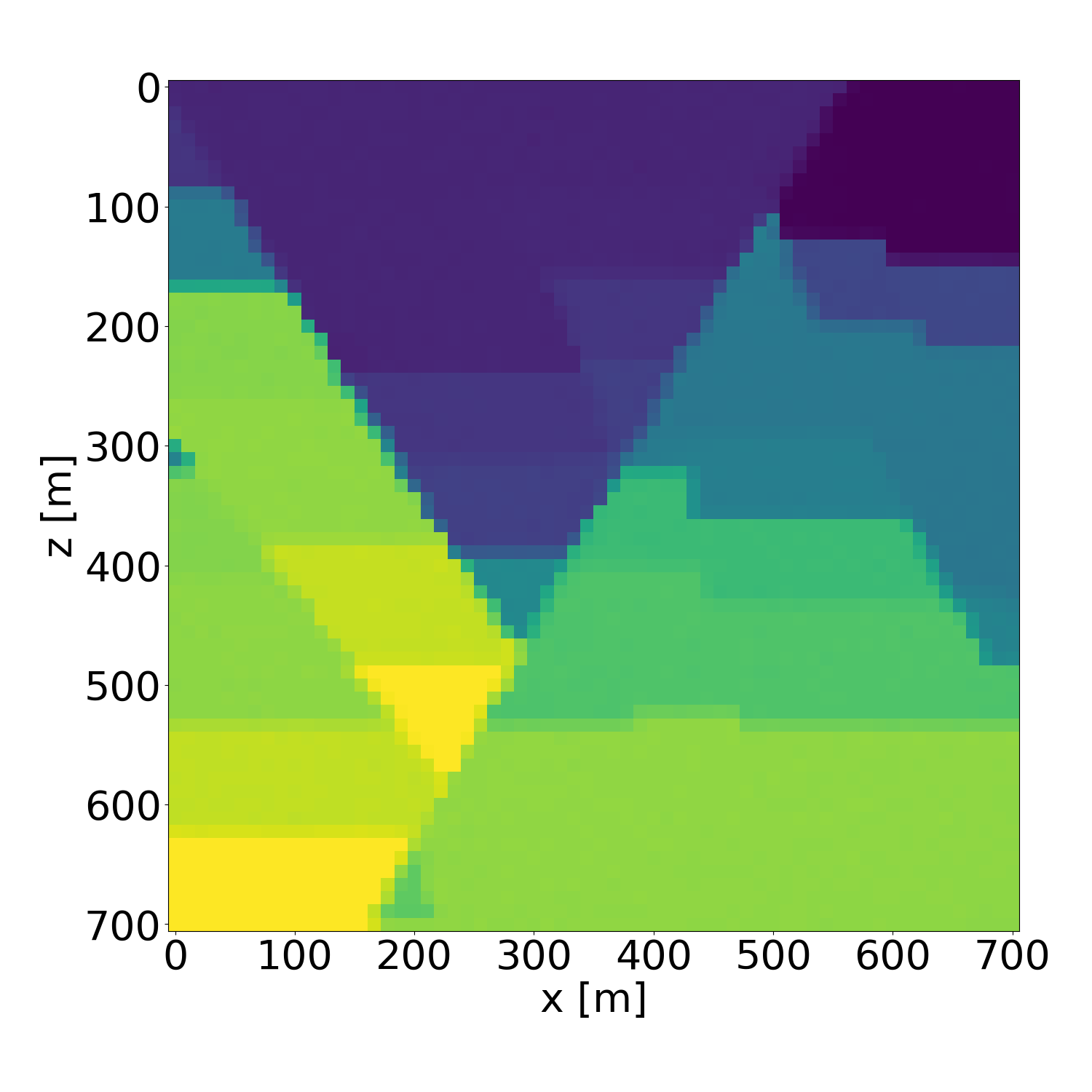}
        \caption{Probabilistic inversion result 5}
    \end{subfigure}
    \hfill
    \begin{subfigure}{0.3\textwidth}
        \includegraphics[width=\textwidth]{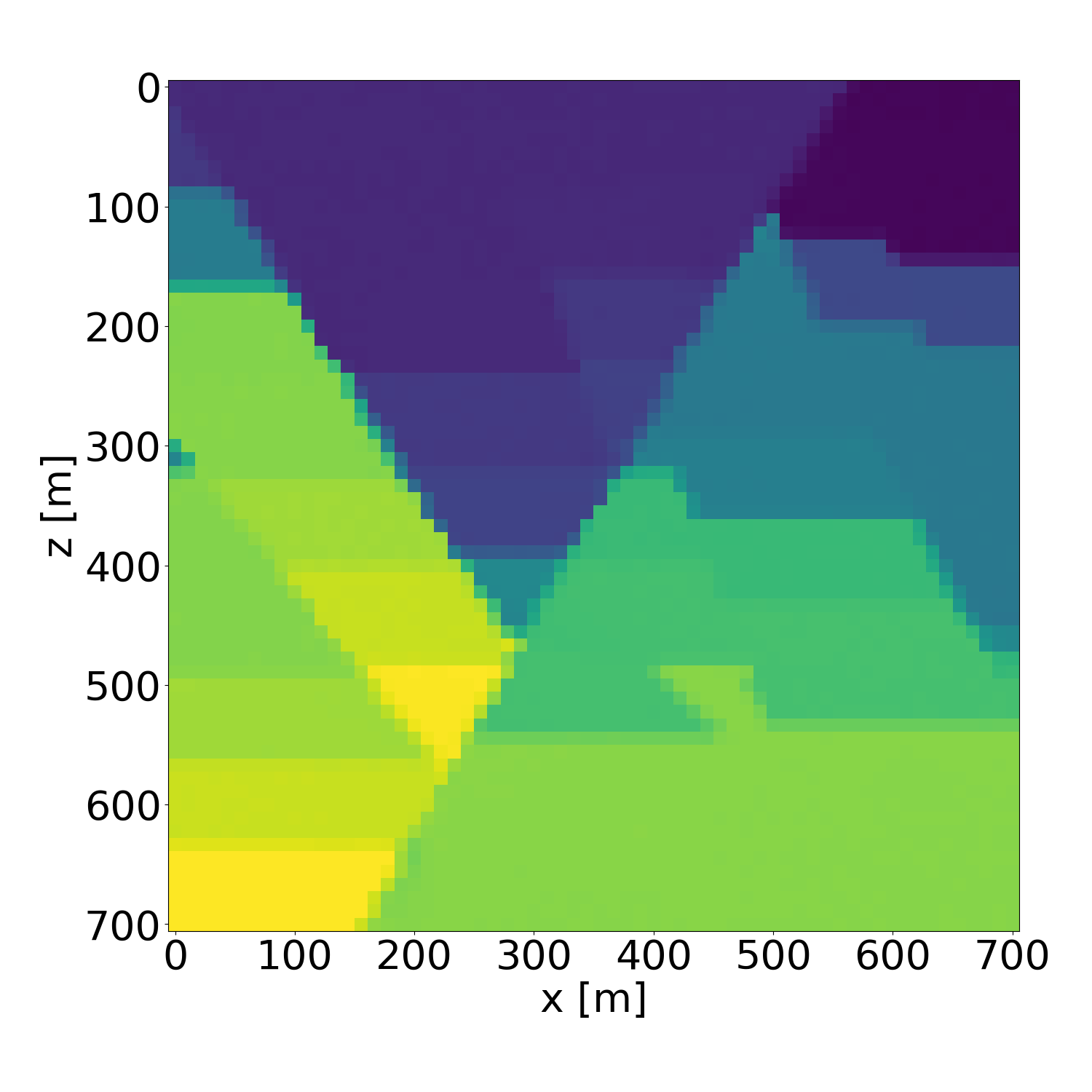}
        \caption{Probabilistic inversion result 6}
    \end{subfigure}
    \hfill
    \begin{subfigure}{0.3\textwidth}
        \includegraphics[width=\textwidth]{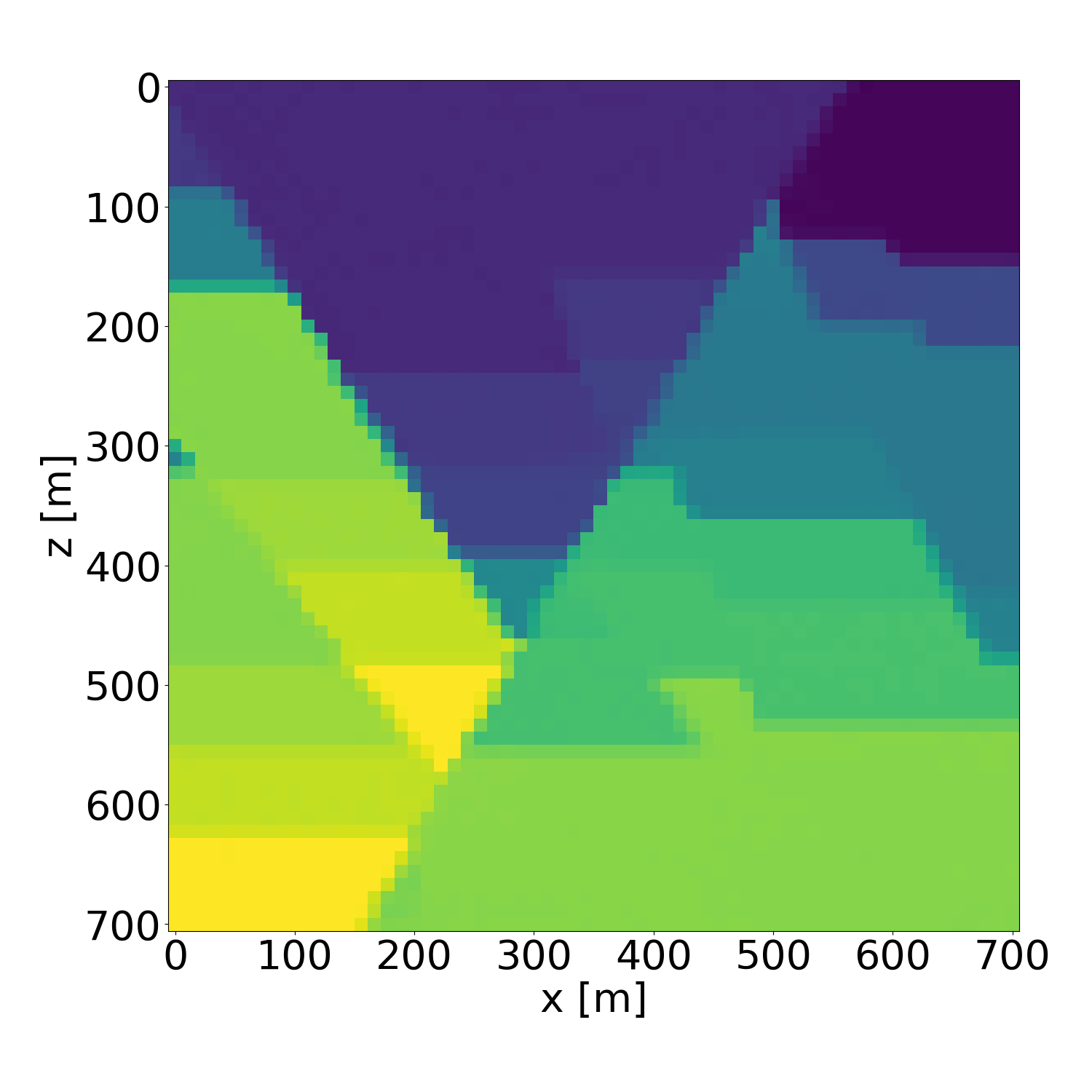}
        \caption{Probabilistic inversion result 7}
    \end{subfigure}
    \hfill
    \begin{subfigure}{0.3\textwidth}
        \includegraphics[width=\textwidth]{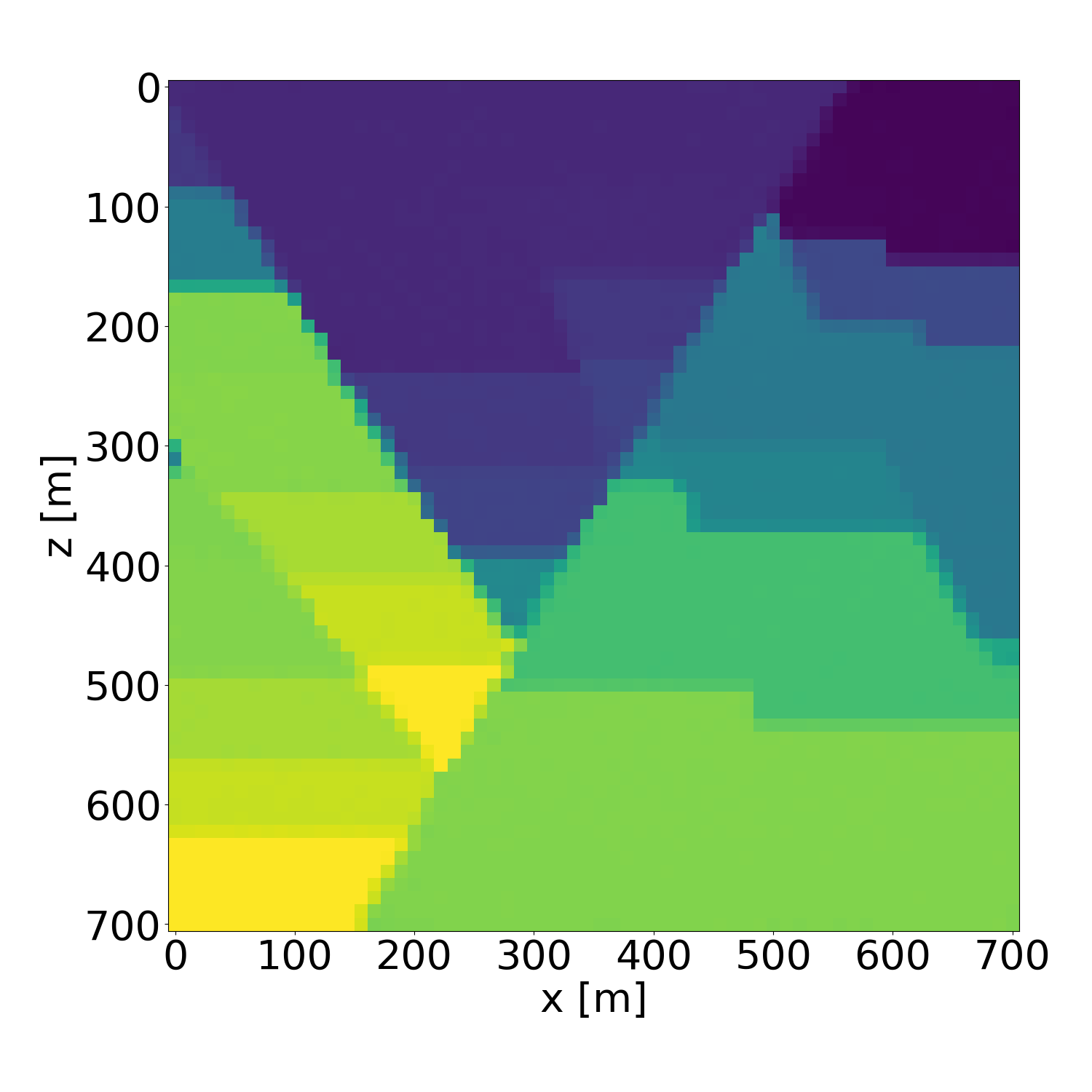}
        \caption{Probabilistic inversion result 8}
    \end{subfigure}
    \hfill
    \begin{subfigure}{0.3\textwidth}
        \includegraphics[width=\textwidth]{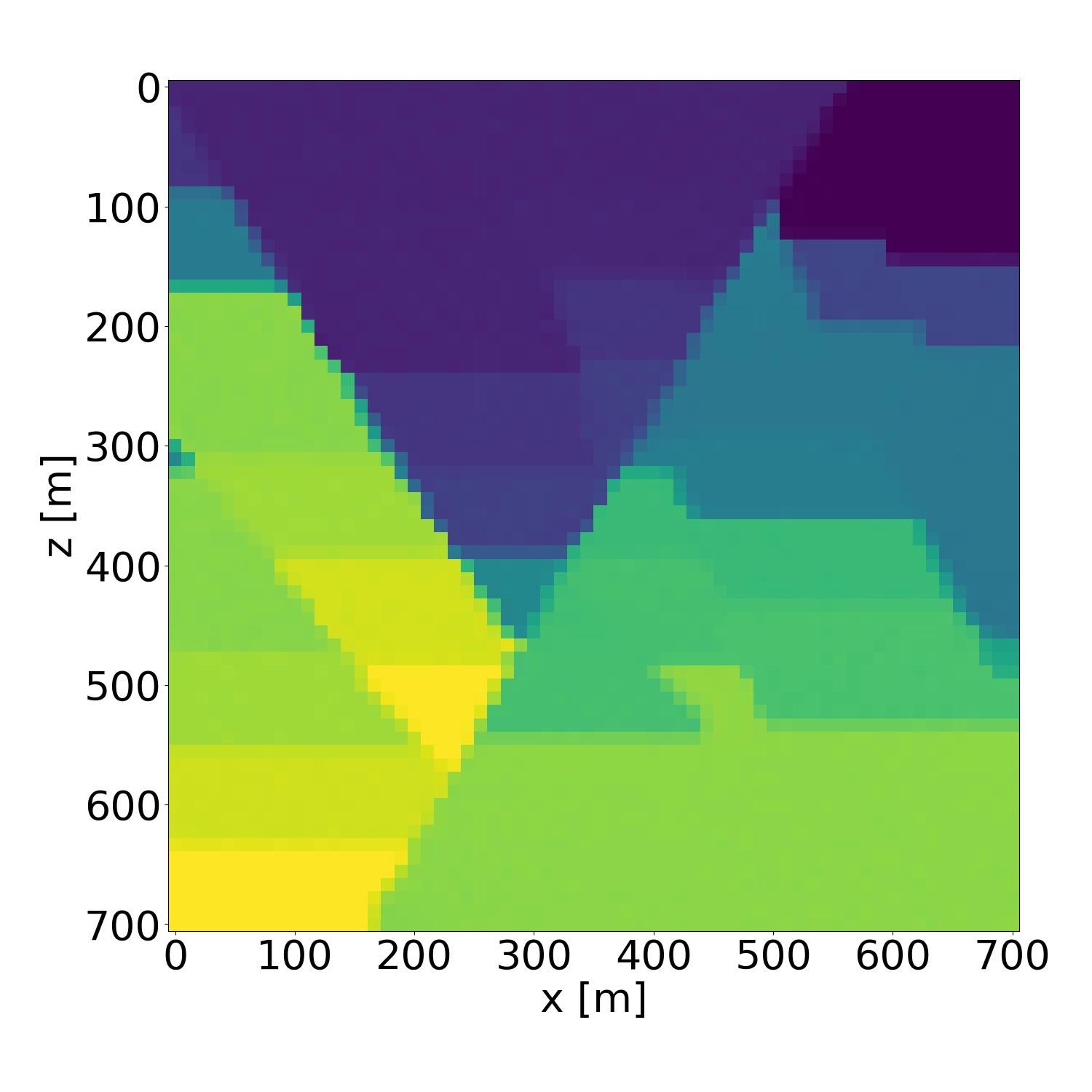}
        \caption{Probabilistic inversion result 9}
    \end{subfigure}
    \hfill
    \caption{Probabilistic inversion results of a isotropic velocity model in the validation dataset. Generated with guidance scale $w=4$.}
    \label{fig-allresult1}
\end{figure}

It is also apparent that the seismic velocity of the probabilistic inversion results are closer to the target velocity near the surface. This is to be expected as there was no geometric correction on the seismogram section, which was used for guidance. Therefore, the trace amplitude for the lower reflections is significantly smaller, and it is more difficult for the encoder to extract the information.\
Still, the probabilistic inversion results show a well-defined structure also in the lower parts (see \autoref{fig-allresult1}), even though there is no or little information on this region. The reason for this is that Flow Matching approximates the target distribution, so the Neural Network generates flows to samples that lie in this distribution. When trained on real data (i.e. geologically realistic velocity models), this has two implications:\
First, that clear boundaries and geologically reasonable velocity models can be predicted without the need of smoothness or boundary constraints, which are necessary in deterministic inversion. Even if there is no information available. It is therefore important to sample multiple times from the predicted posterior distribution and review the standard deviation to assess which boundaries are likely true.\
The second implication is, that whenever there is no guidance information (such as in deeper regions), the Neural Network will predict velocity values that may be likely based on the data it was trained on. It will therefore extrapolate based on the velocity model distribution it learned during training.\
Therefore, it is essential to choose the velocity models in the training dataset carefully to ensure that the training distribution is both as realistic and diverse as possible because it will directly impact how the Neural Network will predict regions with little or no information from the seismic data.

\section{Discussion}
We showed that Flow Matching can be used for probabilistic seismic velocity inversion for simple and complex models. The Flow Matching objective is straightforward to implement in modern Deep Learning frameworks and it resolves in a simple regression problem, which is stable to train. During inference, it is only necessary to solve the ODE.\
Flow Matching may be used to perform probabilistic inversion on high-dimensional distributions, as it does not rely on sampling-based methods.

Flow Matching provides a general framework for probabilistic inversion that can easily be adapted to various geophysical problem settings. There are two key requirements: first, a training dataset must be obtained that accurately represents the target model distribution and includes observed data for each model. Second, a Neural Network architecture has to be designed that suits the data modality.

As an example, we performed a case study using the OpenFWI FlatFaultB dataset. After training, multiple probabilistic inversions were obtained in just a few seconds. However, in the FlatFaultB dataset, there is always the same acquisition geometry, which made it possible to use the pretrained EfficientNet encoder. In real-world scenarios, seismic surveys are generally done with varying acquisition geometry, so the architecture we used in the case study would not be suitable in that case. We believe that it is of high interest to study and research Neural Network architectures that work for arbitrary acquisition geometries. Once such a NN is trained, it may enable the rapid generation of a distribution of plausible high-resolution velocity models.

Furthermore, the fully data-driven approach allows us to use a compressed feature representation of the full wavefield, which not only reduces computational requirements, but also avoids issues like cycle-skipping which are common in physics-driven deterministic inversion methods.

\newpage
\printbibliography

\section{Appendix}

\subsection{More Probabilistic Inversion Results}

In the following figures, the standard deviation of the inversion results has been clipped to $[0, 140]$ for better comparison.

\begin{figure}
    \centering
    \begin{subfigure}{0.3\textwidth}
        \includegraphics[width=\textwidth]{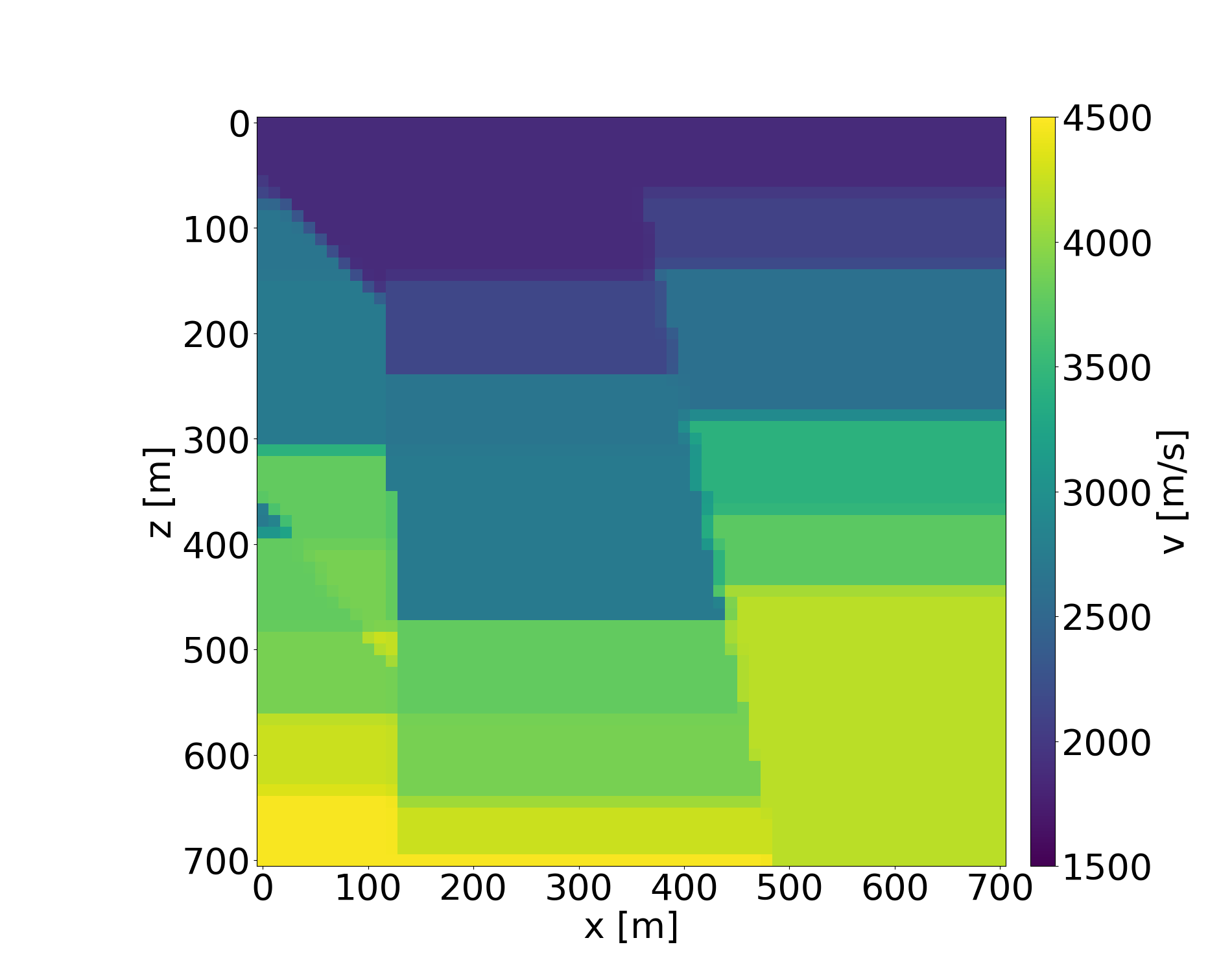}
        \caption{Ground truth (target)}
    \end{subfigure}
    \hfill
    \begin{subfigure}{0.3\textwidth}
        \includegraphics[width=\textwidth]{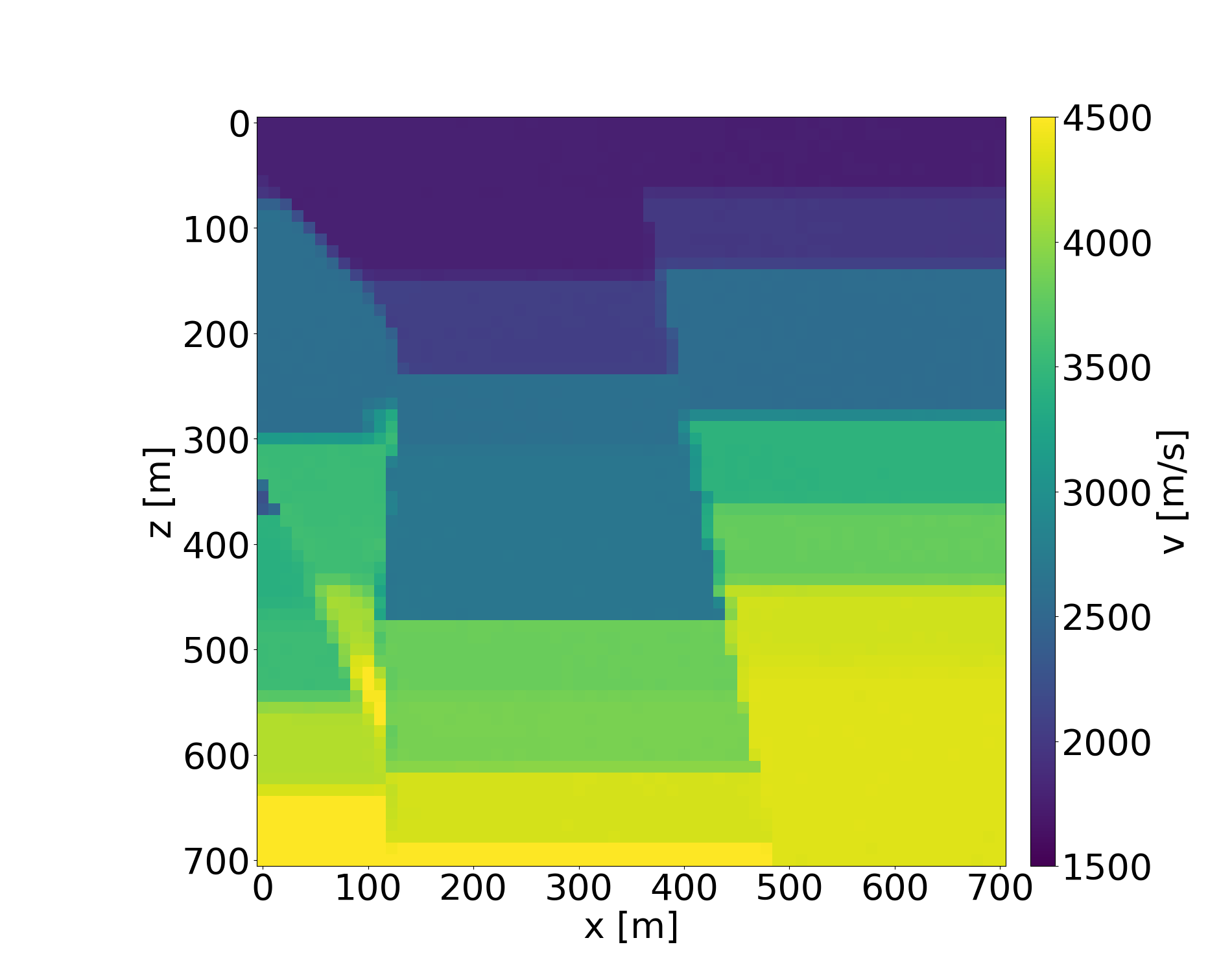}
        \caption{Average inversion result}
    \end{subfigure}
    \hfill
    \begin{subfigure}{0.3\textwidth}
        \includegraphics[width=\textwidth]{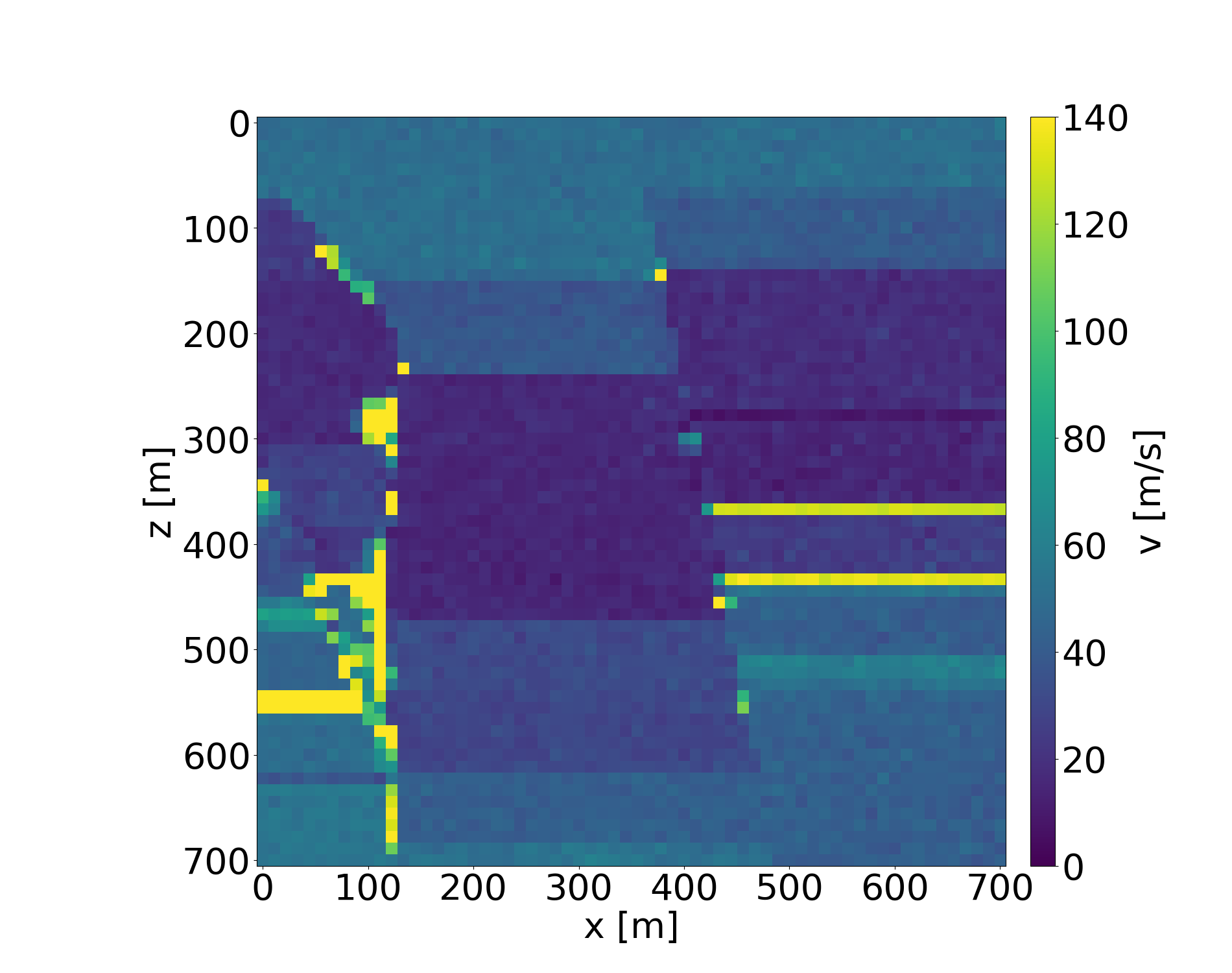}
        \caption{Standard deviation}
    \end{subfigure}
    \hfill
    \begin{subfigure}{0.3\textwidth}
        \includegraphics[width=\textwidth]{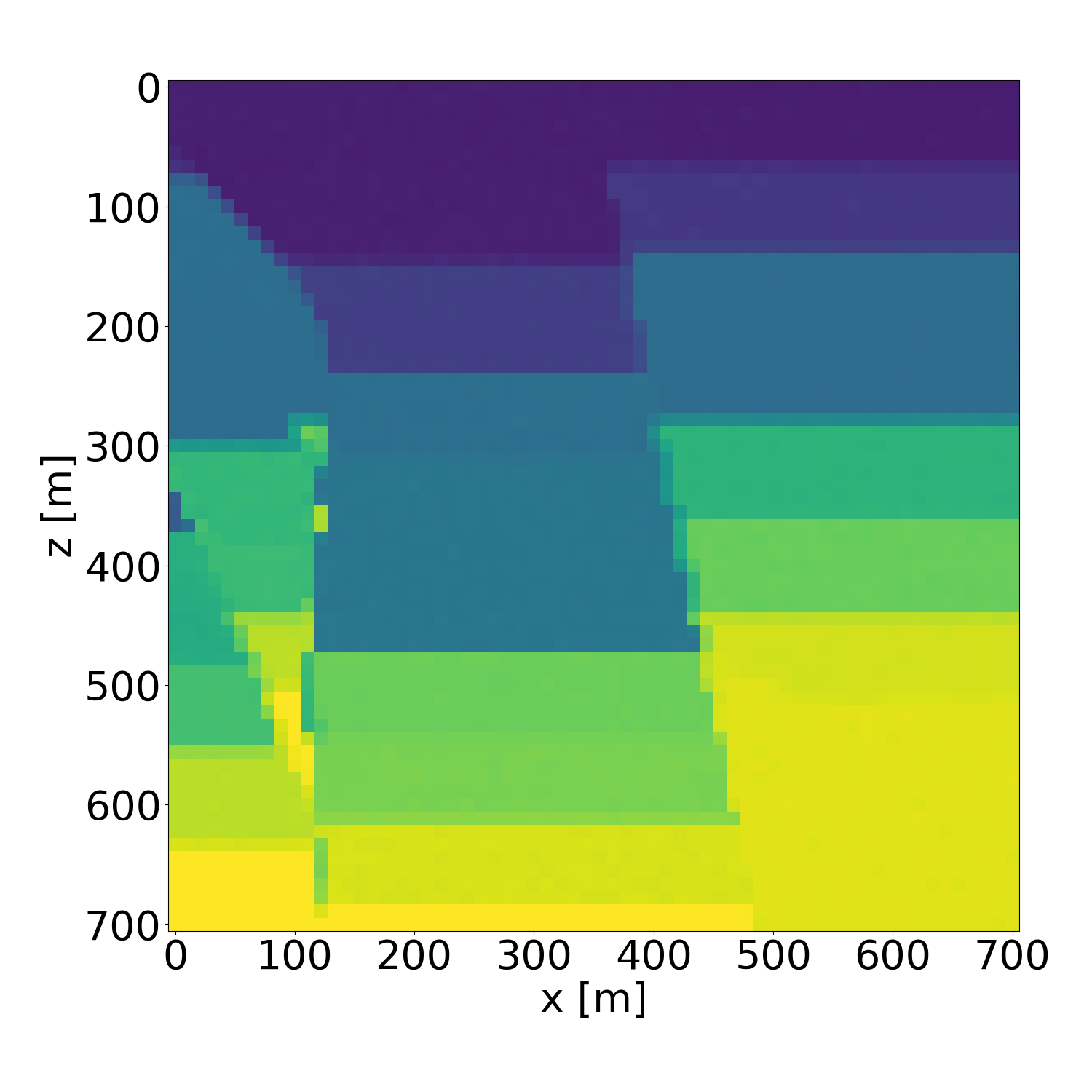}
        \caption{Probabilistic inversion result 1}
    \end{subfigure}
    \hfill
    \begin{subfigure}{0.3\textwidth}
        \includegraphics[width=\textwidth]{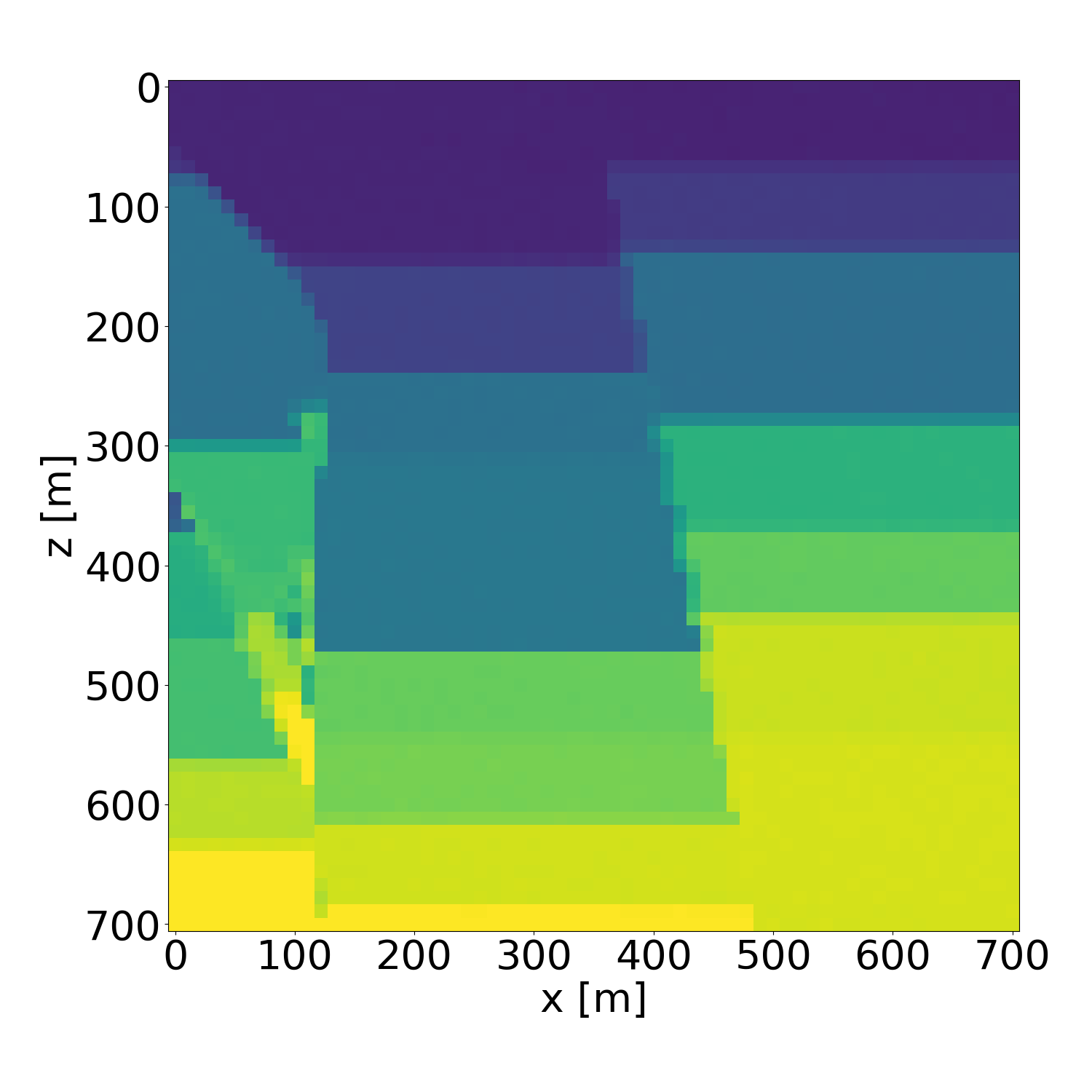}
        \caption{Probabilistic inversion result 2}
    \end{subfigure}
    \hfill
    \begin{subfigure}{0.3\textwidth}
        \includegraphics[width=\textwidth]{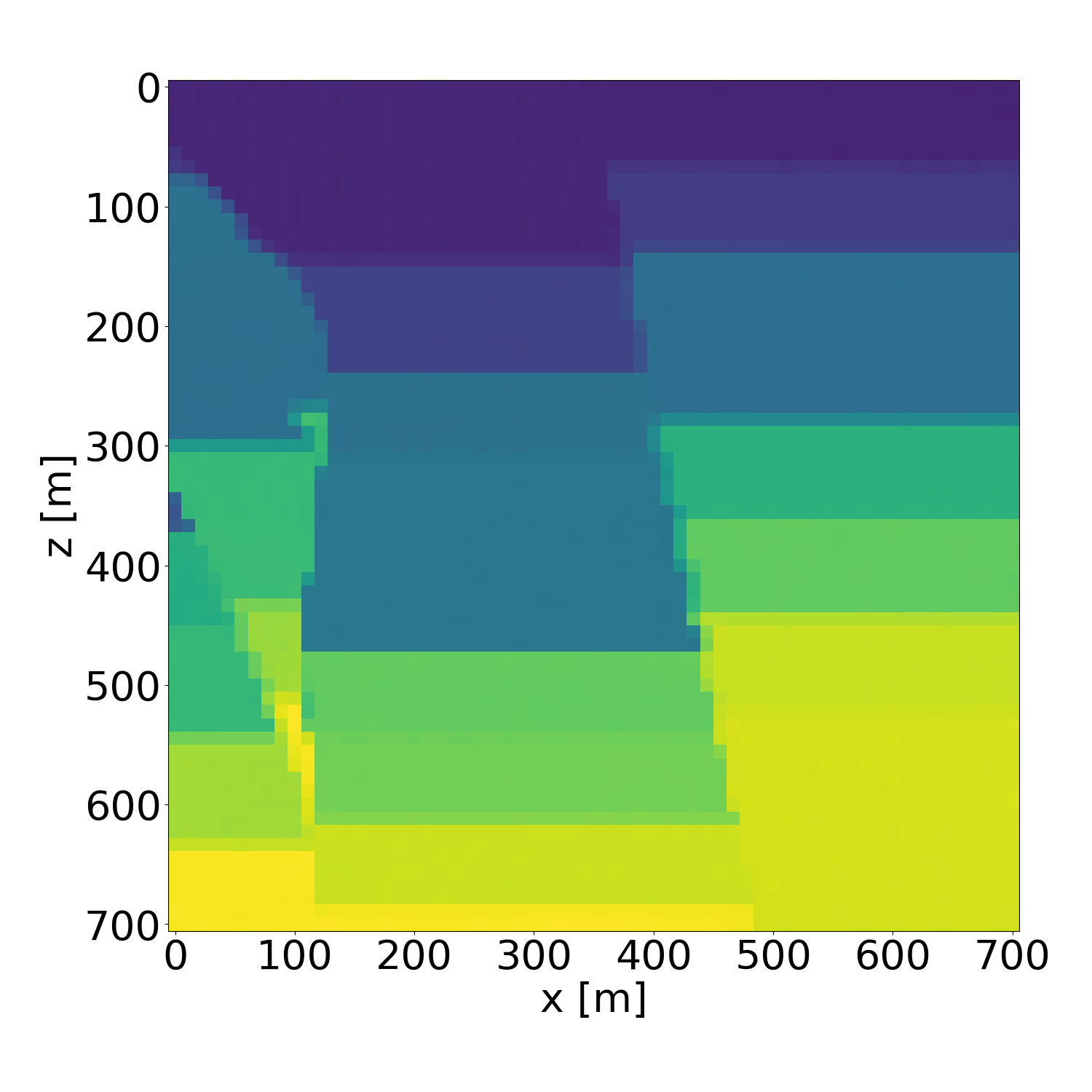}
        \caption{Probabilistic inversion result 3}
    \end{subfigure}
    \hfill
    \begin{subfigure}{0.3\textwidth}
        \includegraphics[width=\textwidth]{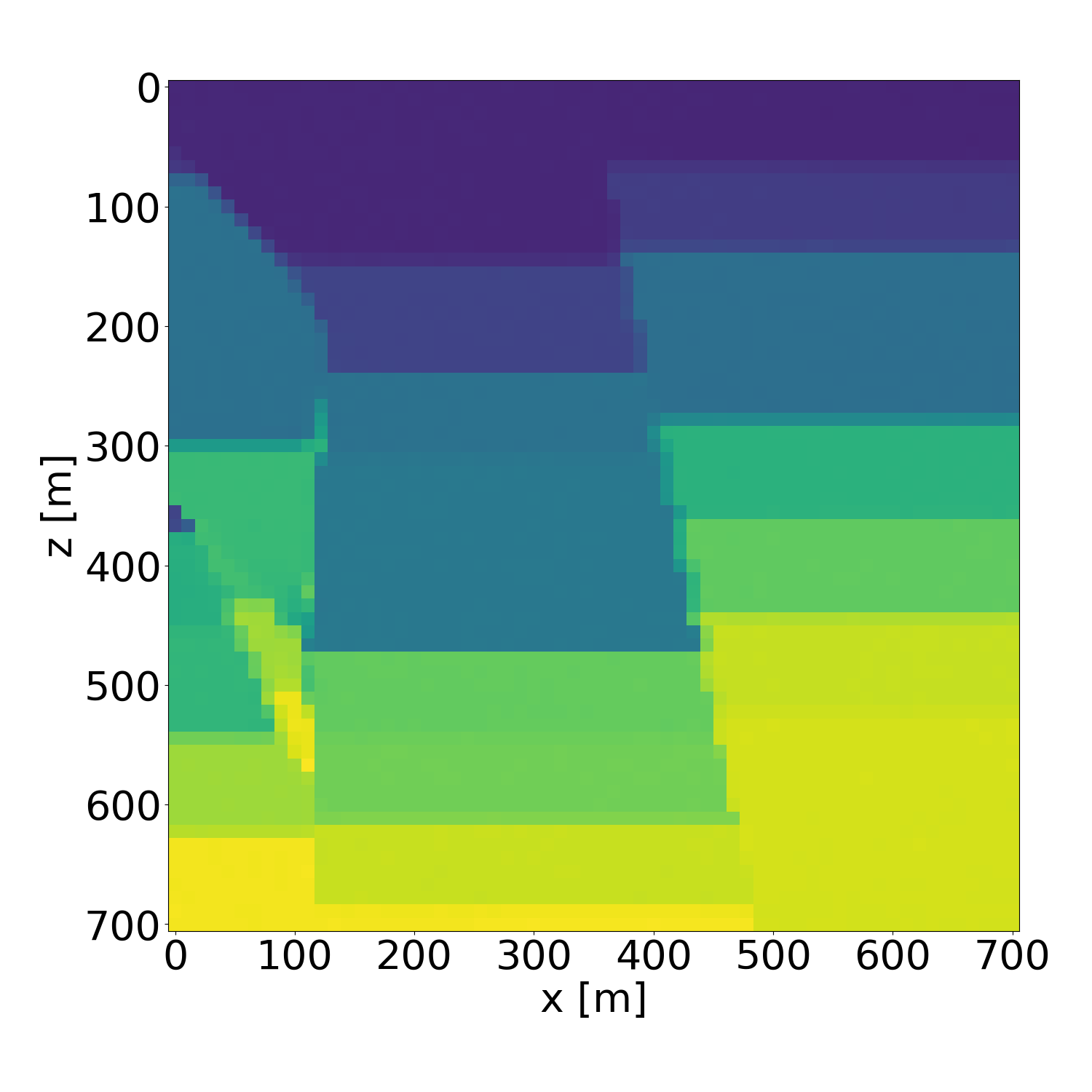}
        \caption{Probabilistic inversion result 4}
    \end{subfigure}
    \hfill
    \begin{subfigure}{0.3\textwidth}
        \includegraphics[width=\textwidth]{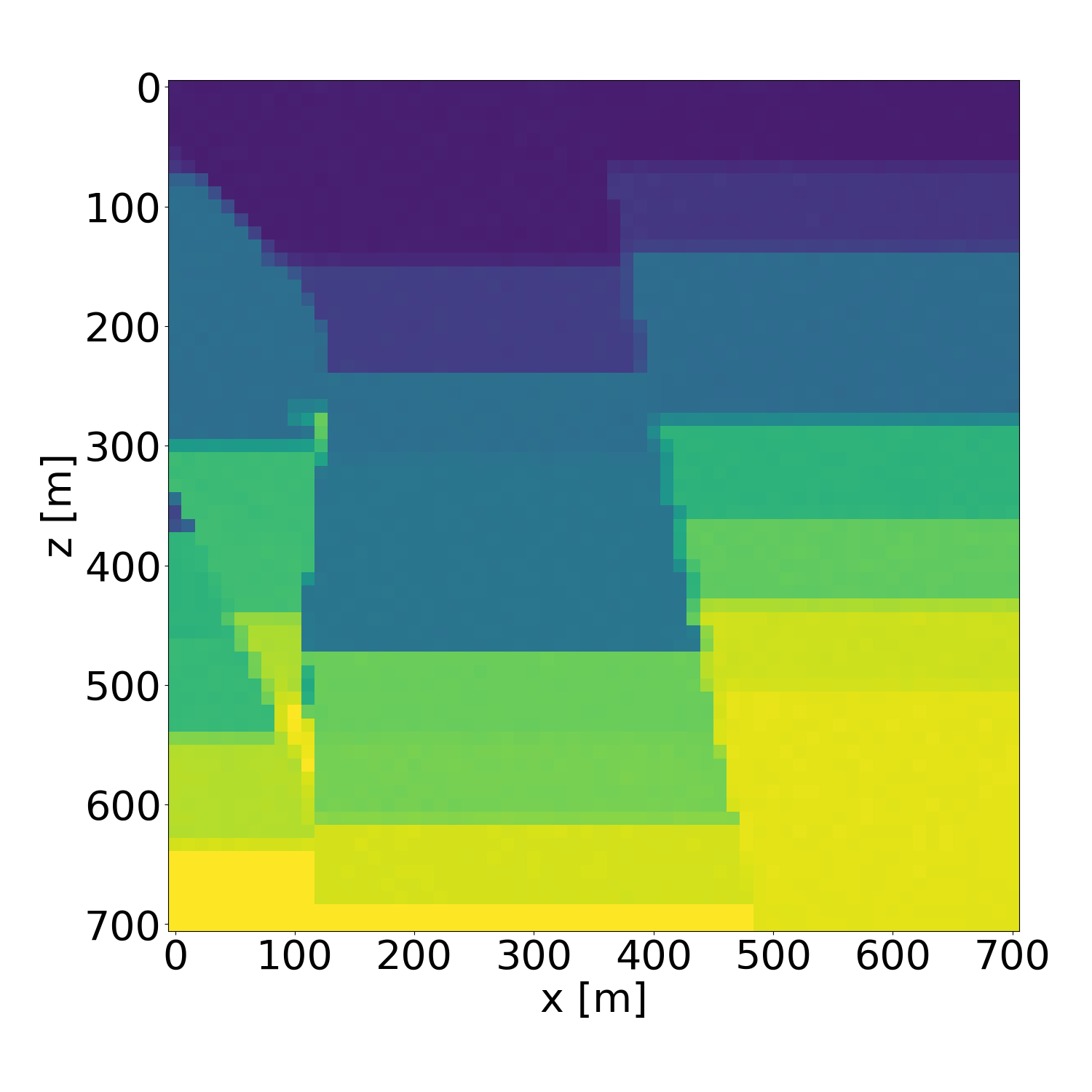}
        \caption{Probabilistic inversion result 5}
    \end{subfigure}
    \hfill
    \begin{subfigure}{0.3\textwidth}
        \includegraphics[width=\textwidth]{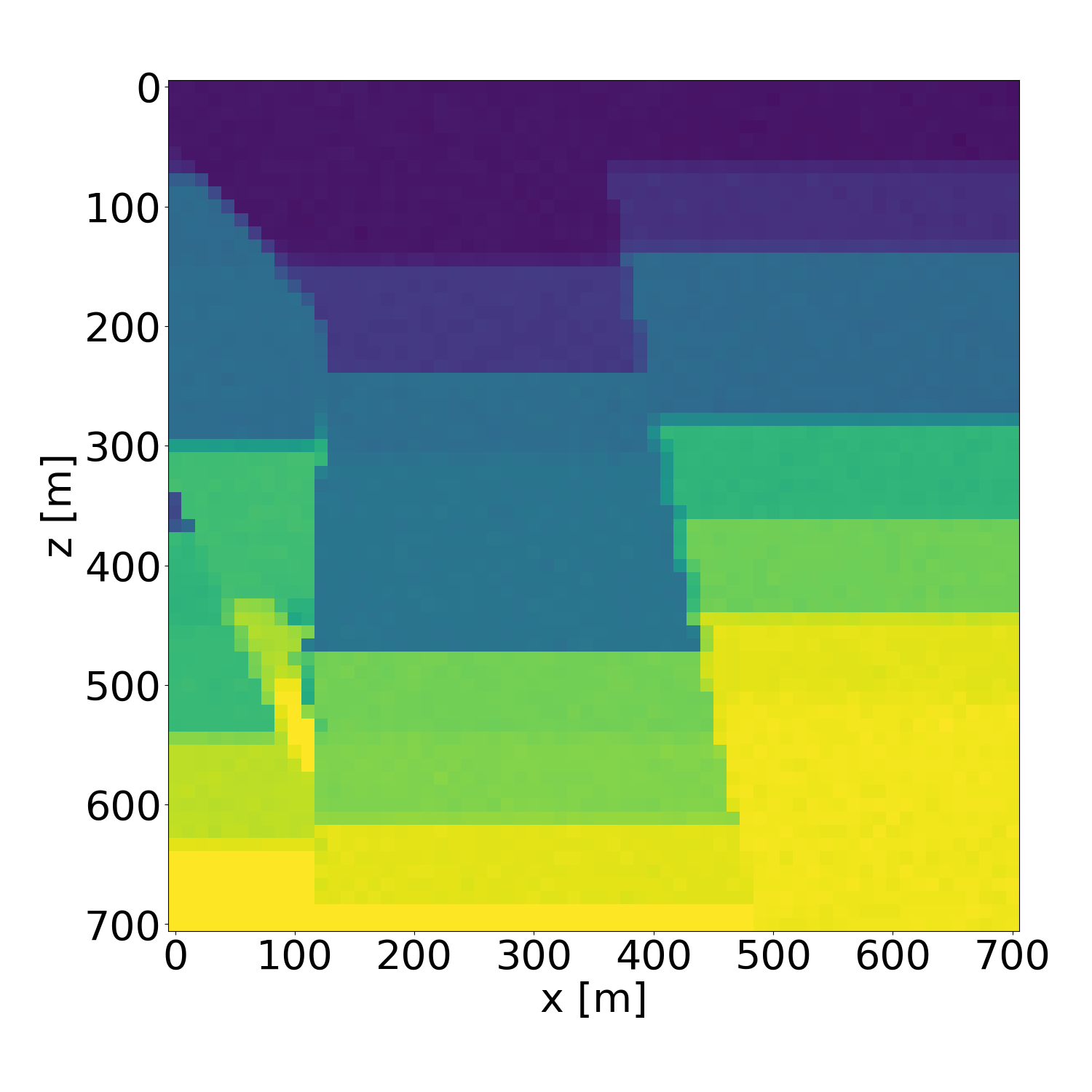}
        \caption{Probabilistic inversion result 6}
    \end{subfigure}
    \hfill
    \begin{subfigure}{0.3\textwidth}
        \includegraphics[width=\textwidth]{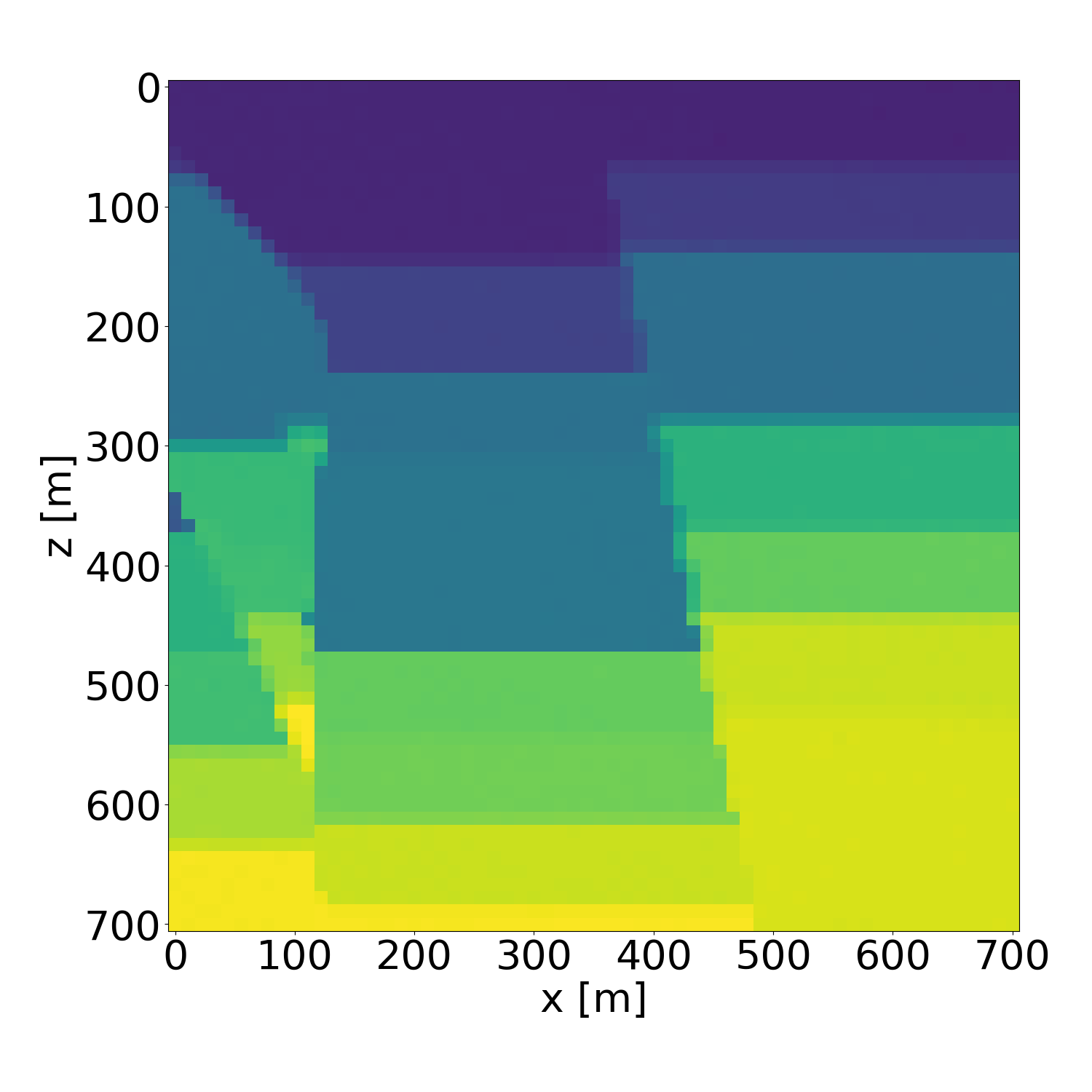}
        \caption{Probabilistic inversion result 7}
    \end{subfigure}
    \hfill
    \begin{subfigure}{0.3\textwidth}
        \includegraphics[width=\textwidth]{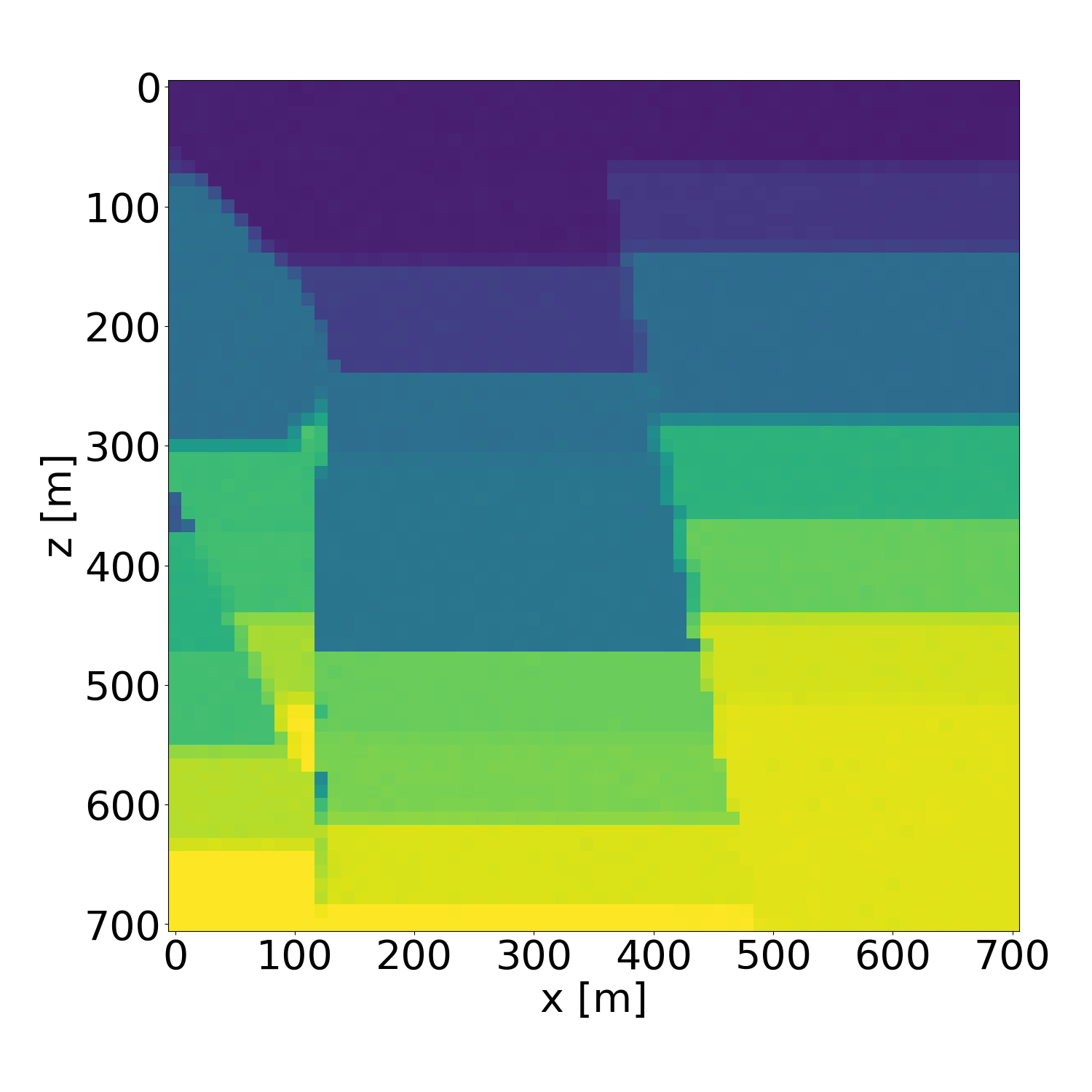}
        \caption{Probabilistic inversion result 8}
    \end{subfigure}
    \hfill
    \begin{subfigure}{0.3\textwidth}
        \includegraphics[width=\textwidth]{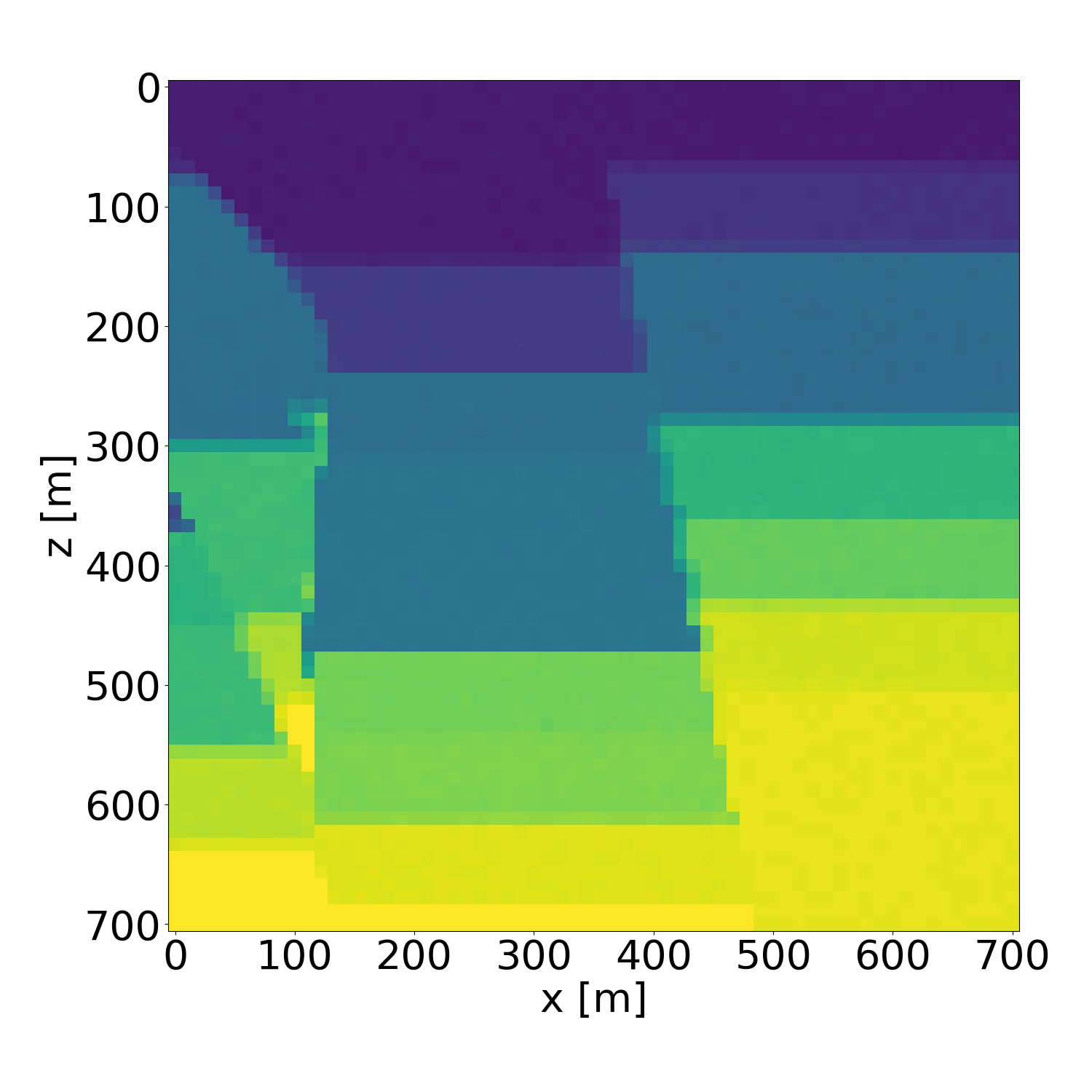}
        \caption{Probabilistic inversion result 9}
    \end{subfigure}
    \hfill
    \caption{Probabilistic inversion results of a isotropic velocity model in the validation dataset. Generated with guidance scale $w=4$.}
    \label{fig-allresult2}
\end{figure}

\begin{figure}
    \centering
    \begin{subfigure}{0.3\textwidth}
        \includegraphics[width=\textwidth]{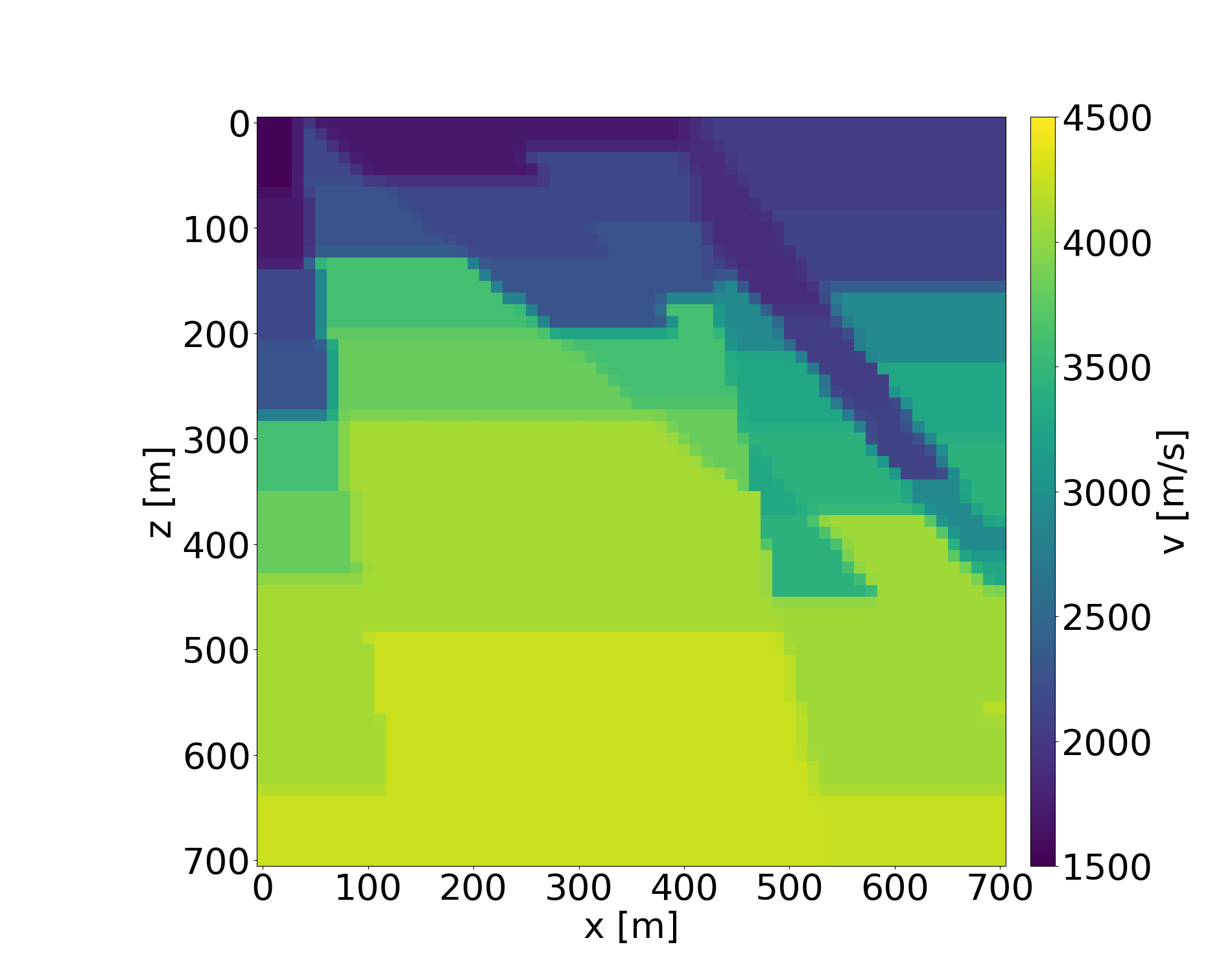}
        \caption{Ground truth (target)}
    \end{subfigure}
    \hfill
    \begin{subfigure}{0.3\textwidth}
        \includegraphics[width=\textwidth]{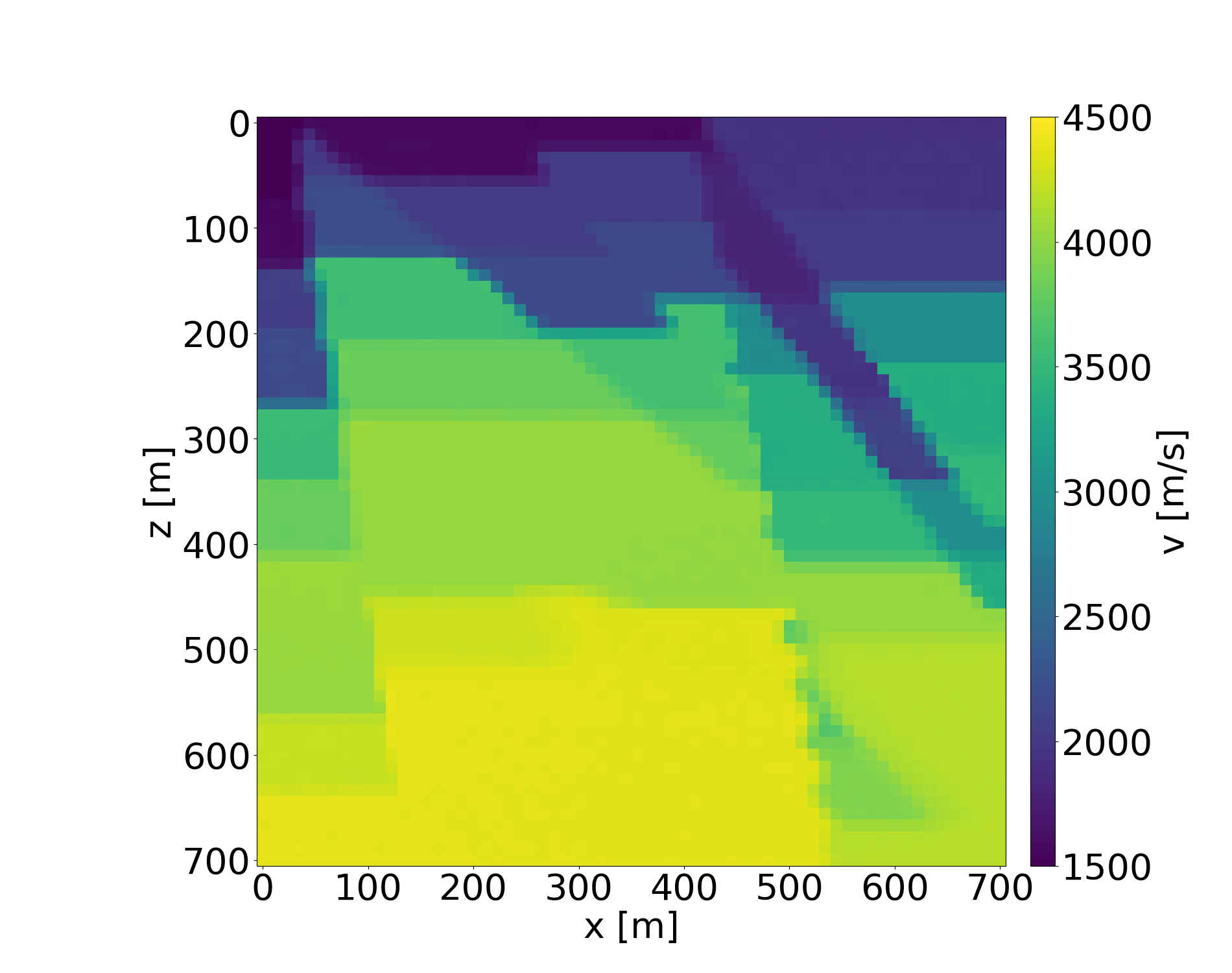}
        \caption{Average inversion result}
    \end{subfigure}
    \hfill
    \begin{subfigure}{0.3\textwidth}
        \includegraphics[width=\textwidth]{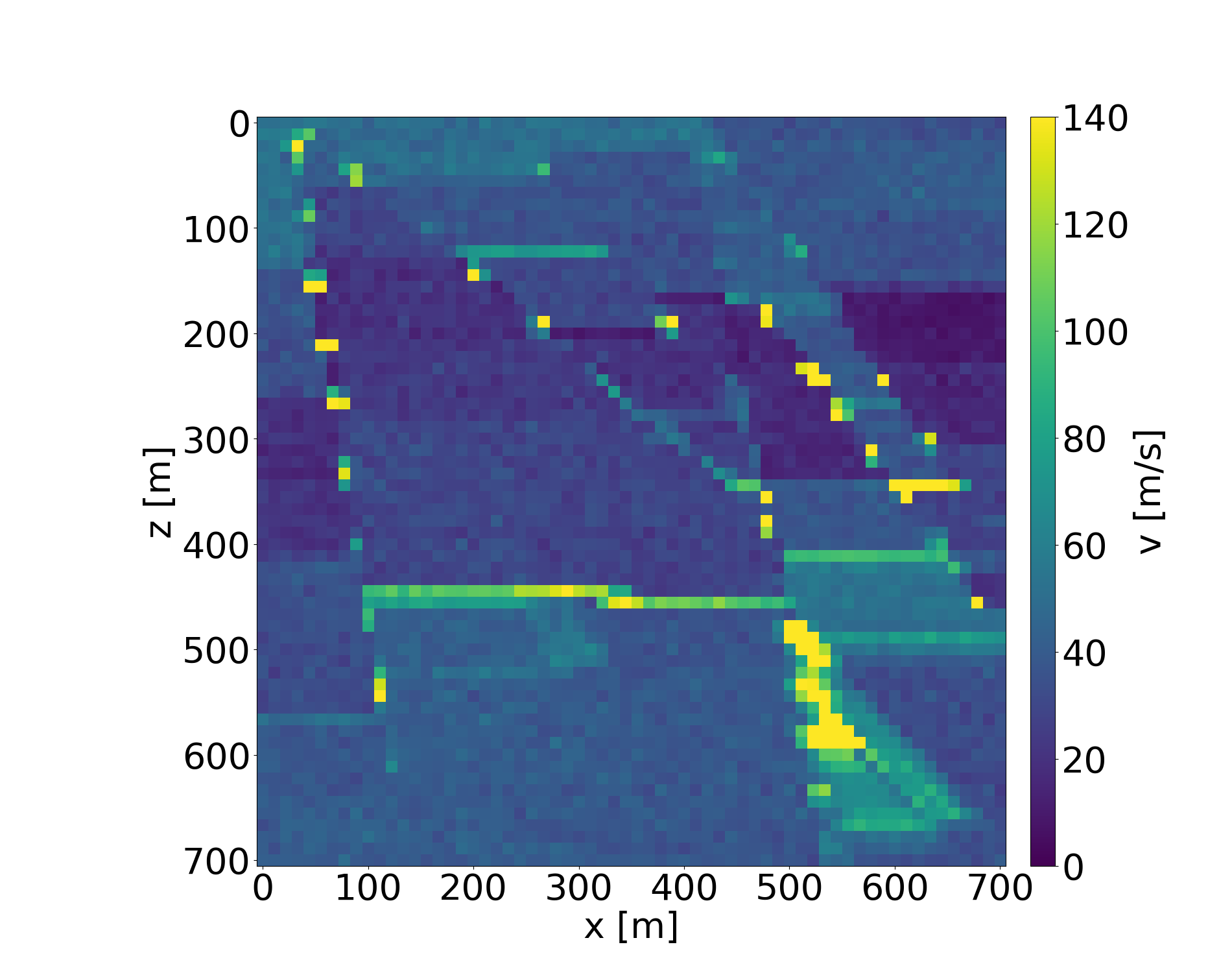}
        \caption{Standard deviation}
    \end{subfigure}
    \hfill
    \begin{subfigure}{0.3\textwidth}
        \includegraphics[width=\textwidth]{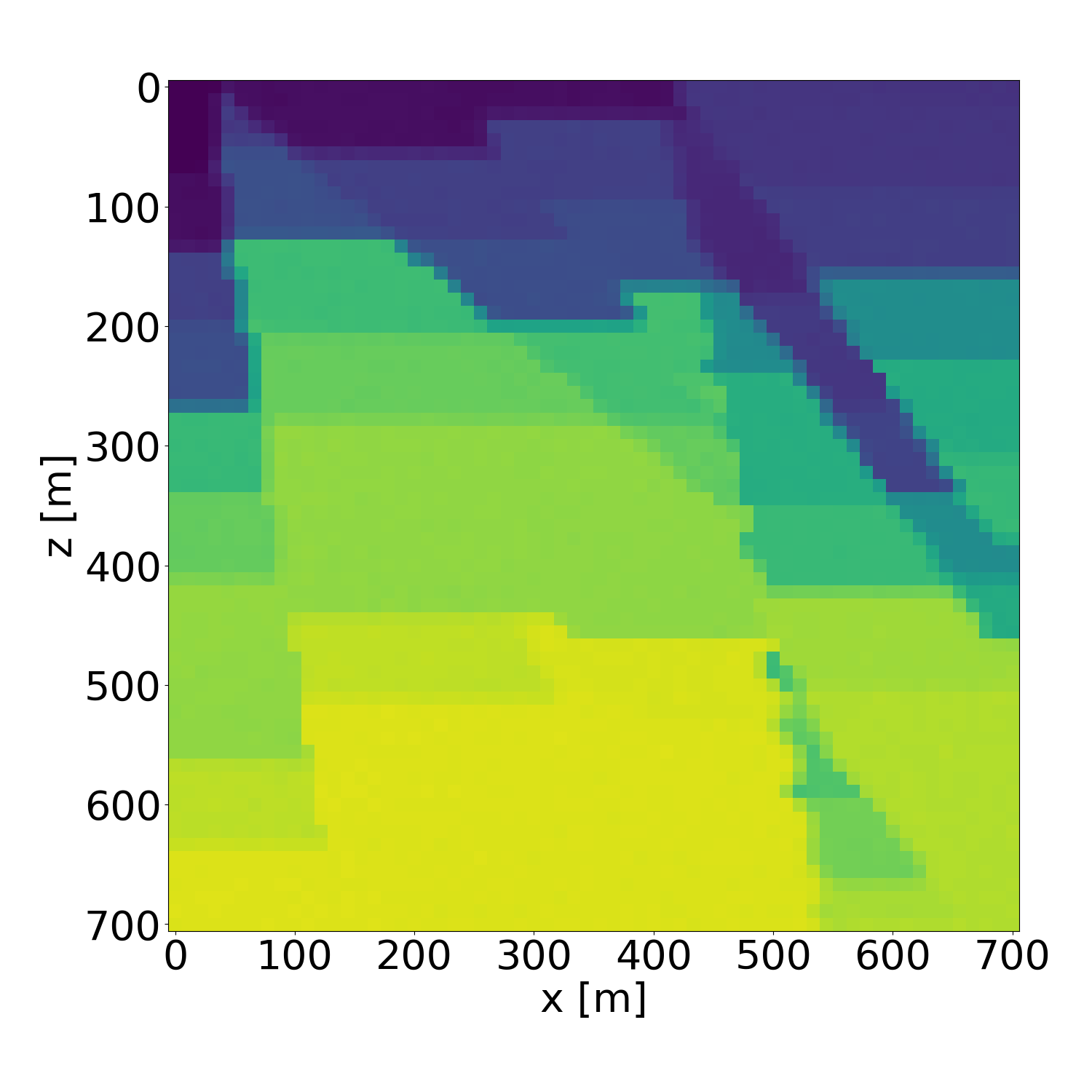}
        \caption{Probabilistic inversion result 1}
    \end{subfigure}
    \hfill
    \begin{subfigure}{0.3\textwidth}
        \includegraphics[width=\textwidth]{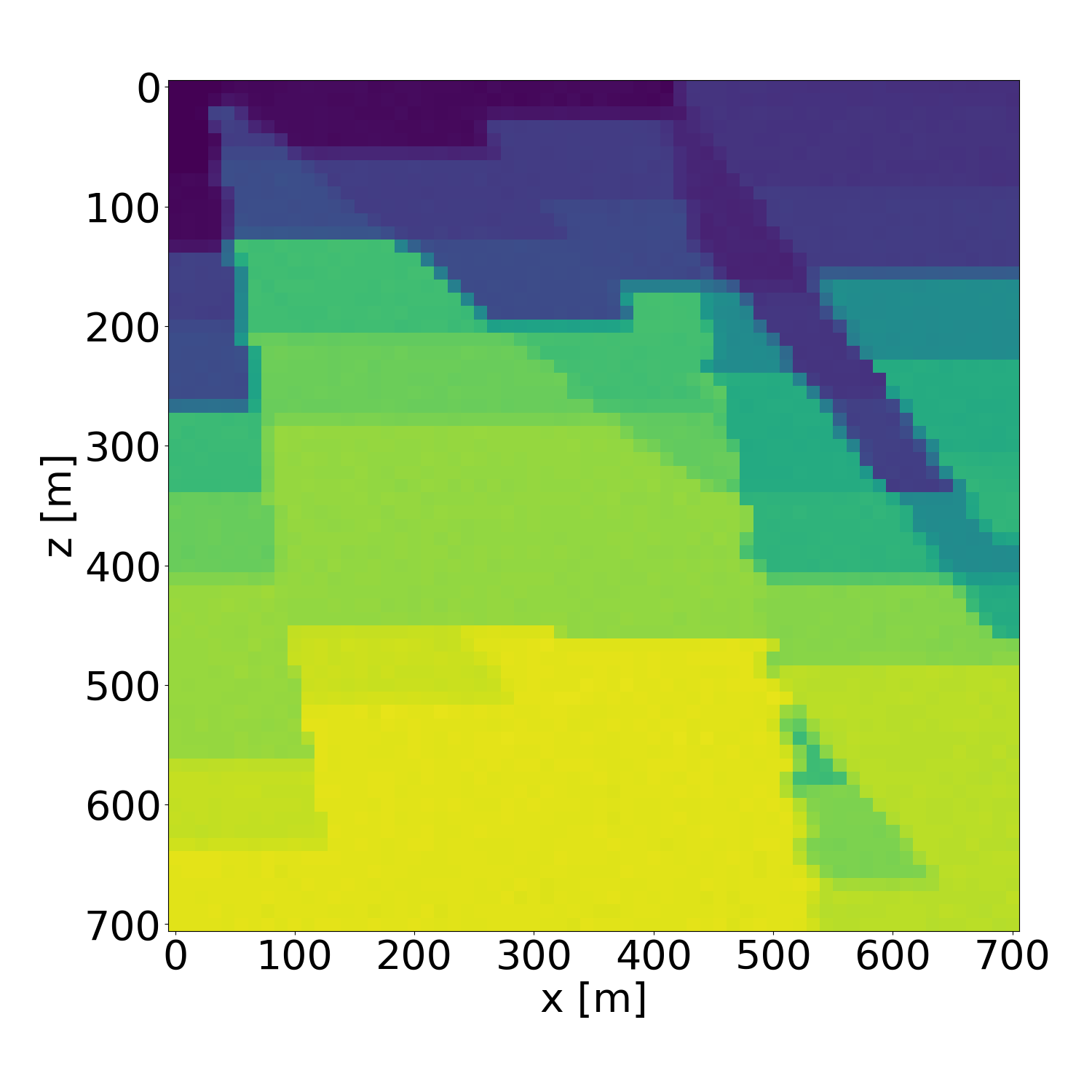}
        \caption{Probabilistic inversion result 2}
    \end{subfigure}
    \hfill
    \begin{subfigure}{0.3\textwidth}
        \includegraphics[width=\textwidth]{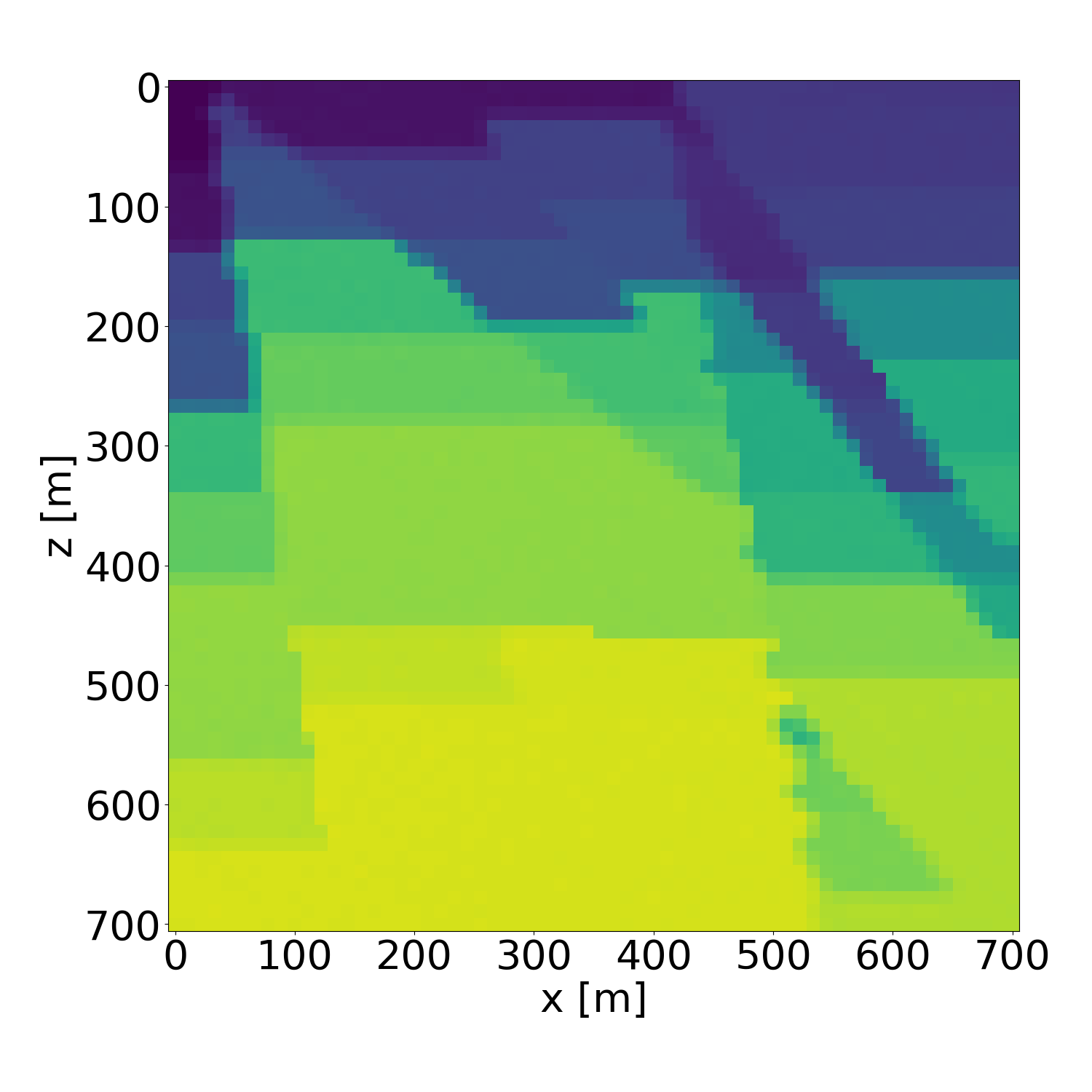}
        \caption{Probabilistic inversion result 3}
    \end{subfigure}
    \hfill
    \begin{subfigure}{0.3\textwidth}
        \includegraphics[width=\textwidth]{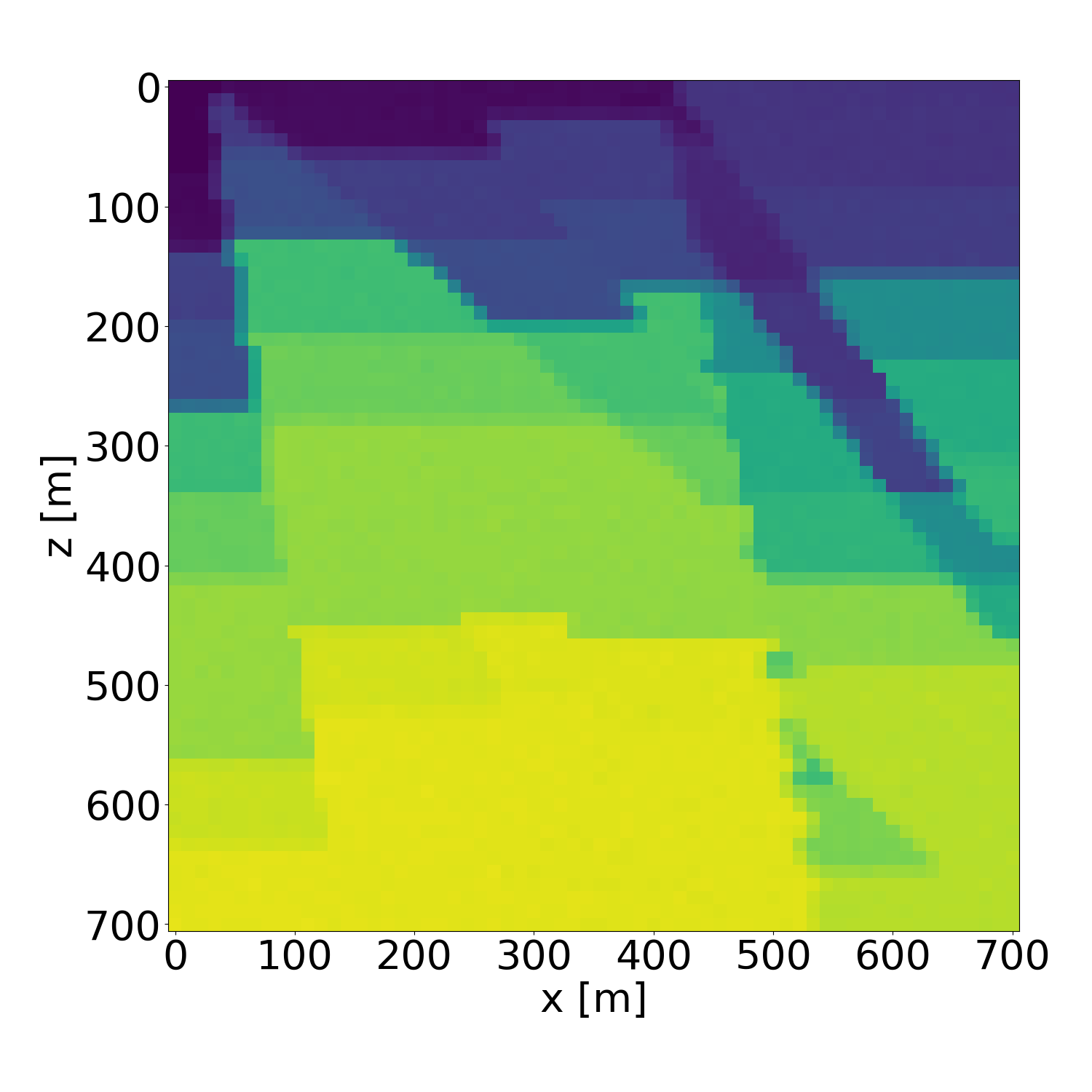}
        \caption{Probabilistic inversion result 4}
    \end{subfigure}
    \hfill
    \begin{subfigure}{0.3\textwidth}
        \includegraphics[width=\textwidth]{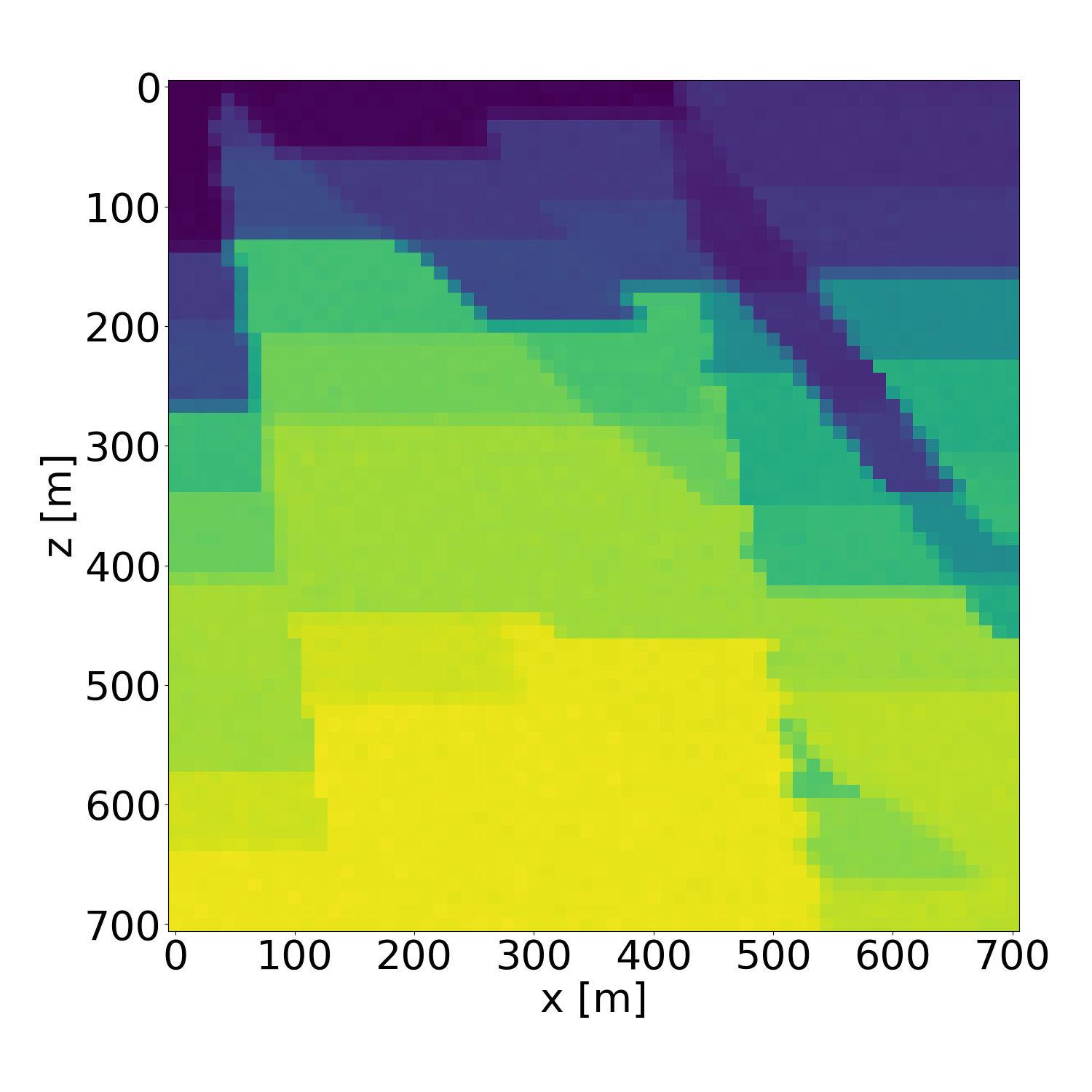}
        \caption{Probabilistic inversion result 5}
    \end{subfigure}
    \hfill
    \begin{subfigure}{0.3\textwidth}
        \includegraphics[width=\textwidth]{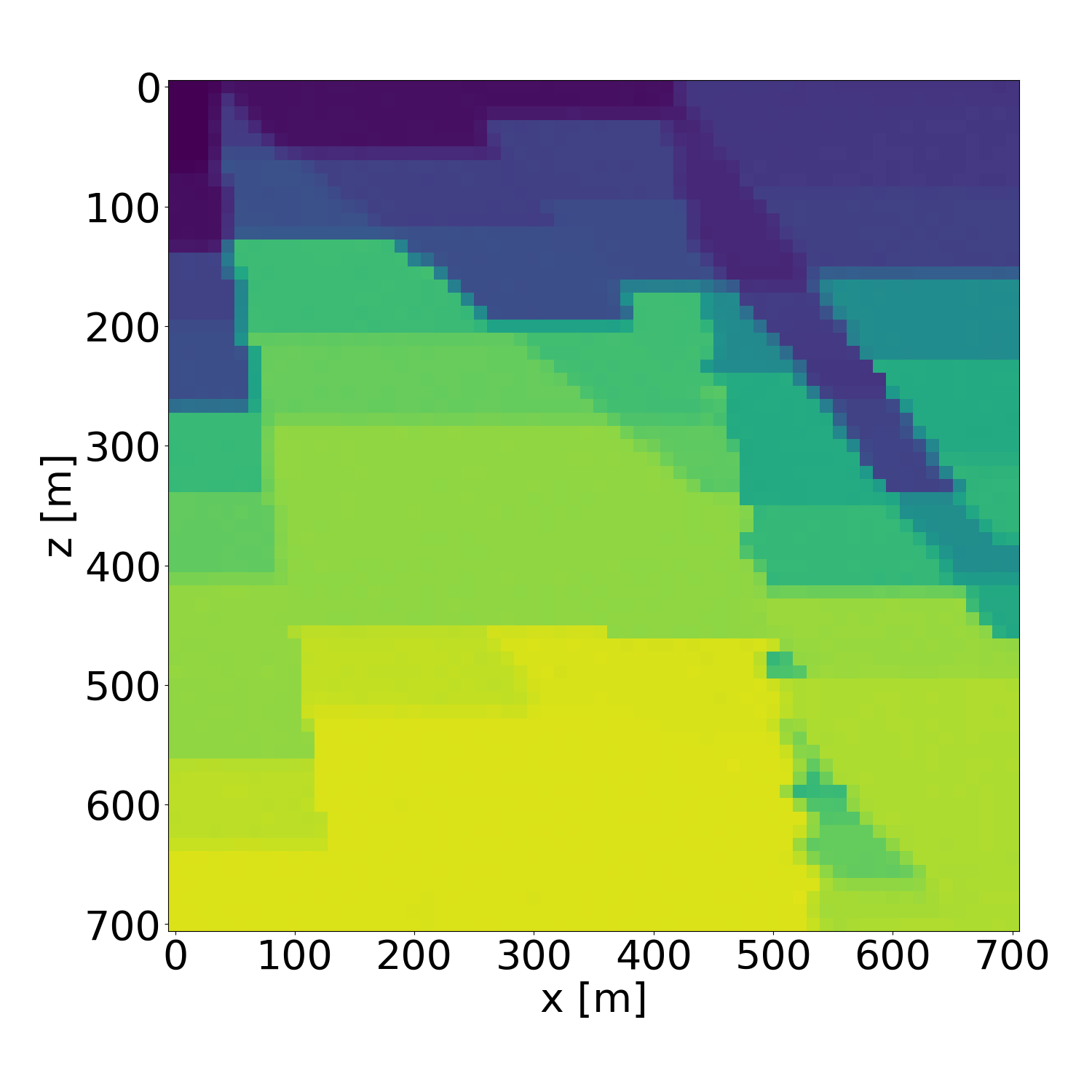}
        \caption{Probabilistic inversion result 6}
    \end{subfigure}
    \hfill
    \begin{subfigure}{0.3\textwidth}
        \includegraphics[width=\textwidth]{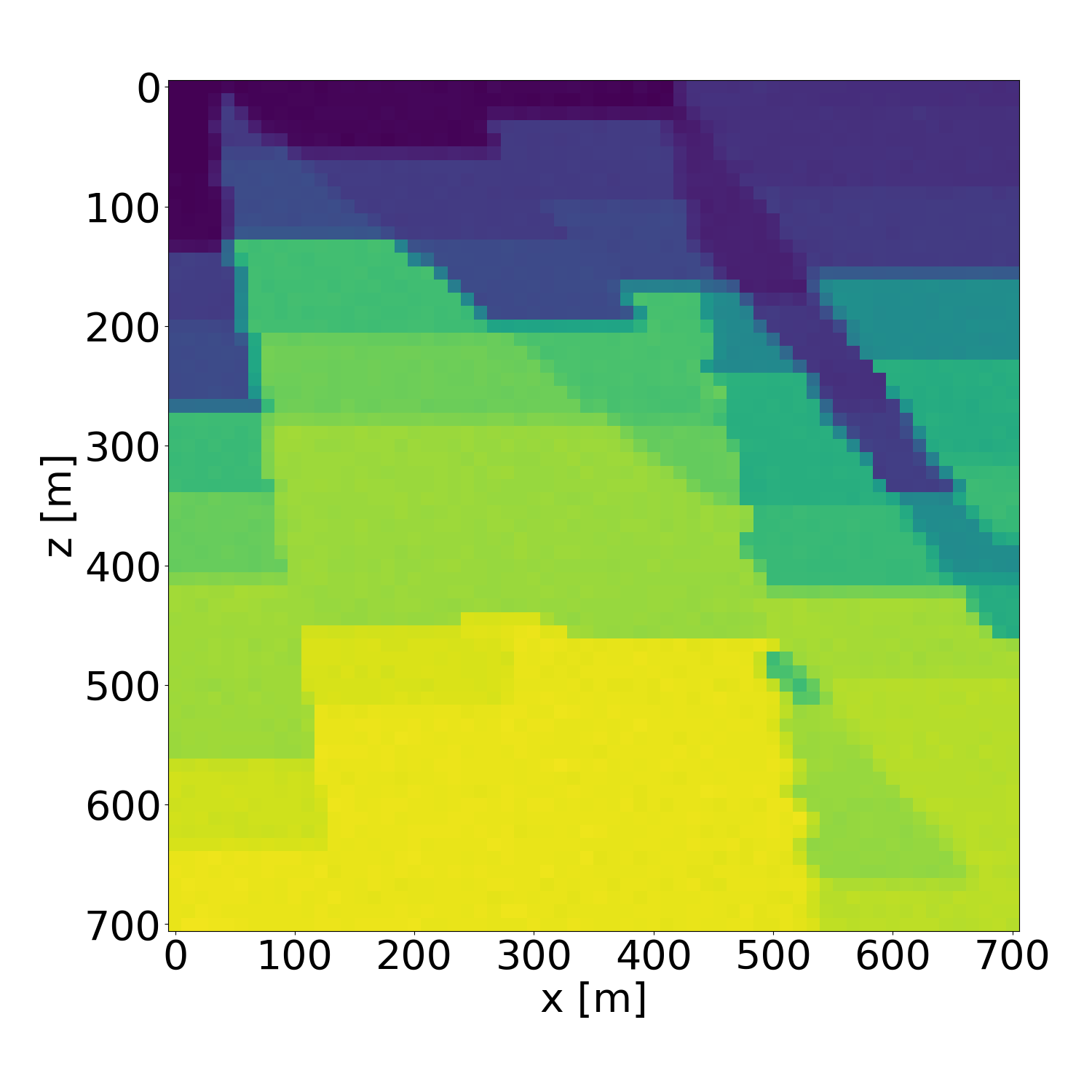}
        \caption{Probabilistic inversion result 7}
    \end{subfigure}
    \hfill
    \begin{subfigure}{0.3\textwidth}
        \includegraphics[width=\textwidth]{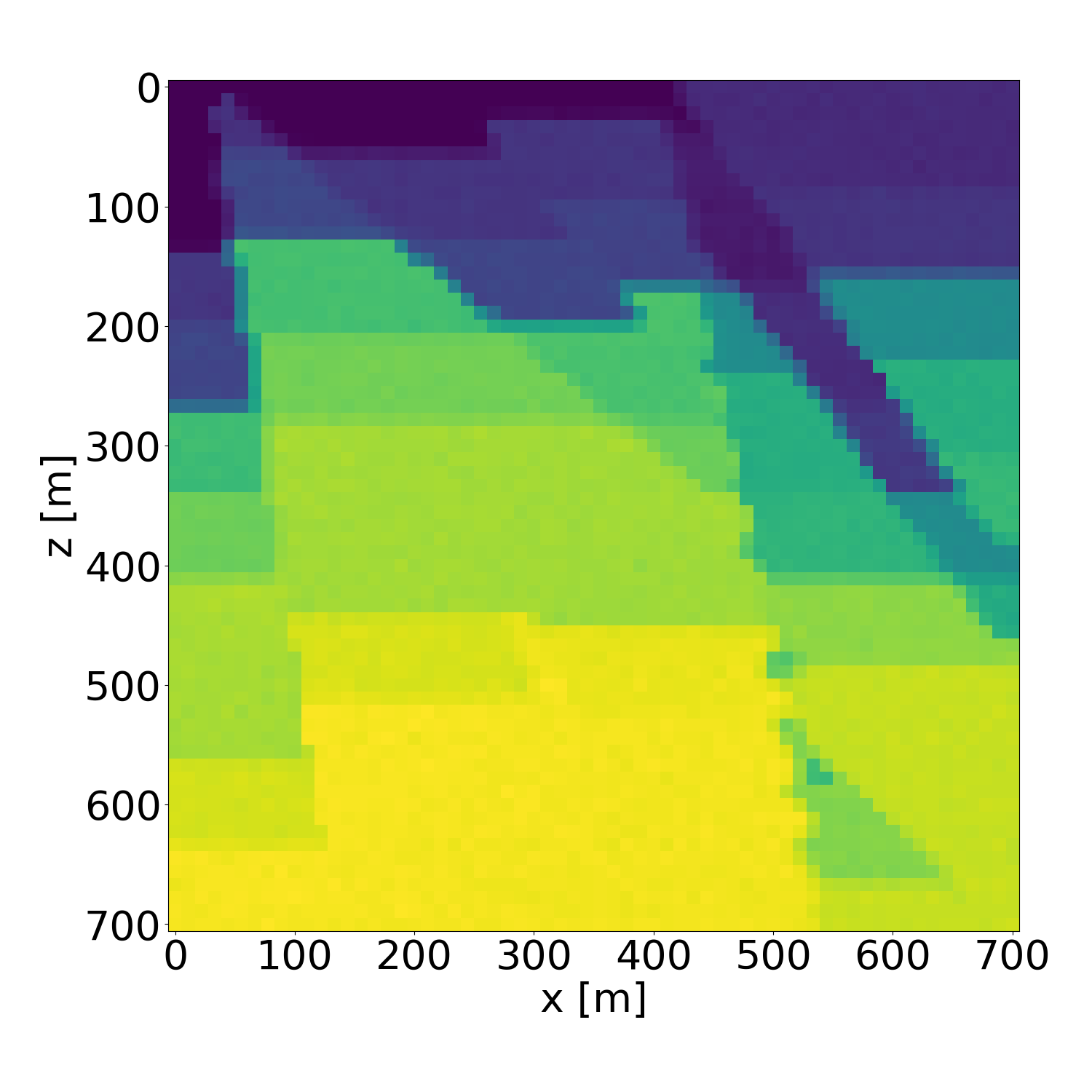}
        \caption{Probabilistic inversion result 8}
    \end{subfigure}
    \hfill
    \begin{subfigure}{0.3\textwidth}
        \includegraphics[width=\textwidth]{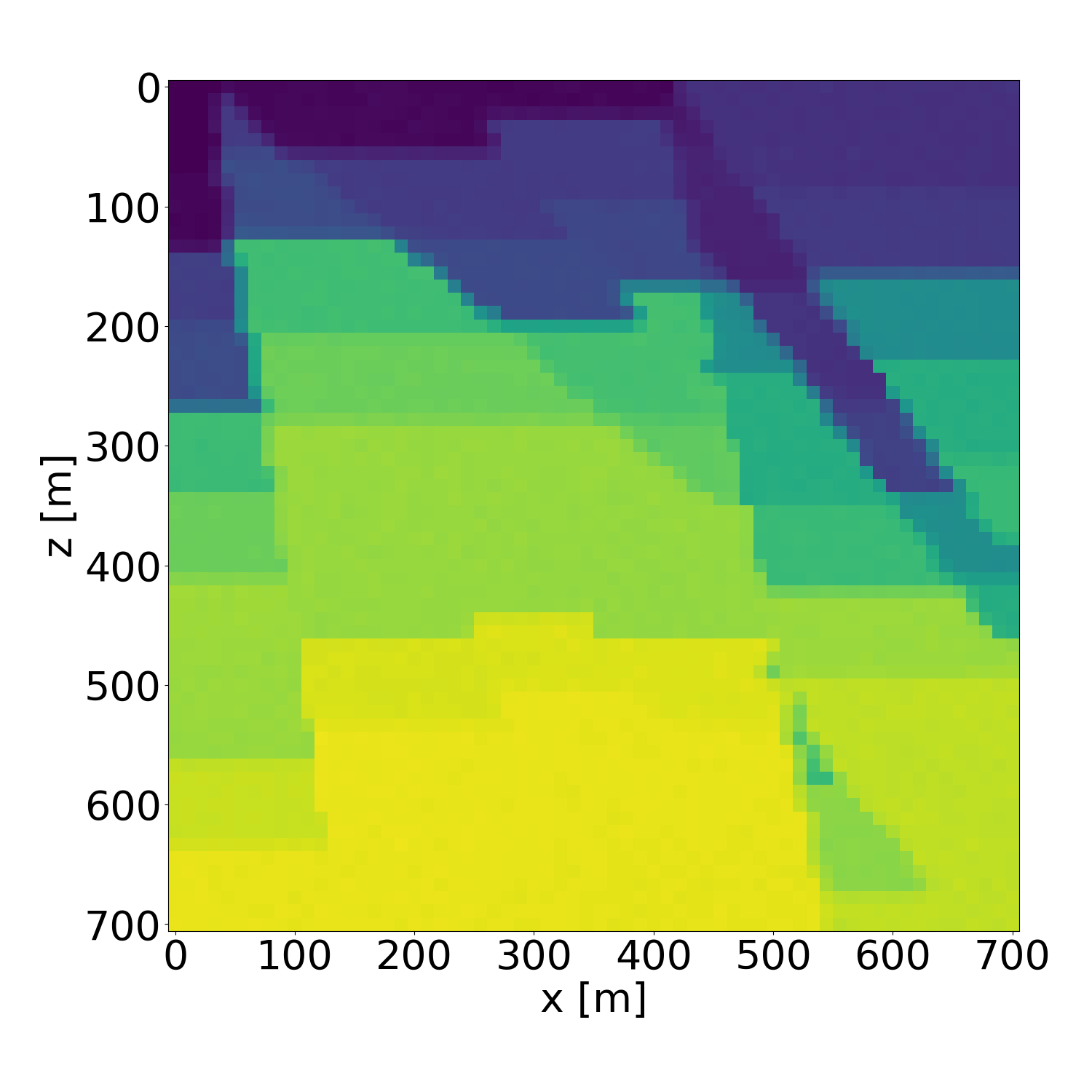}
        \caption{Probabilistic inversion result 9}
    \end{subfigure}
    \hfill
    \caption{Probabilistic inversion results of a isotropic velocity model in the validation dataset. Generated with guidance scale $w=4$.}
    \label{fig-allresult3}
\end{figure}

\begin{figure}
    \centering
    \begin{subfigure}{0.3\textwidth}
        \includegraphics[width=\textwidth]{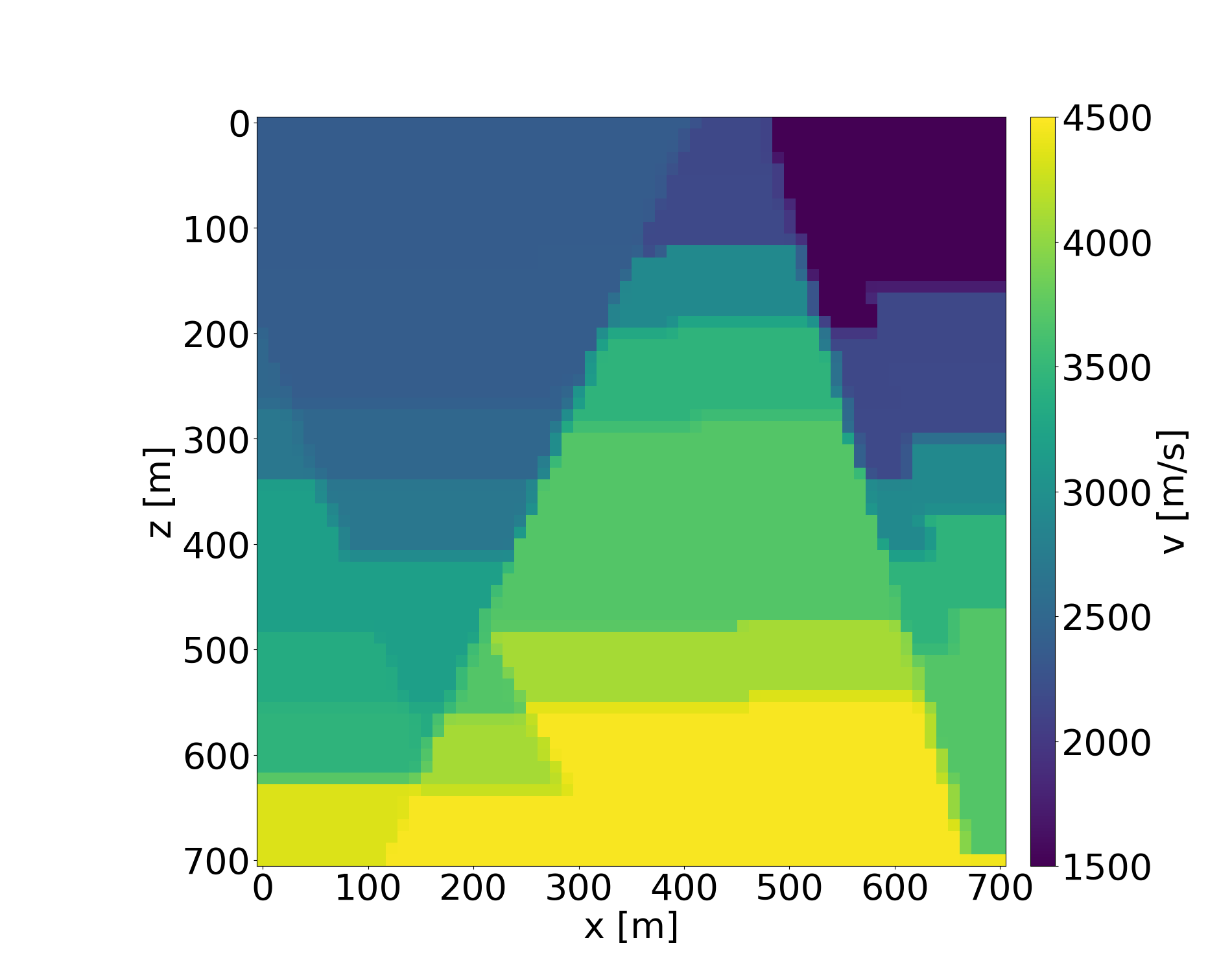}
        \caption{Ground truth (target)}
    \end{subfigure}
    \hfill
    \begin{subfigure}{0.3\textwidth}
        \includegraphics[width=\textwidth]{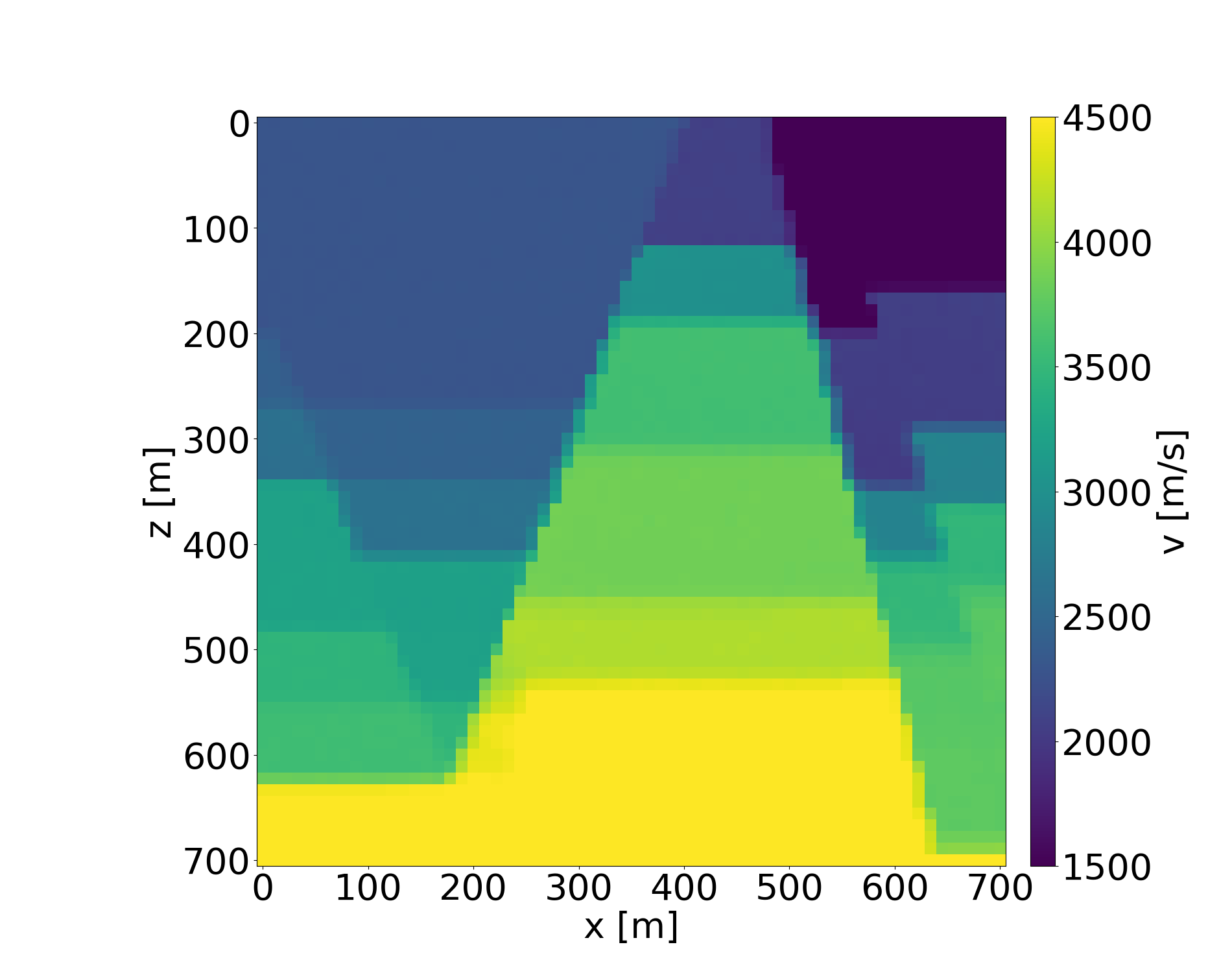}
        \caption{Average inversion result}
    \end{subfigure}
    \hfill
    \begin{subfigure}{0.3\textwidth}
        \includegraphics[width=\textwidth]{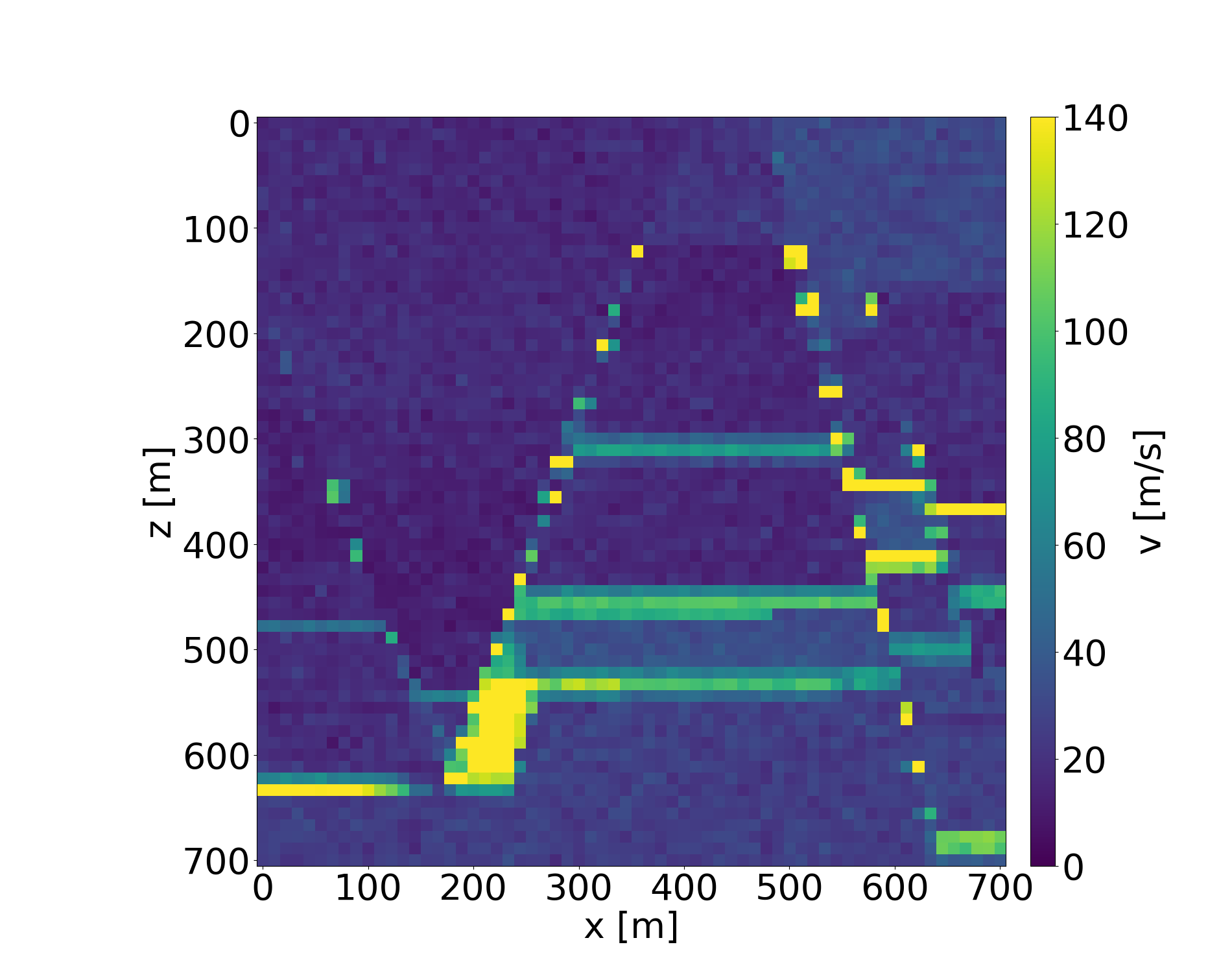}
        \caption{Standard deviation}
    \end{subfigure}
    \hfill
    \begin{subfigure}{0.3\textwidth}
        \includegraphics[width=\textwidth]{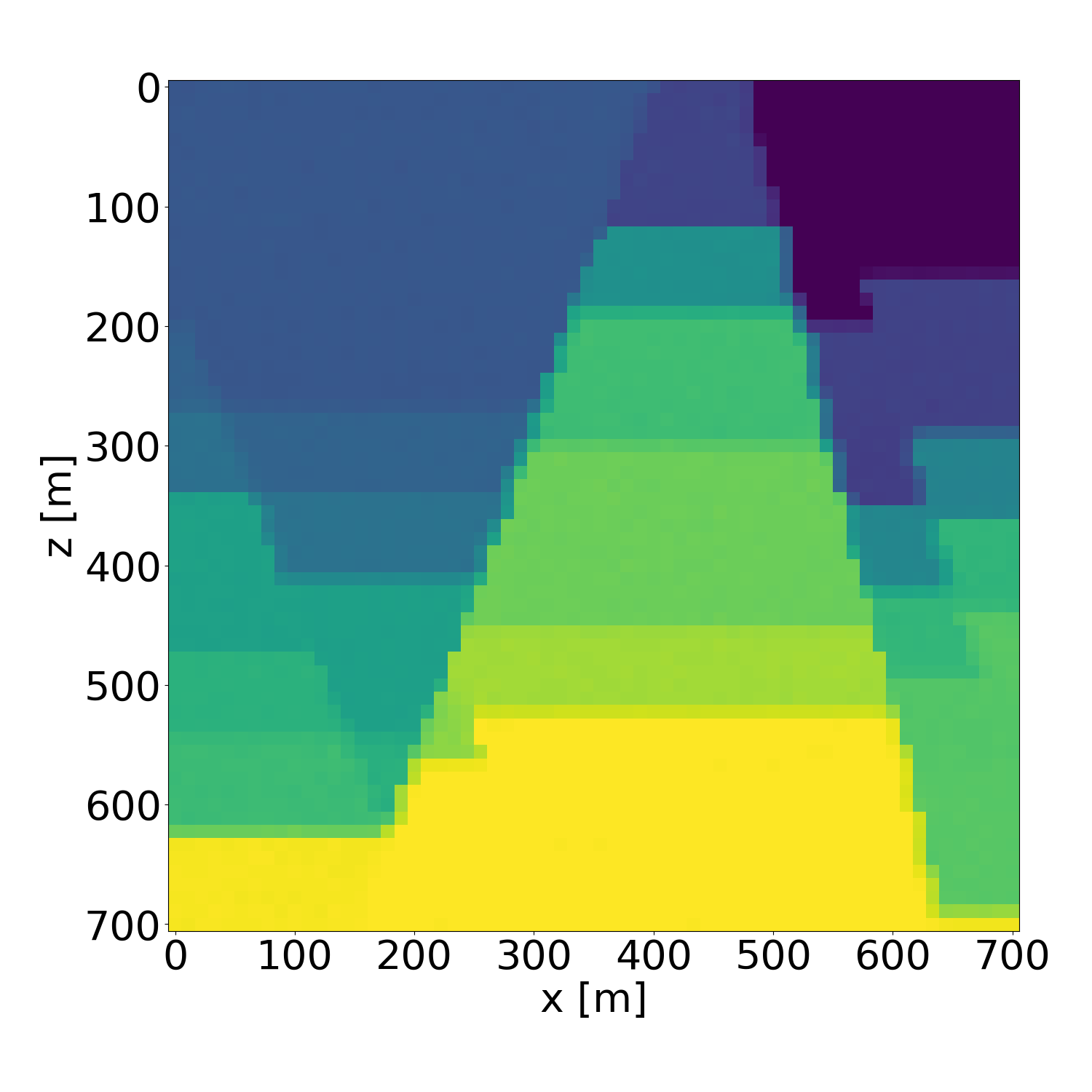}
        \caption{Probabilistic inversion result 1}
    \end{subfigure}
    \hfill
    \begin{subfigure}{0.3\textwidth}
        \includegraphics[width=\textwidth]{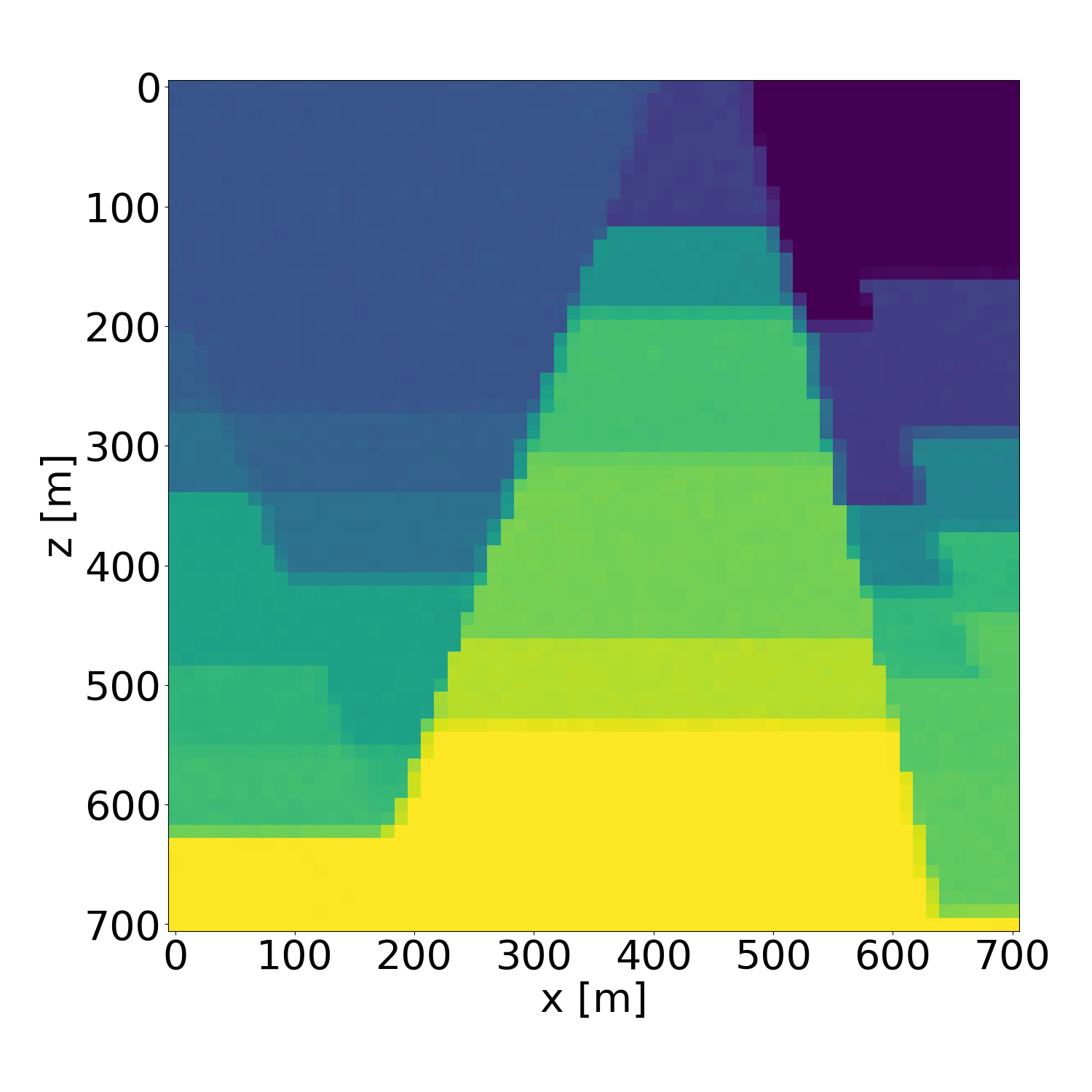}
        \caption{Probabilistic inversion result 2}
    \end{subfigure}
    \hfill
    \begin{subfigure}{0.3\textwidth}
        \includegraphics[width=\textwidth]{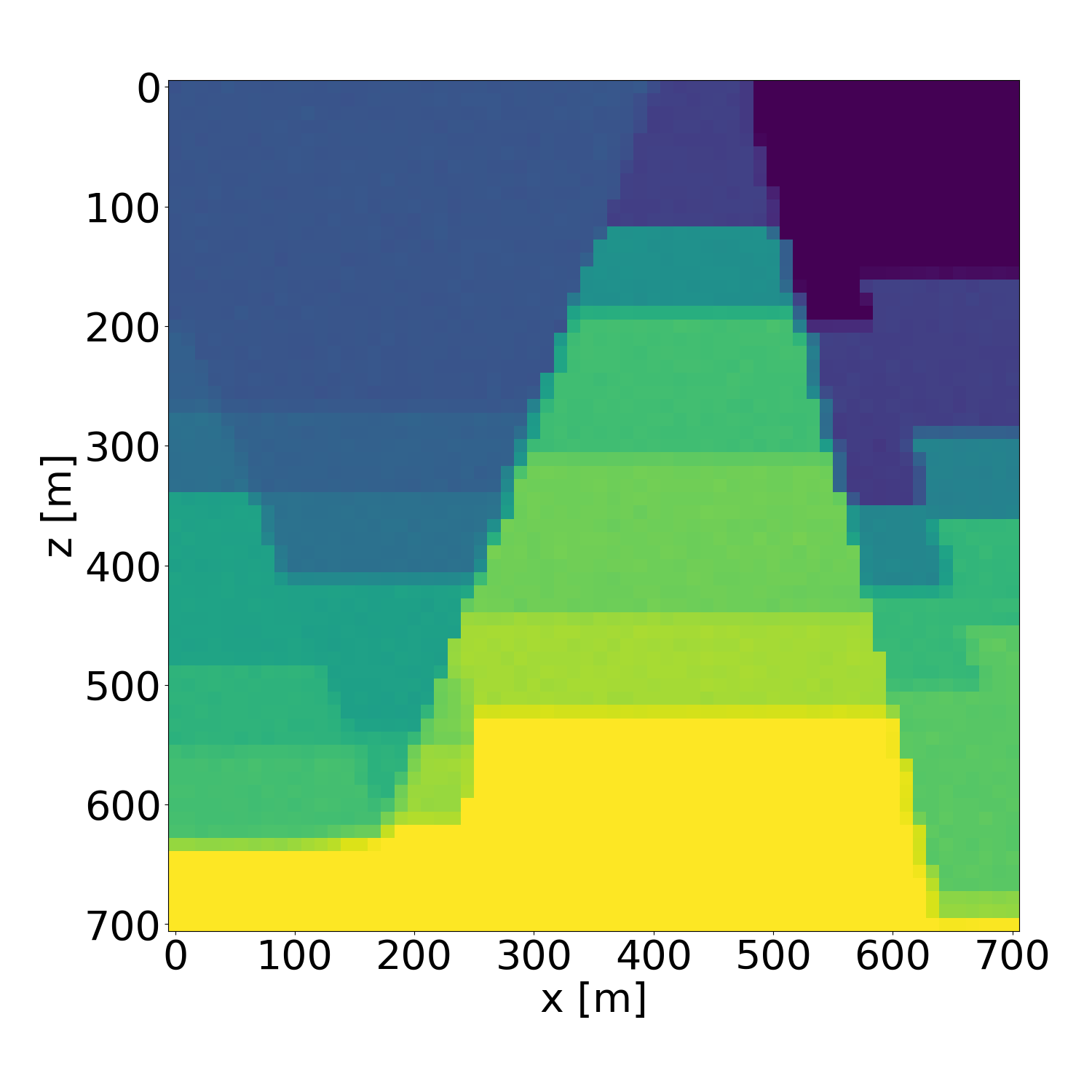}
        \caption{Probabilistic inversion result 3}
    \end{subfigure}
    \hfill
    \begin{subfigure}{0.3\textwidth}
        \includegraphics[width=\textwidth]{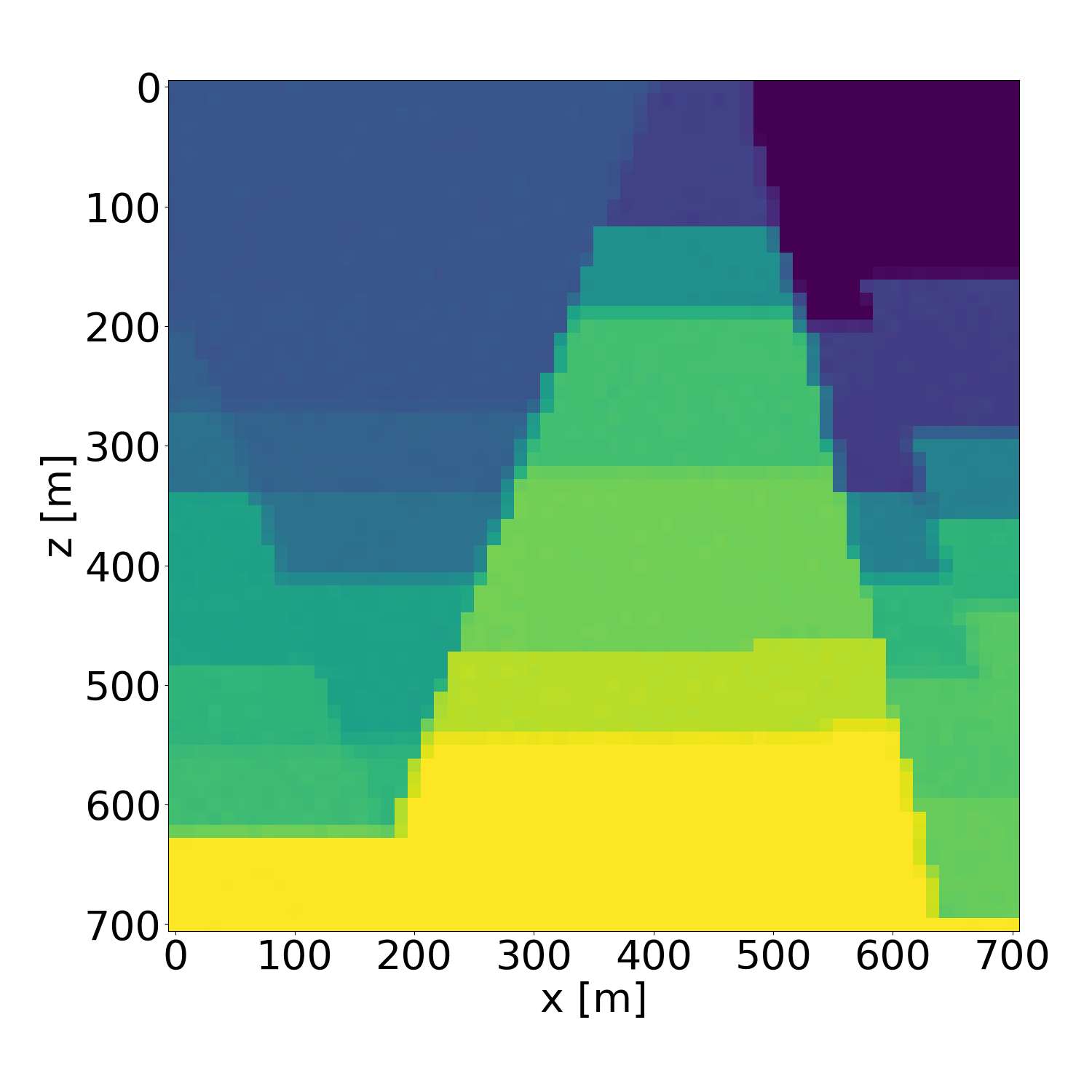}
        \caption{Probabilistic inversion result 4}
    \end{subfigure}
    \hfill
    \begin{subfigure}{0.3\textwidth}
        \includegraphics[width=\textwidth]{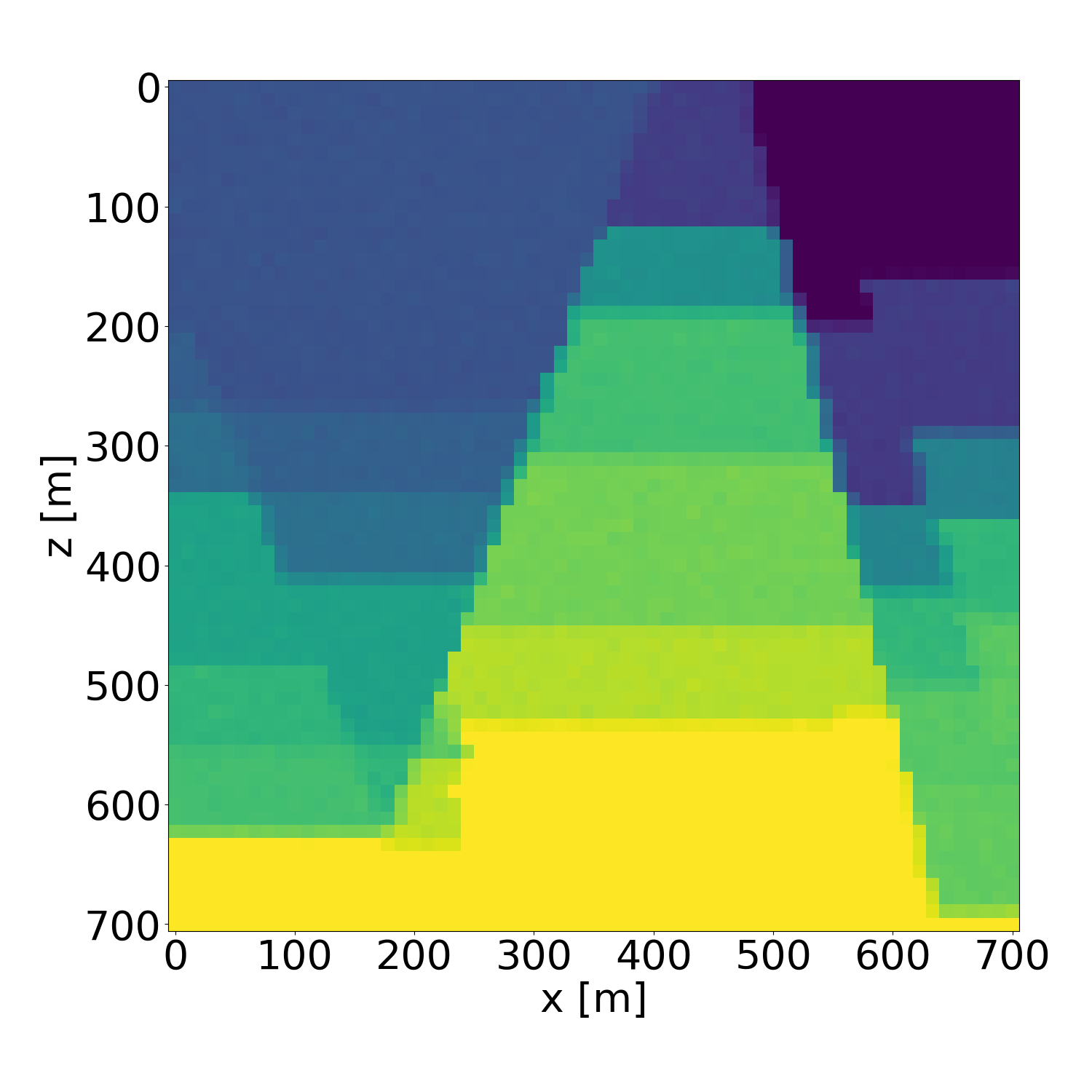}
        \caption{Probabilistic inversion result 5}
    \end{subfigure}
    \hfill
    \begin{subfigure}{0.3\textwidth}
        \includegraphics[width=\textwidth]{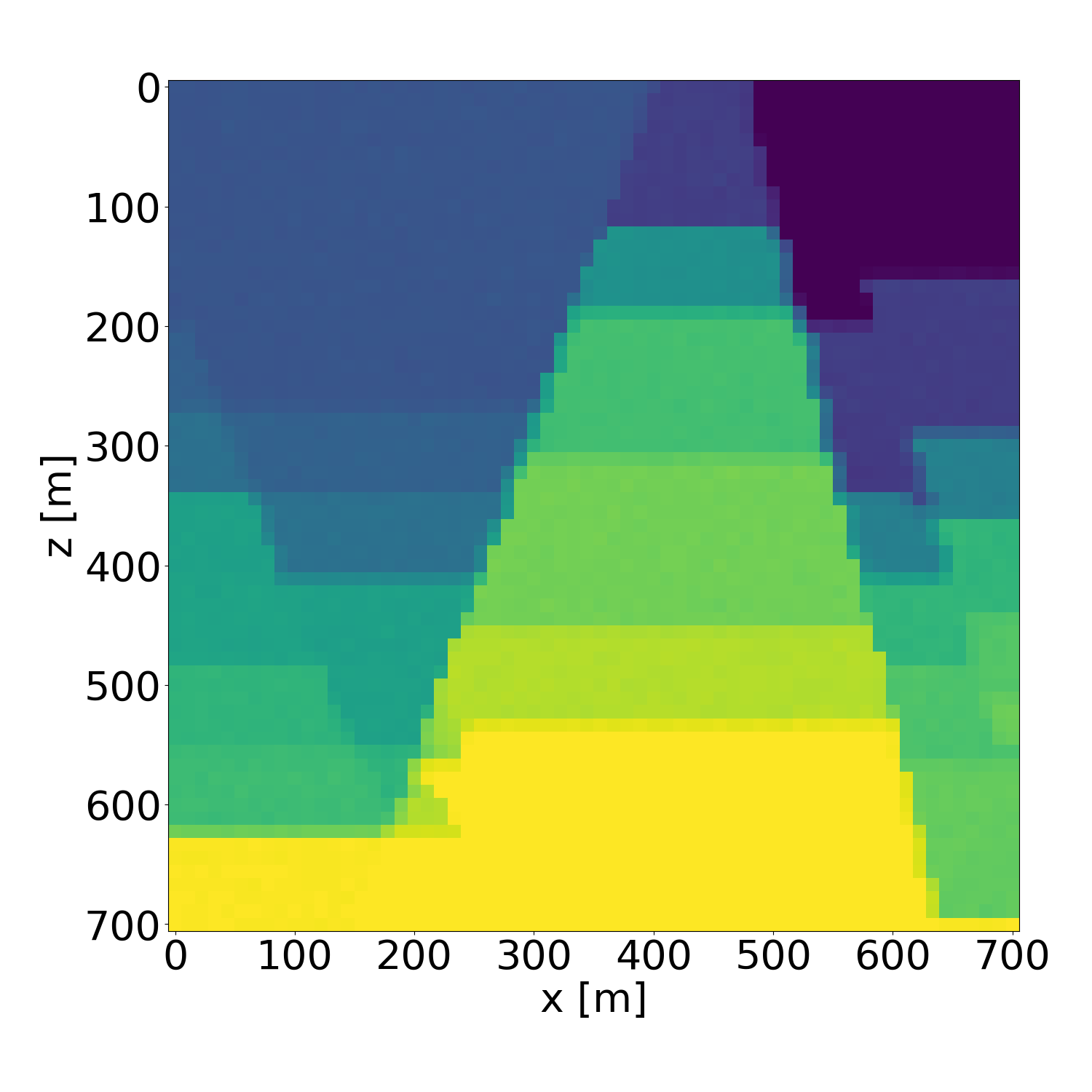}
        \caption{Probabilistic inversion result 6}
    \end{subfigure}
    \hfill
    \begin{subfigure}{0.3\textwidth}
        \includegraphics[width=\textwidth]{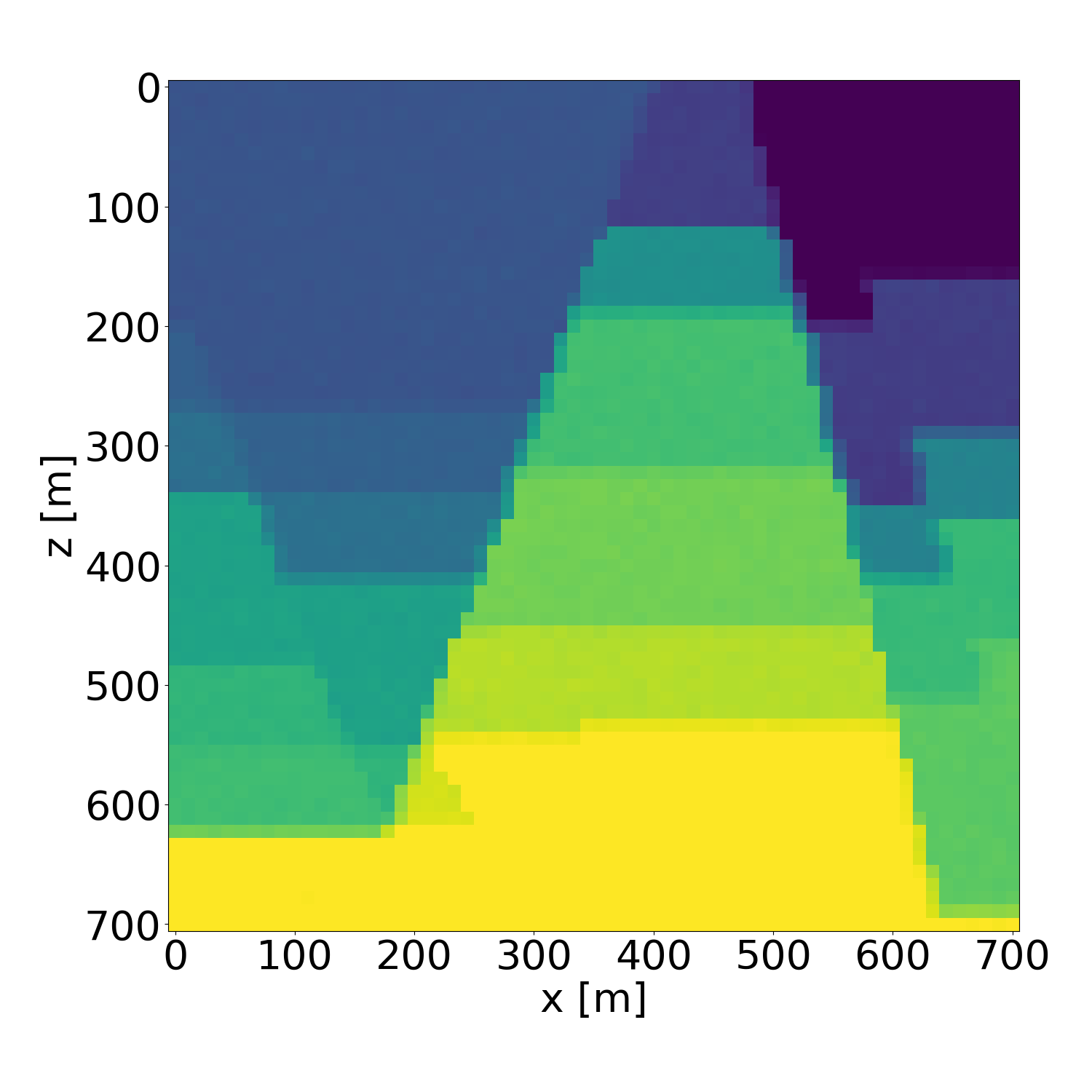}
        \caption{Probabilistic inversion result 7}
    \end{subfigure}
    \hfill
    \begin{subfigure}{0.3\textwidth}
        \includegraphics[width=\textwidth]{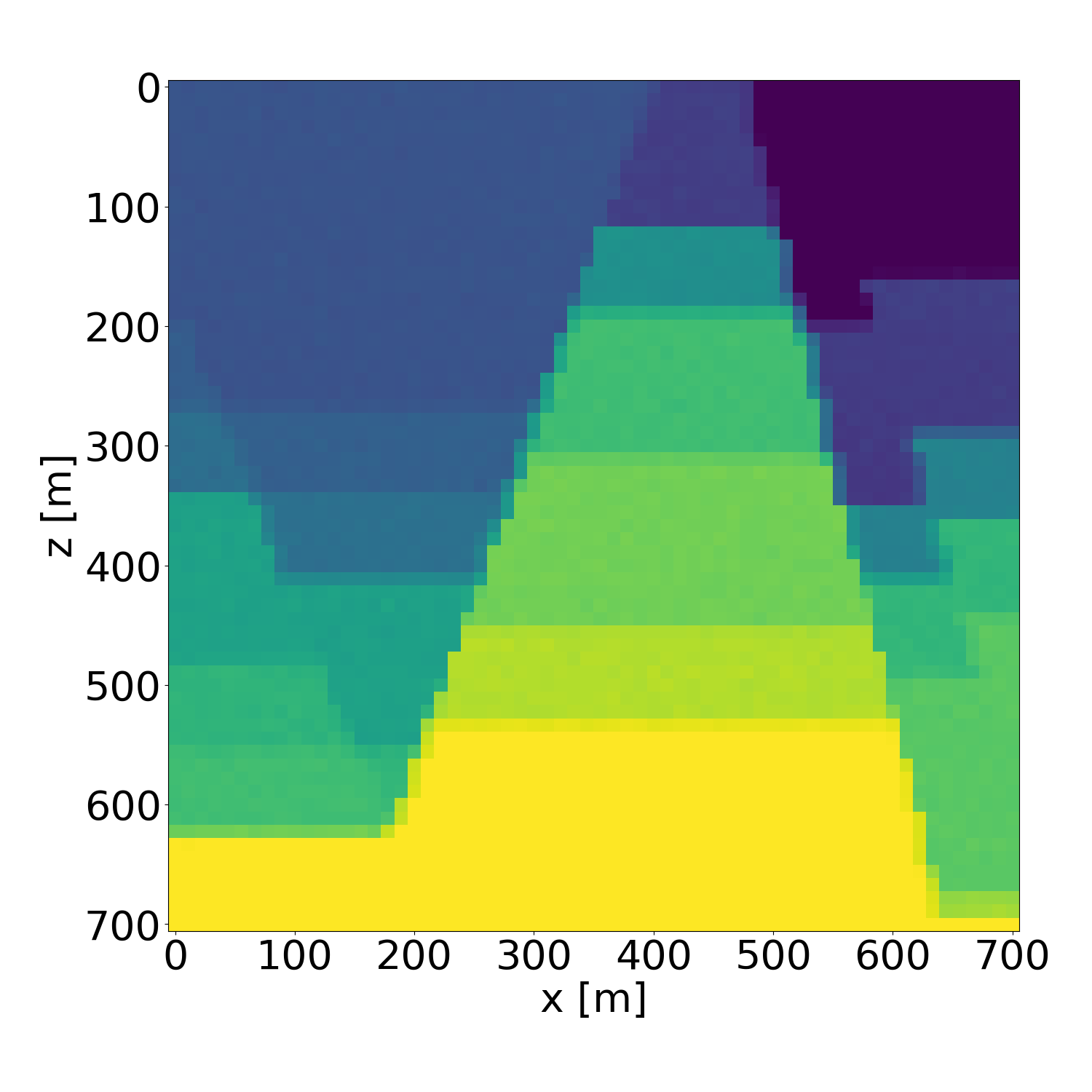}
        \caption{Probabilistic inversion result 8}
    \end{subfigure}
    \hfill
    \begin{subfigure}{0.3\textwidth}
        \includegraphics[width=\textwidth]{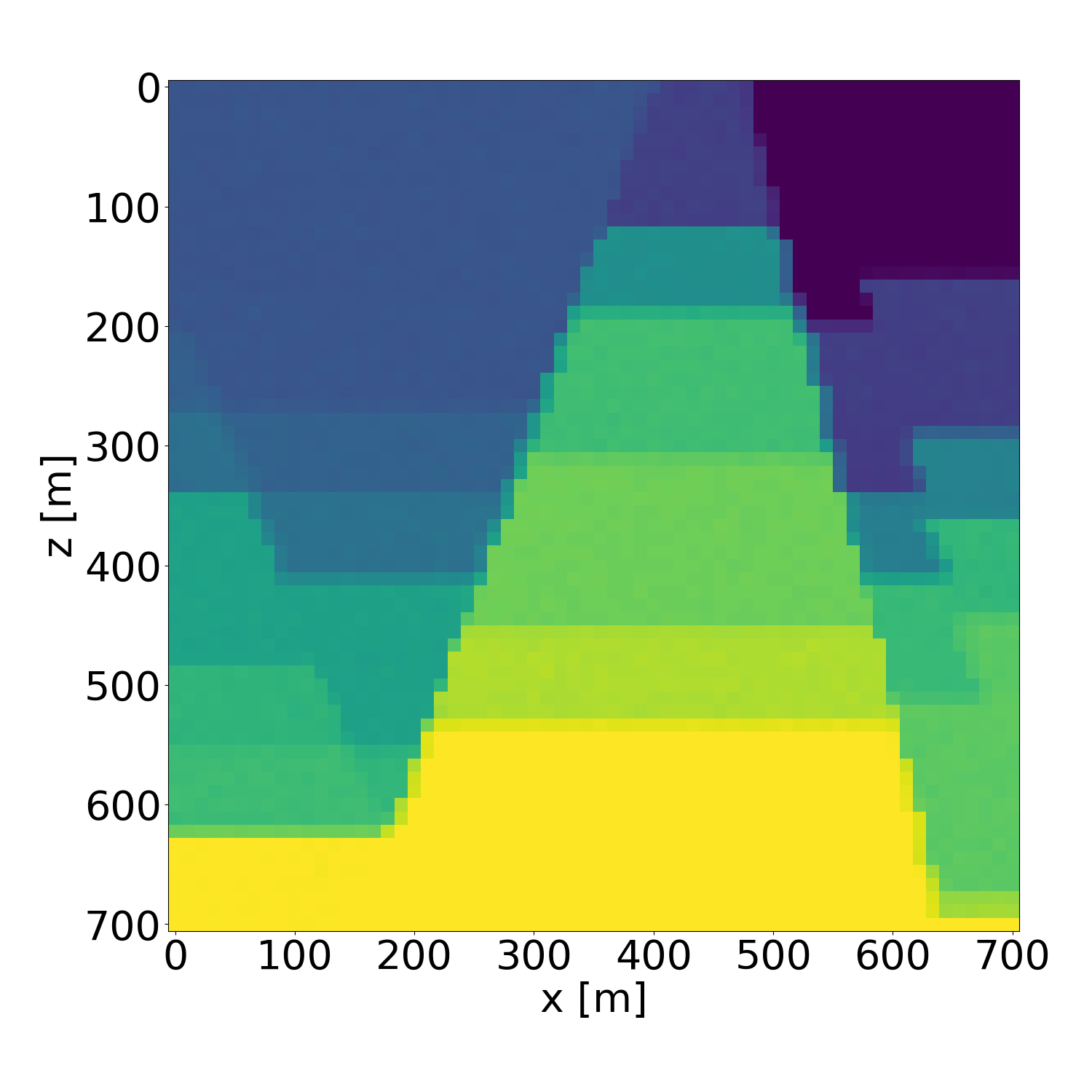}
        \caption{Probabilistic inversion result 9}
    \end{subfigure}
    \hfill
    \caption{Probabilistic inversion results of a isotropic velocity model in the validation dataset. Generated with guidance scale $w=4$.}
    \label{fig-allresult4}
\end{figure}

\begin{figure}
    \centering
    \begin{subfigure}{0.3\textwidth}
        \includegraphics[width=\textwidth]{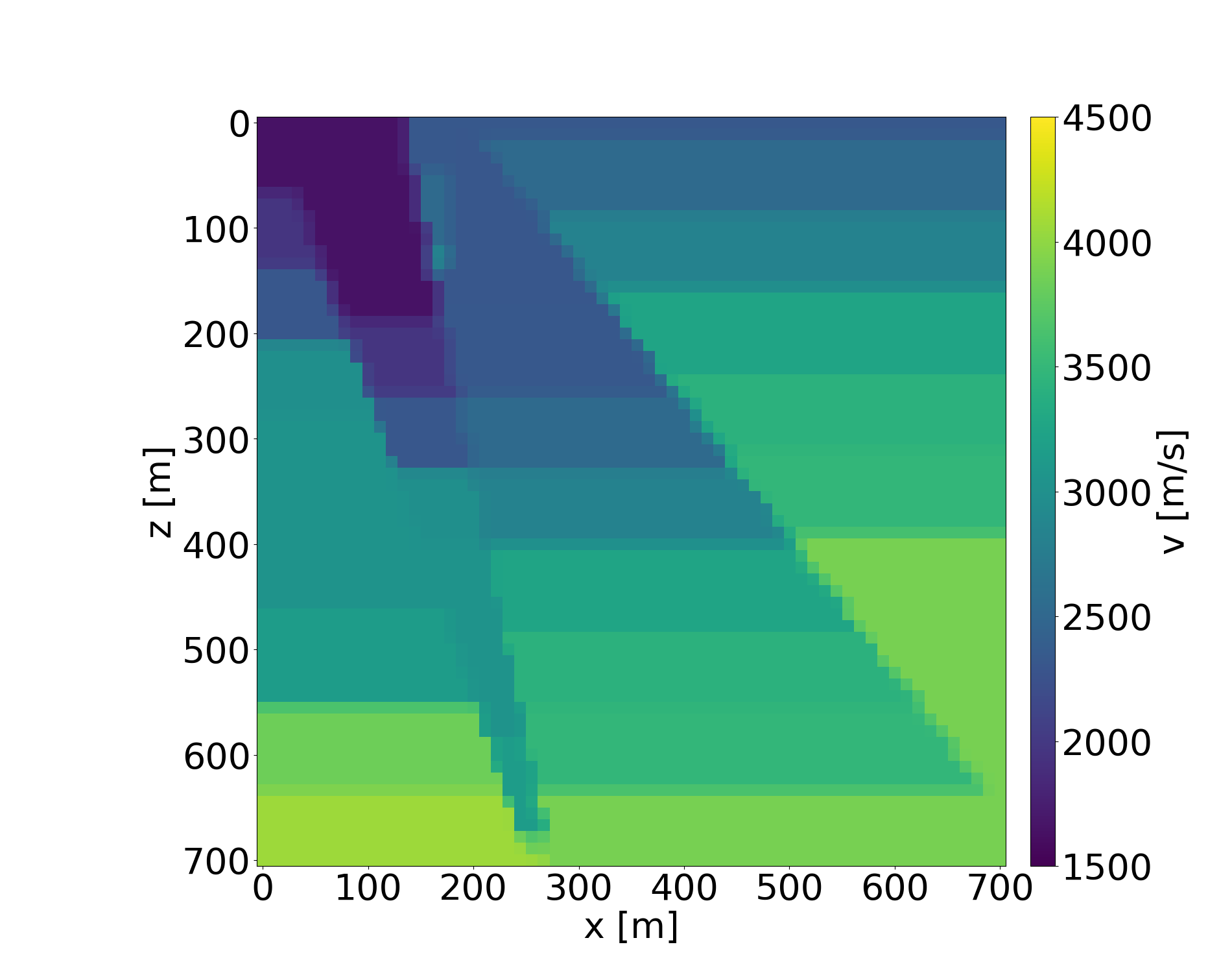}
        \caption{Ground truth (target)}
    \end{subfigure}
    \hfill
    \begin{subfigure}{0.3\textwidth}
        \includegraphics[width=\textwidth]{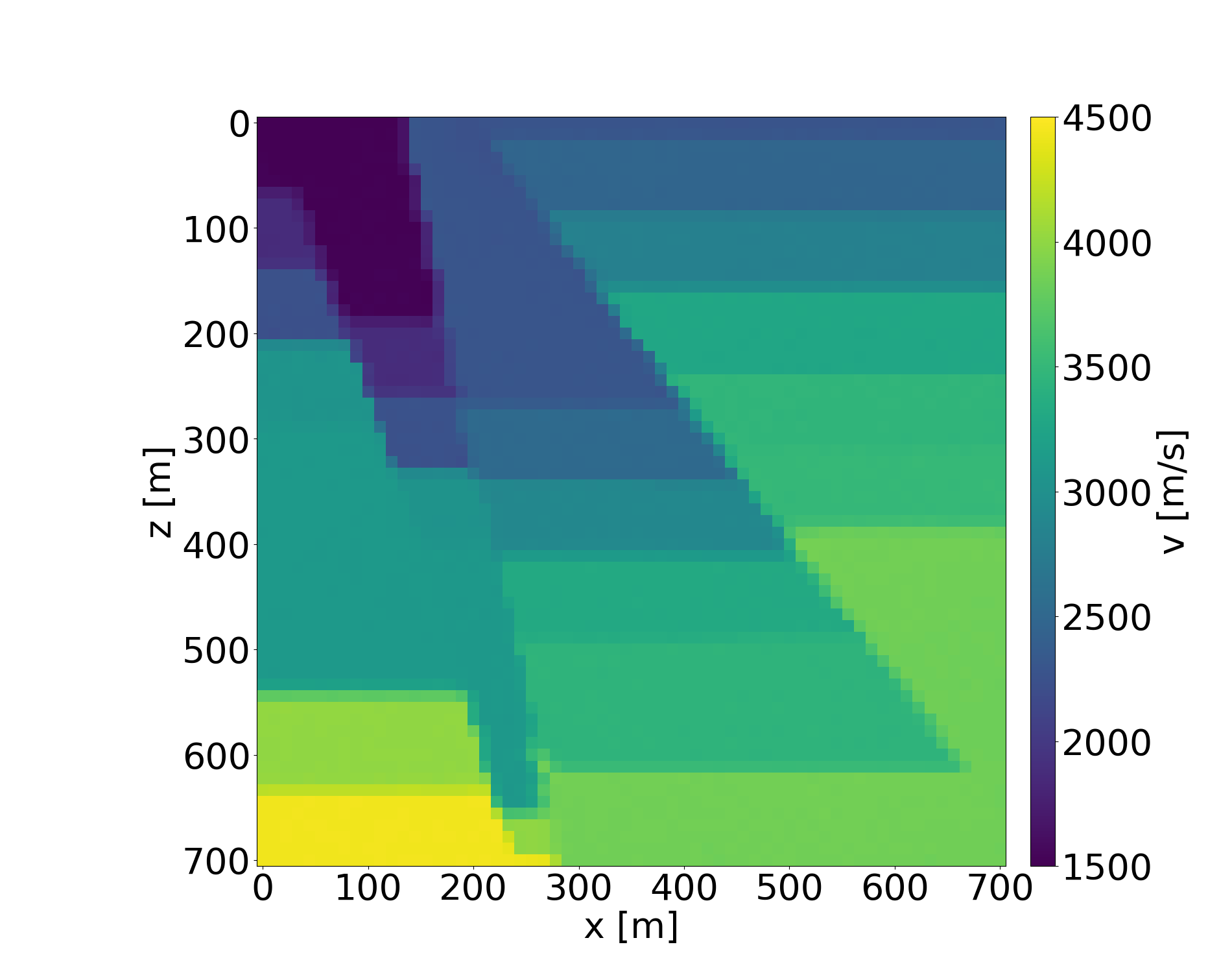}
        \caption{Average inversion result}
    \end{subfigure}
    \hfill
    \begin{subfigure}{0.3\textwidth}
        \includegraphics[width=\textwidth]{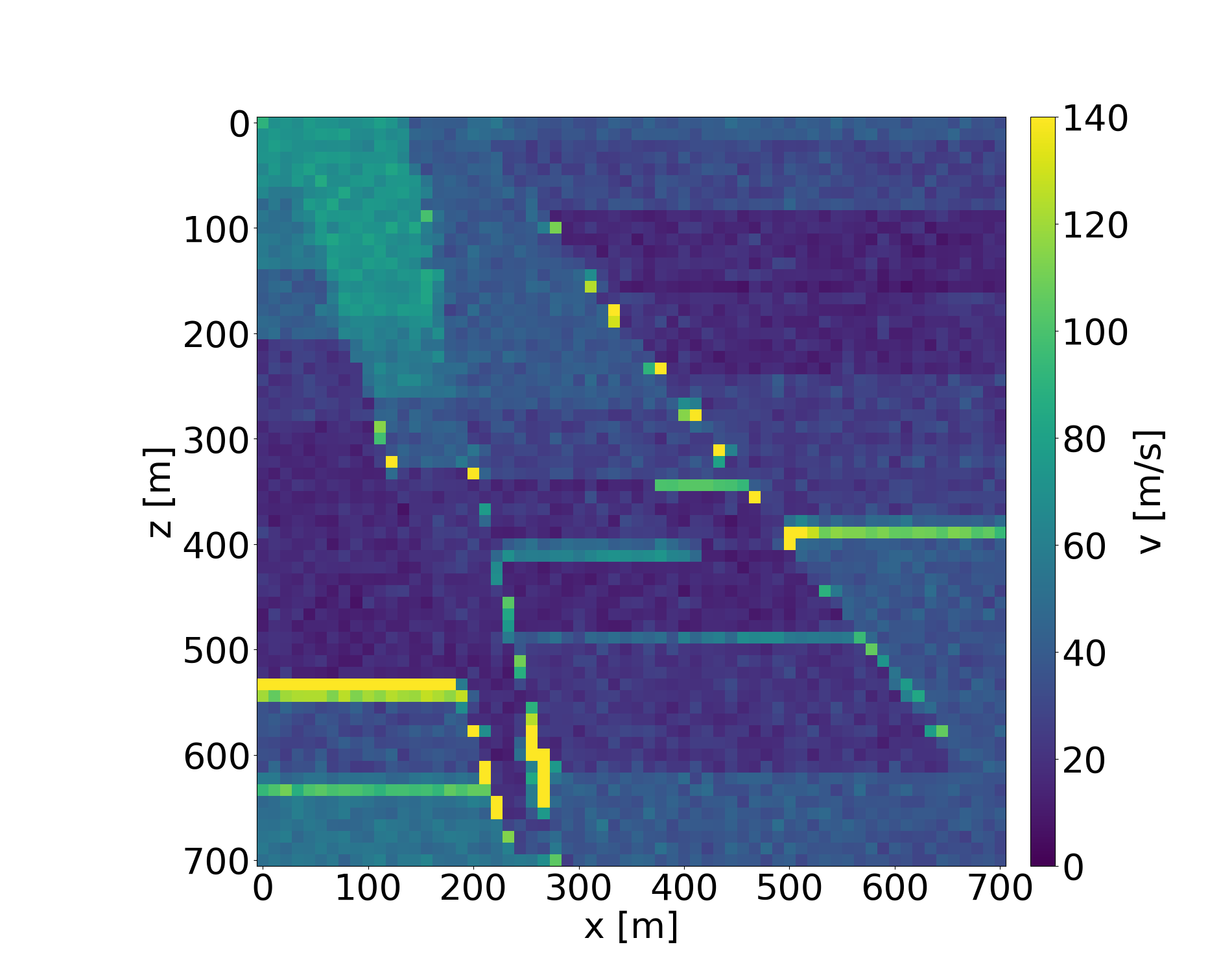}
        \caption{Standard deviation}
    \end{subfigure}
    \hfill
    \begin{subfigure}{0.3\textwidth}
        \includegraphics[width=\textwidth]{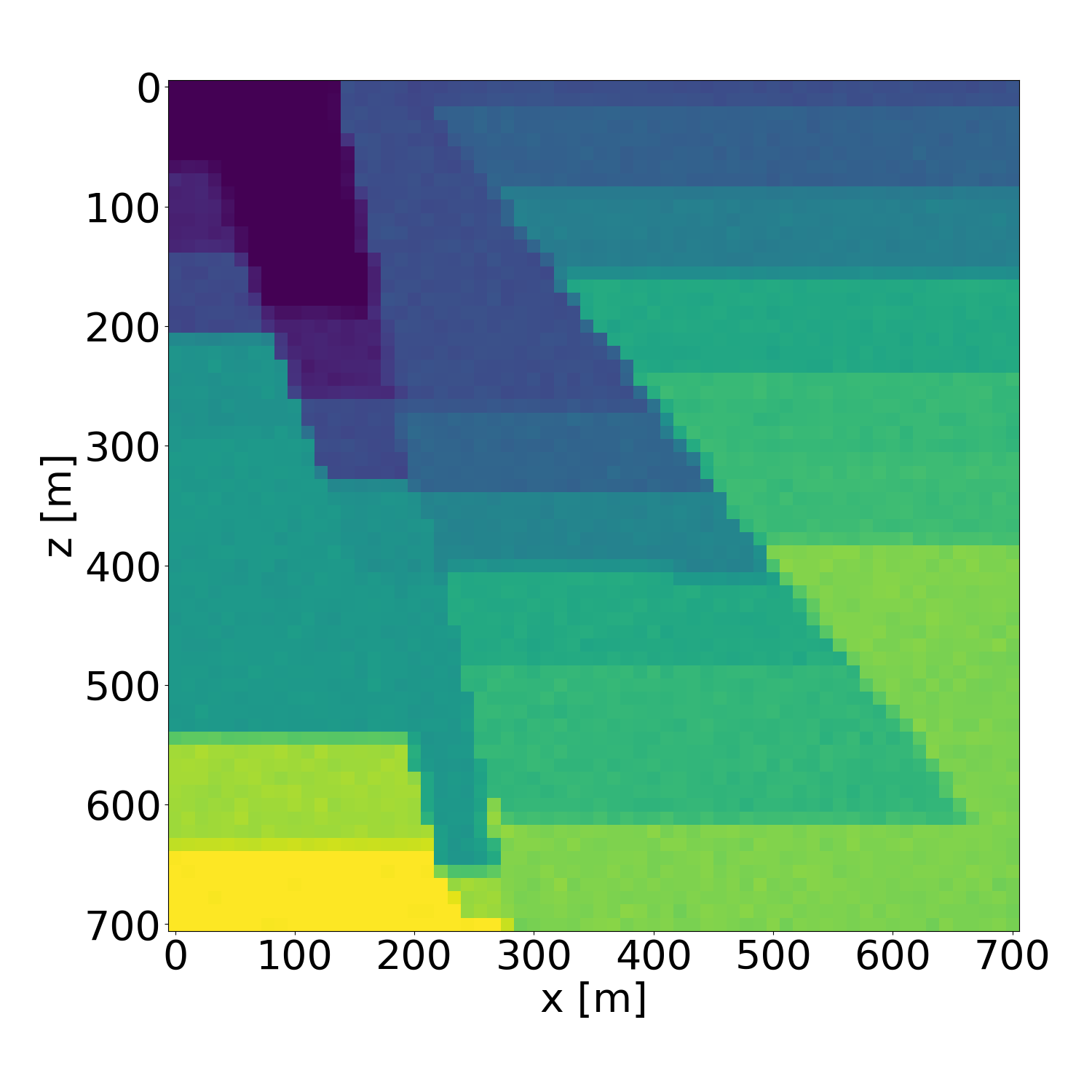}
        \caption{Probabilistic inversion result 1}
    \end{subfigure}
    \hfill
    \begin{subfigure}{0.3\textwidth}
        \includegraphics[width=\textwidth]{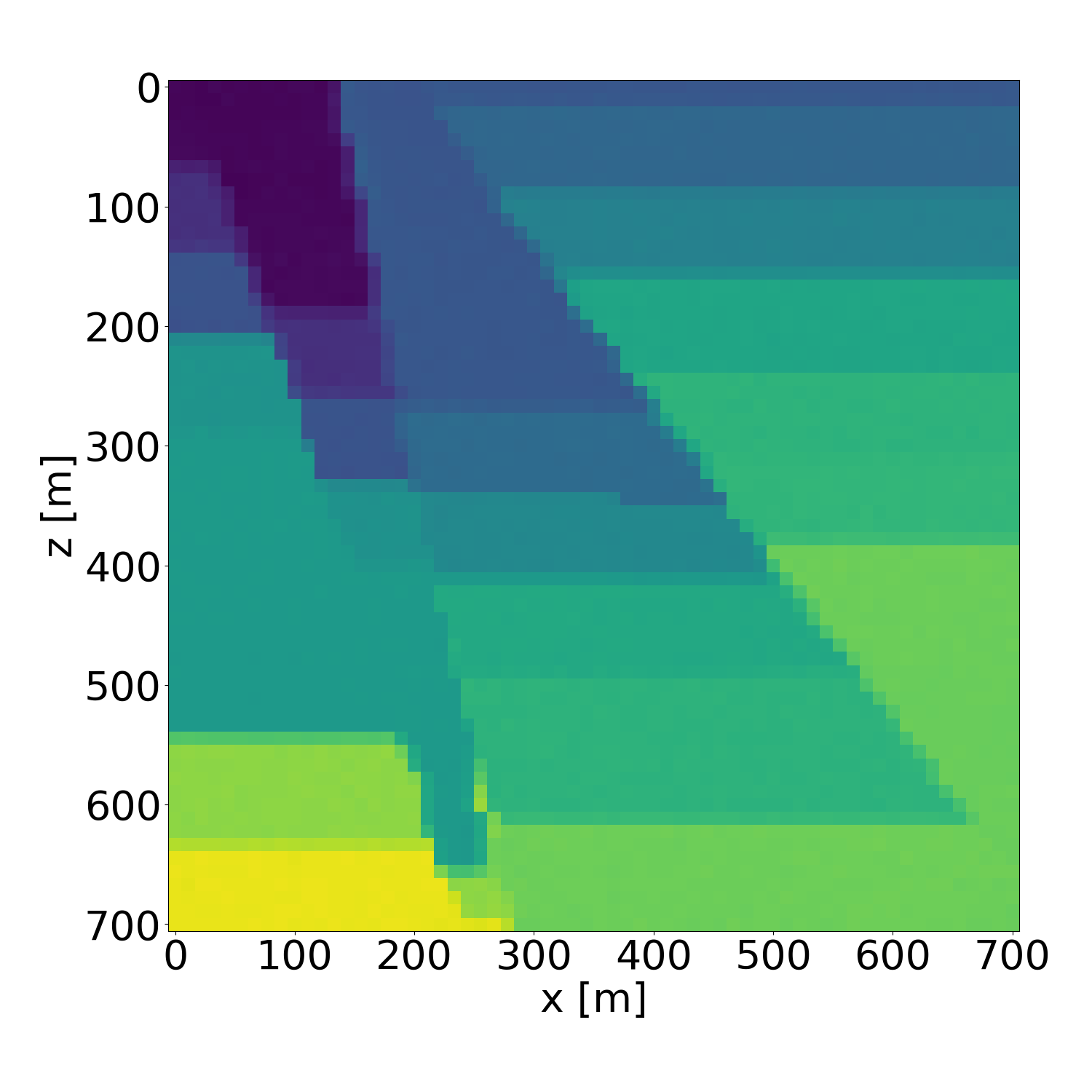}
        \caption{Probabilistic inversion result 2}
    \end{subfigure}
    \hfill
    \begin{subfigure}{0.3\textwidth}
        \includegraphics[width=\textwidth]{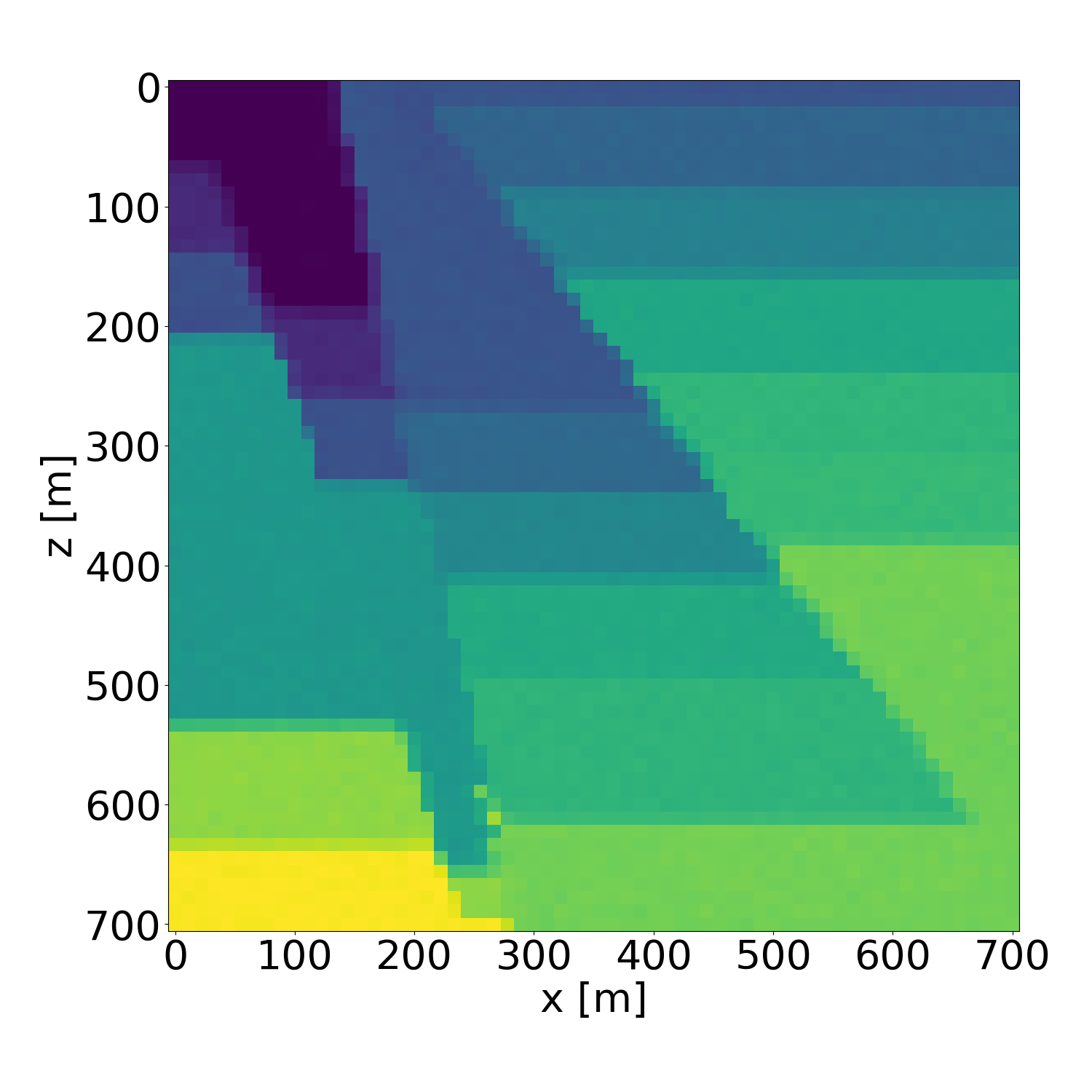}
        \caption{Probabilistic inversion result 3}
    \end{subfigure}
    \hfill
    \begin{subfigure}{0.3\textwidth}
        \includegraphics[width=\textwidth]{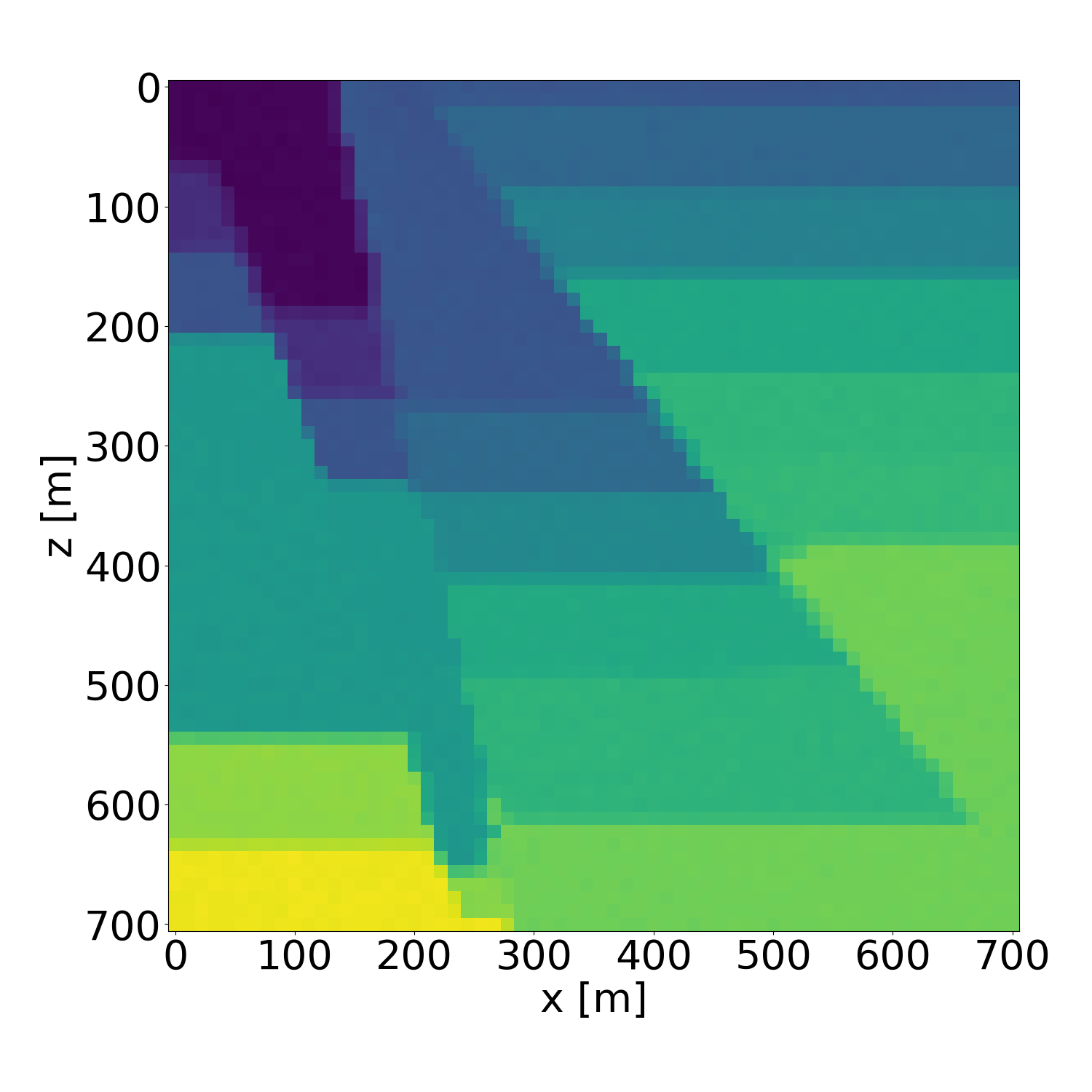}
        \caption{Probabilistic inversion result 4}
    \end{subfigure}
    \hfill
    \begin{subfigure}{0.3\textwidth}
        \includegraphics[width=\textwidth]{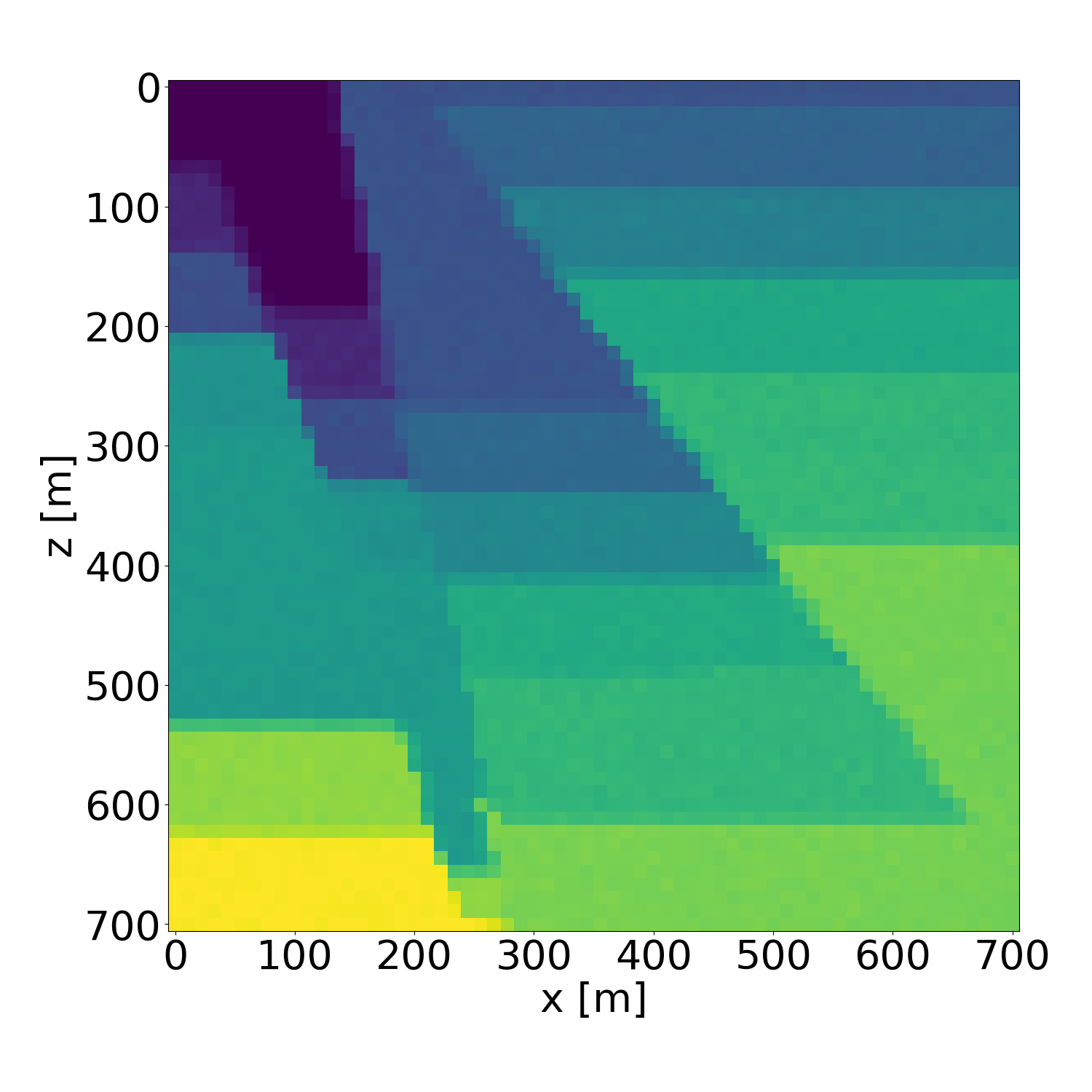}
        \caption{Probabilistic inversion result 5}
    \end{subfigure}
    \hfill
    \begin{subfigure}{0.3\textwidth}
        \includegraphics[width=\textwidth]{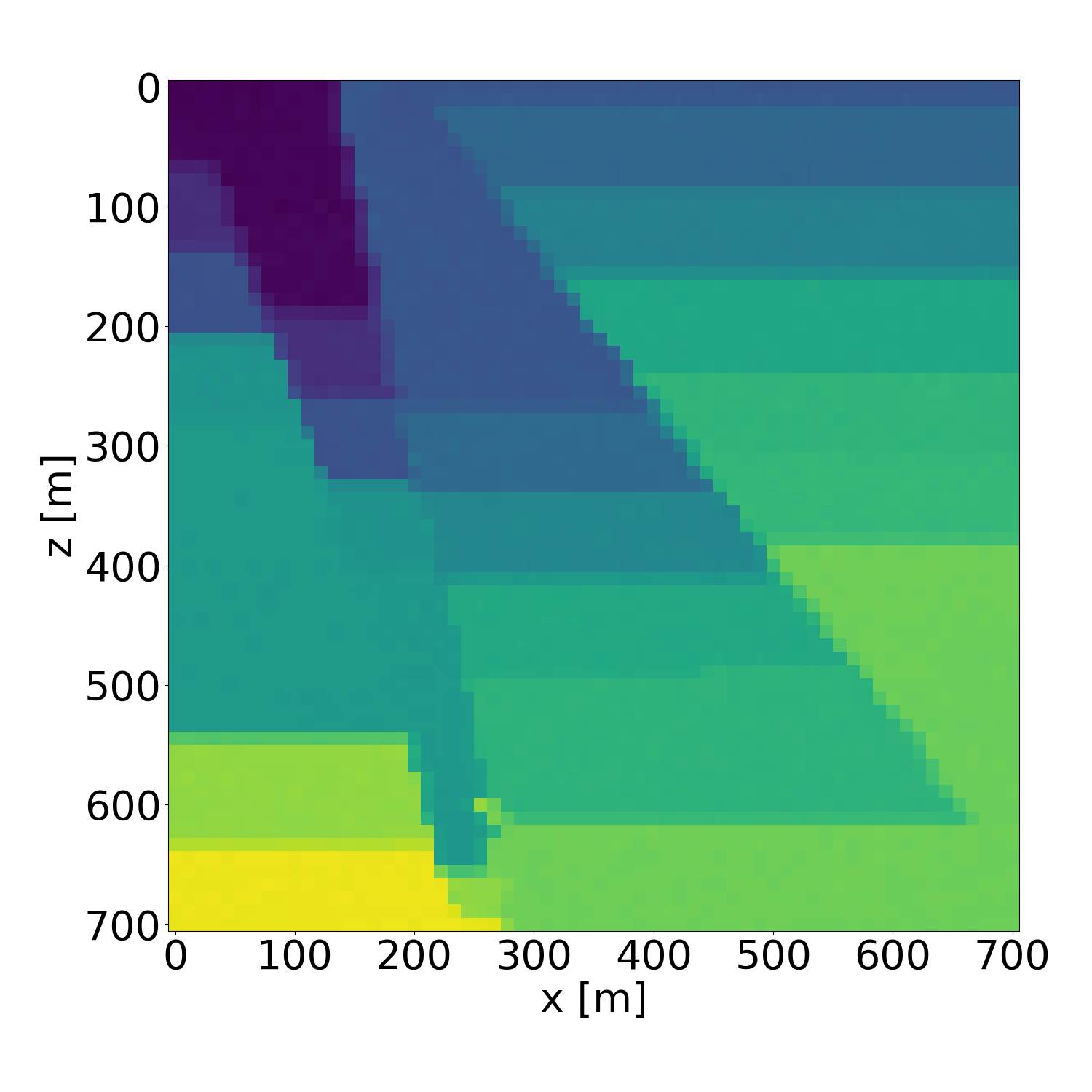}
        \caption{Probabilistic inversion result 6}
    \end{subfigure}
    \hfill
    \begin{subfigure}{0.3\textwidth}
        \includegraphics[width=\textwidth]{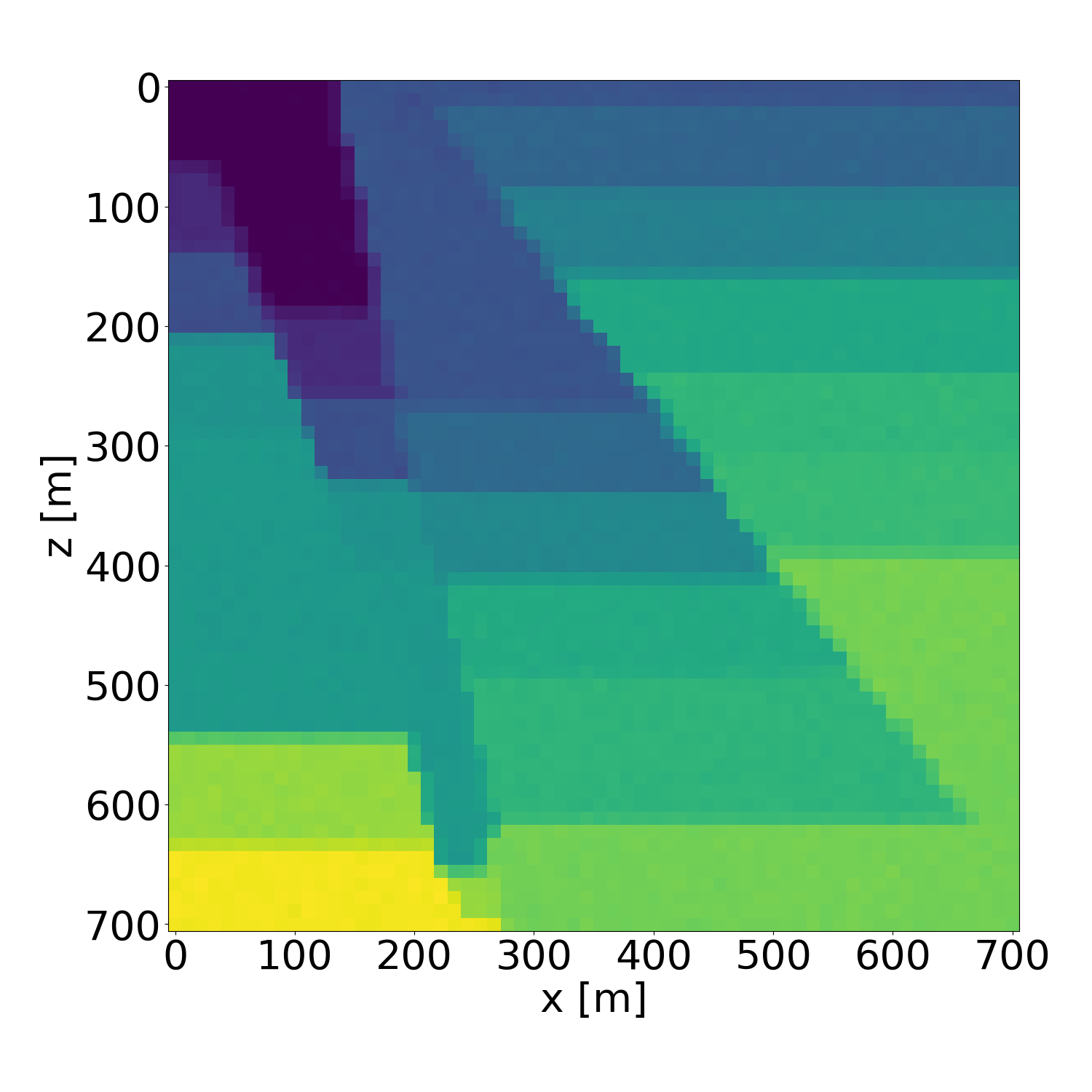}
        \caption{Probabilistic inversion result 7}
    \end{subfigure}
    \hfill
    \begin{subfigure}{0.3\textwidth}
        \includegraphics[width=\textwidth]{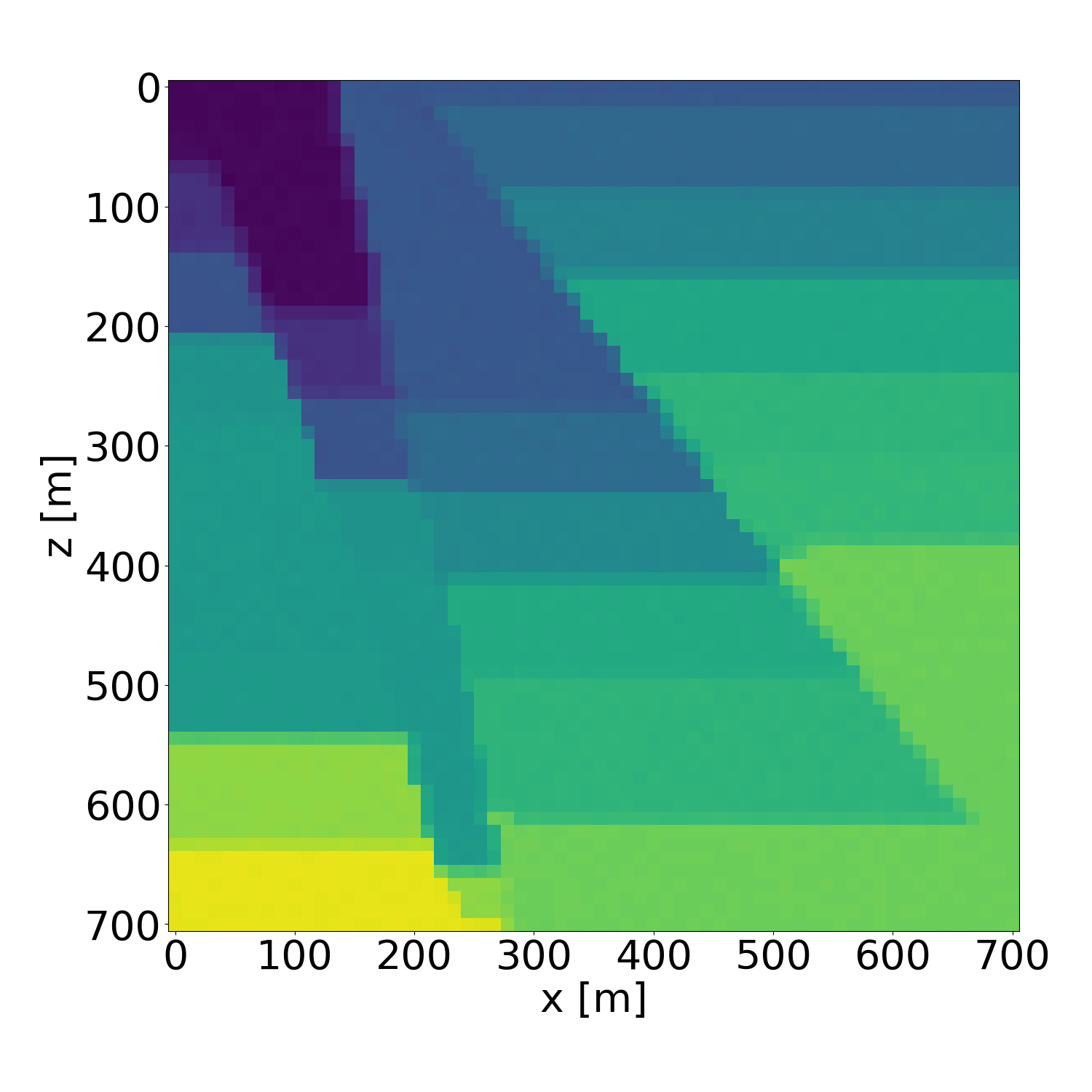}
        \caption{Probabilistic inversion result 8}
    \end{subfigure}
    \hfill
    \begin{subfigure}{0.3\textwidth}
        \includegraphics[width=\textwidth]{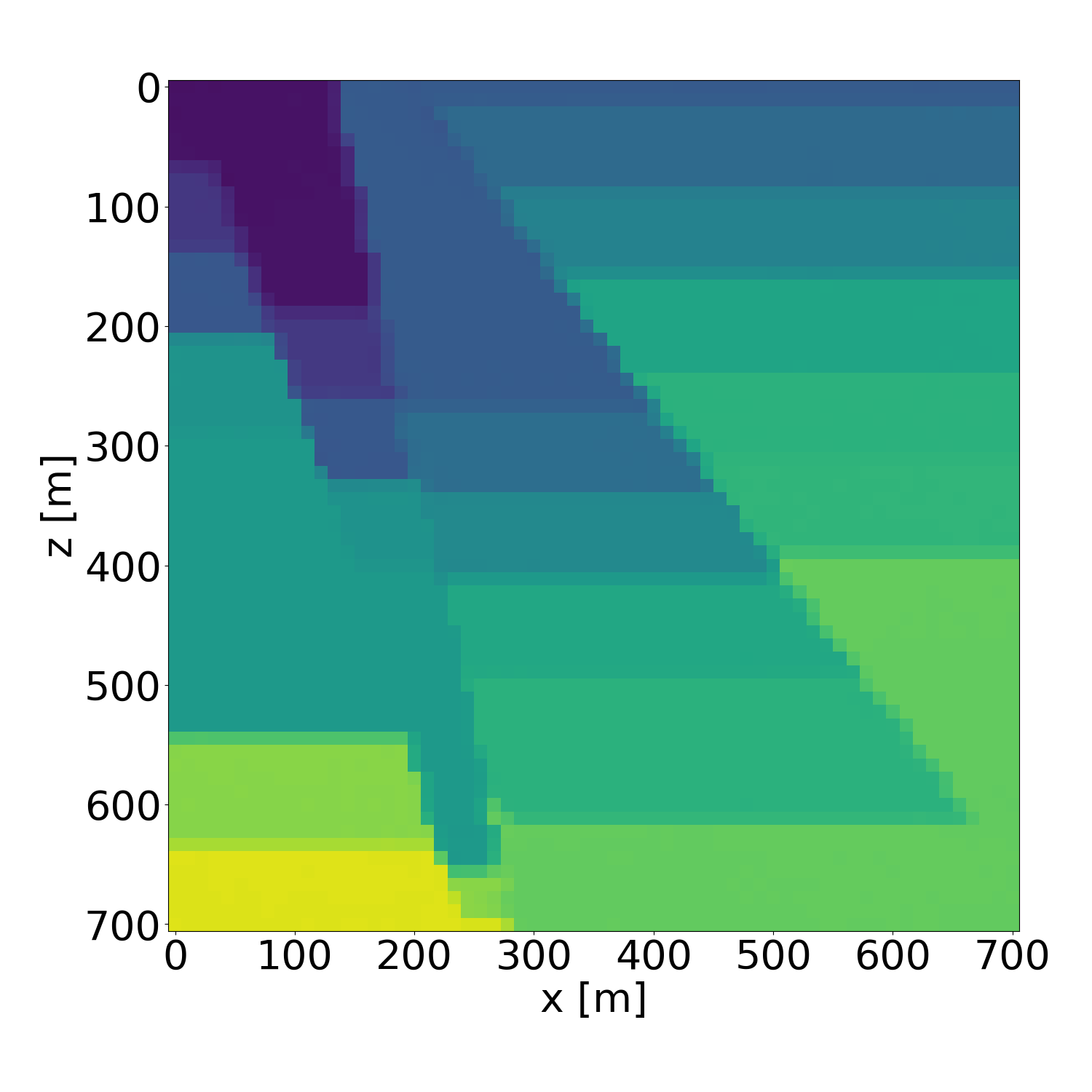}
        \caption{Probabilistic inversion result 9}
    \end{subfigure}
    \hfill
    \caption{Probabilistic inversion results of a isotropic velocity model in the validation dataset. Generated with guidance scale $w=4$.}
    \label{fig-allresult5}
\end{figure}

\begin{figure}
    \centering
    \begin{subfigure}{0.3\textwidth}
        \includegraphics[width=\textwidth]{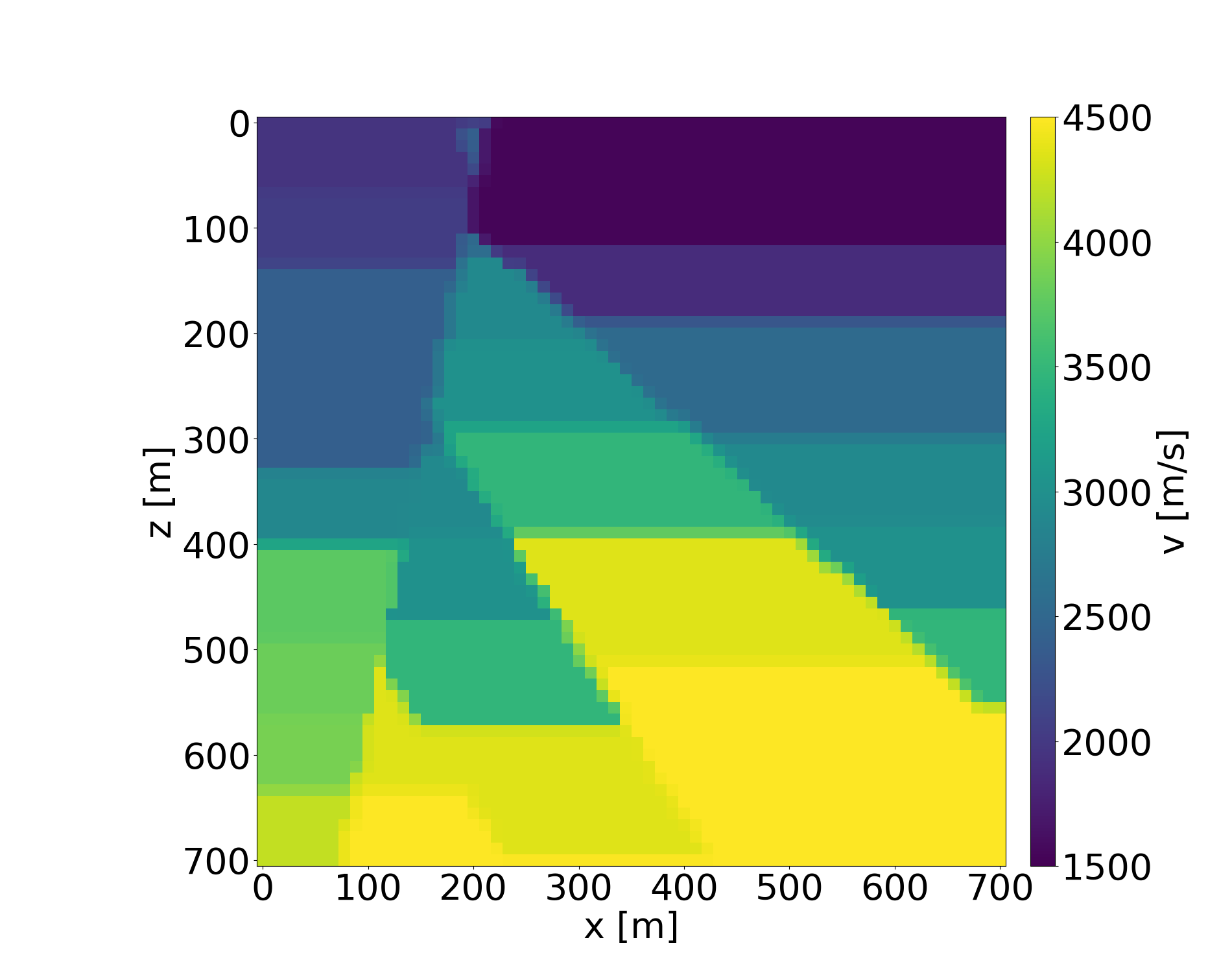}
        \caption{Ground truth (target)}
    \end{subfigure}
    \hfill
    \begin{subfigure}{0.3\textwidth}
        \includegraphics[width=\textwidth]{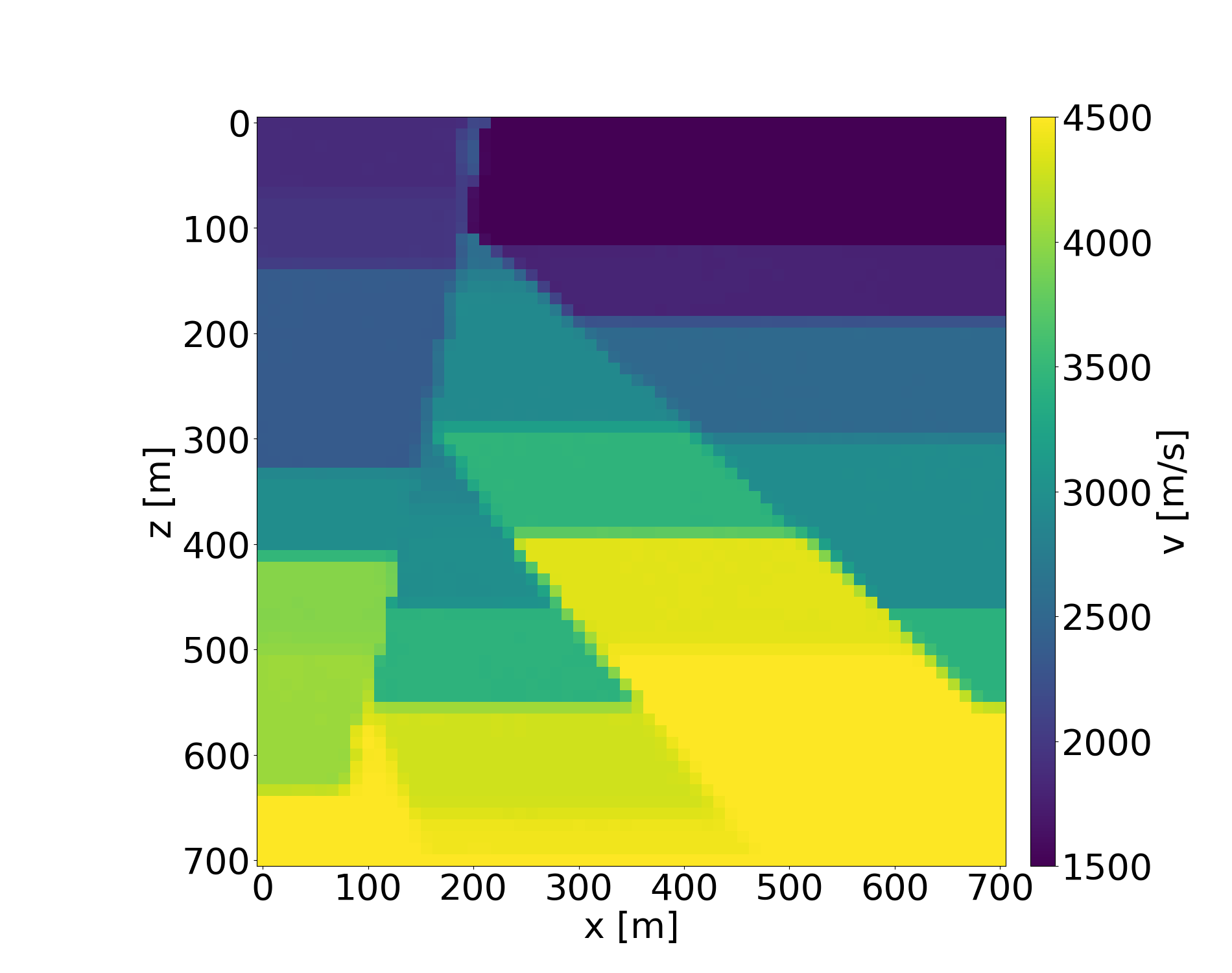}
        \caption{Average inversion result}
    \end{subfigure}
    \hfill
    \begin{subfigure}{0.3\textwidth}
        \includegraphics[width=\textwidth]{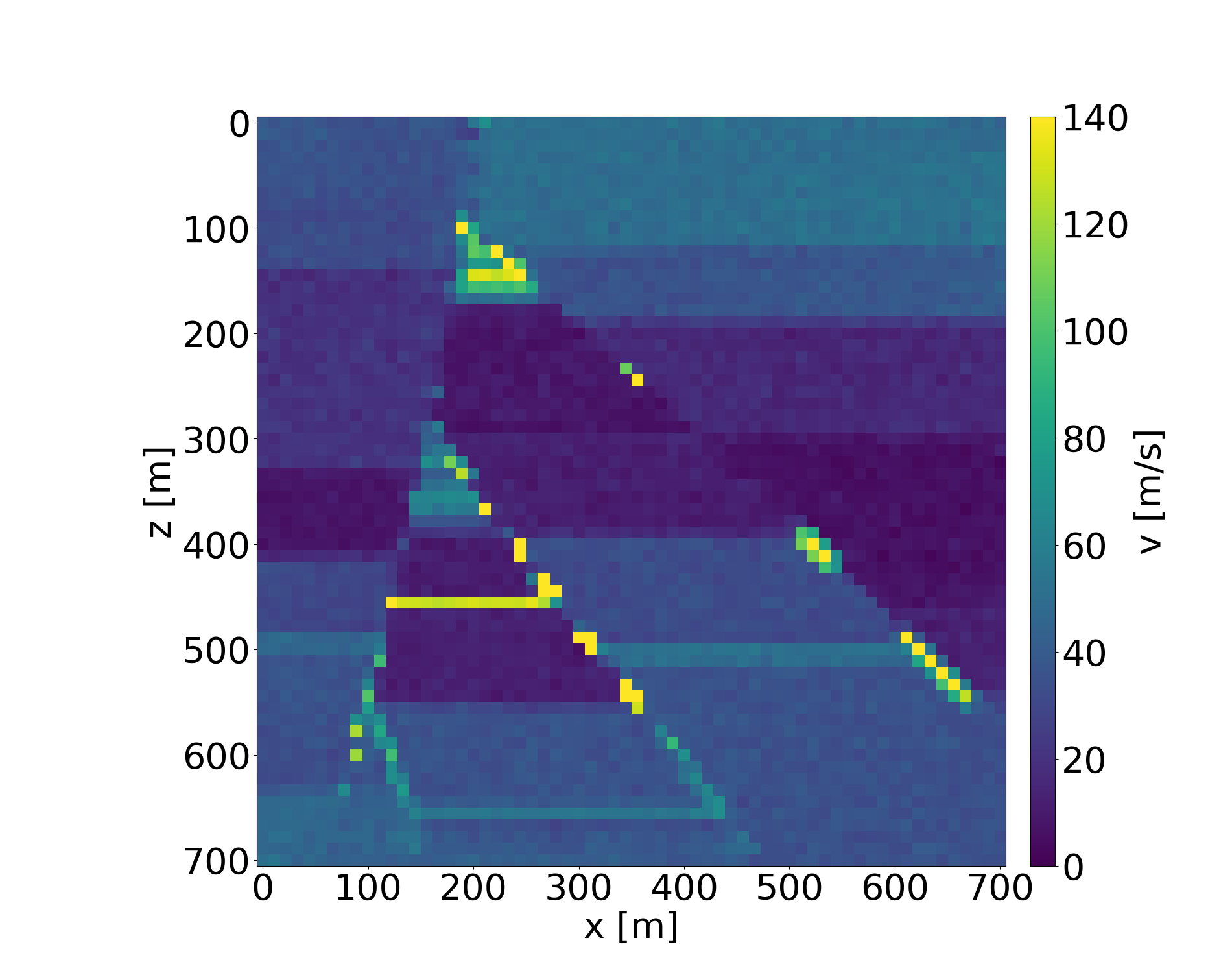}
        \caption{Standard deviation}
    \end{subfigure}
    \hfill
    \begin{subfigure}{0.3\textwidth}
        \includegraphics[width=\textwidth]{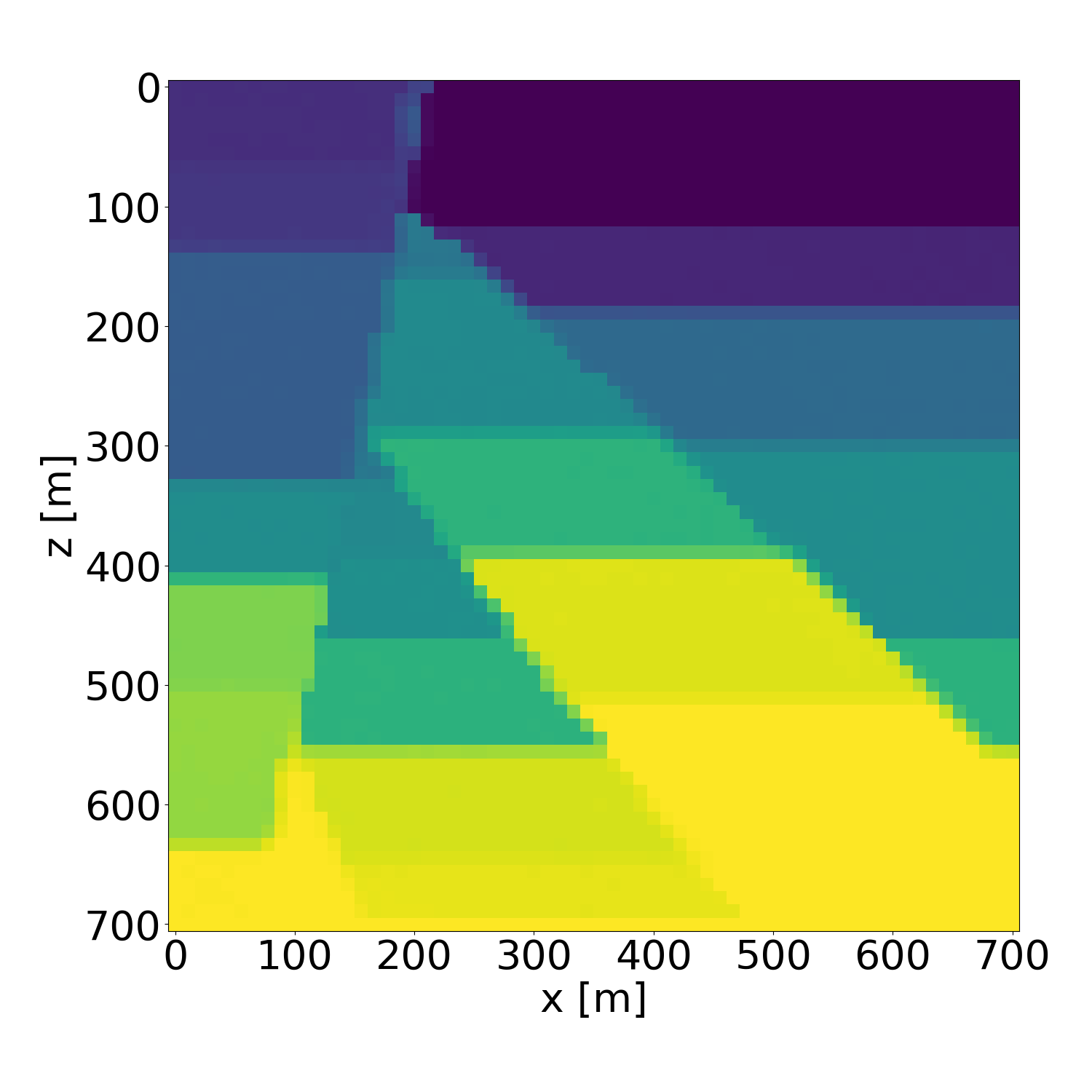}
        \caption{Probabilistic inversion result 1}
    \end{subfigure}
    \hfill
    \begin{subfigure}{0.3\textwidth}
        \includegraphics[width=\textwidth]{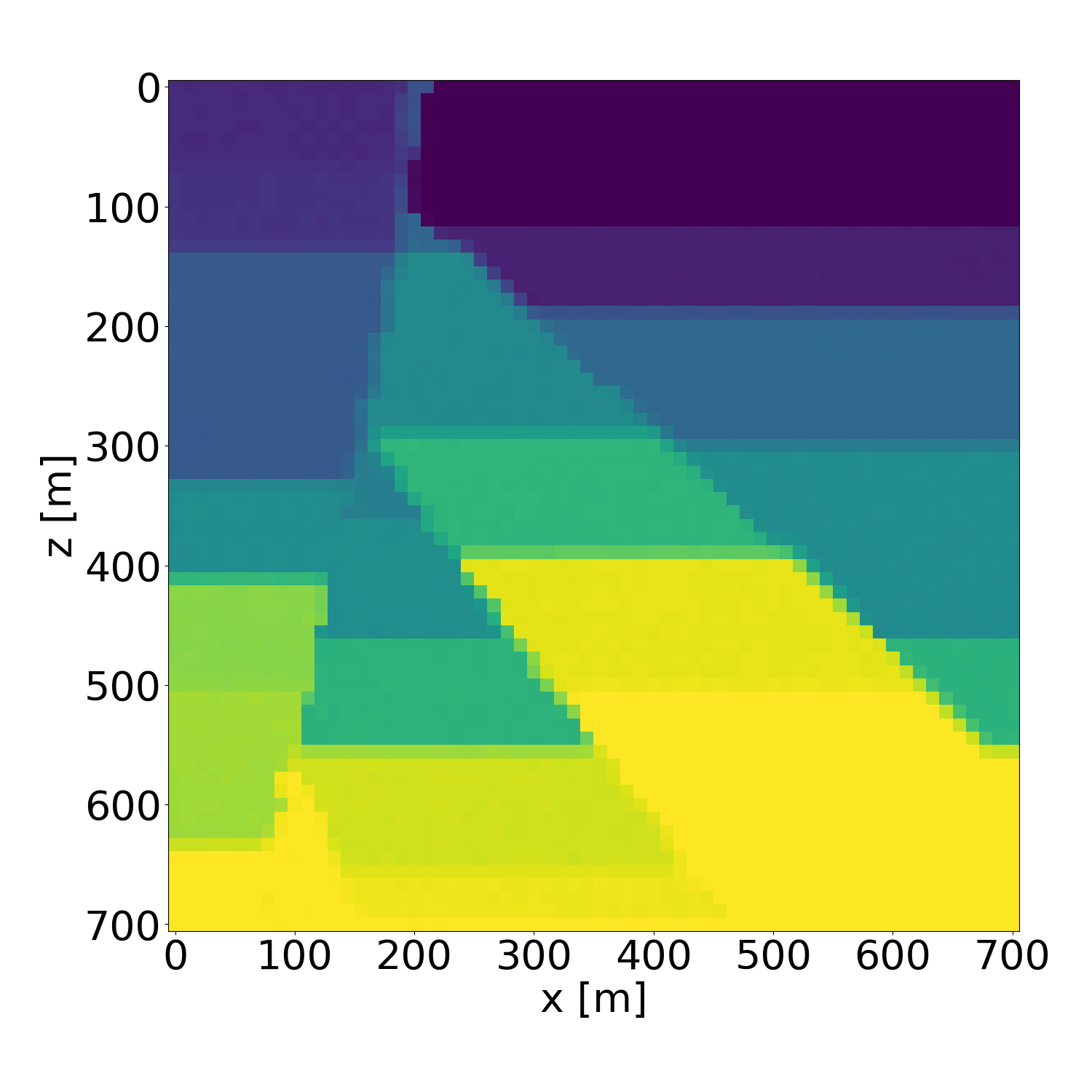}
        \caption{Probabilistic inversion result 2}
    \end{subfigure}
    \hfill
    \begin{subfigure}{0.3\textwidth}
        \includegraphics[width=\textwidth]{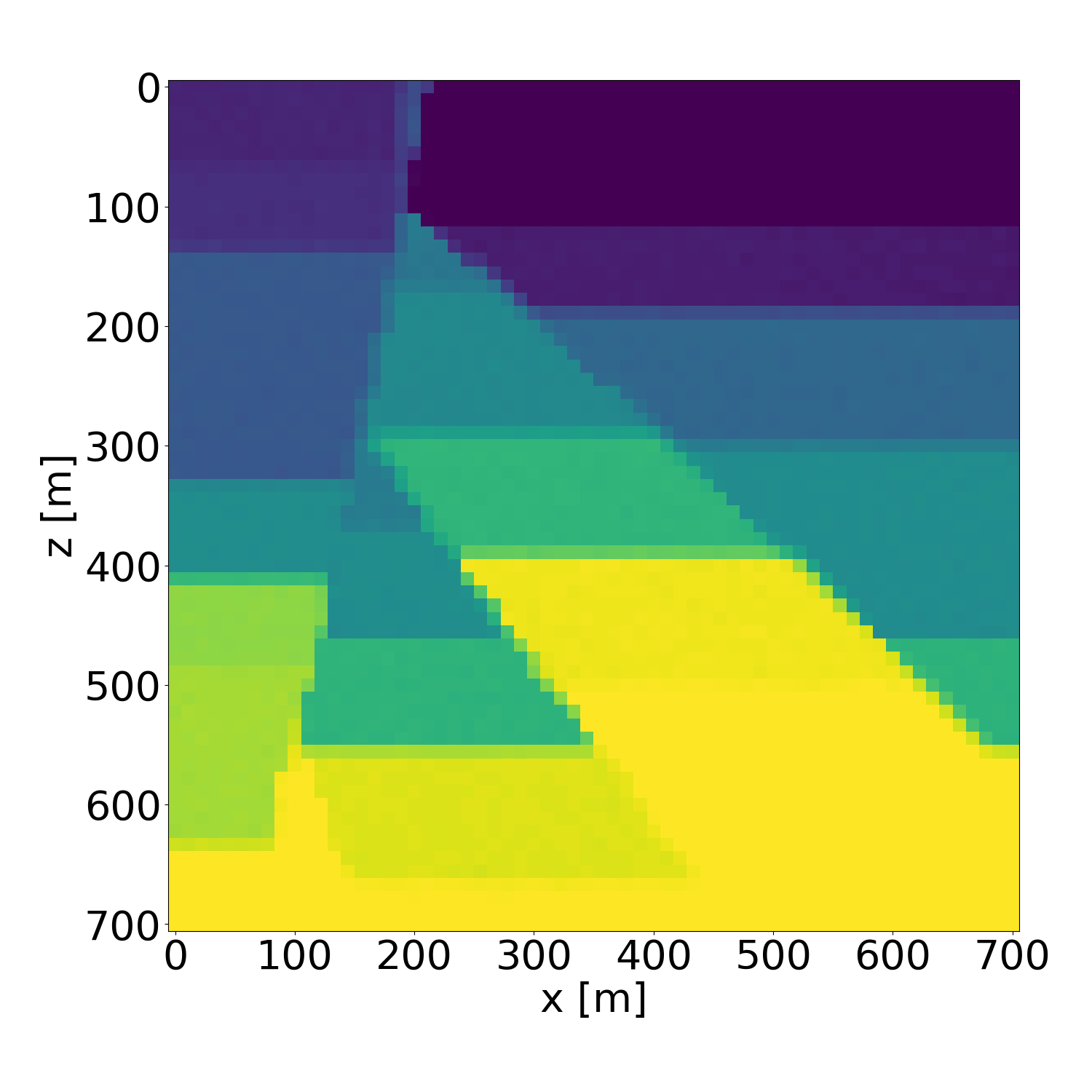}
        \caption{Probabilistic inversion result 3}
    \end{subfigure}
    \hfill
    \begin{subfigure}{0.3\textwidth}
        \includegraphics[width=\textwidth]{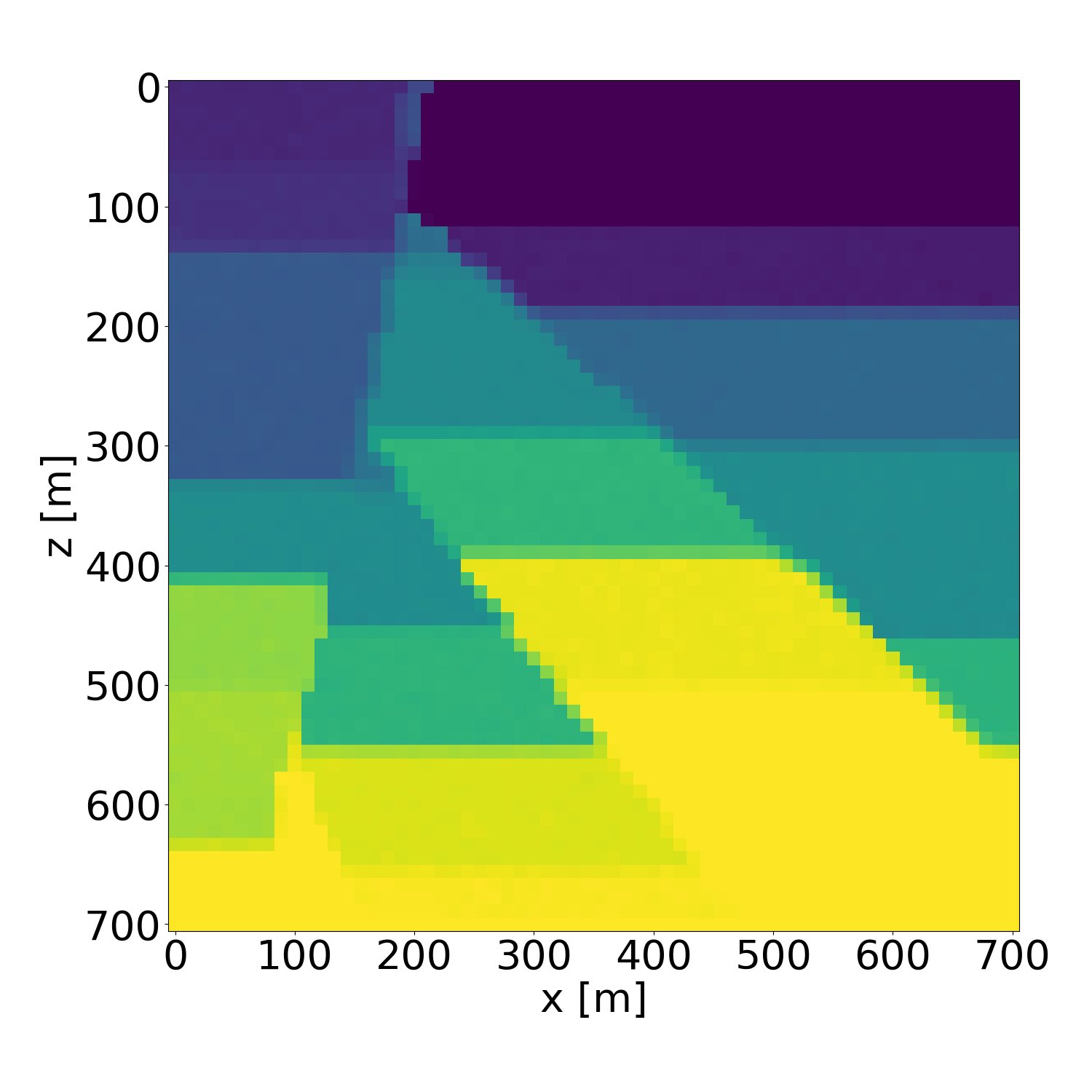}
        \caption{Probabilistic inversion result 4}
    \end{subfigure}
    \hfill
    \begin{subfigure}{0.3\textwidth}
        \includegraphics[width=\textwidth]{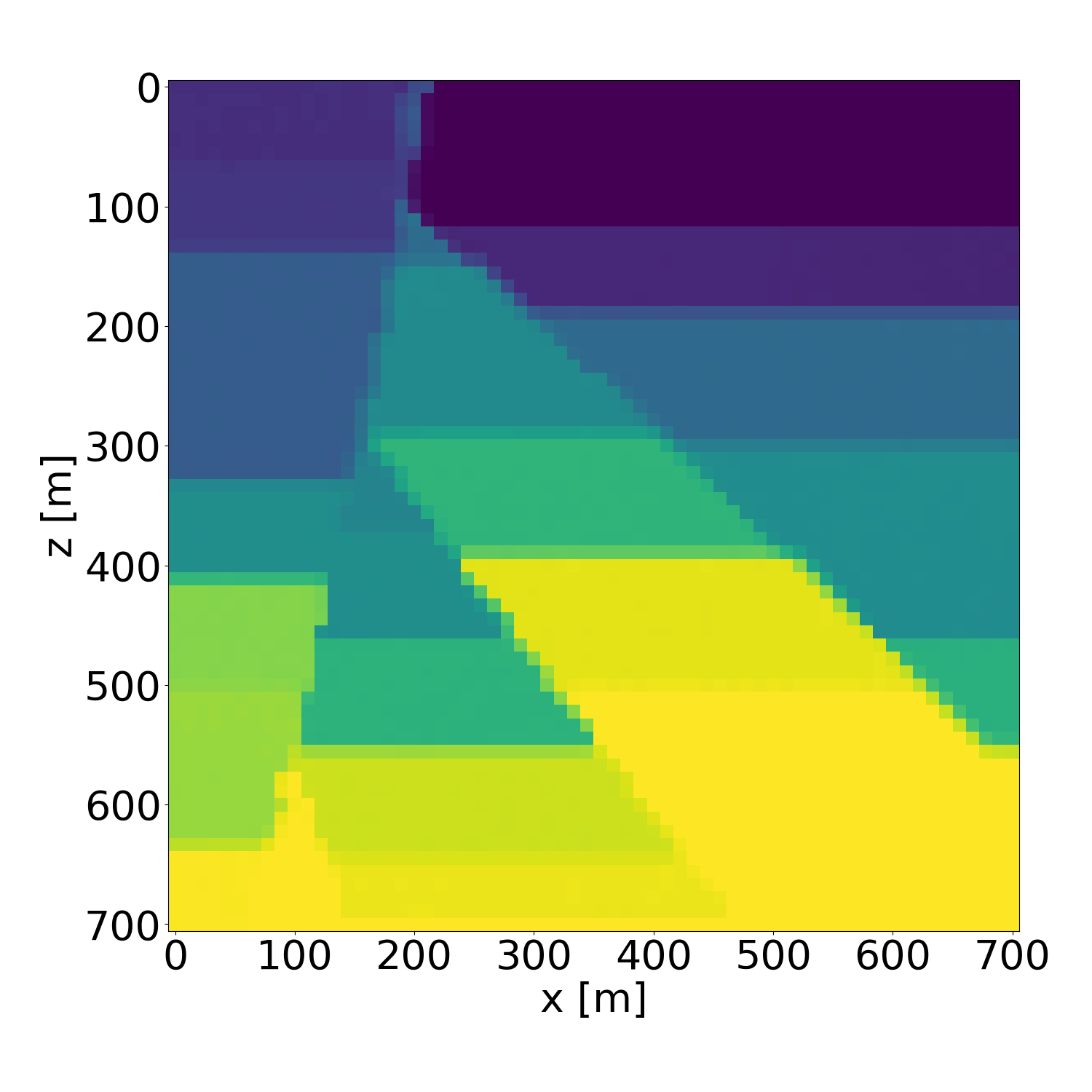}
        \caption{Probabilistic inversion result 5}
    \end{subfigure}
    \hfill
    \begin{subfigure}{0.3\textwidth}
        \includegraphics[width=\textwidth]{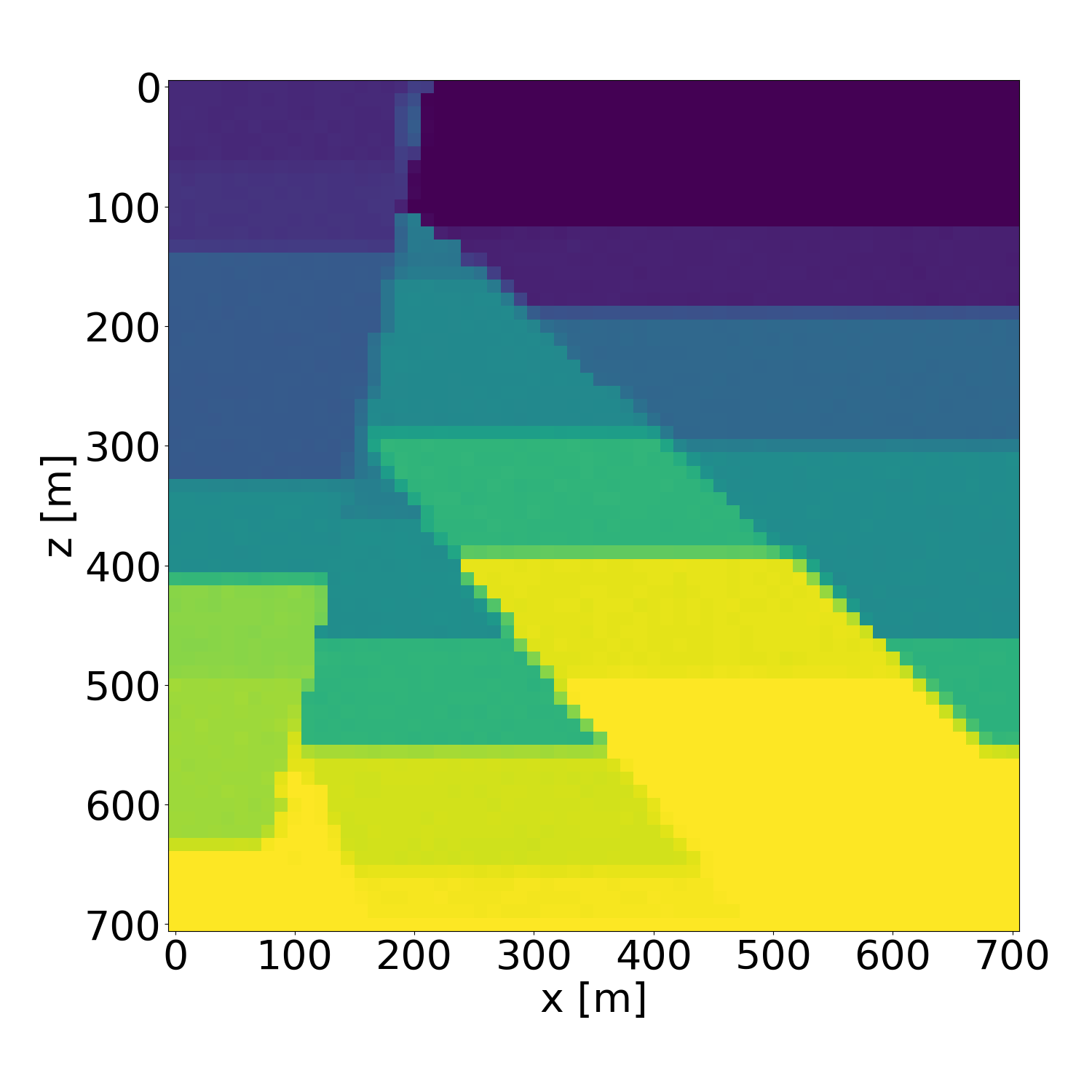}
        \caption{Probabilistic inversion result 6}
    \end{subfigure}
    \hfill
    \begin{subfigure}{0.3\textwidth}
        \includegraphics[width=\textwidth]{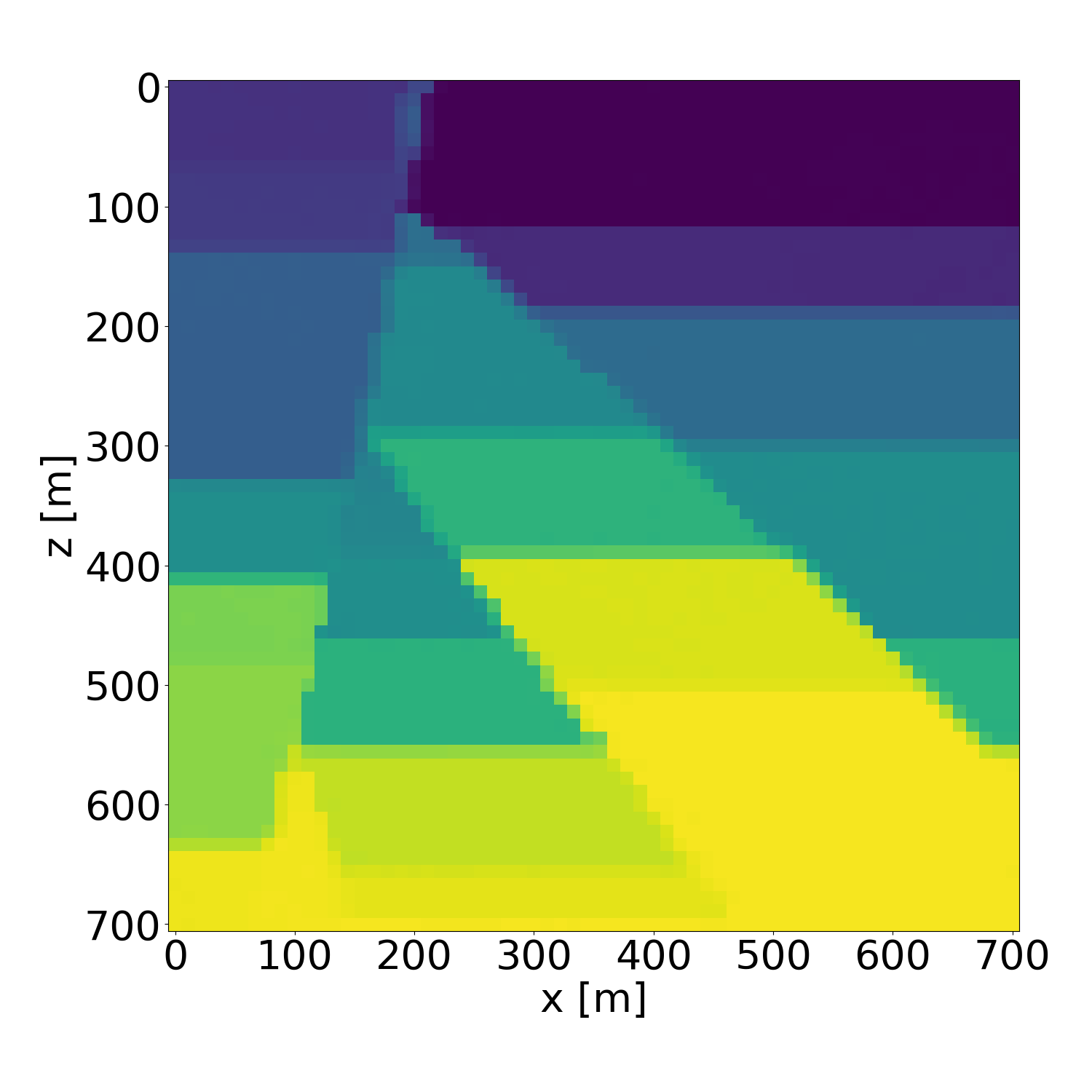}
        \caption{Probabilistic inversion result 7}
    \end{subfigure}
    \hfill
    \begin{subfigure}{0.3\textwidth}
        \includegraphics[width=\textwidth]{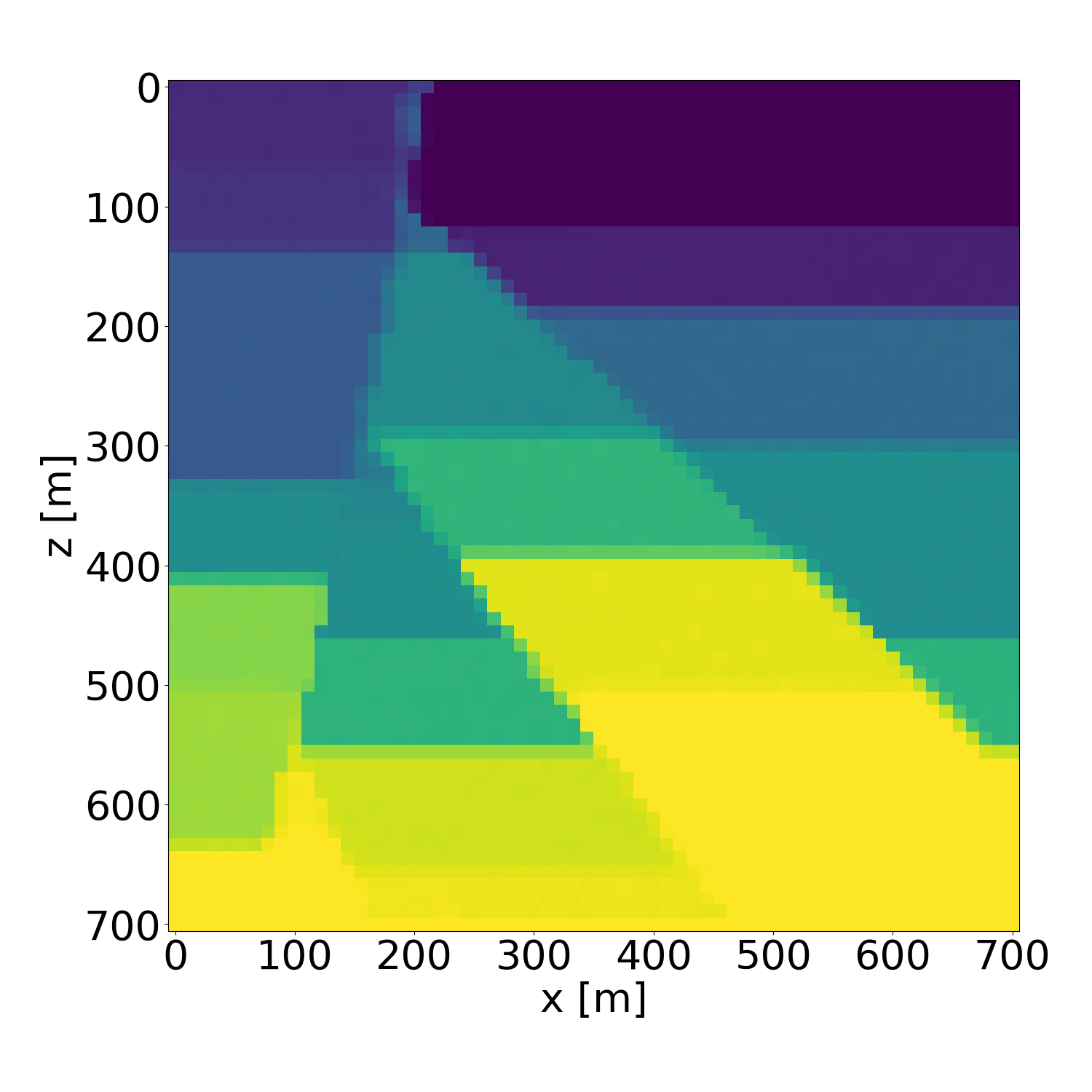}
        \caption{Probabilistic inversion result 8}
    \end{subfigure}
    \hfill
    \begin{subfigure}{0.3\textwidth}
        \includegraphics[width=\textwidth]{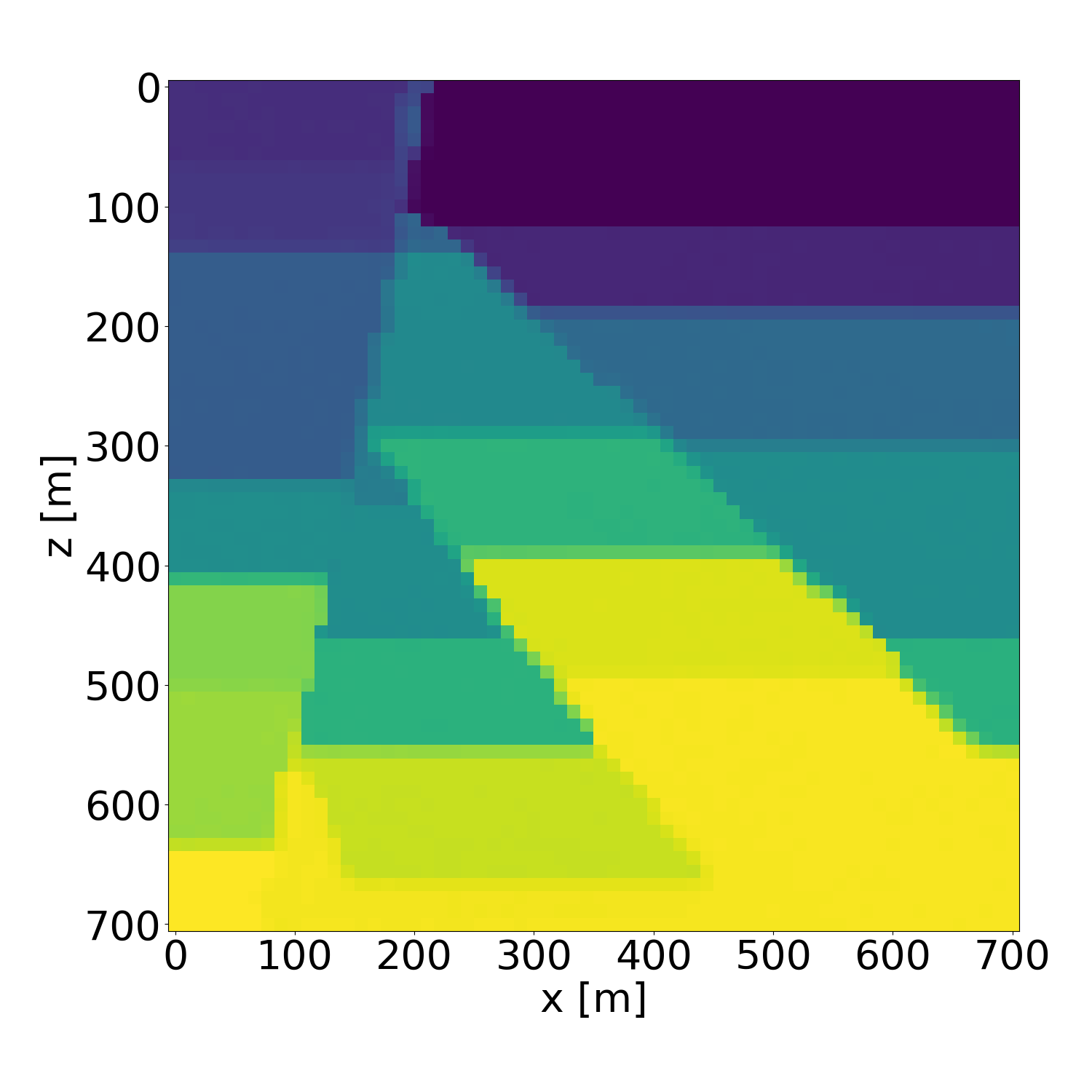}
        \caption{Probabilistic inversion result 9}
    \end{subfigure}
    \hfill
    \caption{Probabilistic inversion results of a isotropic velocity model in the validation dataset. Generated with guidance scale $w=4$.}
    \label{fig-allresult6}
\end{figure}

\begin{figure}
    \centering
    \begin{subfigure}{0.3\textwidth}
        \includegraphics[width=\textwidth]{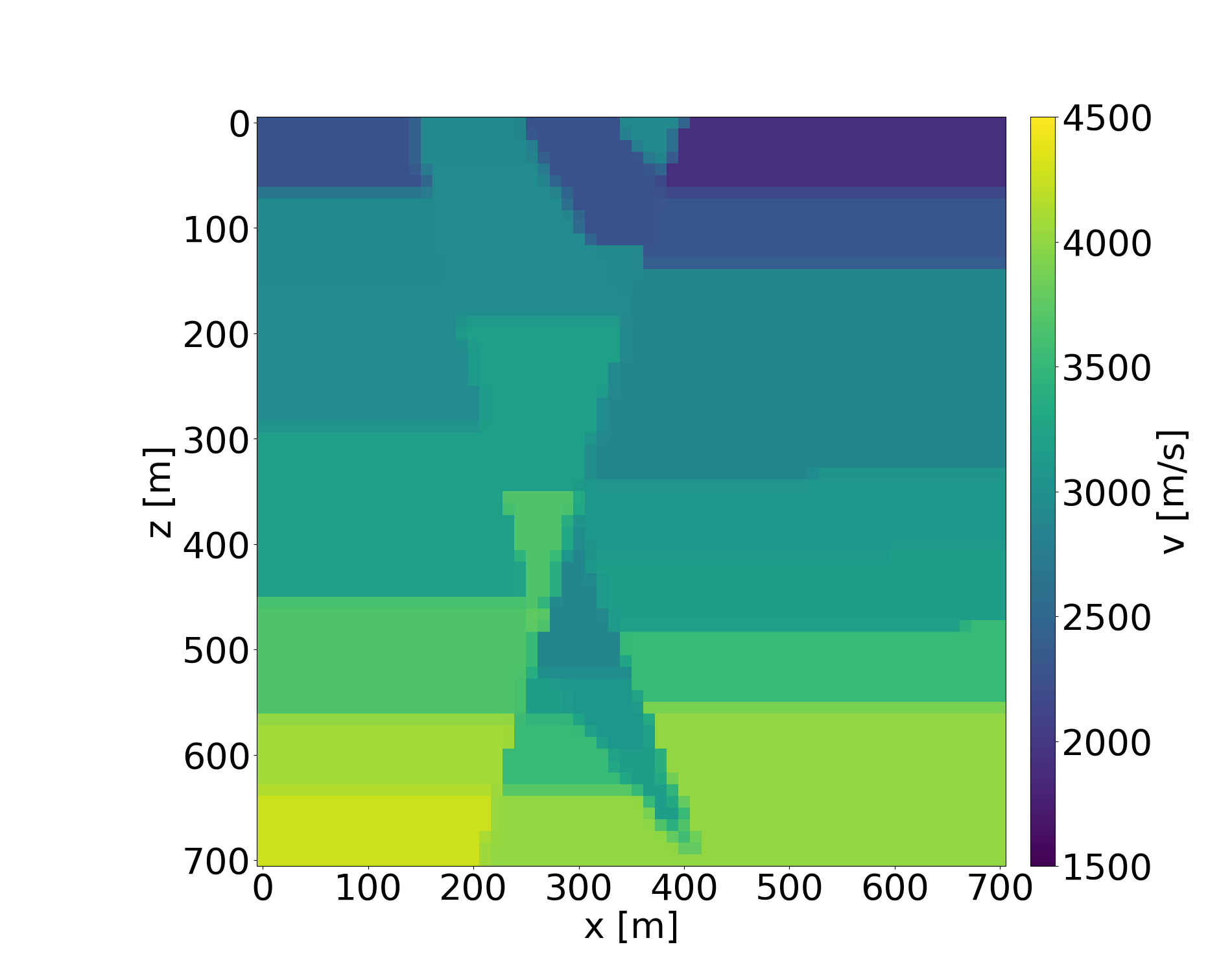}
        \caption{Ground truth (target)}
    \end{subfigure}
    \hfill
    \begin{subfigure}{0.3\textwidth}
        \includegraphics[width=\textwidth]{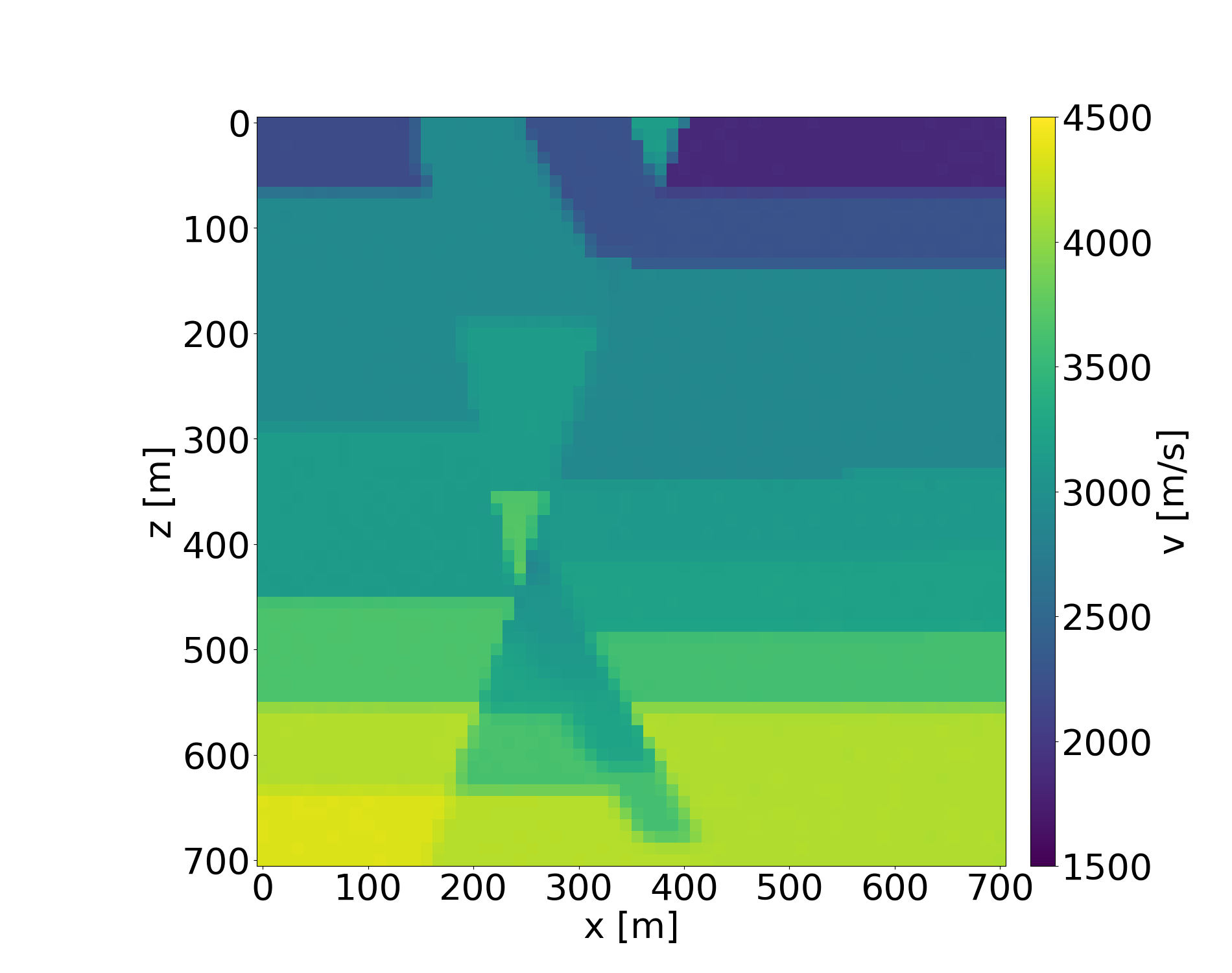}
        \caption{Average inversion result}
    \end{subfigure}
    \hfill
    \begin{subfigure}{0.3\textwidth}
        \includegraphics[width=\textwidth]{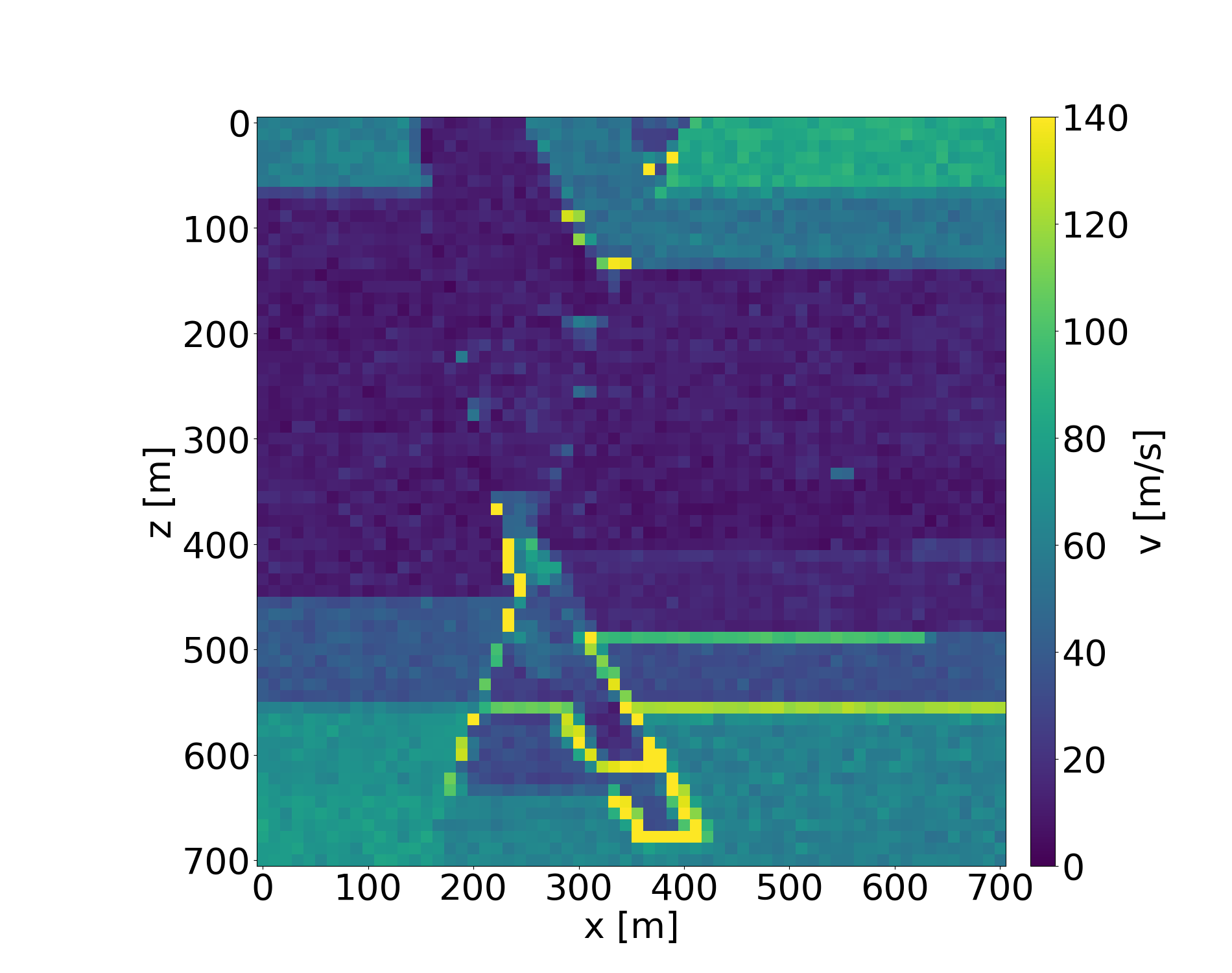}
        \caption{Standard deviation}
    \end{subfigure}
    \hfill
    \begin{subfigure}{0.3\textwidth}
        \includegraphics[width=\textwidth]{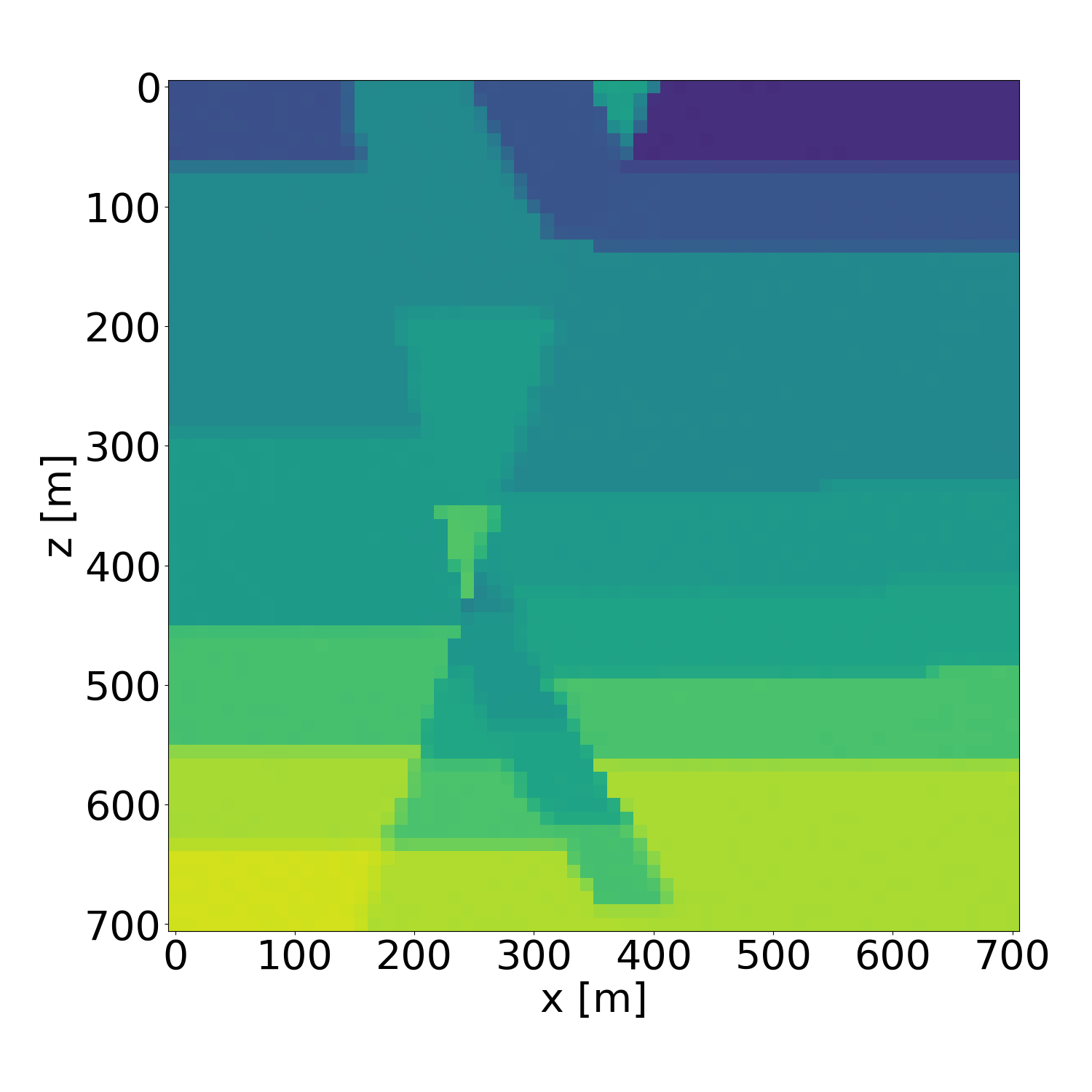}
        \caption{Probabilistic inversion result 1}
    \end{subfigure}
    \hfill
    \begin{subfigure}{0.3\textwidth}
        \includegraphics[width=\textwidth]{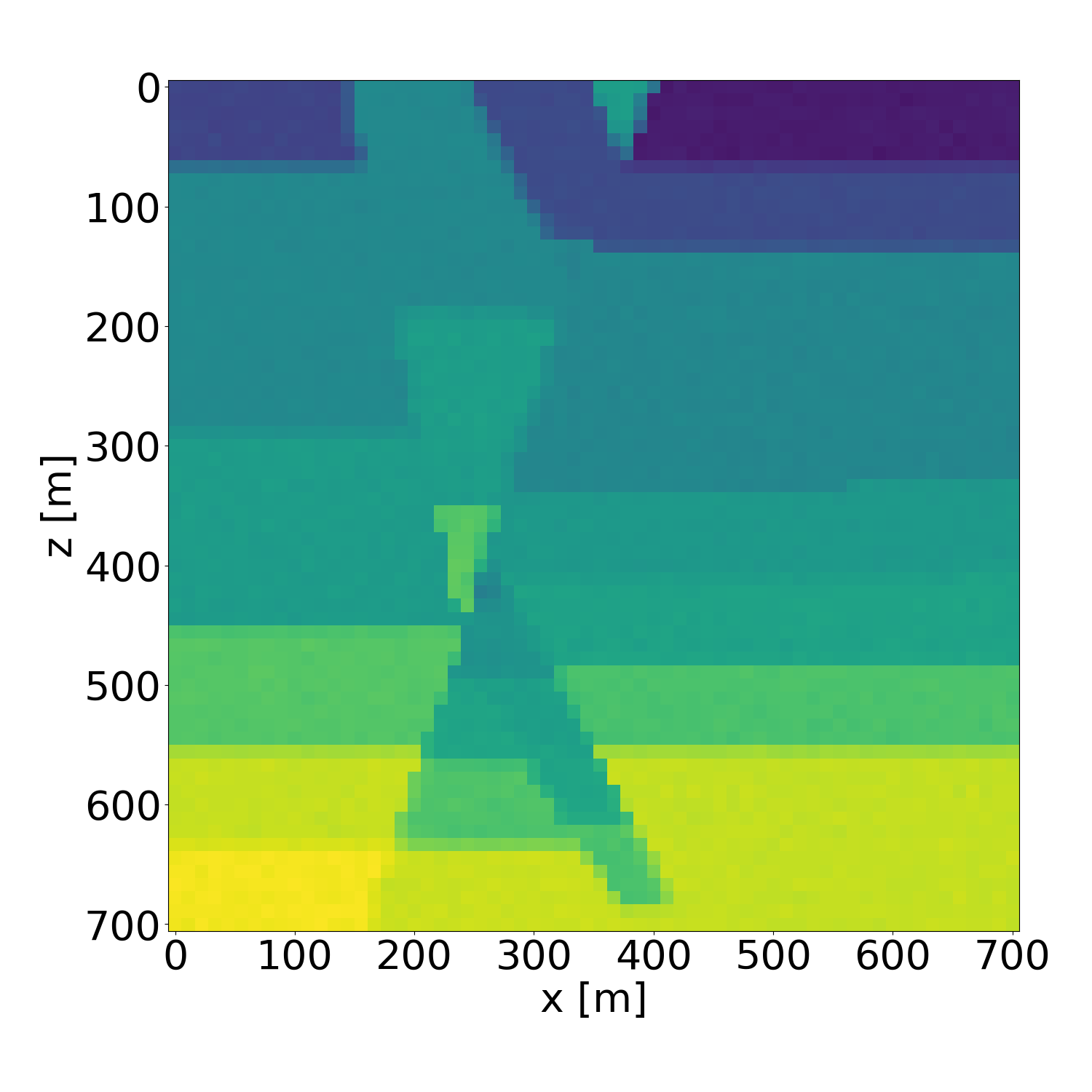}
        \caption{Probabilistic inversion result 2}
    \end{subfigure}
    \hfill
    \begin{subfigure}{0.3\textwidth}
        \includegraphics[width=\textwidth]{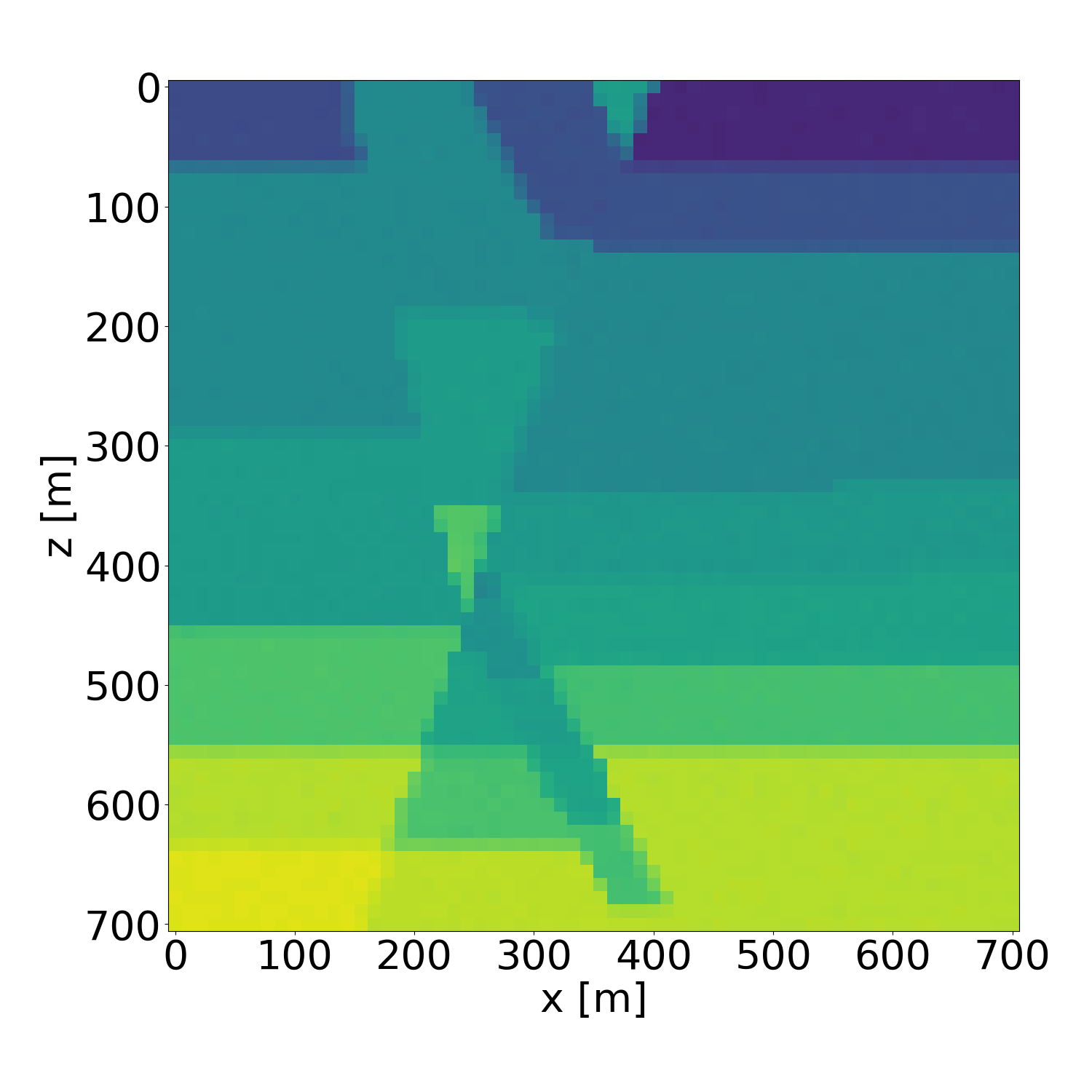}
        \caption{Probabilistic inversion result 3}
    \end{subfigure}
    \hfill
    \begin{subfigure}{0.3\textwidth}
        \includegraphics[width=\textwidth]{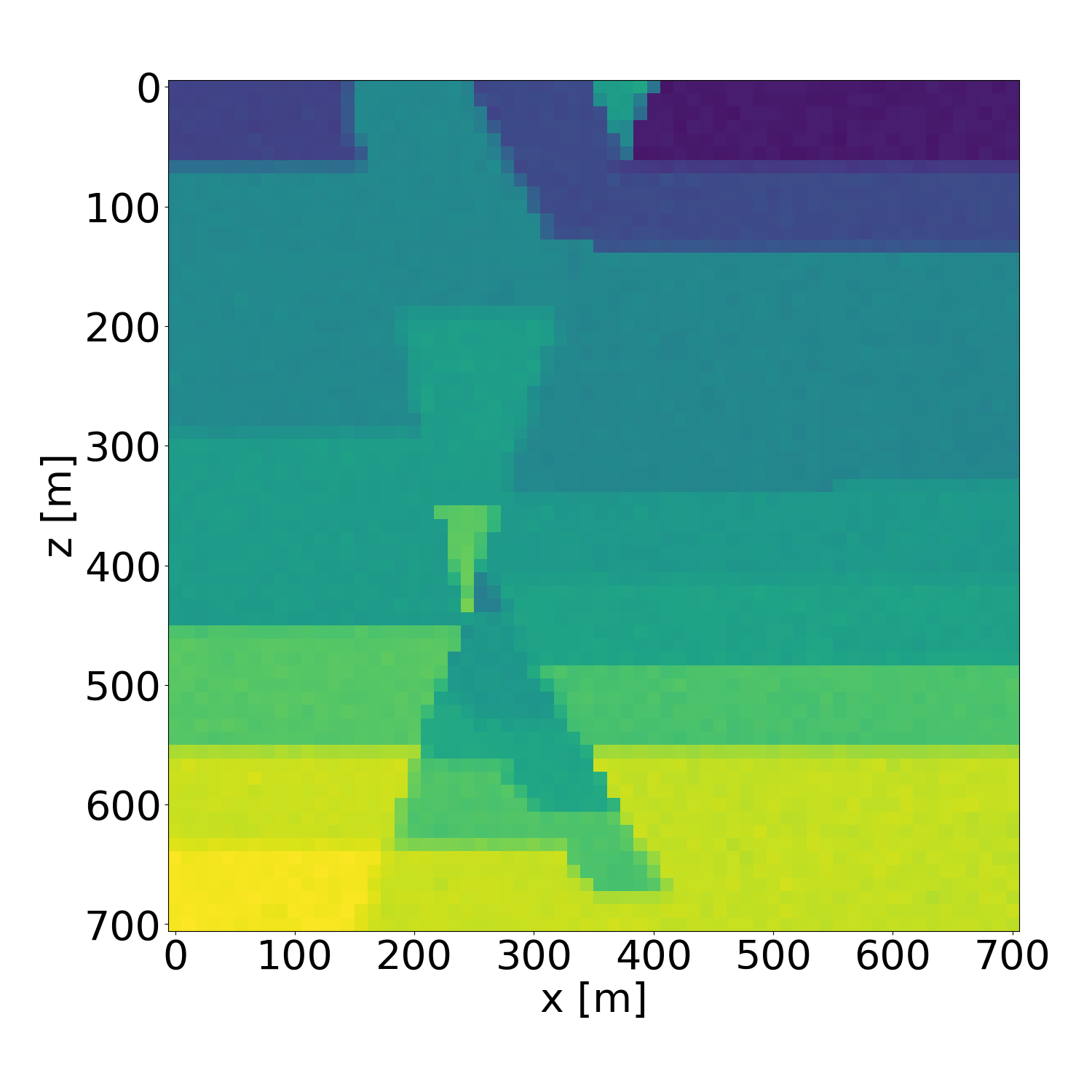}
        \caption{Probabilistic inversion result 4}
    \end{subfigure}
    \hfill
    \begin{subfigure}{0.3\textwidth}
        \includegraphics[width=\textwidth]{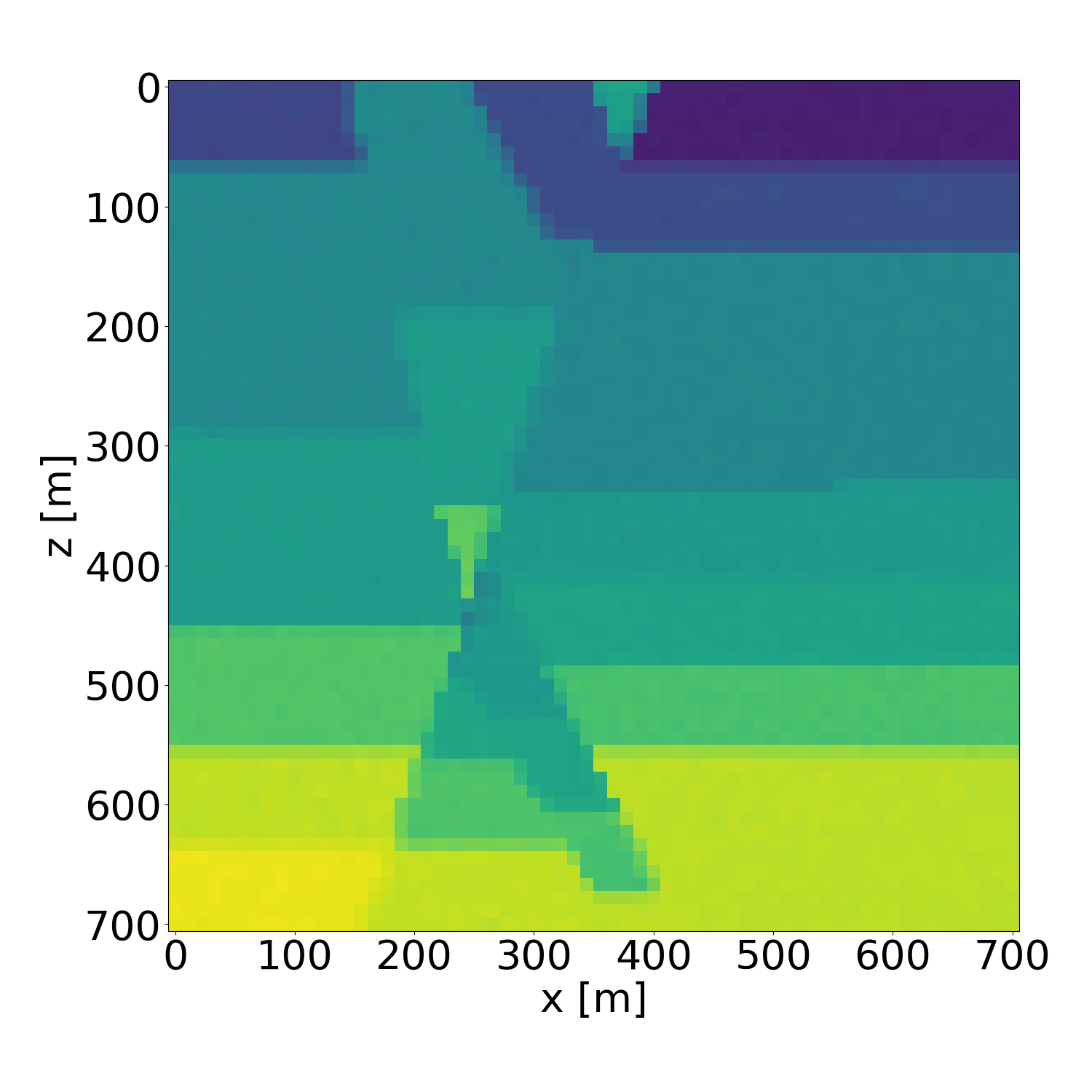}
        \caption{Probabilistic inversion result 5}
    \end{subfigure}
    \hfill
    \begin{subfigure}{0.3\textwidth}
        \includegraphics[width=\textwidth]{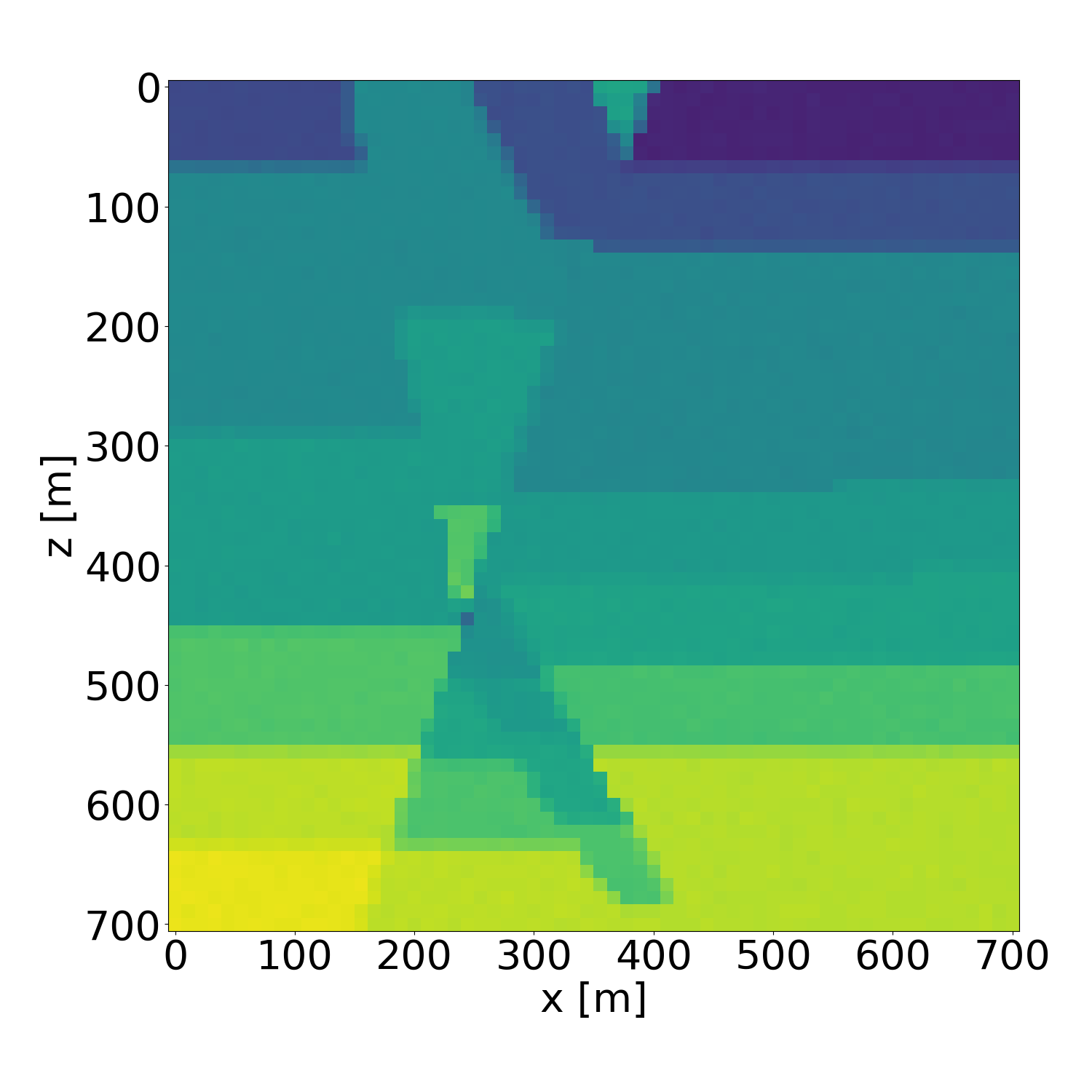}
        \caption{Probabilistic inversion result 6}
    \end{subfigure}
    \hfill
    \begin{subfigure}{0.3\textwidth}
        \includegraphics[width=\textwidth]{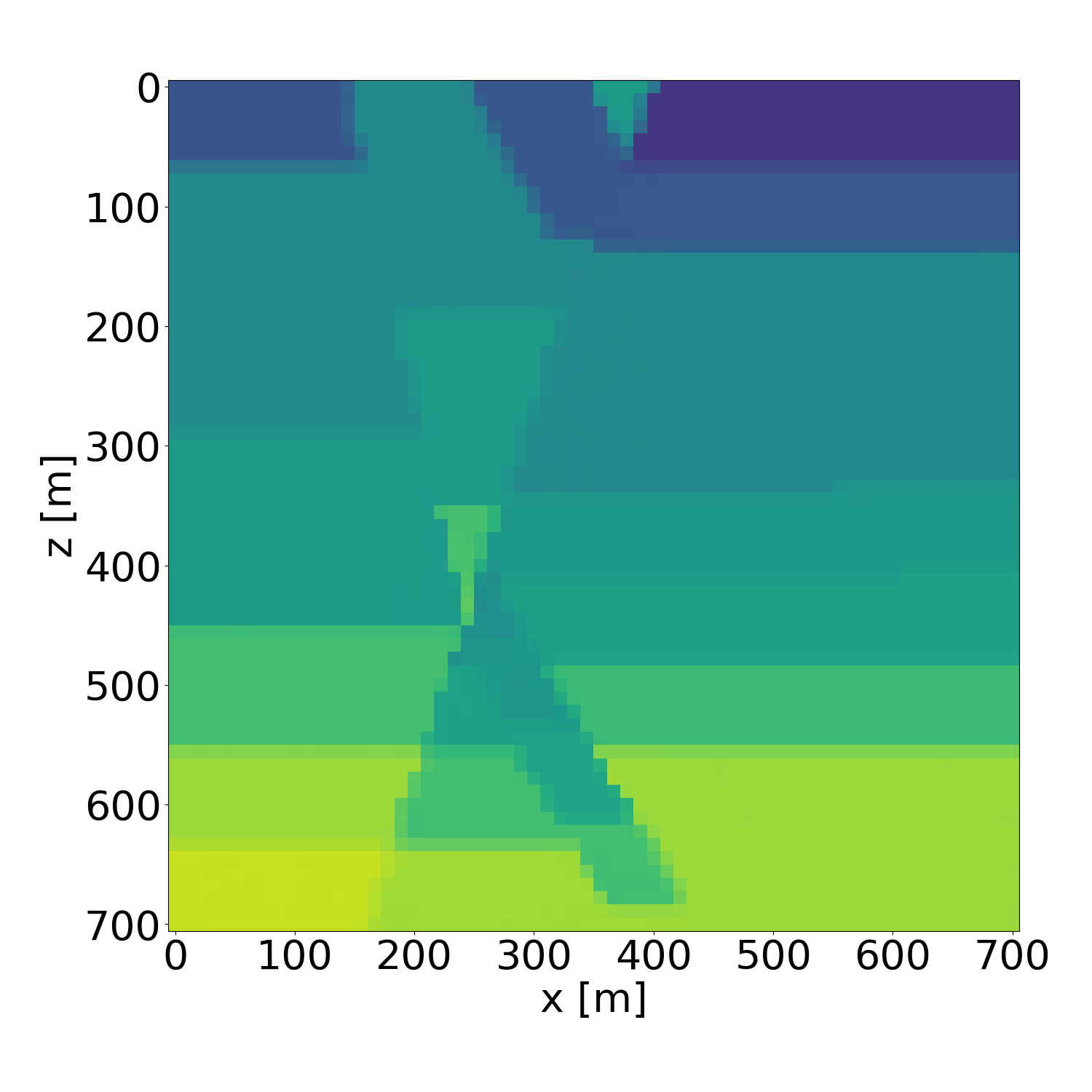}
        \caption{Probabilistic inversion result 7}
    \end{subfigure}
    \hfill
    \begin{subfigure}{0.3\textwidth}
        \includegraphics[width=\textwidth]{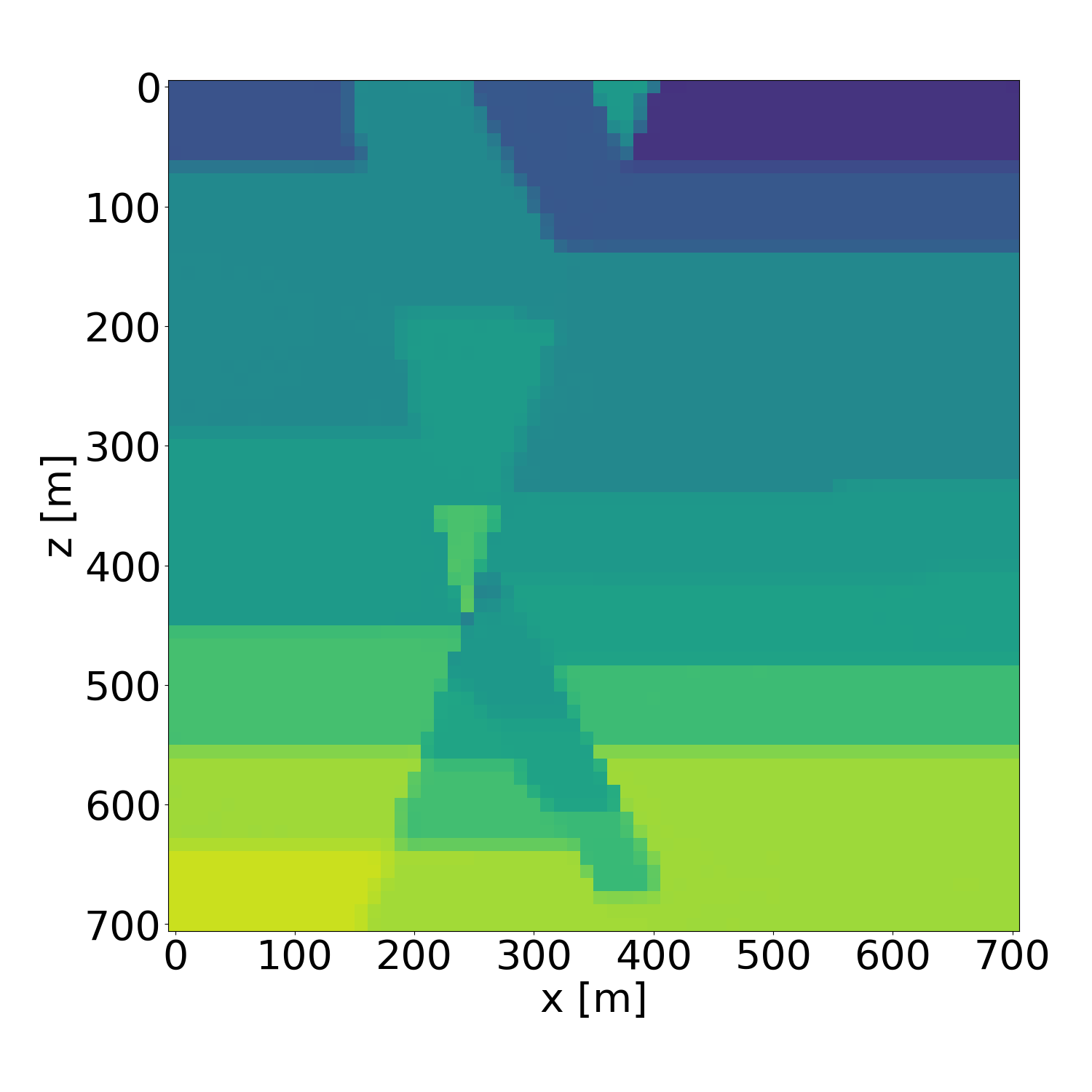}
        \caption{Probabilistic inversion result 8}
    \end{subfigure}
    \hfill
    \begin{subfigure}{0.3\textwidth}
        \includegraphics[width=\textwidth]{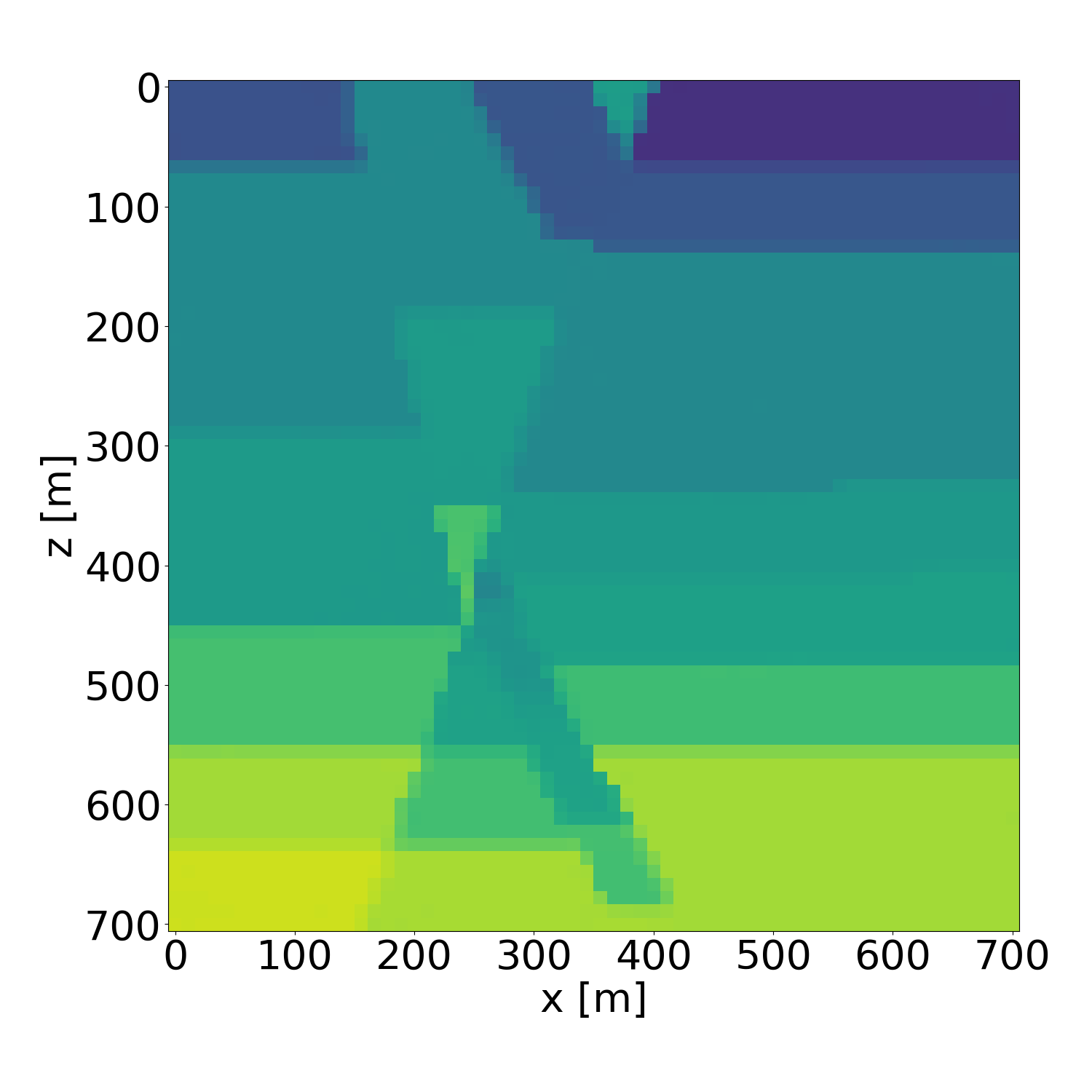}
        \caption{Probabilistic inversion result 9}
    \end{subfigure}
    \hfill
    \caption{Probabilistic inversion results of a isotropic velocity model in the validation dataset. Generated with guidance scale $w=4$.}
    \label{fig-allresult7}
\end{figure}

\begin{figure}
    \centering
    \begin{subfigure}{0.3\textwidth}
        \includegraphics[width=\textwidth]{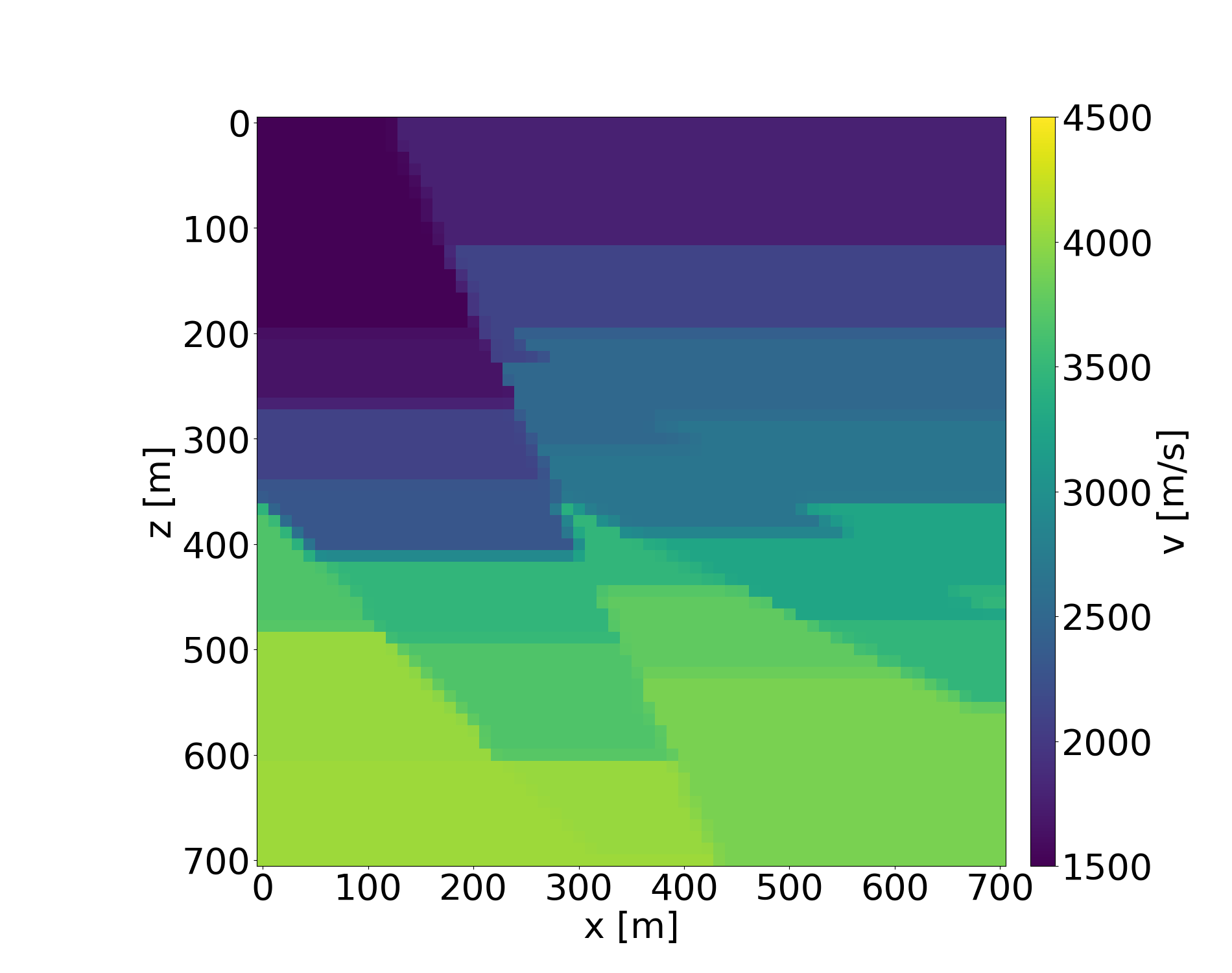}
        \caption{Ground truth (target)}
    \end{subfigure}
    \hfill
    \begin{subfigure}{0.3\textwidth}
        \includegraphics[width=\textwidth]{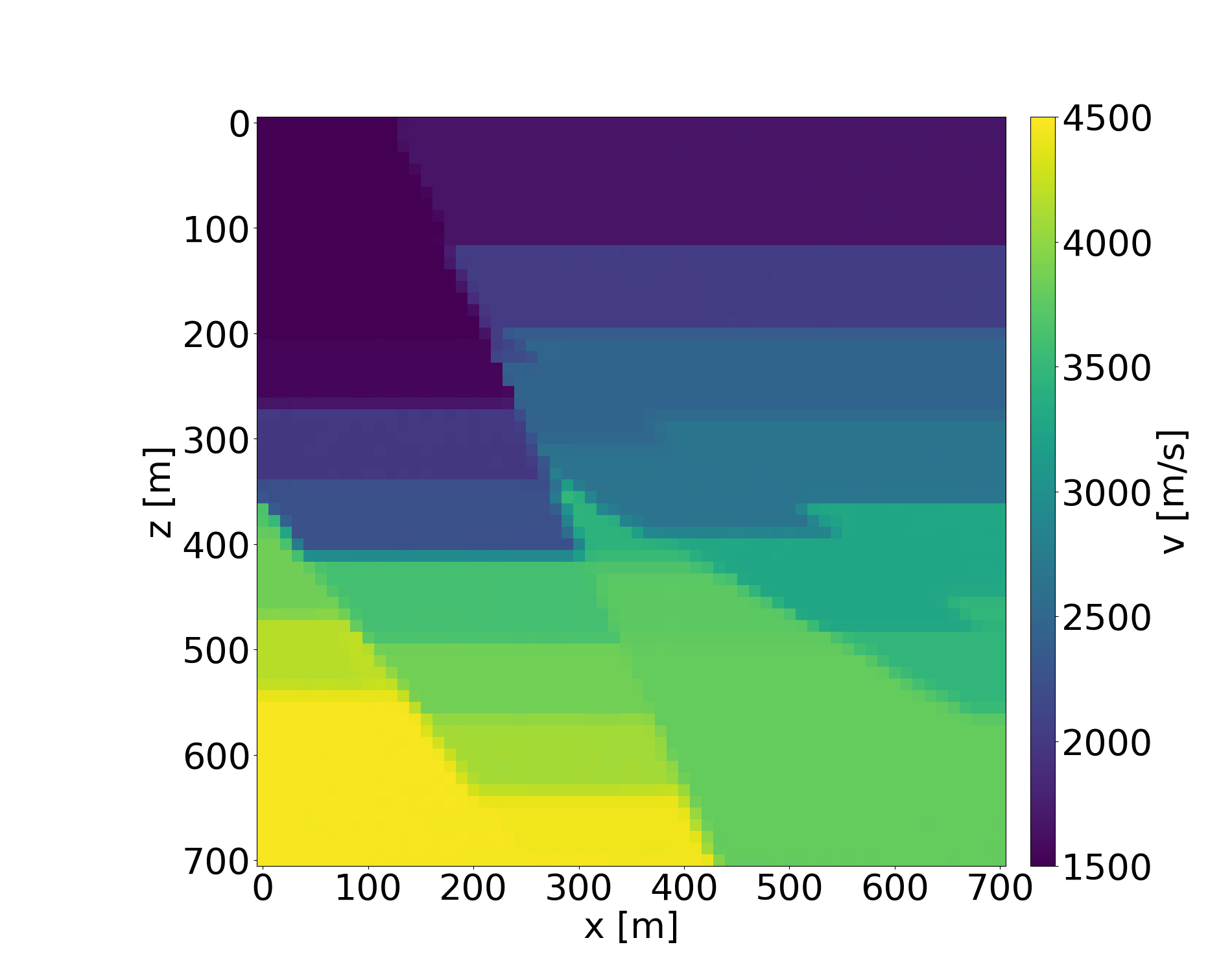}
        \caption{Average inversion result}
    \end{subfigure}
    \hfill
    \begin{subfigure}{0.3\textwidth}
        \includegraphics[width=\textwidth]{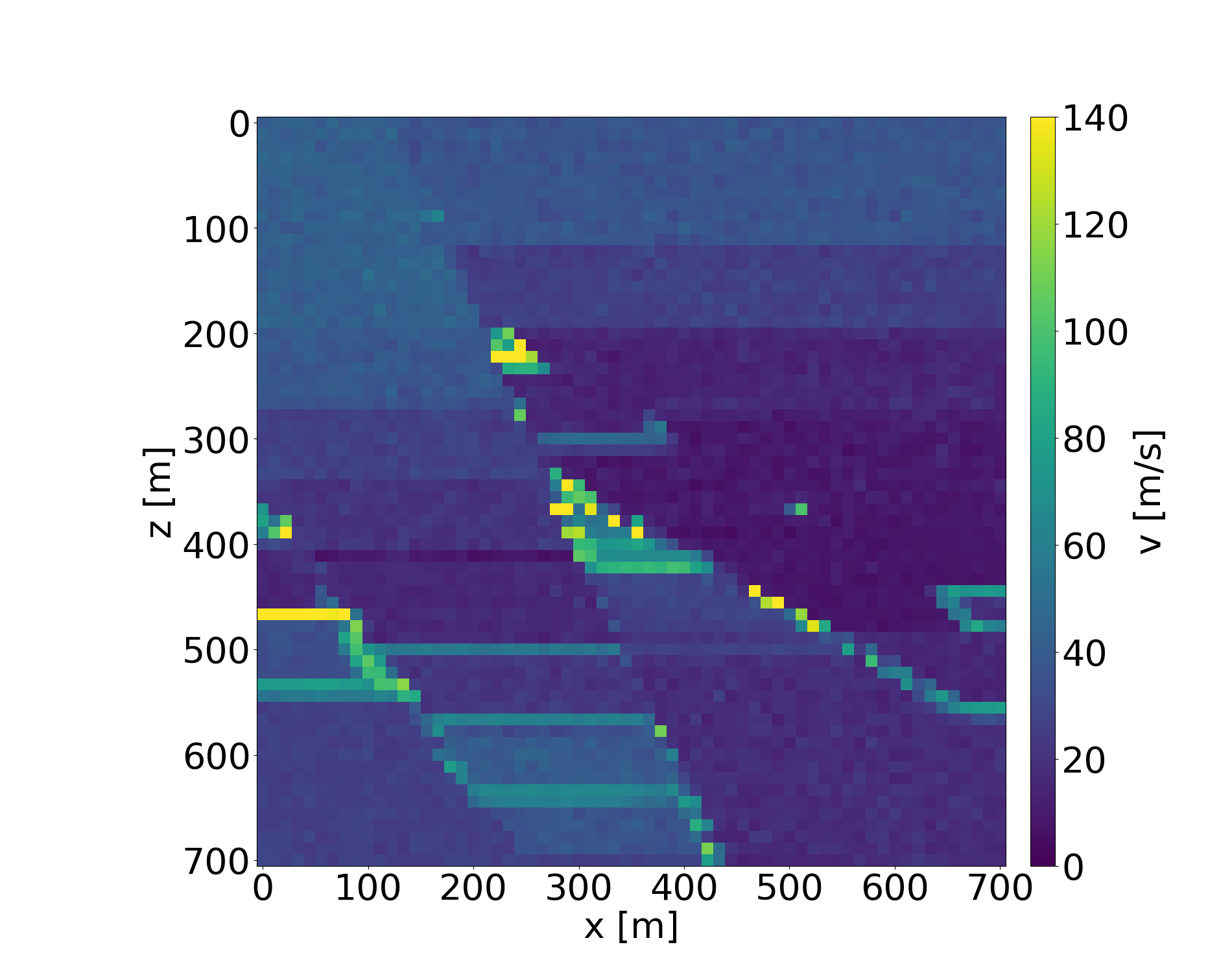}
        \caption{Standard deviation}
    \end{subfigure}
    \hfill
    \begin{subfigure}{0.3\textwidth}
        \includegraphics[width=\textwidth]{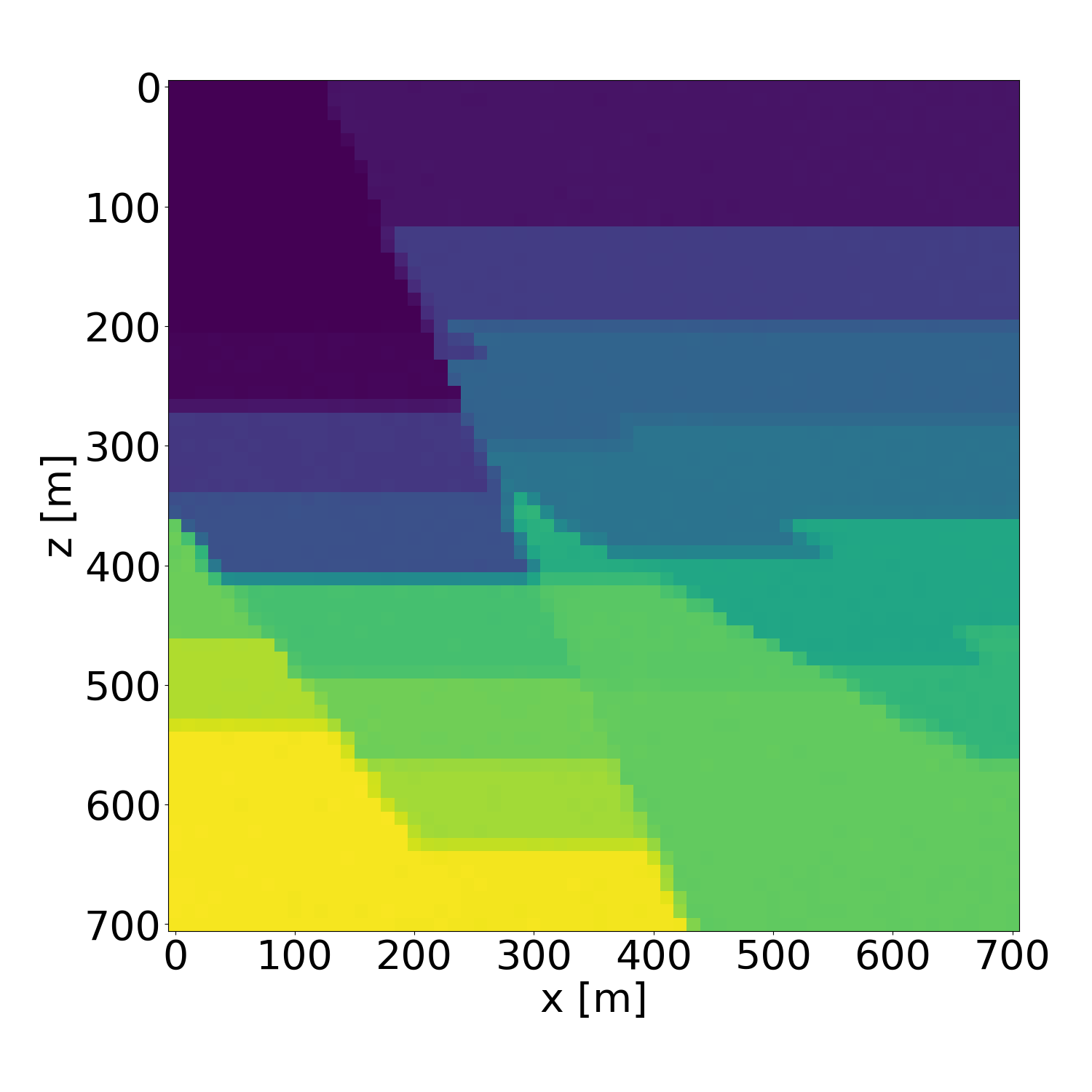}
        \caption{Probabilistic inversion result 1}
    \end{subfigure}
    \hfill
    \begin{subfigure}{0.3\textwidth}
        \includegraphics[width=\textwidth]{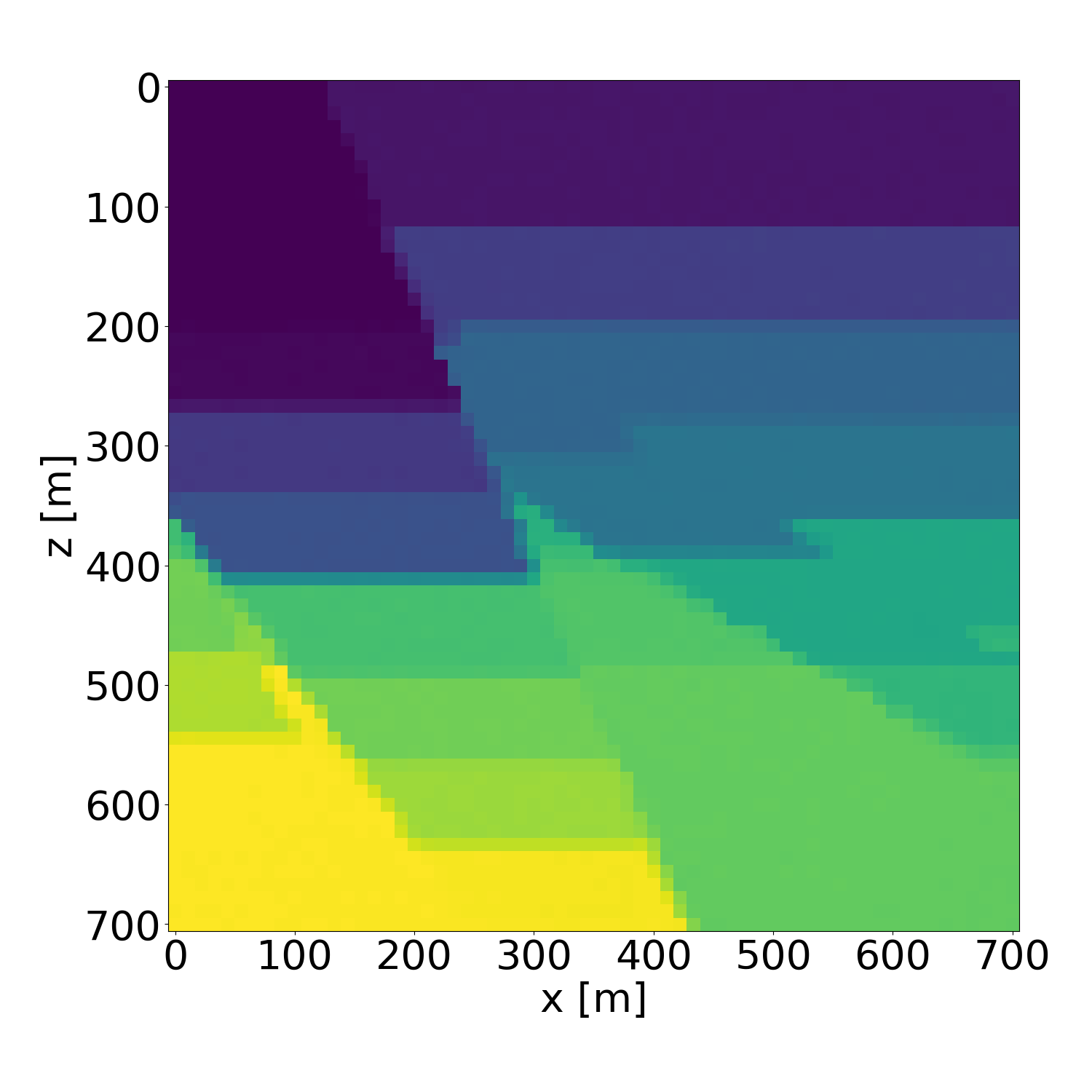}
        \caption{Probabilistic inversion result 2}
    \end{subfigure}
    \hfill
    \begin{subfigure}{0.3\textwidth}
        \includegraphics[width=\textwidth]{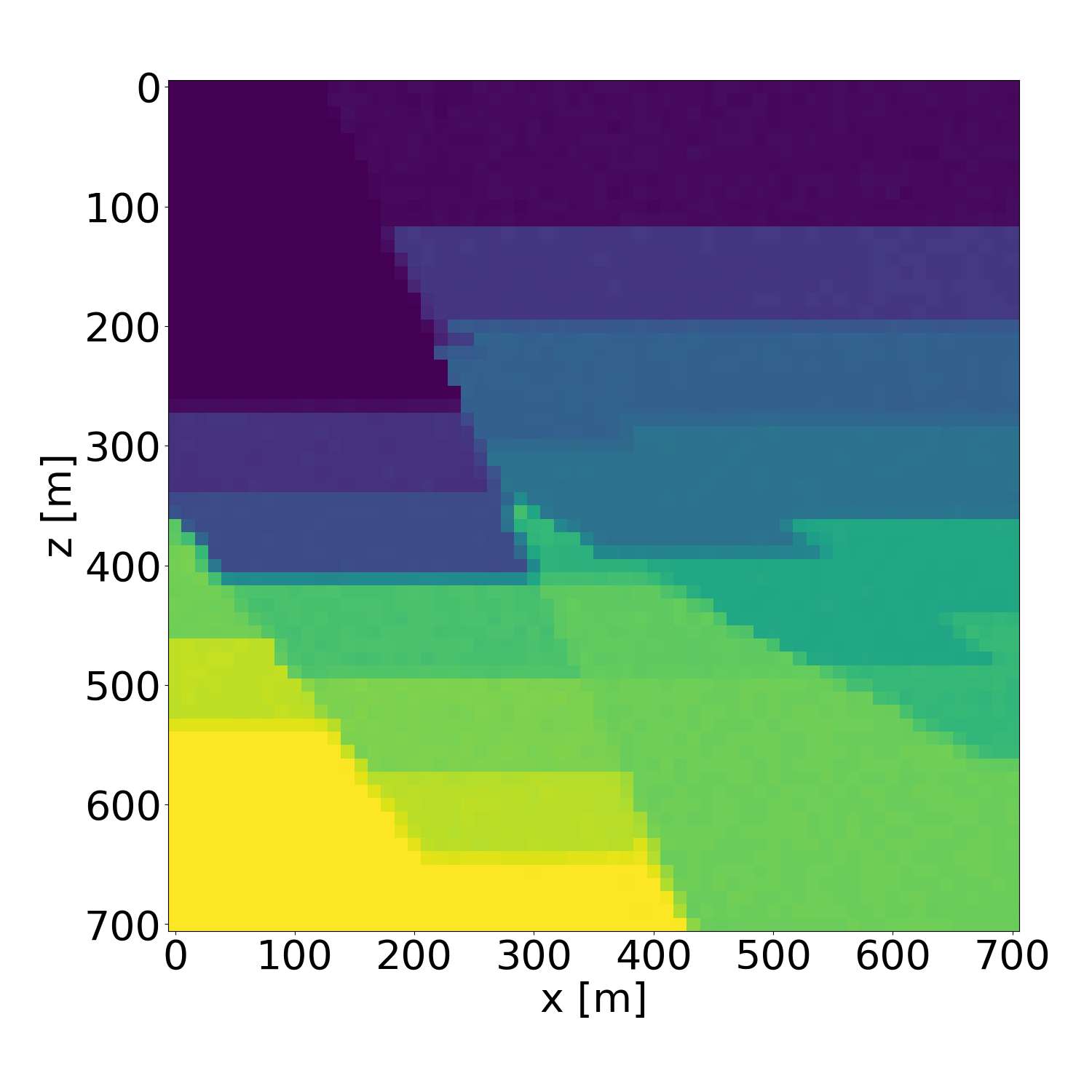}
        \caption{Probabilistic inversion result 3}
    \end{subfigure}
    \hfill
    \begin{subfigure}{0.3\textwidth}
        \includegraphics[width=\textwidth]{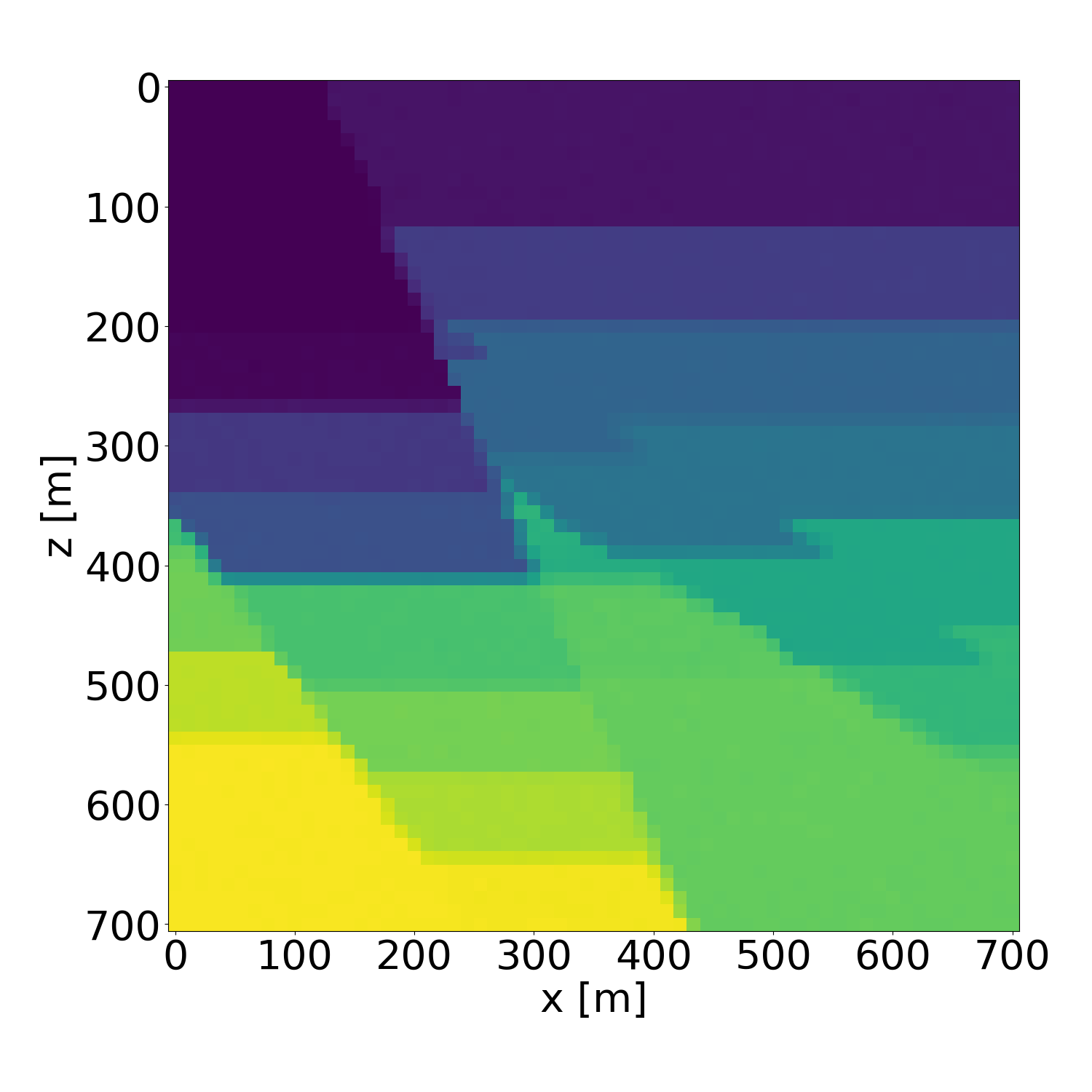}
        \caption{Probabilistic inversion result 4}
    \end{subfigure}
    \hfill
    \begin{subfigure}{0.3\textwidth}
        \includegraphics[width=\textwidth]{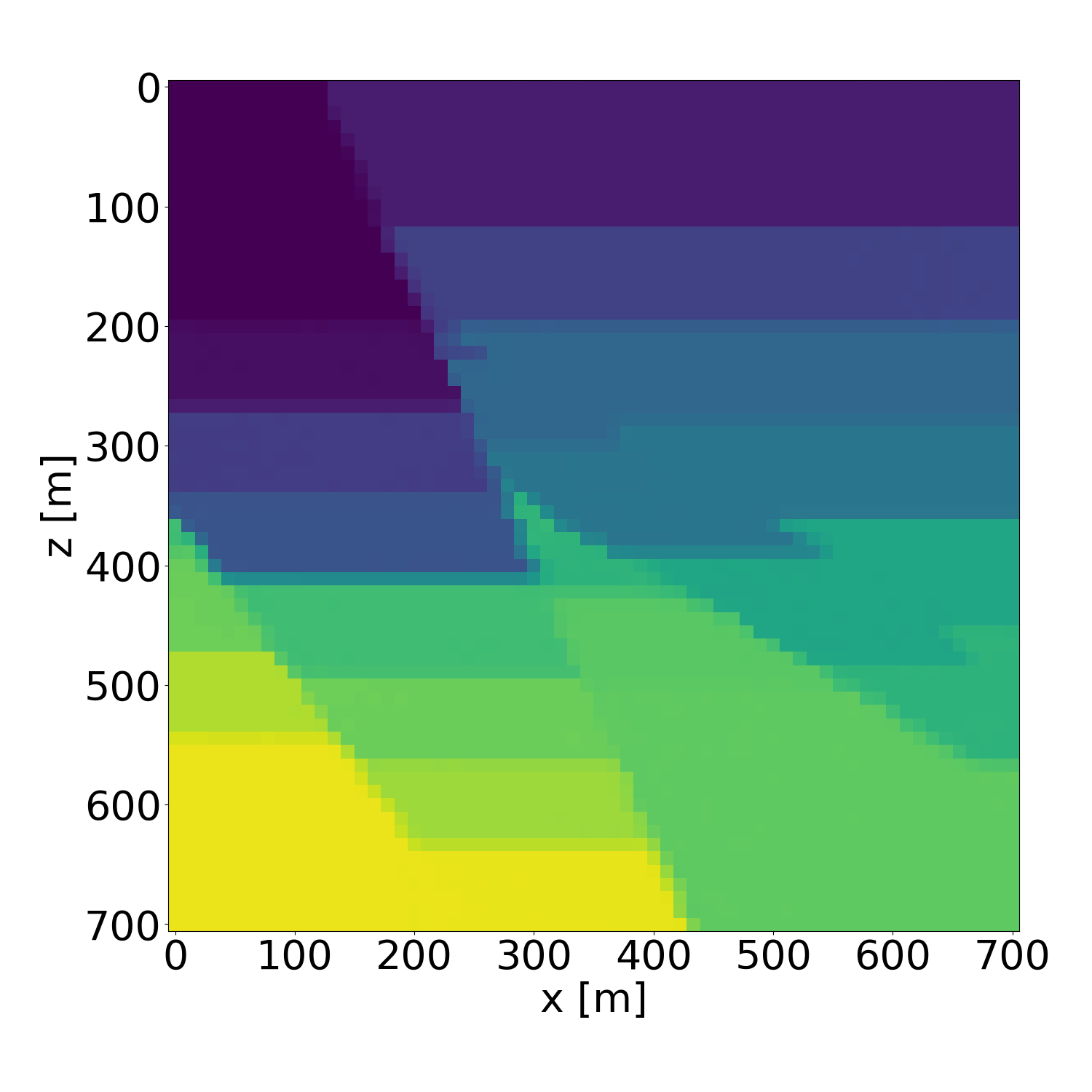}
        \caption{Probabilistic inversion result 5}
    \end{subfigure}
    \hfill
    \begin{subfigure}{0.3\textwidth}
        \includegraphics[width=\textwidth]{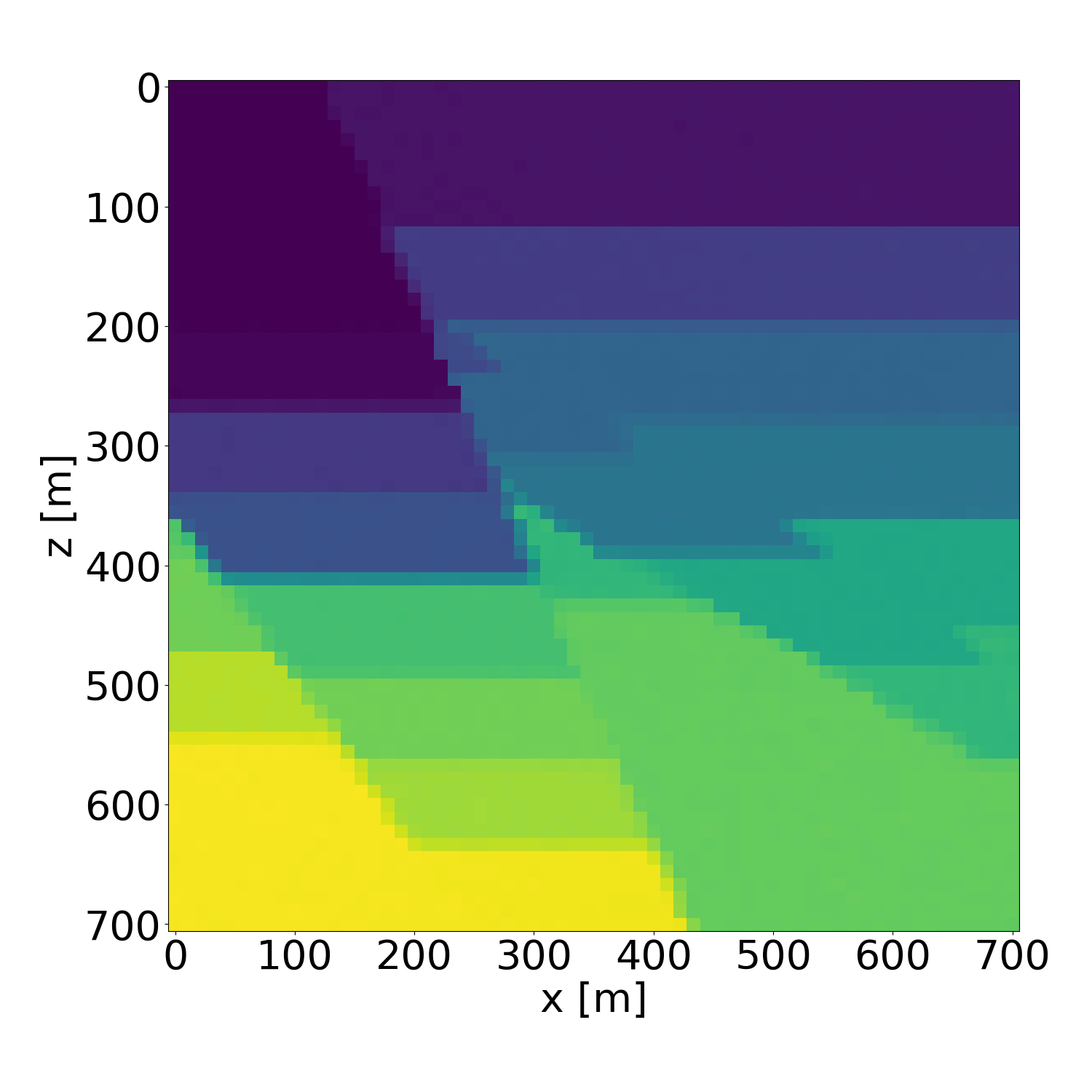}
        \caption{Probabilistic inversion result 6}
    \end{subfigure}
    \hfill
    \begin{subfigure}{0.3\textwidth}
        \includegraphics[width=\textwidth]{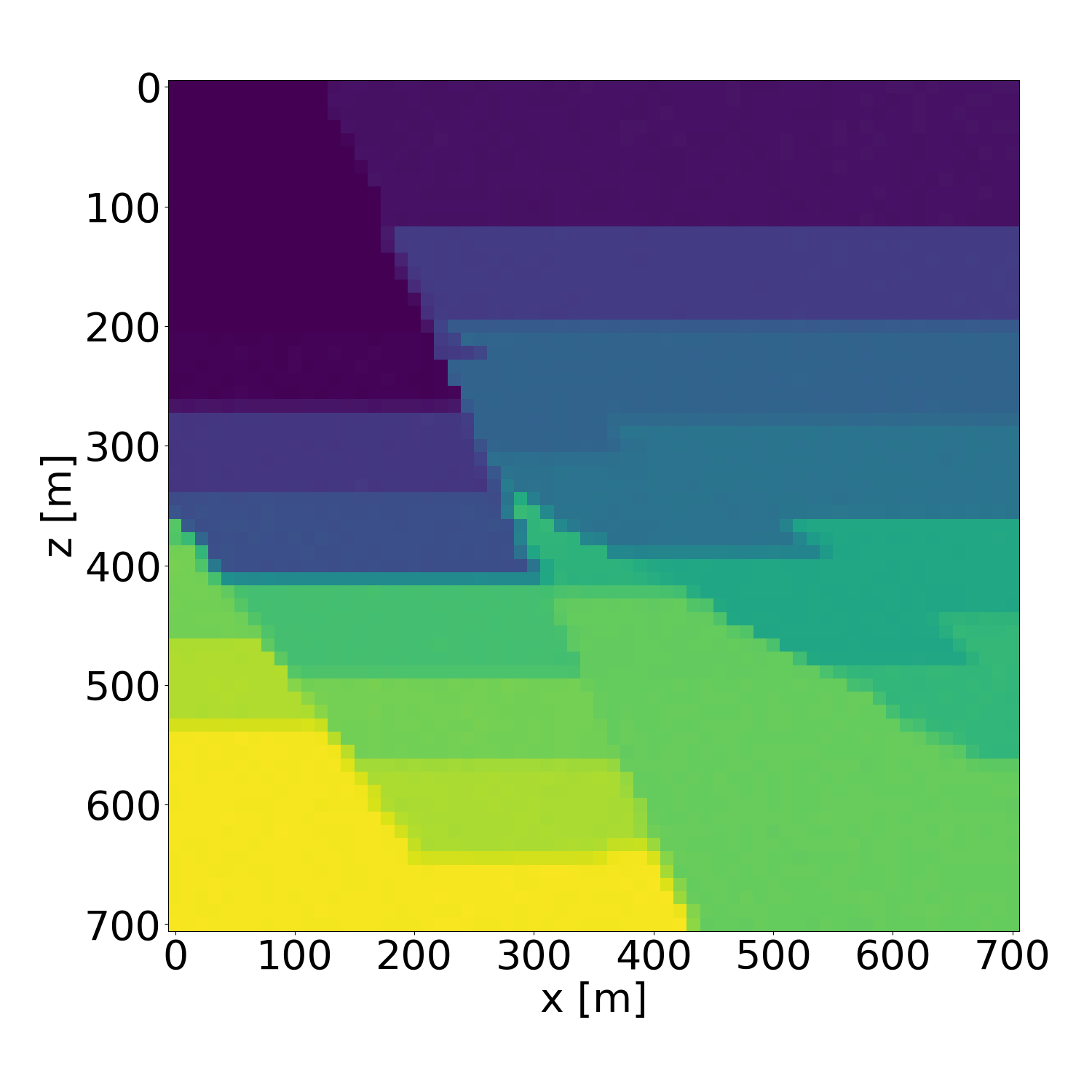}
        \caption{Probabilistic inversion result 7}
    \end{subfigure}
    \hfill
    \begin{subfigure}{0.3\textwidth}
        \includegraphics[width=\textwidth]{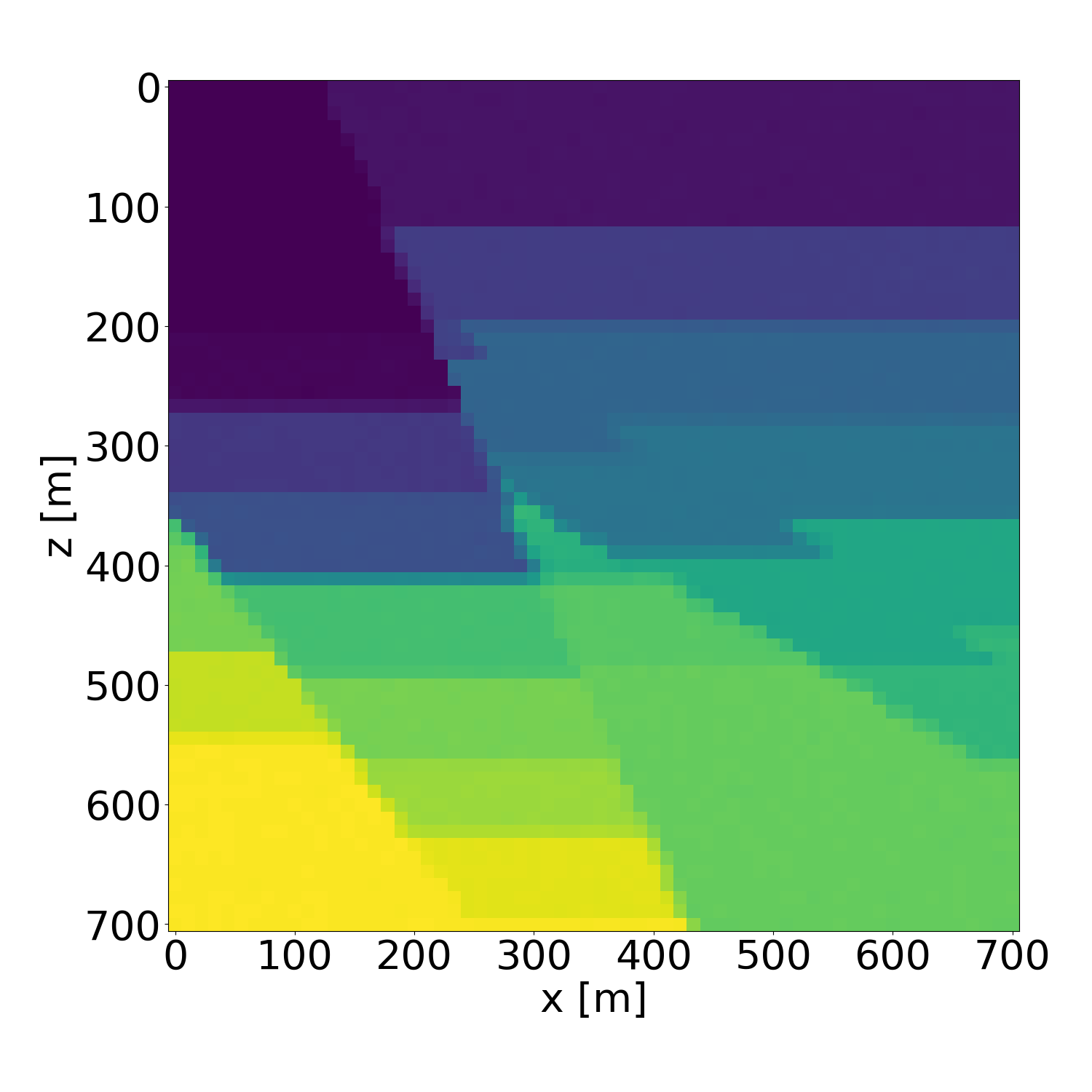}
        \caption{Probabilistic inversion result 8}
    \end{subfigure}
    \hfill
    \begin{subfigure}{0.3\textwidth}
        \includegraphics[width=\textwidth]{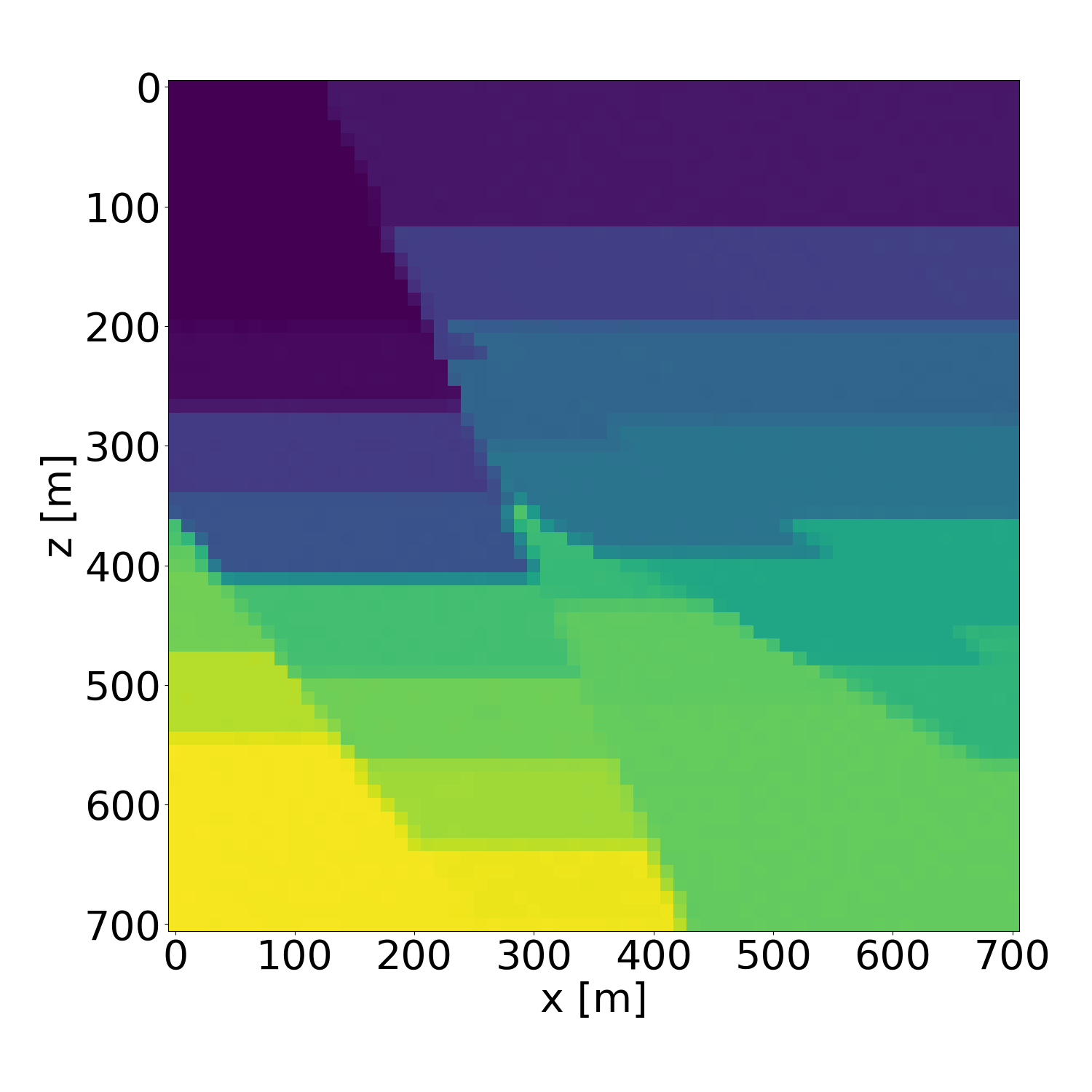}
        \caption{Probabilistic inversion result 9}
    \end{subfigure}
    \hfill
    \caption{Probabilistic inversion results of a isotropic velocity model in the validation dataset. Generated with guidance scale $w=4$.}
    \label{fig-allresult8}
\end{figure}

\begin{figure}
    \centering
    \begin{subfigure}{0.3\textwidth}
        \includegraphics[width=\textwidth]{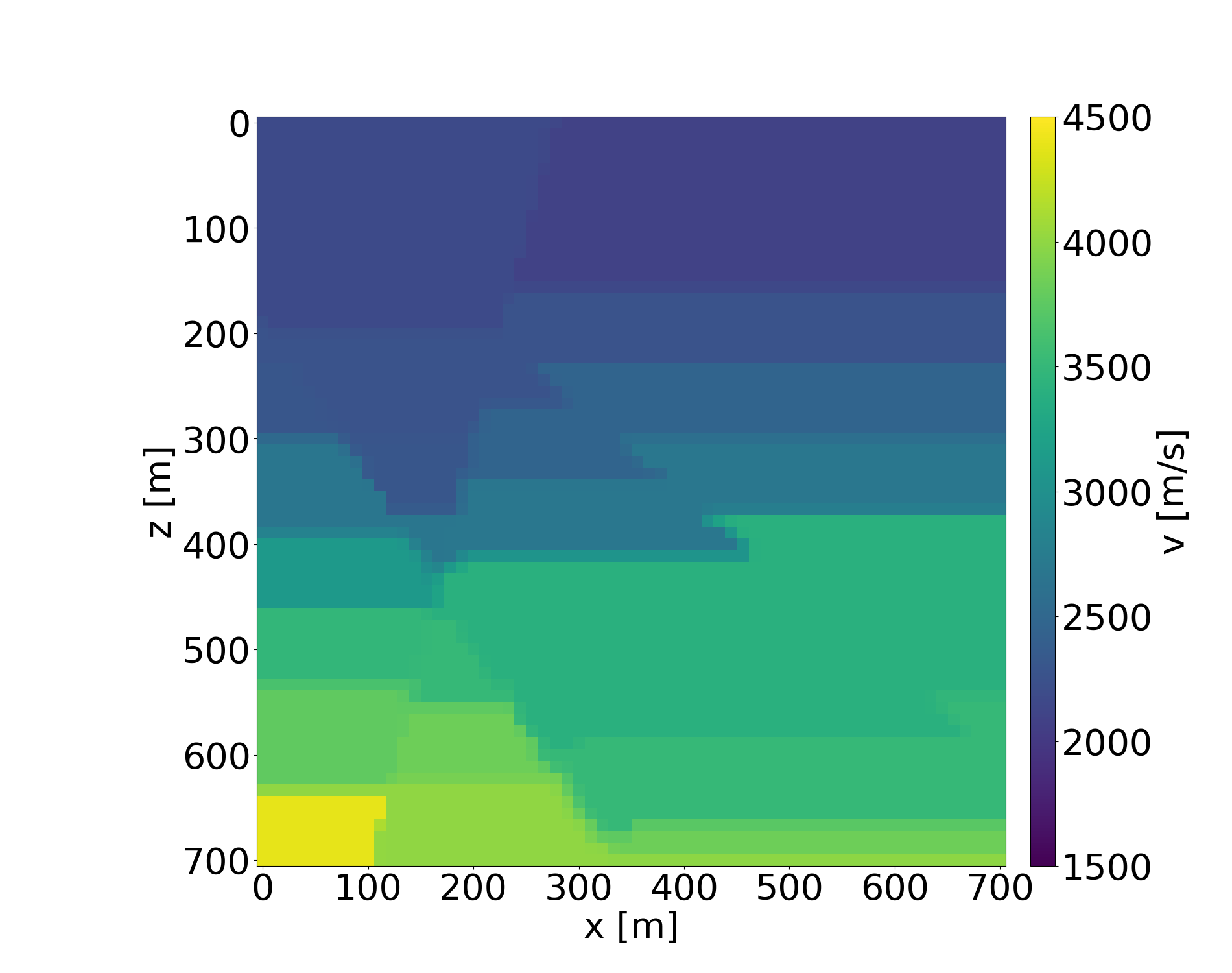}
        \caption{Ground truth (target)}
    \end{subfigure}
    \hfill
    \begin{subfigure}{0.3\textwidth}
        \includegraphics[width=\textwidth]{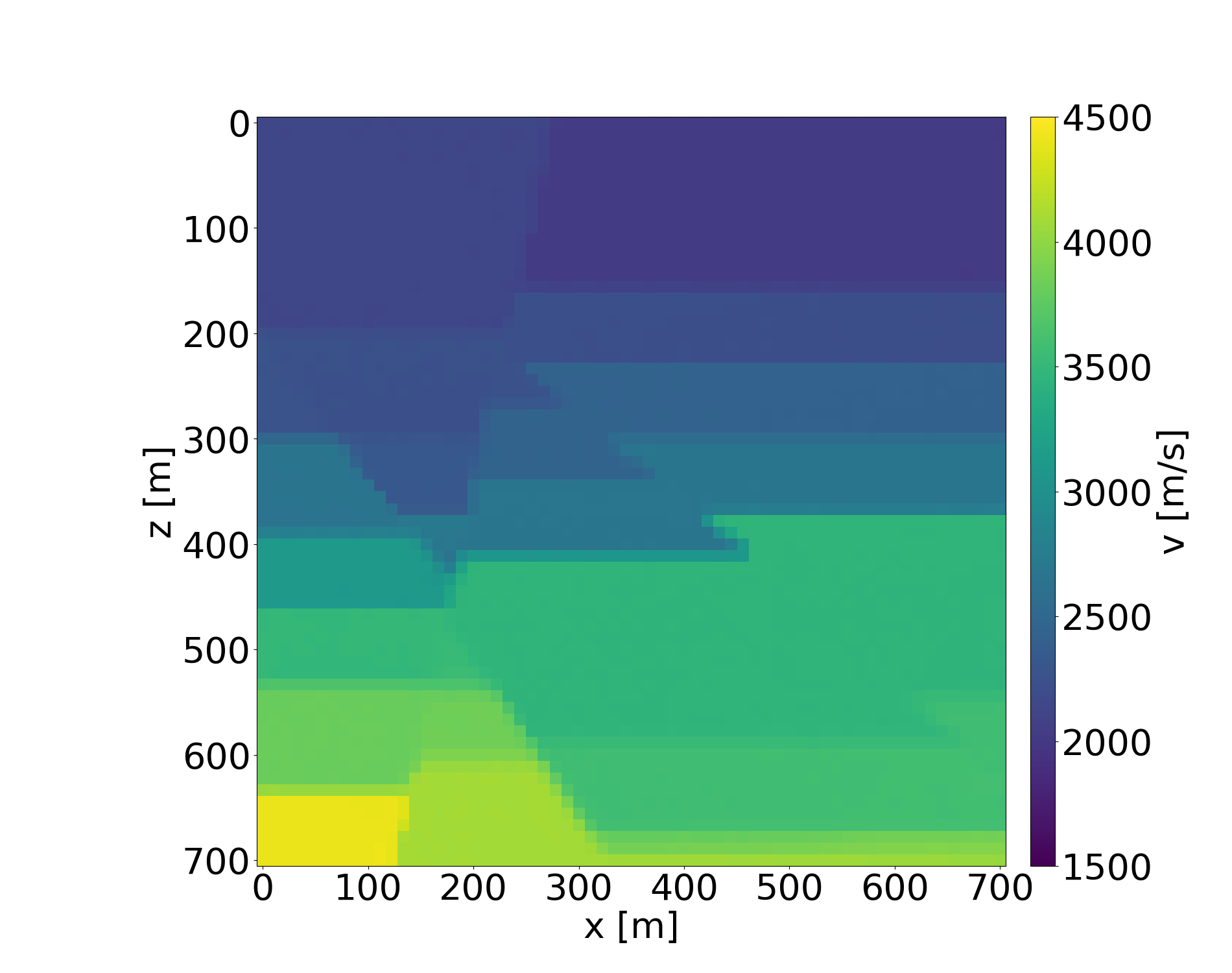}
        \caption{Average inversion result}
    \end{subfigure}
    \hfill
    \begin{subfigure}{0.3\textwidth}
        \includegraphics[width=\textwidth]{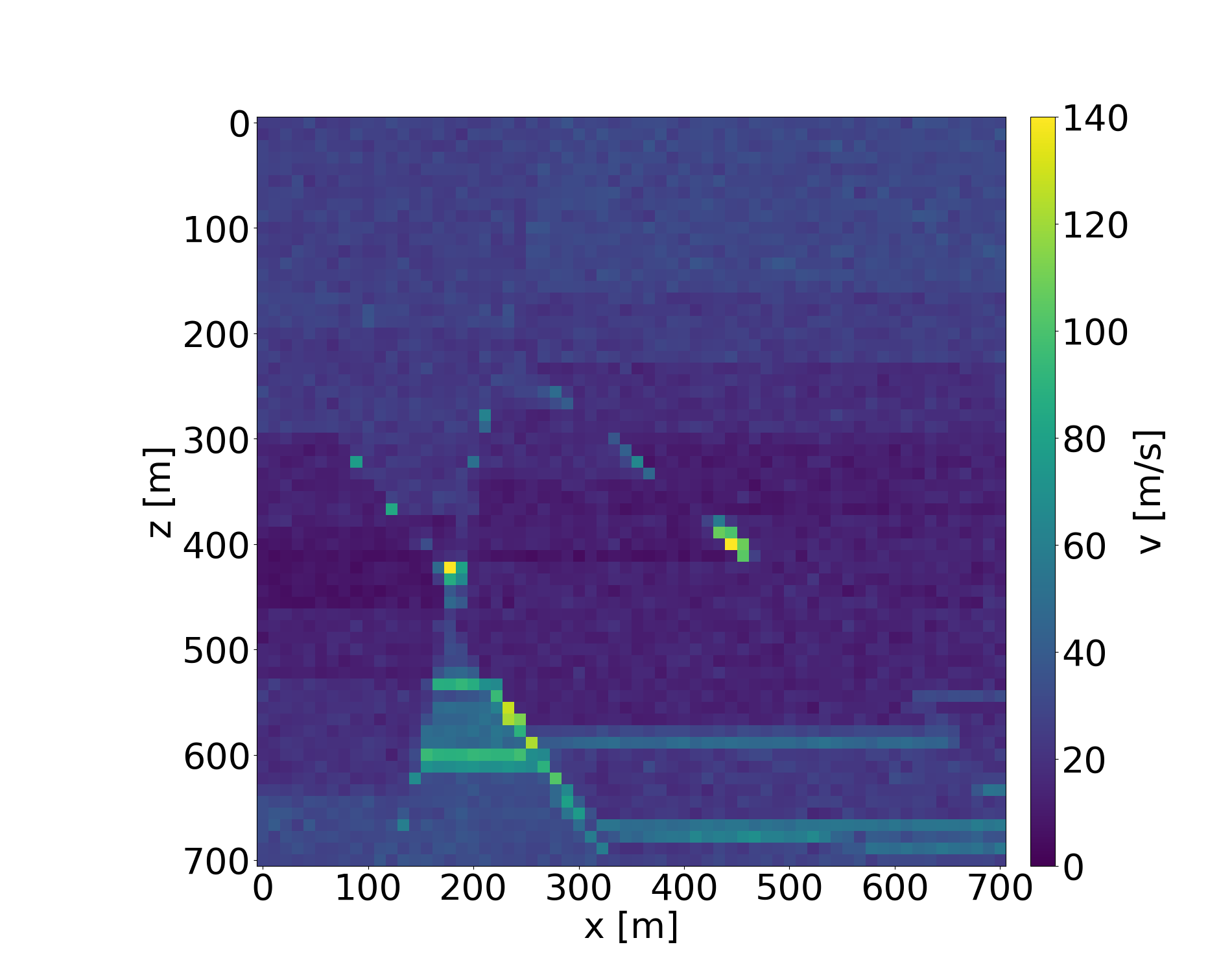}
        \caption{Standard deviation}
    \end{subfigure}
    \hfill
    \begin{subfigure}{0.3\textwidth}
        \includegraphics[width=\textwidth]{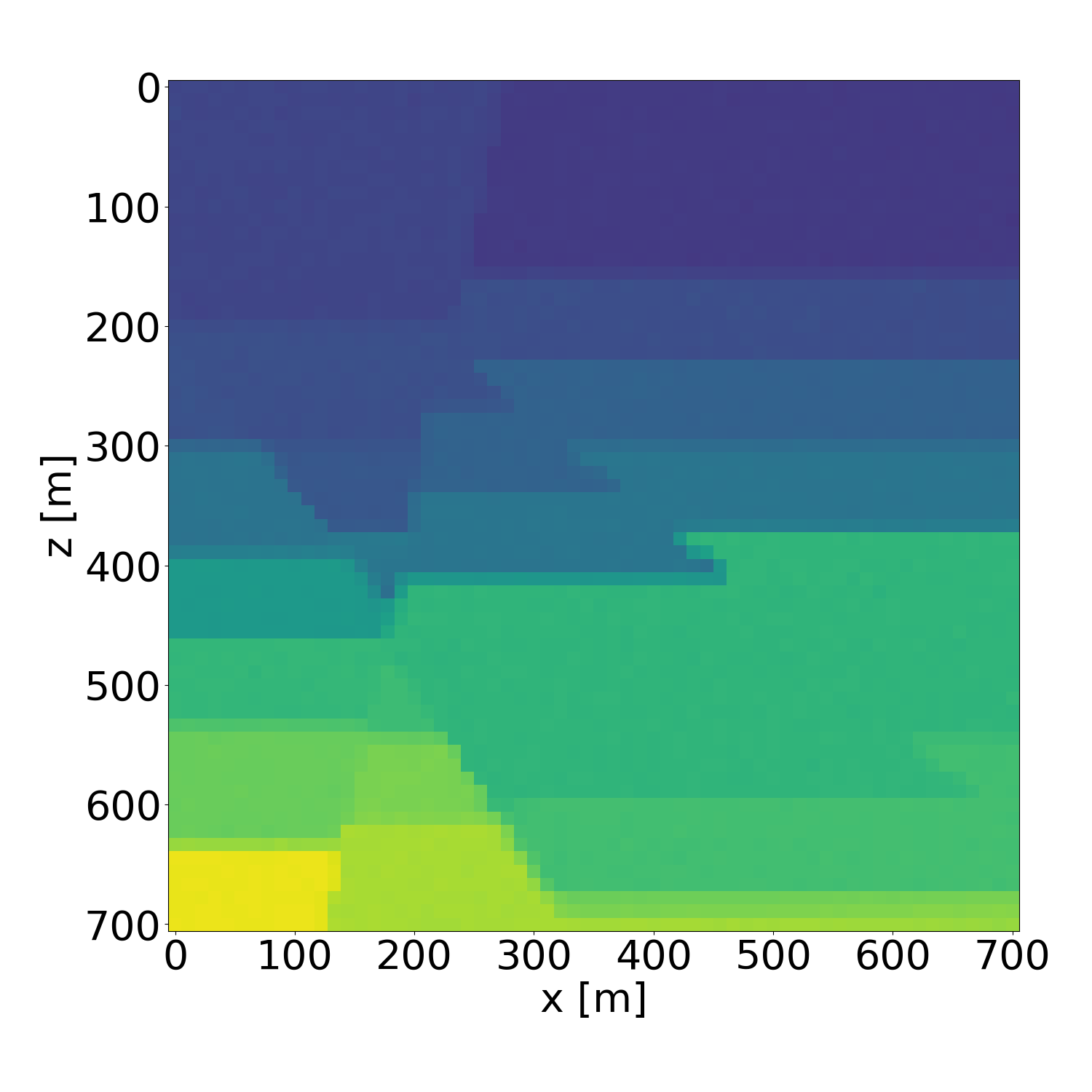}
        \caption{Probabilistic inversion result 1}
    \end{subfigure}
    \hfill
    \begin{subfigure}{0.3\textwidth}
        \includegraphics[width=\textwidth]{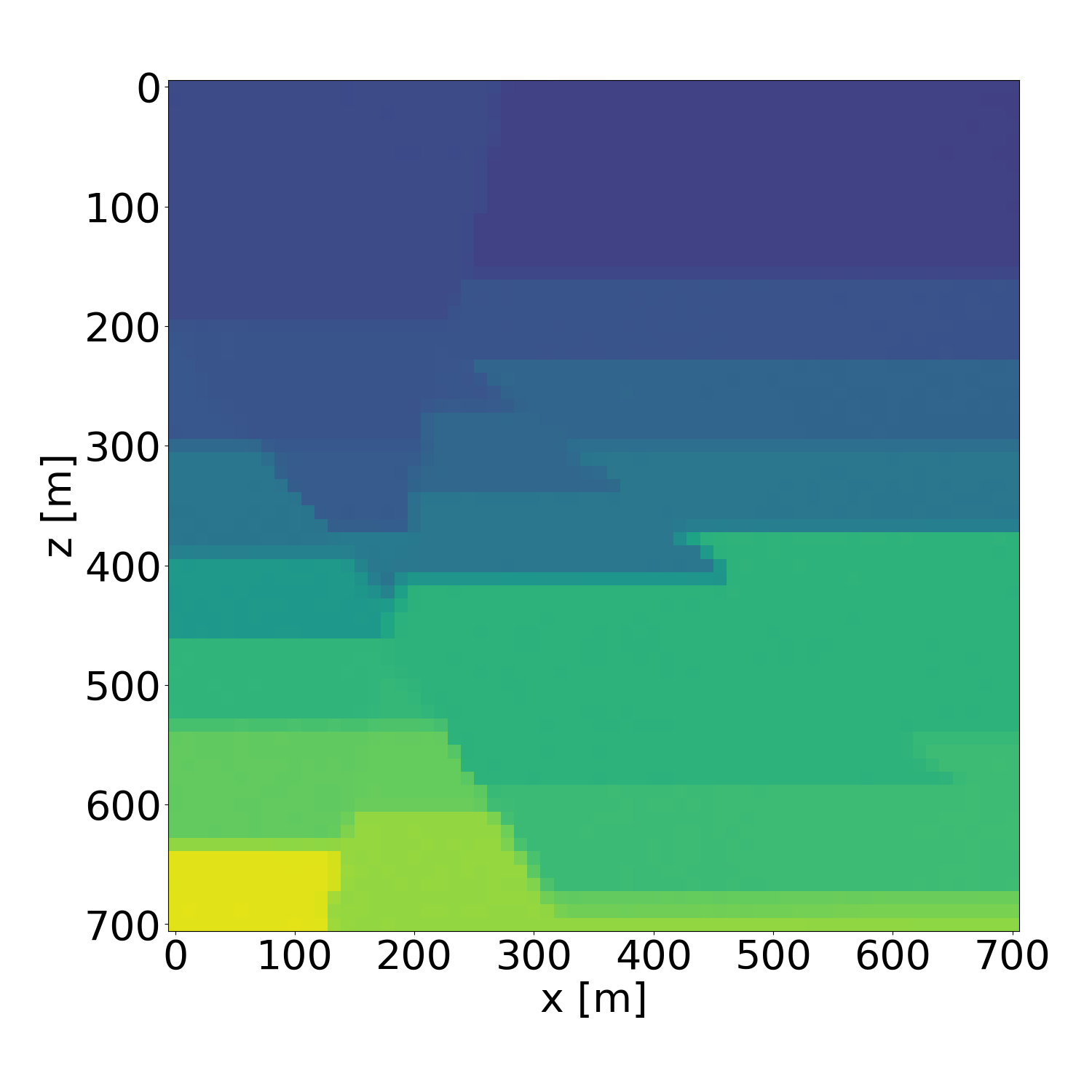}
        \caption{Probabilistic inversion result 2}
    \end{subfigure}
    \hfill
    \begin{subfigure}{0.3\textwidth}
        \includegraphics[width=\textwidth]{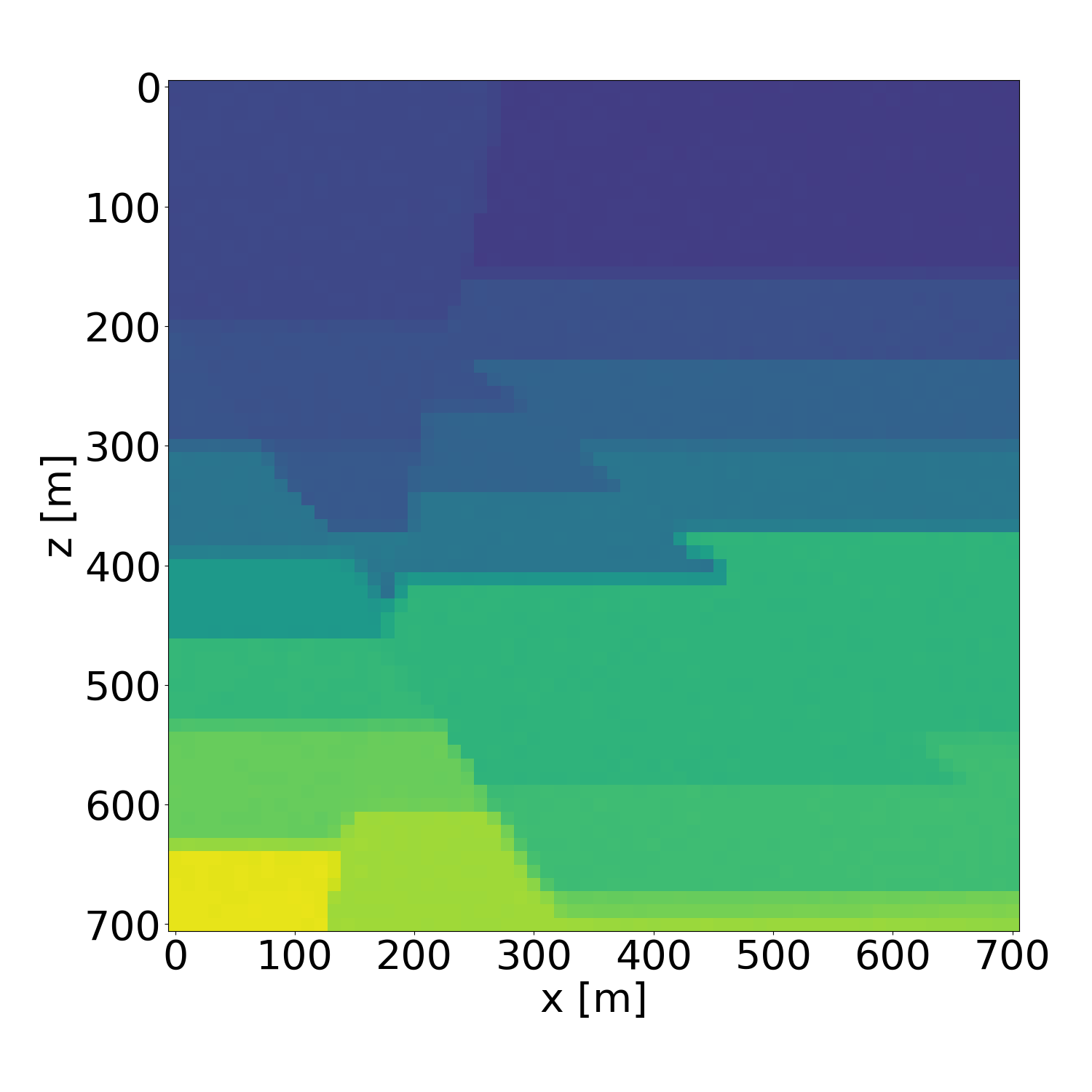}
        \caption{Probabilistic inversion result 3}
    \end{subfigure}
    \hfill
    \begin{subfigure}{0.3\textwidth}
        \includegraphics[width=\textwidth]{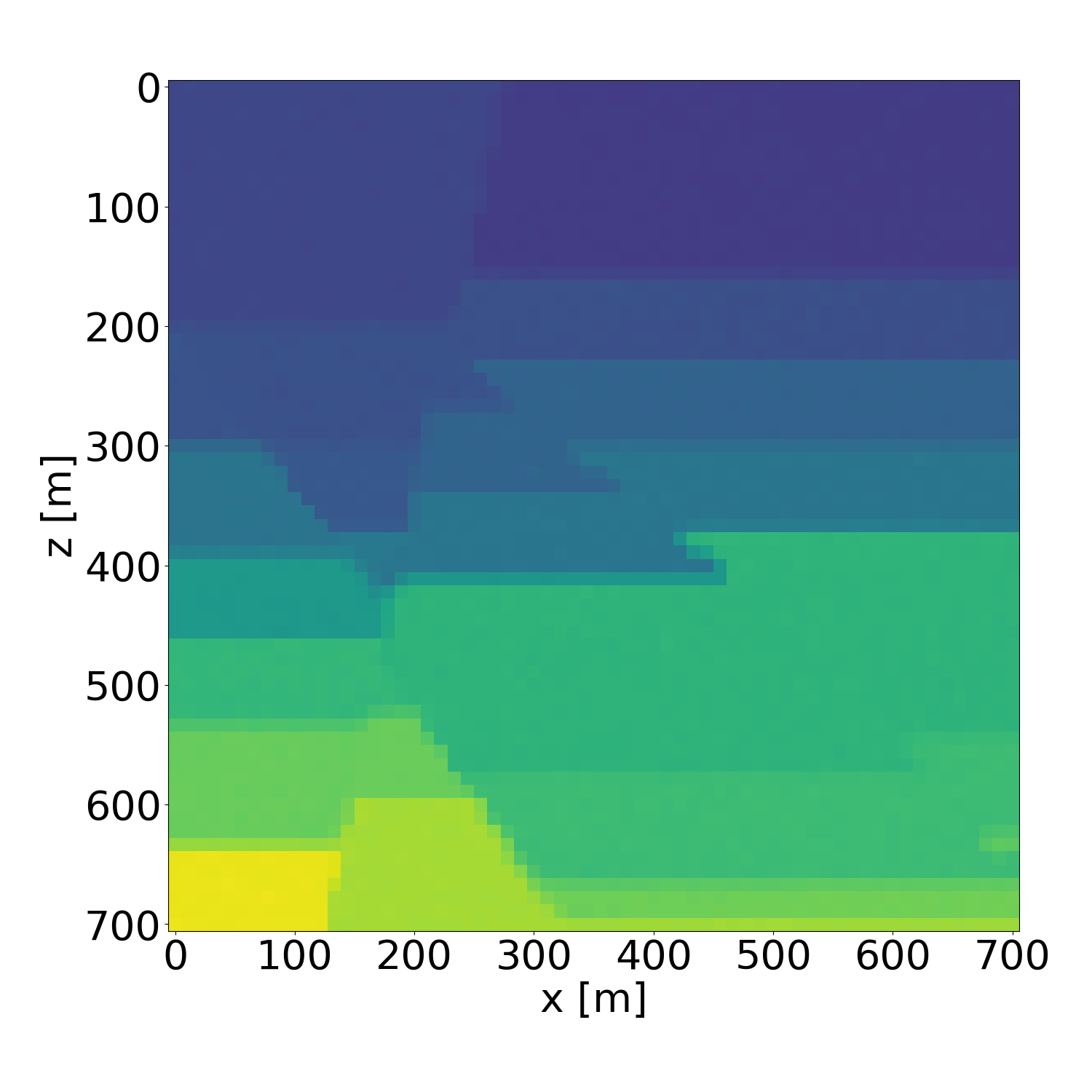}
        \caption{Probabilistic inversion result 4}
    \end{subfigure}
    \hfill
    \begin{subfigure}{0.3\textwidth}
        \includegraphics[width=\textwidth]{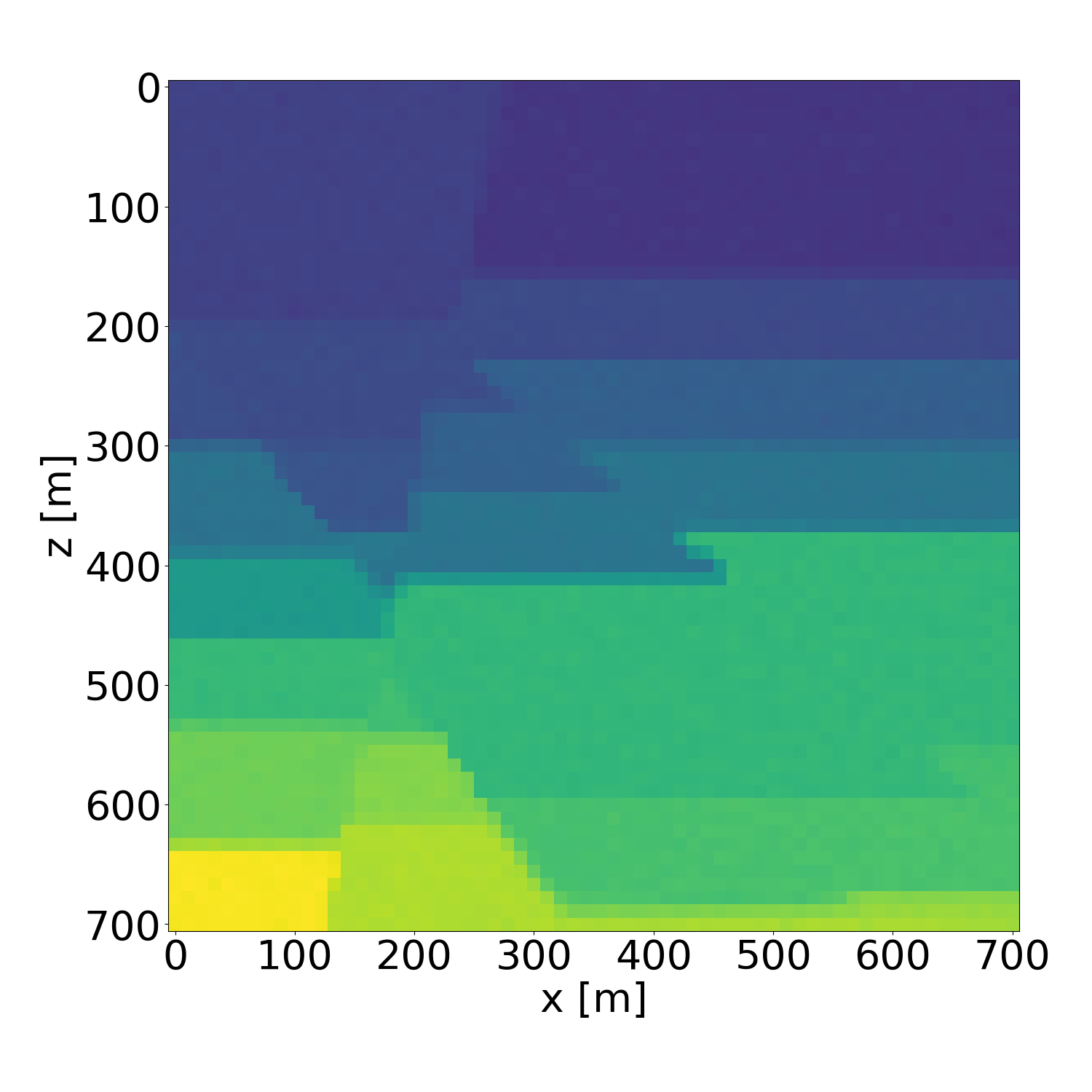}
        \caption{Probabilistic inversion result 5}
    \end{subfigure}
    \hfill
    \begin{subfigure}{0.3\textwidth}
        \includegraphics[width=\textwidth]{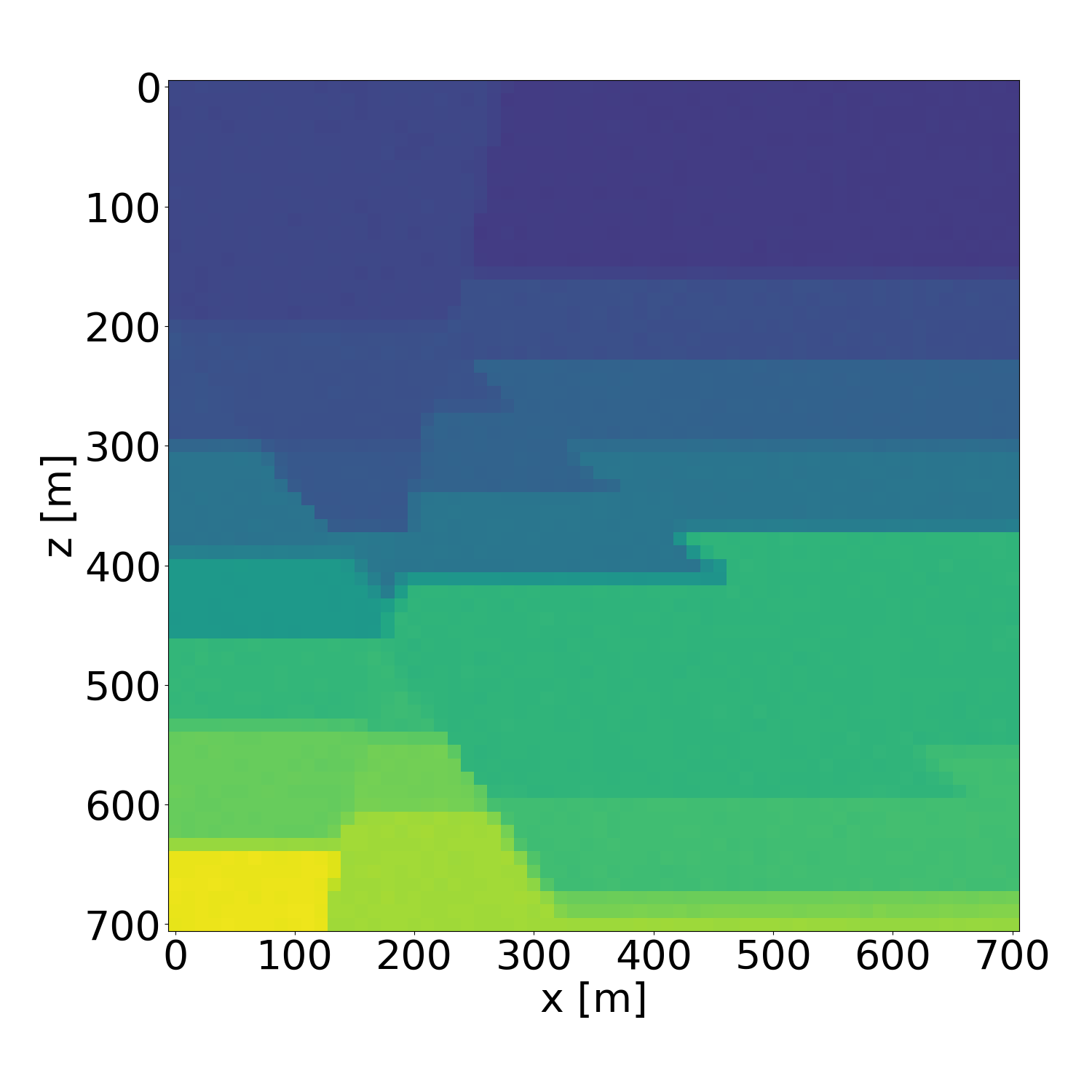}
        \caption{Probabilistic inversion result 6}
    \end{subfigure}
    \hfill
    \begin{subfigure}{0.3\textwidth}
        \includegraphics[width=\textwidth]{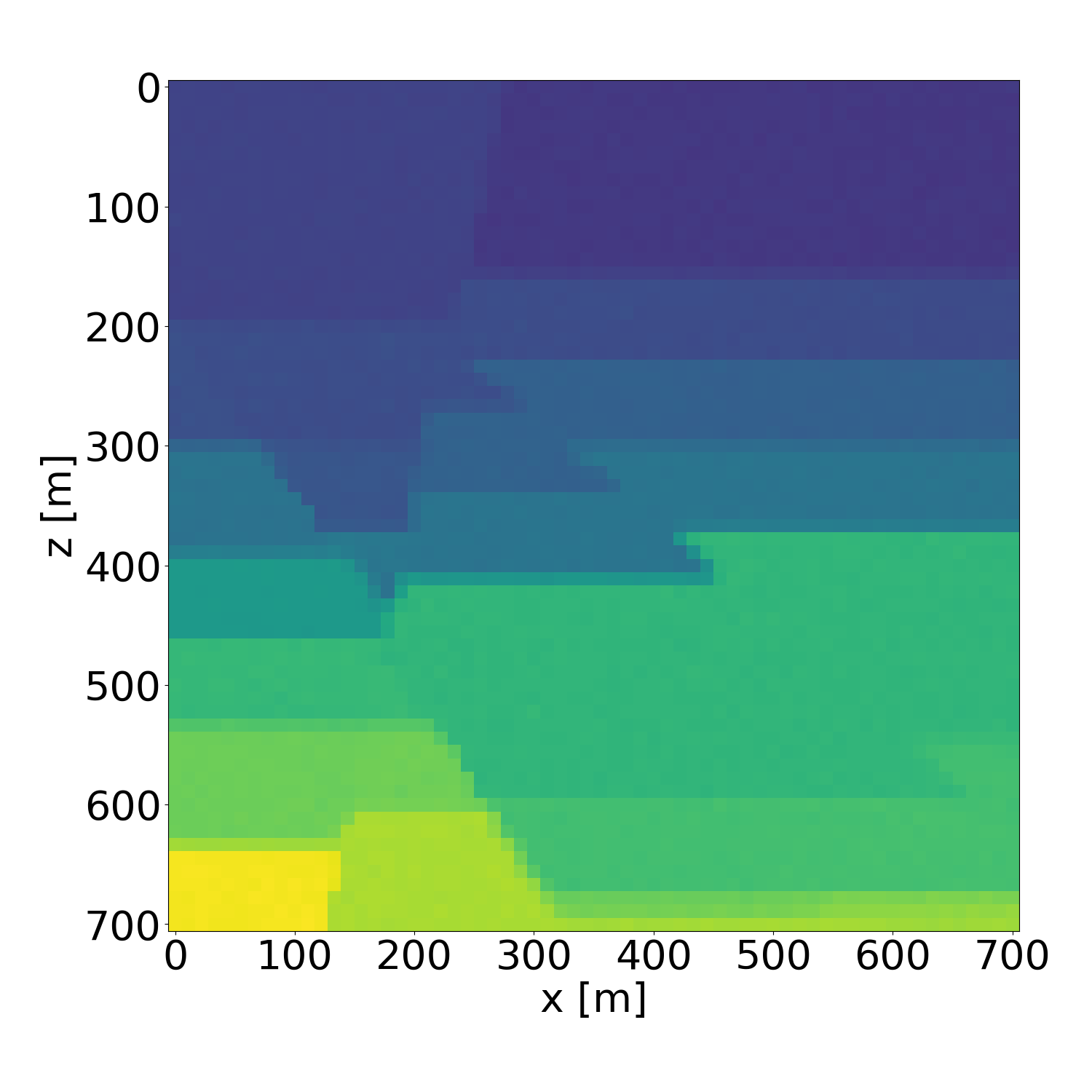}
        \caption{Probabilistic inversion result 7}
    \end{subfigure}
    \hfill
    \begin{subfigure}{0.3\textwidth}
        \includegraphics[width=\textwidth]{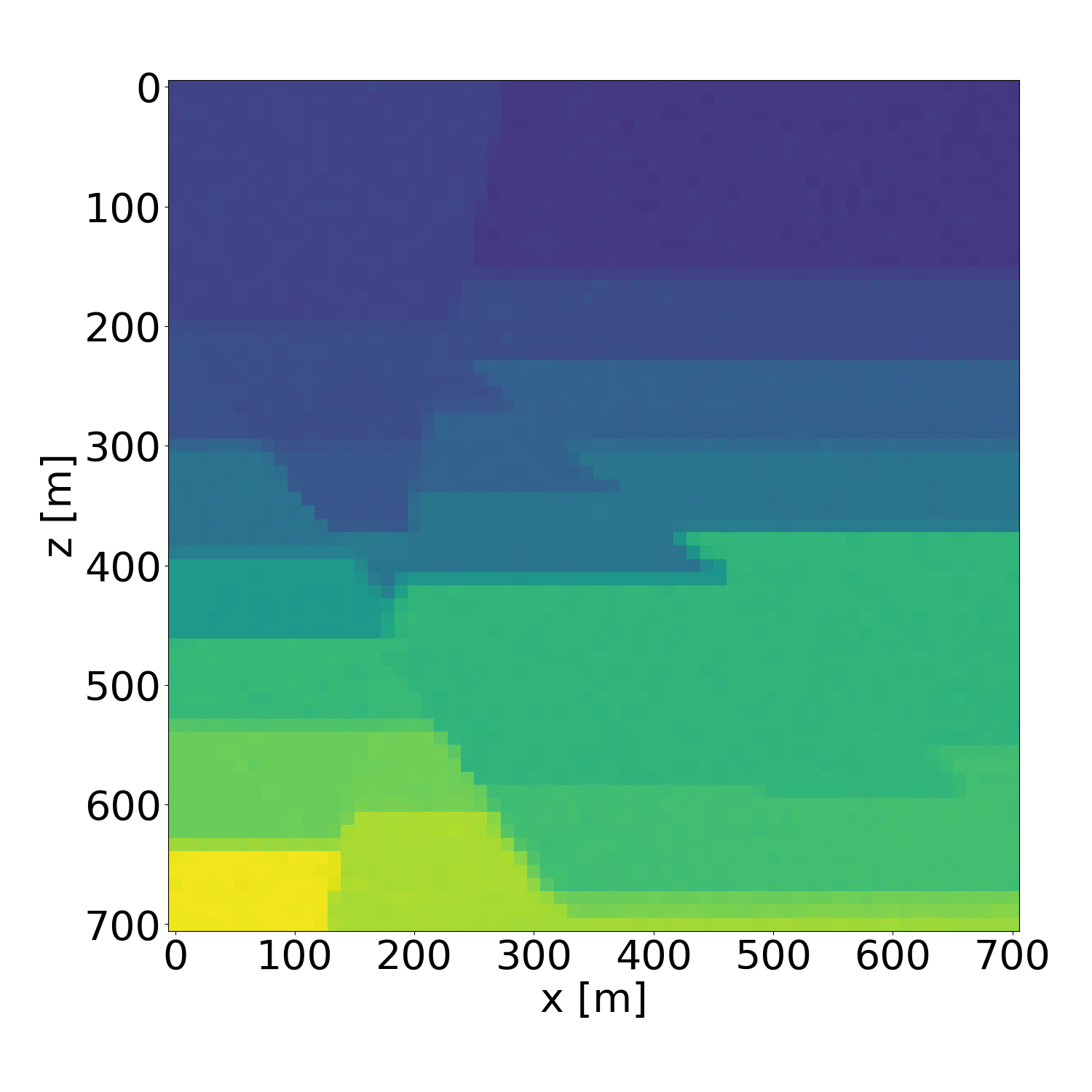}
        \caption{Probabilistic inversion result 8}
    \end{subfigure}
    \hfill
    \begin{subfigure}{0.3\textwidth}
        \includegraphics[width=\textwidth]{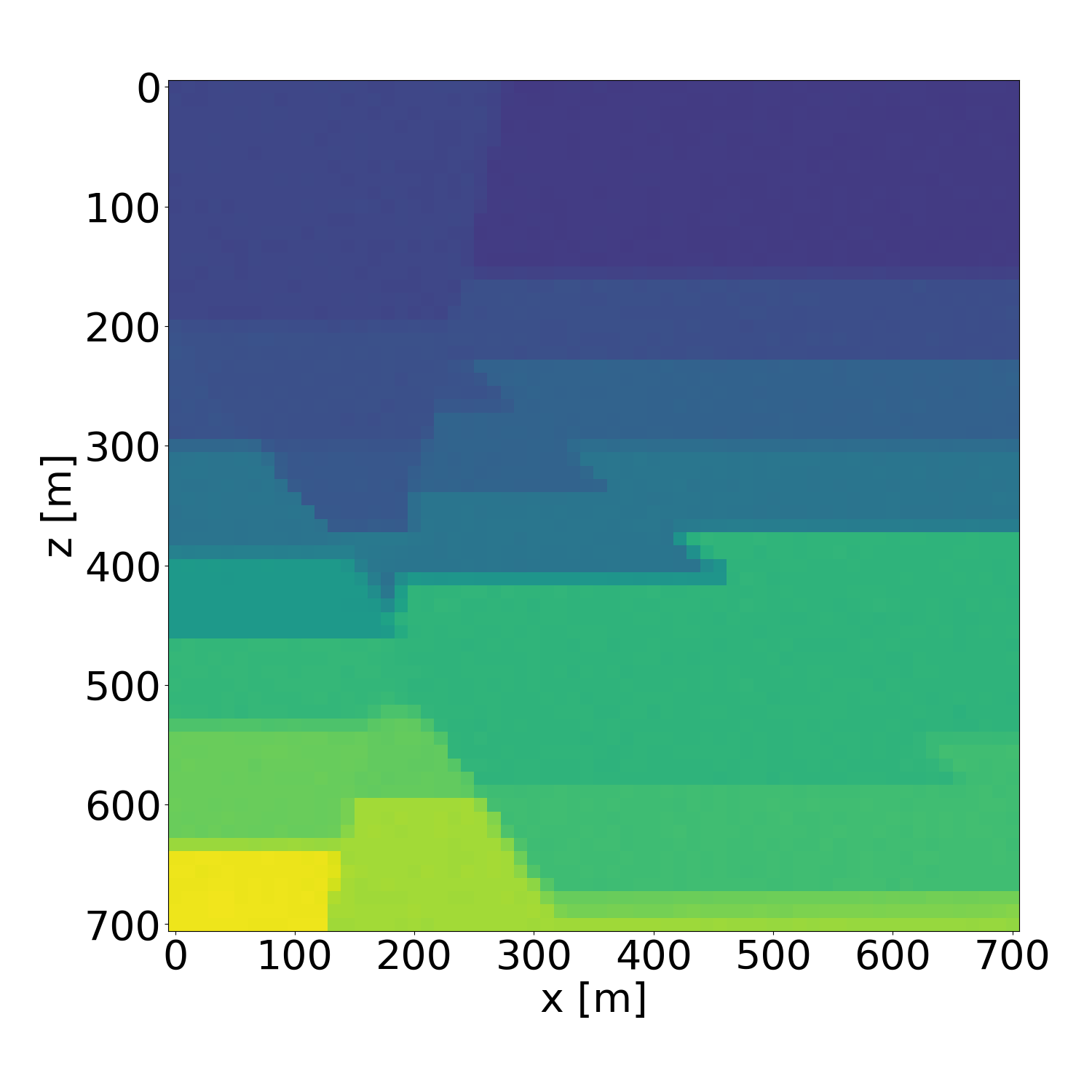}
        \caption{Probabilistic inversion result 9}
    \end{subfigure}
    \hfill
    \caption{Probabilistic inversion results of a isotropic velocity model in the validation dataset. Generated with guidance scale $w=4$.}
    \label{fig-allresult9}
\end{figure}

\begin{figure}
    \centering
    \begin{subfigure}{0.3\textwidth}
        \includegraphics[width=\textwidth]{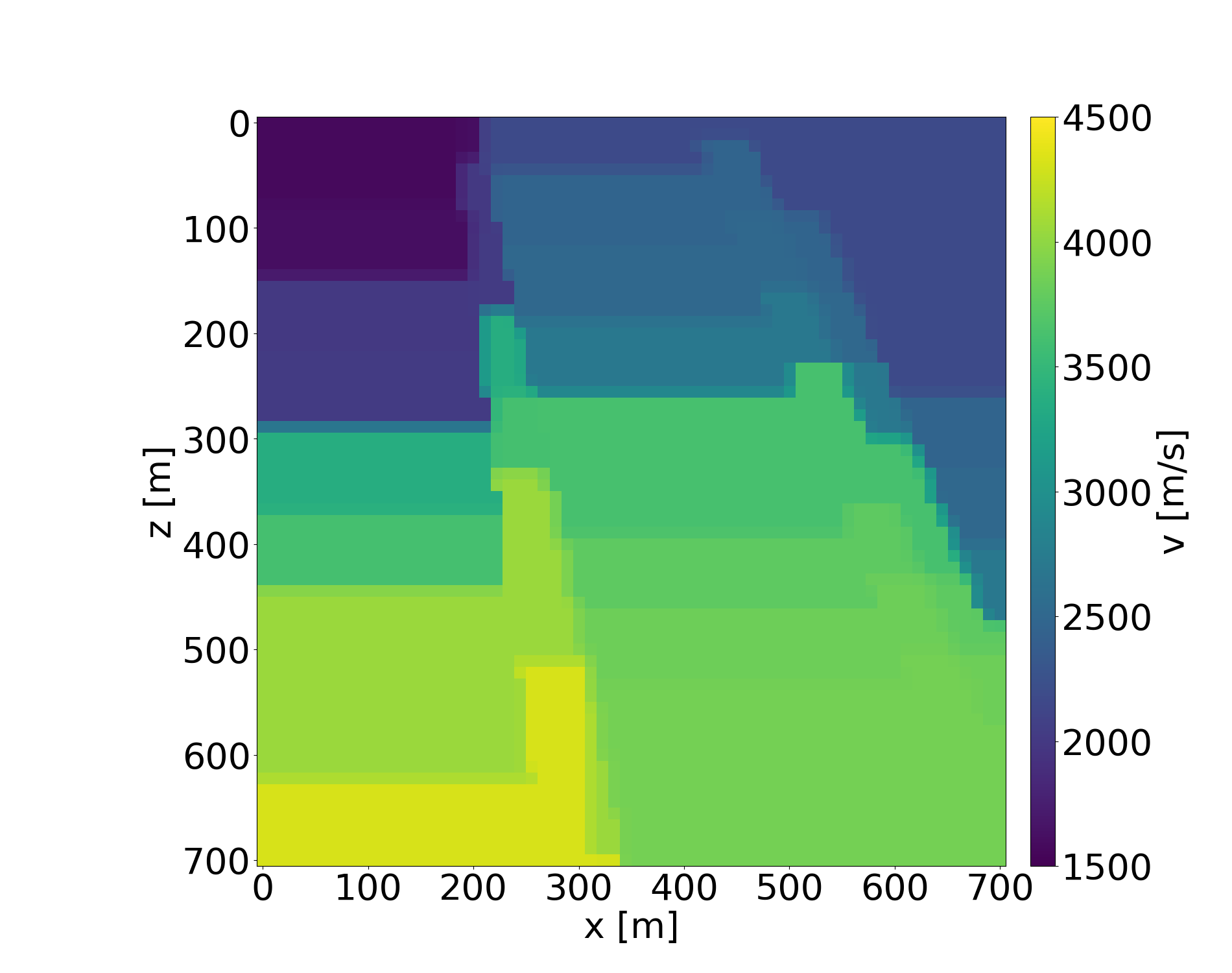}
        \caption{Ground truth (target)}
    \end{subfigure}
    \hfill
    \begin{subfigure}{0.3\textwidth}
        \includegraphics[width=\textwidth]{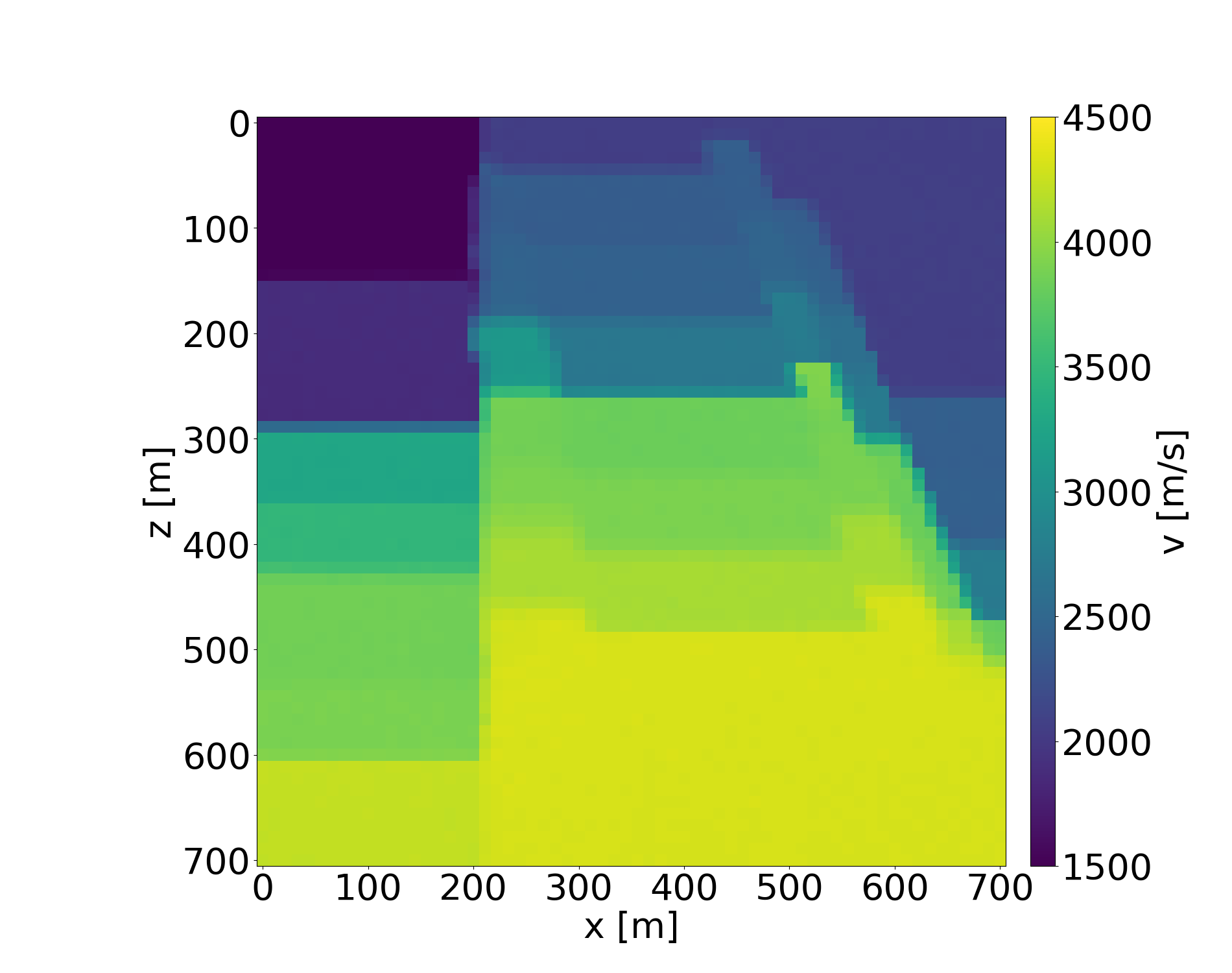}
        \caption{Average inversion result}
    \end{subfigure}
    \hfill
    \begin{subfigure}{0.3\textwidth}
        \includegraphics[width=\textwidth]{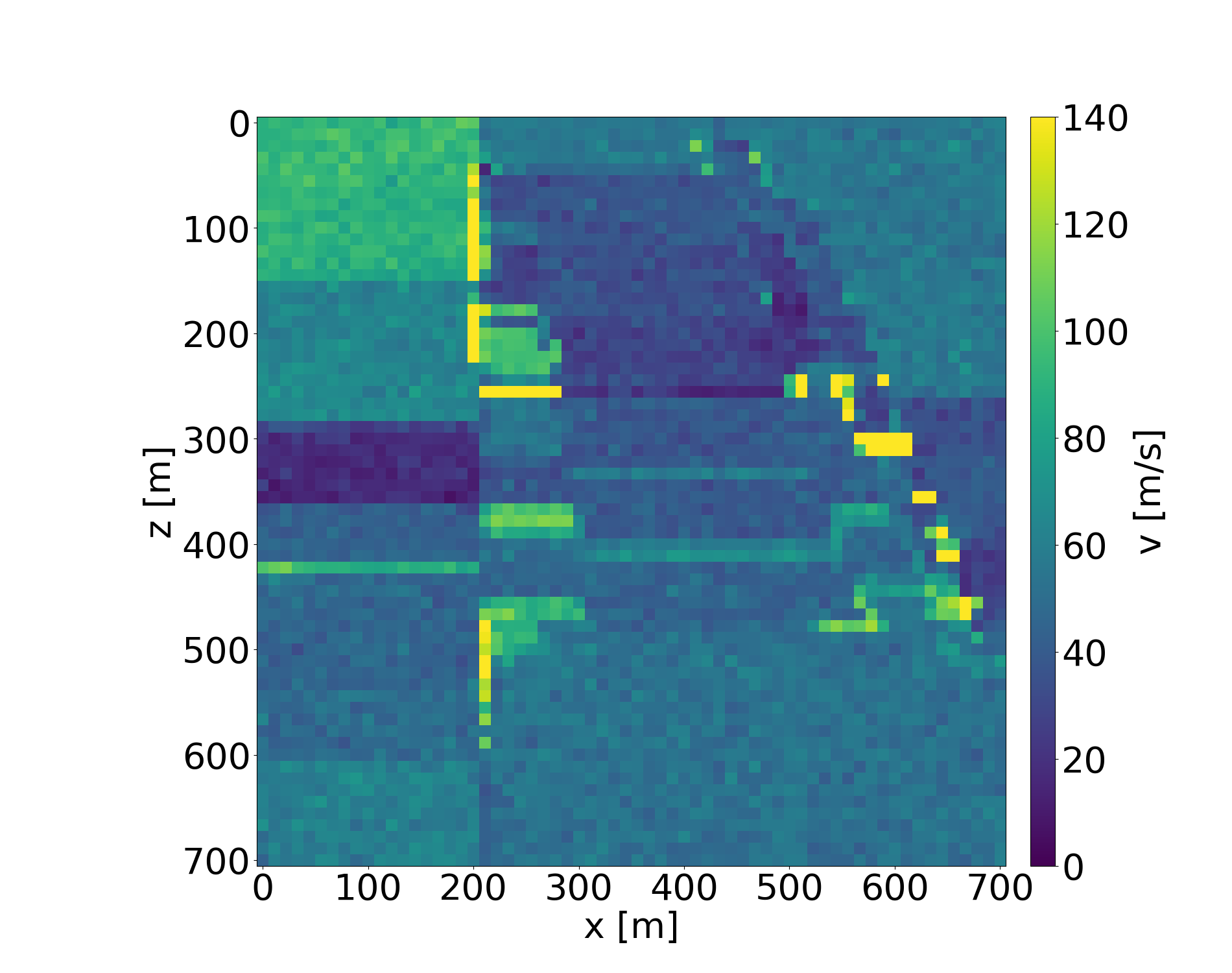}
        \caption{Standard deviation}
    \end{subfigure}
    \hfill
    \begin{subfigure}{0.3\textwidth}
        \includegraphics[width=\textwidth]{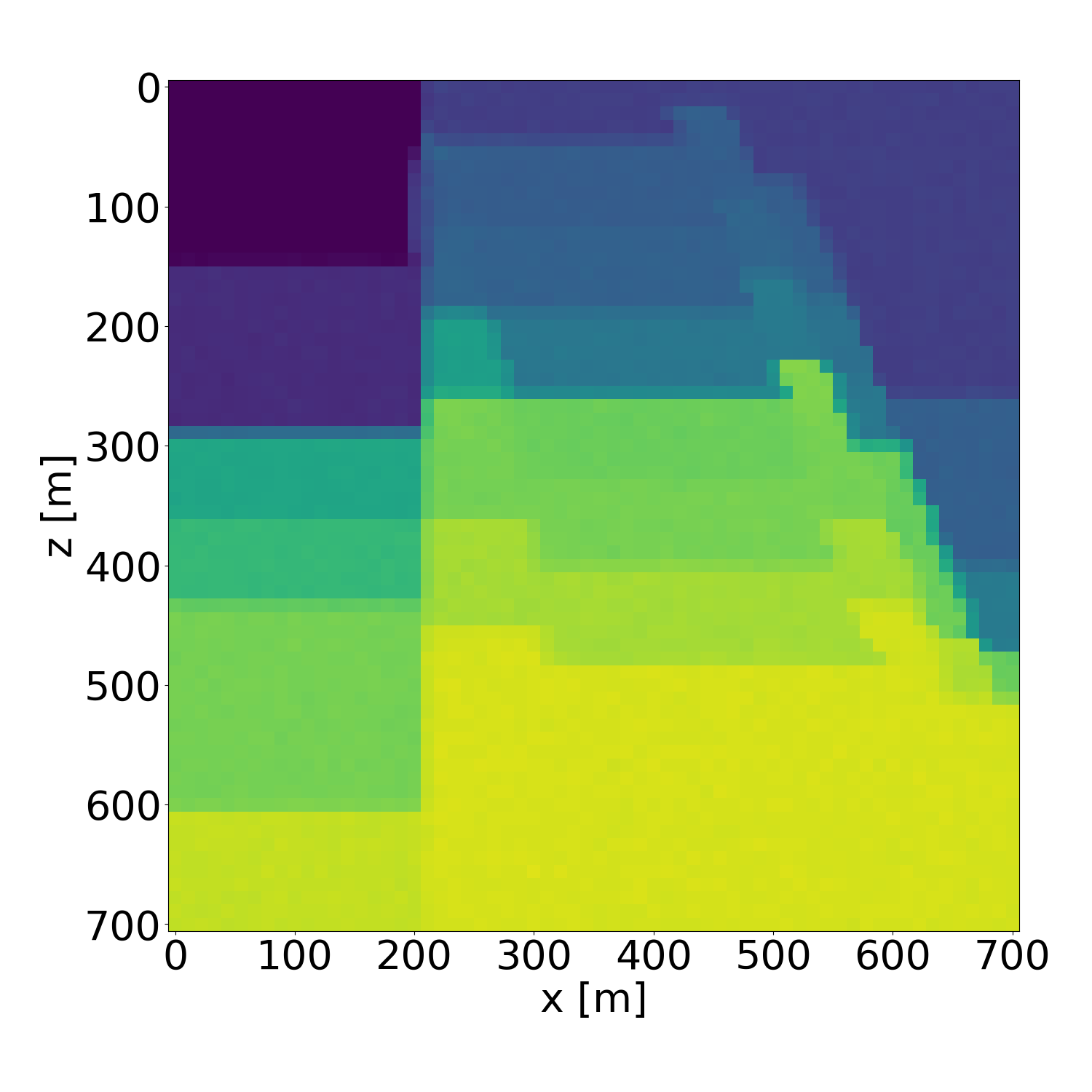}
        \caption{Probabilistic inversion result 1}
    \end{subfigure}
    \hfill
    \begin{subfigure}{0.3\textwidth}
        \includegraphics[width=\textwidth]{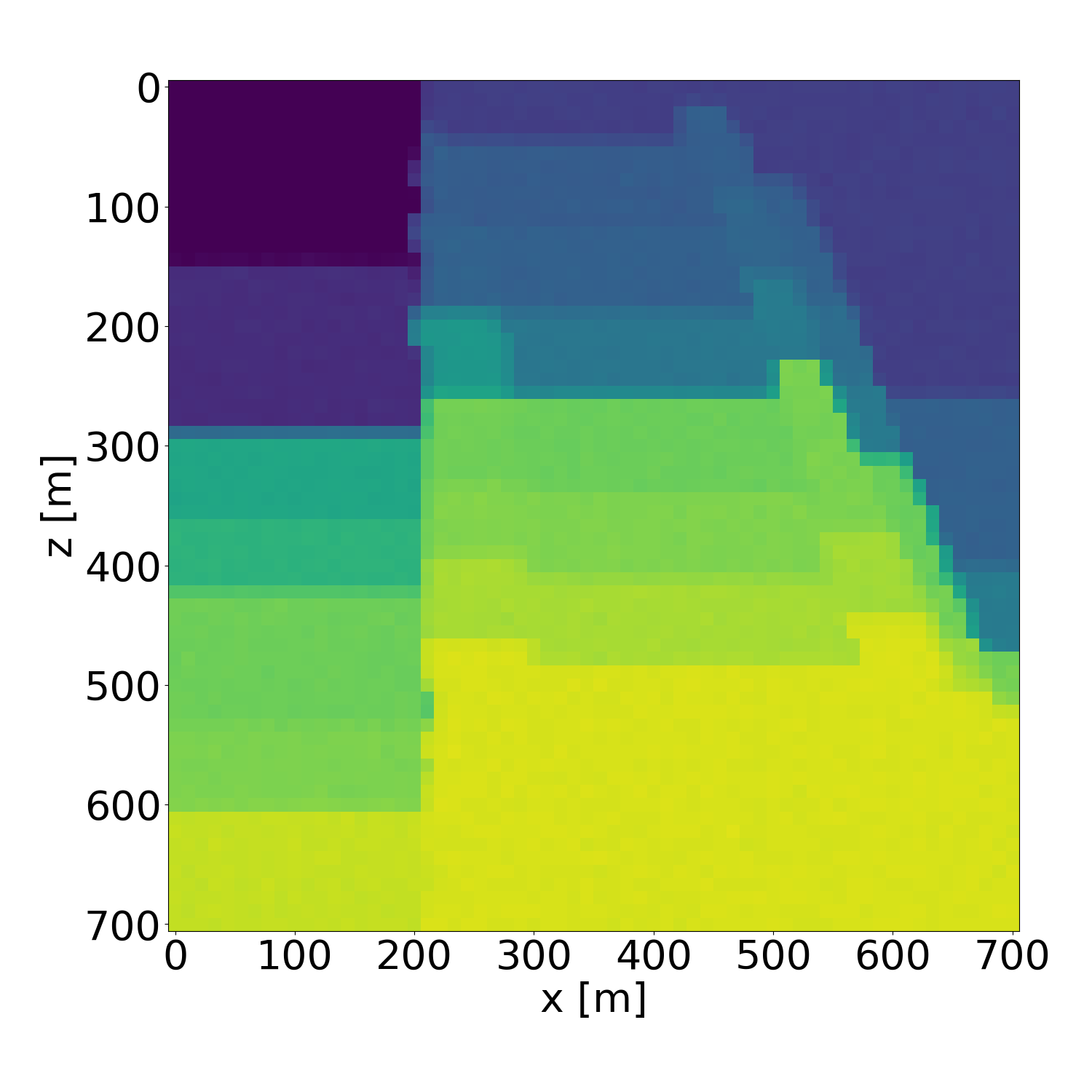}
        \caption{Probabilistic inversion result 2}
    \end{subfigure}
    \hfill
    \begin{subfigure}{0.3\textwidth}
        \includegraphics[width=\textwidth]{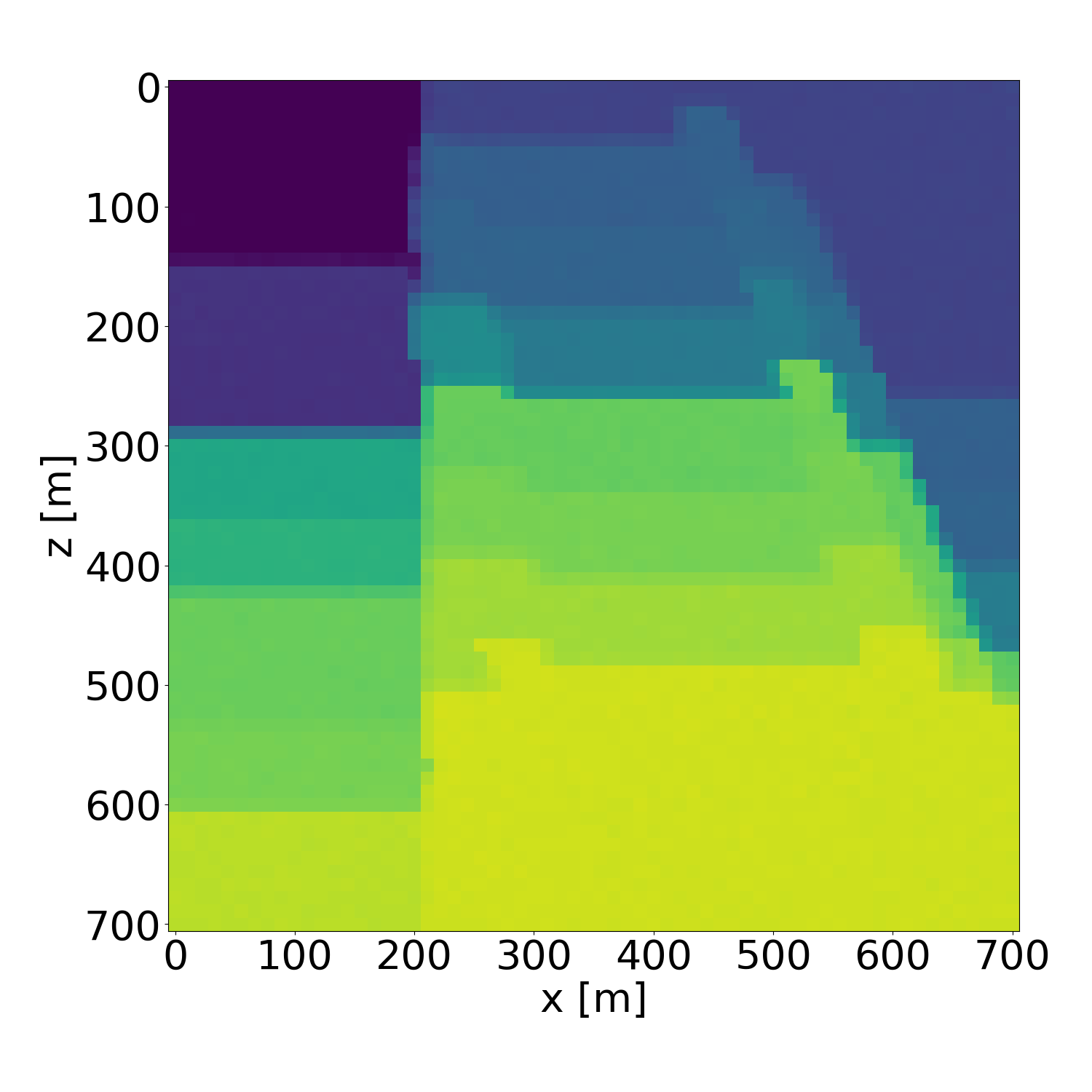}
        \caption{Probabilistic inversion result 3}
    \end{subfigure}
    \hfill
    \begin{subfigure}{0.3\textwidth}
        \includegraphics[width=\textwidth]{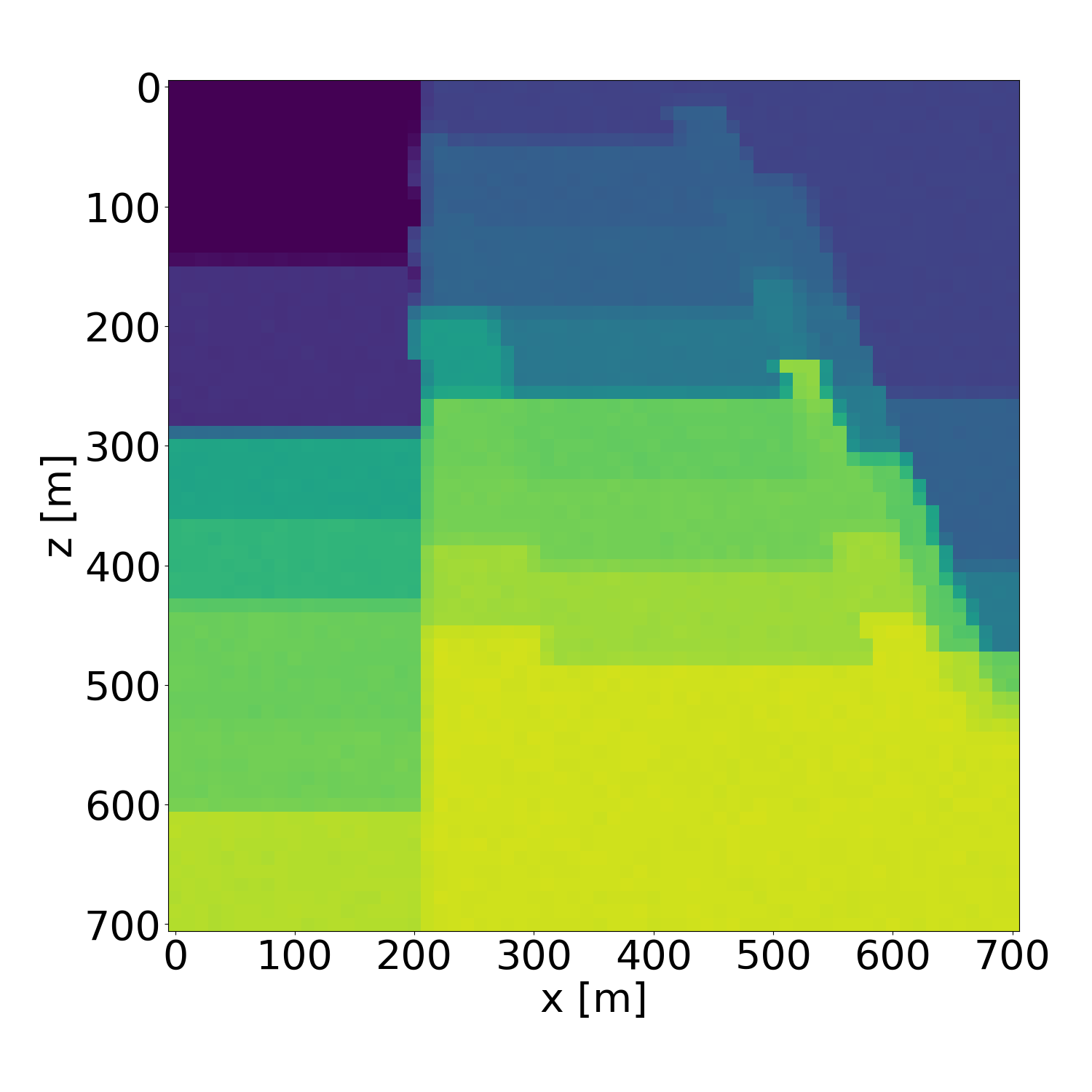}
        \caption{Probabilistic inversion result 4}
    \end{subfigure}
    \hfill
    \begin{subfigure}{0.3\textwidth}
        \includegraphics[width=\textwidth]{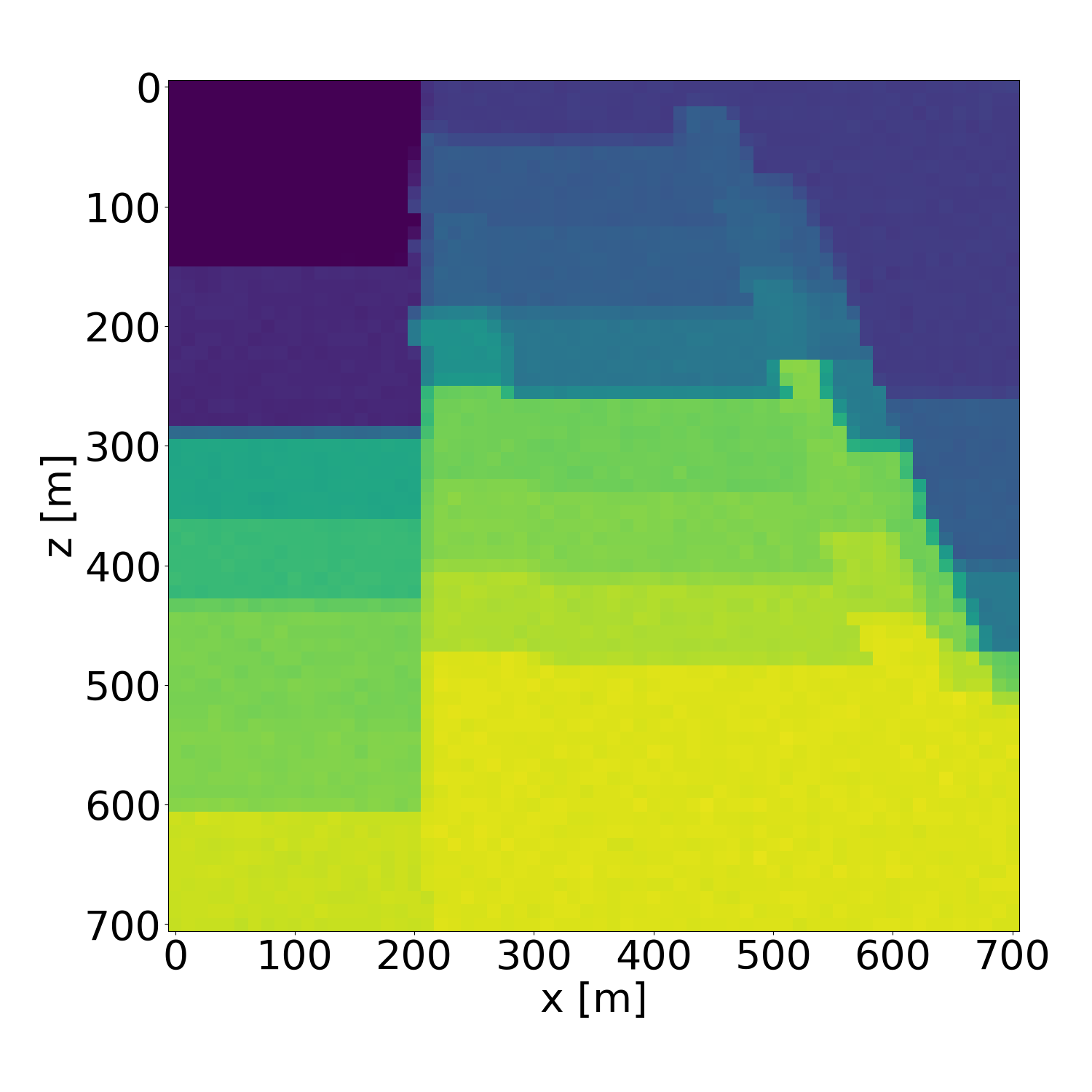}
        \caption{Probabilistic inversion result 5}
    \end{subfigure}
    \hfill
    \begin{subfigure}{0.3\textwidth}
        \includegraphics[width=\textwidth]{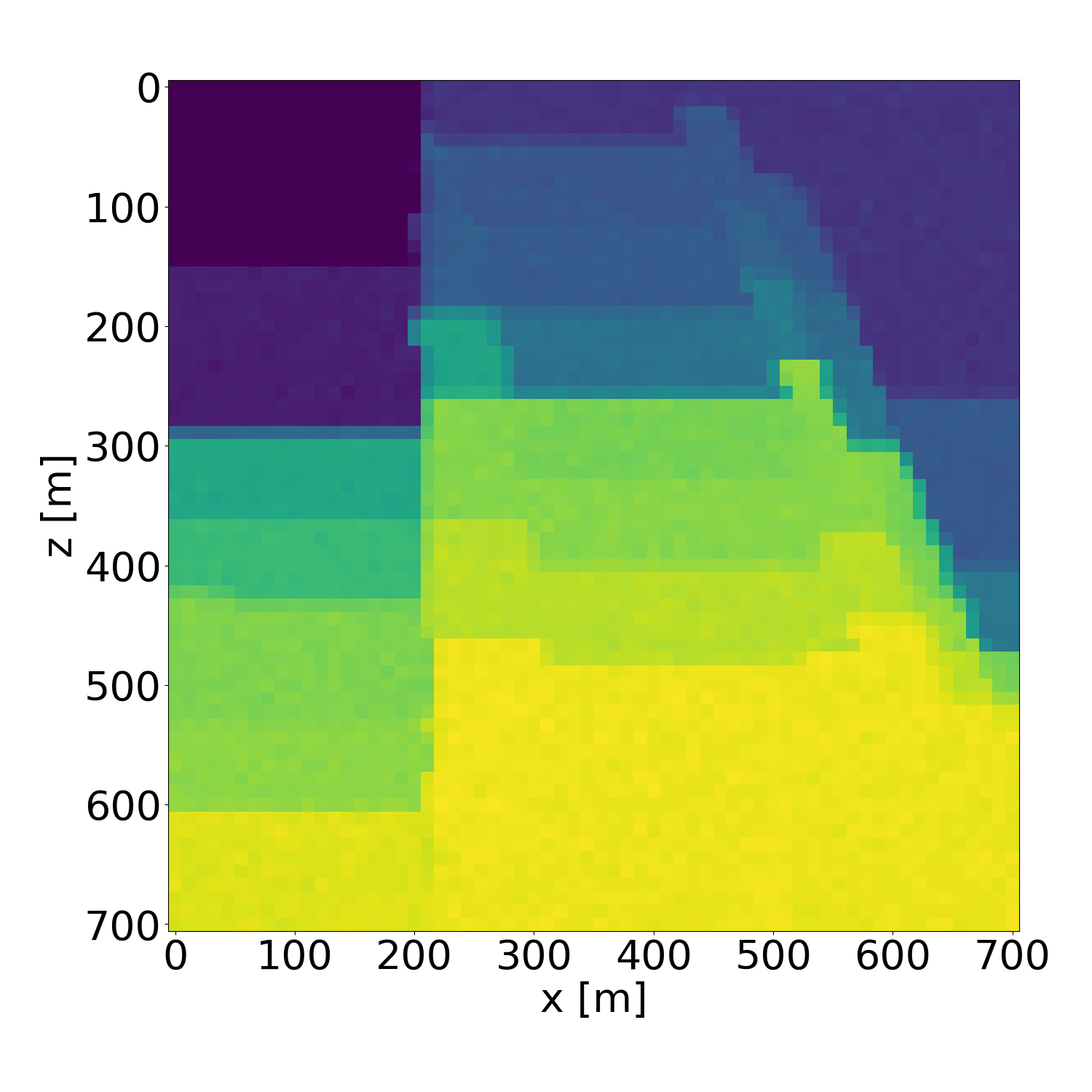}
        \caption{Probabilistic inversion result 6}
    \end{subfigure}
    \hfill
    \begin{subfigure}{0.3\textwidth}
        \includegraphics[width=\textwidth]{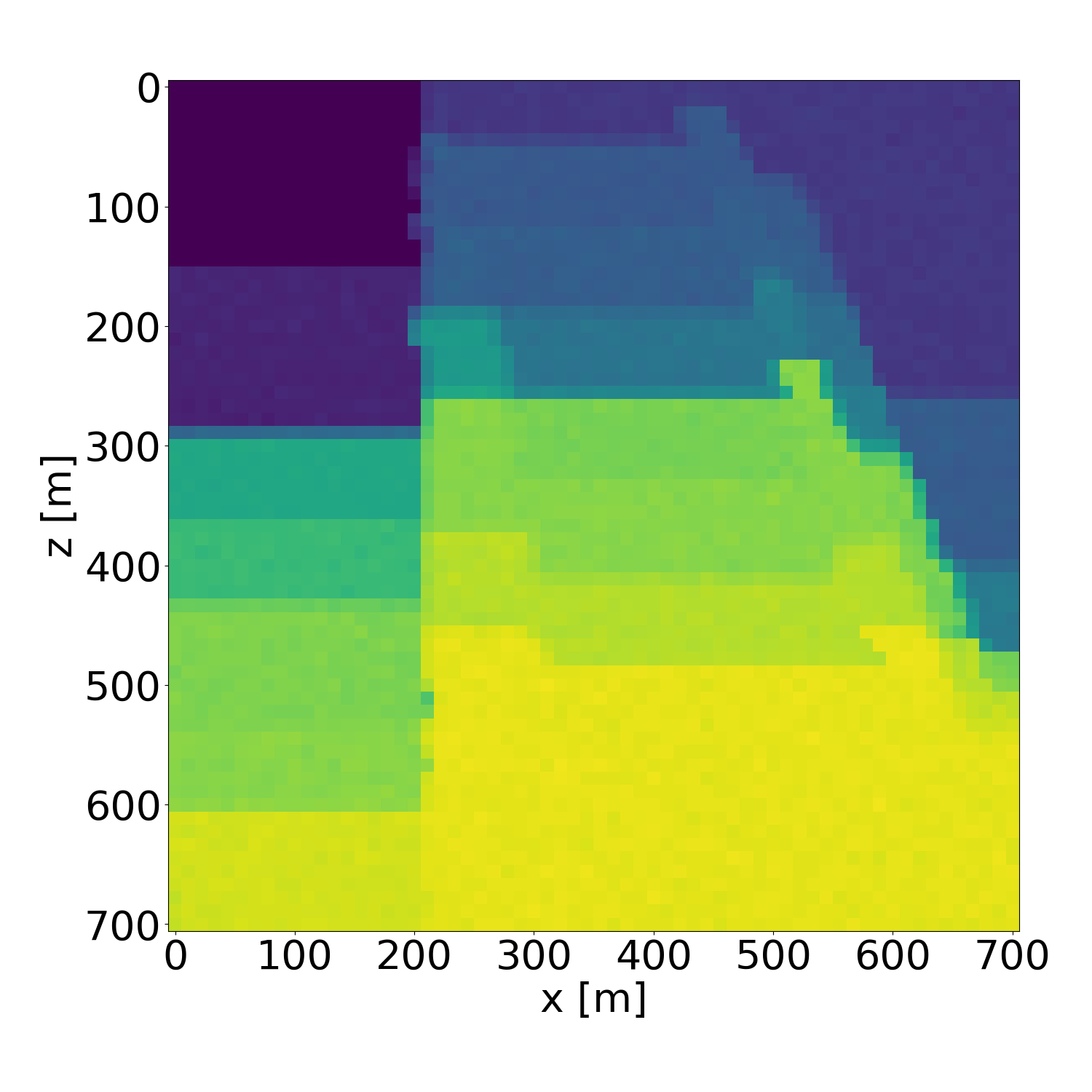}
        \caption{Probabilistic inversion result 7}
    \end{subfigure}
    \hfill
    \begin{subfigure}{0.3\textwidth}
        \includegraphics[width=\textwidth]{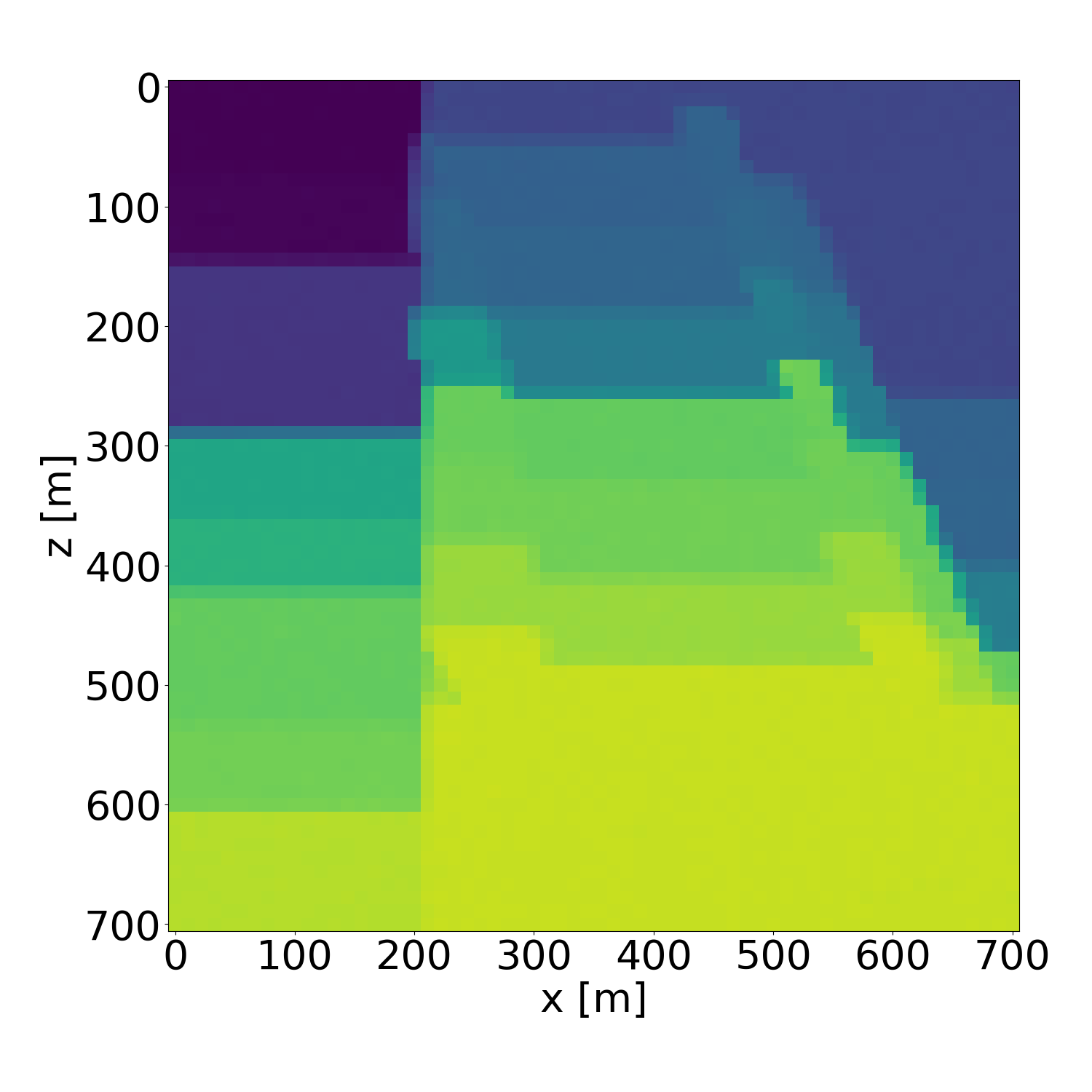}
        \caption{Probabilistic inversion result 8}
    \end{subfigure}
    \hfill
    \begin{subfigure}{0.3\textwidth}
        \includegraphics[width=\textwidth]{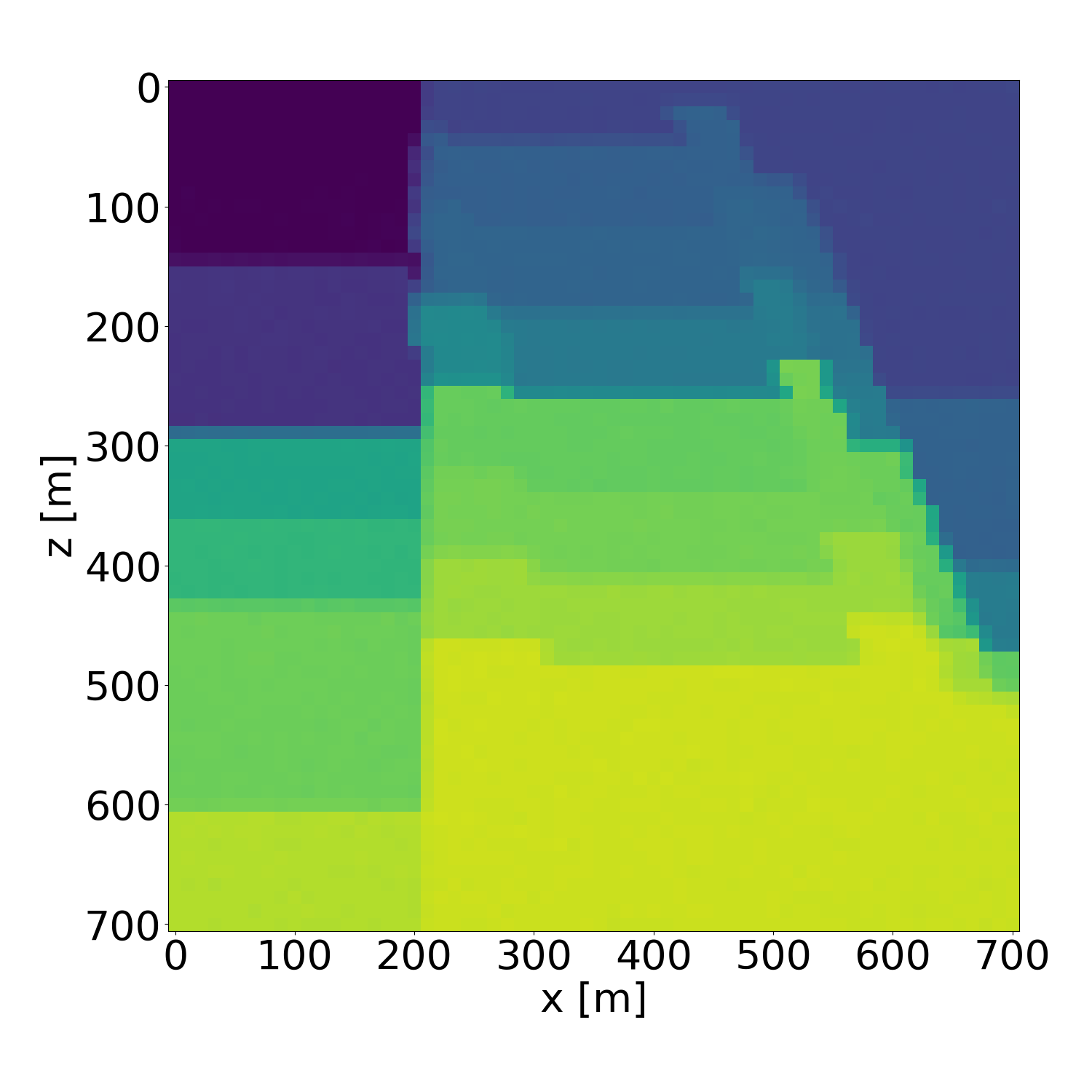}
        \caption{Probabilistic inversion result 9}
    \end{subfigure}
    \hfill
    \caption{Probabilistic inversion results of a isotropic velocity model in the validation dataset. Generated with guidance scale $w=4$.}
    \label{fig-allresult10}
\end{figure}

\label{lastpage}

\end{document}